\documentclass[preprint,3p,authoryear]{elsarticle}

\usepackage{amsmath,amssymb,amsfonts}
\usepackage{algorithmic}
\usepackage{graphicx}
\usepackage[]{algorithm2e}
\usepackage{textcomp}
\usepackage{xcolor}
\usepackage{epstopdf}
\usepackage{multirow}
\usepackage{mathtools}
\usepackage[small]{caption}
\usepackage{subcaption}
\usepackage{tabularx}
\usepackage{pbox}
\usepackage{setspace}
\usepackage{relsize}
\usepackage{amsbsy}
\usepackage{mathtools}
\usepackage[referable]{threeparttablex}
\DeclareMathOperator*{\argmin}{argmin}
\DeclareMathOperator*{\argmax}{argmax}

\journal{Neural Network}

\begin{document}
	
\begin{frontmatter}

		\title{A novel method for extracting interpretable knowledge from a spiking neural classifier with time-varying synaptic weights}

		\author[label1,label2]{Abeegithan Jeyasothy\corref{correspondingauthor}}
		\cortext[correspondingauthor]{Corresponding author}
		\ead{abeegith1@e.ntu.edu.sg}
		\author[label3]{Suresh Sundaram}
		\author[label2]{Savitha Ramasamy}
		\author[label1]{Narasimhan Sundararajan}

		\address[label1]{School of Computer Science and Engineering, Nanyang Technological University, Singapore}
		\address[label3]{Department of Aerospace Engineering, Indian Institute of Science, Bangalore, India}
		\address[label2]{Institute for Infocomm Research, Agency for Science, Technology and Research, Singapore}
		
\begin{abstract}
	
This paper presents a novel method for information interpretability in a Multi-Class Synaptic Efficacy Function based leaky-integrate-fire neuRON (MC-SEFRON) classifier that uses time-varying synaptic weights. To develop a method to extract knowledge stored in a trained multi-class classifier, first, the binary-class SEFRON classifier developed earlier is extended to handle multi-class problems. For a given input sample, MC-SEFRON uses the population encoding scheme to encode the real-valued input data into spike patterns. MC-SEFRON is trained using the same supervised learning rule given in the binary-class SEFRON classifier. After training, the proposed knowledge encoding method extracts the knowledge for a given class stored in the classifier by mapping the weighted postsynaptic potential in the time domain to the feature domain as Feature Strength Functions (FSFs). A set of FSFs corresponding to each output class represents the extracted knowledge from the MC-SEFRON classifier. This knowledge encoding method is derived to maintain consistency between the classification in the time domain and the feature domain. Also, the correctness of the extracted knowledge is quantitatively measured by using the FSFs directly for classification tasks. For a given input, each FSF is sampled at the input value to obtain the corresponding feature strength value. Then the aggregated feature strength values obtained for each class are used to determine the output class labels during classification. For a given input, feature strength values are used to interpret the predictions during the classification task. Using ten benchmark UCI machine learning datasets and the MNIST dataset, the knowledge extraction method, interpretation and the reliability of the extracted knowledge are demonstrated. Based on the studies, it can be seen that on an average, the difference in the classification accuracies using the extracted knowledge directly and those obtained by MC-SEFRON classifier is only around $0.9\%~\&~0.1\%$ for UCI machine learning datasets and the MNIST dataset respectively. This clearly shows that the knowledge represented by the FSFs of the MC-SEFRON has acceptable reliability and the interpretability of classification using the classifier's knowledge has been justified.
			
\end{abstract}
		
		
\end{frontmatter}

\section{Introduction}

Although artificial neural networks have great potential for prediction and pattern recognition in several applications, they have remained mostly as black-boxes that are difficult to interpret the reasons for their predictions. This renders them mostly uncertain and unreliable, especially, for decision making in sensitive applications like healthcare. Only recently, there has been considerable research attention towards developing interpretable machine learning approaches.

Earlier methods on deriving interpretability from machine learning classifiers focussed on gradient descent based sensitivity analysis methods tailored to specific classification techniques. For example, gradient propagation from the output layer to the input layer in a convolutional neural network enables one to visualize class sensitive input regions in~\cite{SA}. On the other hand, gradients are propagated through a deconvolution method in~\cite{Deconv} to map the relationship between specific regions of the input to the inferences of a convolutional neural network. During the sensitivity analysis, the gradient values coming into the Rectified Linear Unit (ReLU) are set to zero if the input value to the ReLU is negative during the forward pass in~\cite{SA}, whereas the gradient values coming into the ReLU are set to zero only if it is negative in~\cite{Deconv}. Guided backpropagation combines both the above approaches in~\cite{SA,Deconv} and is used in~\cite{Guideprop}, where the gradient values coming into ReLU are set to zero if either the gradient value or input during the forward pass is negative. However, these methods suffer in determining the input regions that are negatively influencing the inference. Propagation of importance score is conserved within the layers of the network in~\cite{LRP} to include the negatively influencing input regions. As there are class dependencies in these important scores, Deep Learning Important FeaTures (DeepLIFT)~\cite{DeepLift}  computes the changes in the importance scores for input regions for each class. 

In addition to these gradient based sensitivity analysis methods, there exist input perturbation based sensitivity analysis methods to interpret the significance of individual feature to the inferences in each class~\cite{LIME,SHAP}. These methods are often validated by eliminating the ranked features in sequence to establish their empirical significance to the obtained inference~\cite{LRPcomp,DeepLift,SHAP}. Although all the above methods help to establish empirical relationships between the set of input features to their corresponding model-based predictions, they do not quantify the uncertainty measures of their interpretations. 

An alternative approach in explaining the prediction of a classifier is by expressing the knowledge acquired after training the neural network in an easily explainable form ~\cite{Distil,transDistil,che}. In~\cite{SoftTree}, soft targets from the deep neural network are used to train a soft decision tree to interpret the predictions. A decision tree is learned to provide semantic level explanations for convolutional neural network predictions in~\cite{CNNtree}. In these methods,  the classification accuracy of the new explainable model is used to show the reliability of the explanations for the classifier's predictions. However, these explainable models are not extracted from the trained classifier rather they are also a newly trained classifier that produces similar outputs as the base-classifier. Due to the fundamental differences in the classifiers, the interpretation of the prediction inferred from these new classifiers may not be aligned with the base-classifier's predictions.  

Although there are several studies on interpreting decisions made by deep neural networks as indicated above, these methods do not have a framework to validate the consistency of the explanation with the classifier's prediction. Also due to the difference in the activation unit (spiking neuron), these methods may require major modifications for use in Spiking Neural Networks (SNNs). Even though SNNs are energy efficient and hardware friendly compared to other artificial neural networks as mentioned in the recent review~\cite{DeepSNN}, SNNs still remain as black-boxes and hardly any study on deriving an interpretation from the trained SNN classifiers exist. To overcome these problems, we propose a new method herein to extract the knowledge stored in a spiking neural classifier with time-varying weights that is also reliable and consistent.

In this paper, a multi-class classifier for the spiking neuron with time-varying weight model~\cite{Abeegithan}, referred (hereafter) to as a Multi-Class-SEFRON (MC-SEFRON) is first developed. MC-SEFRON classifier is trained using the modified  Spike-Timing-Dependent Plasticity (STDP) rule developed earlier in~\cite{Abeegithan}. In a trained MC-SEFRON classifier, the input information is encoded in synaptic efficacy functions (time-varying weight functions). Interpreting the predictions made by MC-SEFRON in the time domain is very challenging. However, interpreting and visualizing the predictions are much easier in the feature domain. Hence, in this paper, we propose a new knowledge encoding method to extract knowledge from a trained MC-SEFRON classifier by mapping the weighted postsynaptic potentials in the time domain into the actual feature space as functions of the features, referred to as Feature Strength Functions (FSF). FSF is the new interpretable form of an MC-SEFRON classifier in the feature domain. A set of FSFs provides the representation of the knowledge learned by a trained MC-SEFRON classifier. This new knowledge encoding method is derived to maintain the consistency between the classifications made by the MC-SEFRON classifier and the FSF both in the time domain and the feature domain respectively. FSF is then used to demonstrate the interpretability of MC-SEFRON's predictions during classification. 

The correctness of using the FSF for classification directly is measured by its performance (classification accuracy) on the same classification task used for the MC-SEFRON classifier. For a given input, sampling the FSFs for a given output class at the input feature values gives the corresponding feature strength values. During the classification using FSF, output class label for a given input sample is predicted by the output class that corresponds to the highest aggregated feature strength value. The explanation for prediction during the classification is provided based on both the given input sample's individual and aggregated feature strength values. Performance of both the MC-SEFRON and FSF are evaluated using ten benchmark data sets from the UCI machine learning repository and the MNIST dataset. Based on the study results, it can be seen that on an average the difference between the classification accuracy of MC-SEFRON and FSF (for the same classification problem) is around $0.9\%$ for the datasets from UCI machine learning repository and is $0.1\%$ for the MNIST dataset.

The paper is organized as follows: first, the multi-class classification problem formulation and the learning algorithm for an MC-SEFRON classifier are presented in Section~\ref{section:Sefron}. In Section~\ref{section:Knowledge_Extraction}, the detailed framework for extracting interpretable knowledge from the MC-SEFRON classifier is presented. Section~\ref{section:experiment} presents the performance and the interpretability of the MC-SEFRON classifier using both the ten UCI machine learning datasets and the MNIST dataset. Finally, the conclusions from the study are summarised in Section~\ref{section:conclusion}.

\section{MC-SEFRON classifier for multi-class problems} 
\label{section:Sefron}

\begin{figure}[htb!]	
	\centering
	\includegraphics[width=0.6\linewidth]{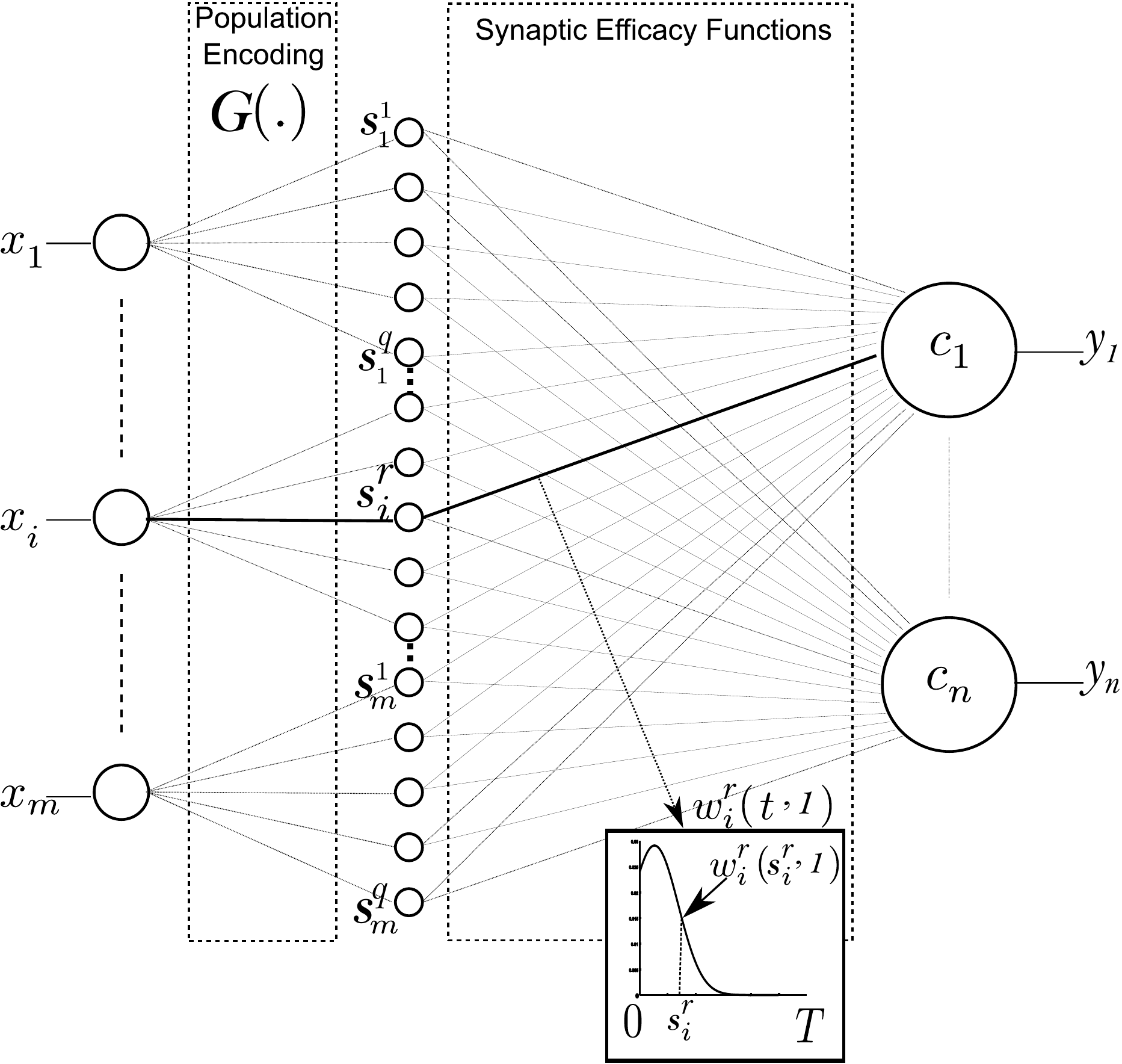}
	\caption{Architecture of MC-SEFRON with 5 RF neurons in population encoding scheme. Zoomed in view of Synaptic efficacy function between $c_1$ output neuron and the $r^{th}$ RF neuron of the $i^{th}$ input feature shows the sampling of momentary weight $w_i^r\big(s_i^r,1\big)$ from $w_i^r\big(t,1\big)$.}
	\label{fig:MC_SEFRON2}
\end{figure}

MC-SEFRON is a spiking neural classifier without any hidden layers, where input neurons are directly connected to output neurons via time-varying weight models (synaptic efficacy functions). The architecture of an MC-SEFRON classifier is shown in Fig~\ref{fig:MC_SEFRON2}. Here the weight between an input and an output neuron is a time-varying function instead of a fixed value. The time-varying weight model is represented by a sum of multiple time-varying kernels, here Gaussian kernels (time-varying functions) are used and they are learned by the supervised learning rule as given in~\cite{Abeegithan}. In MC-SEFRON, a modified STDP~\cite{STDP_overview} rule is used to determine the required weight updates corresponding to each presynaptic spike. The amplitudes and the centers of the time-varying kernels are determined by the amplitudes of the weight updates and the time of the corresponding presynaptic spikes, respectively.  

\subsection{Multi-class classification problem}

Here the function $S(.)$ represents an ideal MC-SEFRON classifier. For an input $\textbf{x}=[x_1, x_2,..., x_m]$ where $\textbf{x}\in[0,1]^m$ (here  $m$ is the number of features in the input data), $S(\textbf{x})$ produces an output $\textbf{y}=[y_1, y_2,..., y_n]$ where $\textbf{y}$ is the vector of first postsynaptic spike time from each output classes and $\textbf{y}\in[0,T+\delta T]^n$. Here $n$ is the total number of output classes and $T$ is the presynaptic spike interval limit and $\delta T$  represents an incremental smaller time after $T$ allow late postsynaptic spikes. 

The ideal functional relationship between the input $\textbf{x}$ and the output $\textbf{y}$ is denoted as $\textbf{y} = S(\textbf{x})$. For an input $\textbf{x}^k$ belonging to the class $c_k$, an ideal classifier produces an output $\textbf{y}$ given by,
\begin{equation}
y_j=\begin{dcases}
\hat{t}_d & \text{if $j = k$} \\ 
\hat{t}_d +t_m & \text{otherwise} 
\end{dcases}
\label{equation:output_y}
\end{equation}

Here $j~(j\in \{1,n\})$ is the index of the output class neuron, $\hat{t}_d$ is the desired postsynaptic spike time and $t_m$ represents a smaller margin time.

The classification rule for MC-SEFRON is the same rule as in other multi-class SNN classifiers~\cite{SpikeProp}. In this classification rule, output class label for an input pattern is determined by the postsynaptic neuron (output neuron) that fires first and the class label $\hat{c}$ is predicted as,
\begin{equation}
\hat{c}=\argmin_j \{\textbf{y} \}
\label{equation:classification_rule}
\end{equation}

By examining equation~\ref{equation:output_y} and~\ref{equation:classification_rule}, it can be observed that an ideal classifier will always predict the correct class for a given input sample. In this paper, MC-SEFRON is trained to approximate the ideal classifier function $S(.)$. It may be noted that the input $\textbf{x}$ is in the feature domain and the output $\textbf{y}$ is in the time domain. Hence, coded output class labels for a supervised learning framework are chosen as some desired postsynaptic spike times. In this paper, the same supervised learning rule proposed in~\cite{Abeegithan} is used to determine a reference postsynaptic spike time $\hat{t}^\text{rf}_j$ for each class as the coded output class label. The approximated classifier function is denoted as $\hat{S}(.)$, where the $\hat{S}(\textbf{x})$ produces an output $\hat{\textbf{y}}$  \big( $\textbf{y}=\hat{S}(\textbf{x})+\varepsilon=\hat{\textbf{y}}+\varepsilon$\big). The approximation error $\varepsilon$ is generally caused by differences within the same class input patterns.

\subsection{Population encoding scheme}
\label{section:Population_coding}

Since the input $\textbf{x}$ is real-valued, first, an encoding scheme is required to convert the real values to spike trains. A real-valued input $\textbf{x}$ is encoded into a spike pattern $\textbf{s}$ using the population encoding scheme~\cite{SpikeProp}. In the population encoding scheme, multiple Gaussian functions with evenly spaced centres are used as Receptive Field (RF) neurons and the input data is projected into a higher dimensional space ($\textbf{R}^{m\times q}$, where $q$ is the number of RF neurons used in the population encoding) as shown in Fig~\ref{fig:MC_SEFRON2}. 

For a given input $\textbf{x}$, each RF neuron produces a firing strength $\phi_i^r$, where $i$ ($i\in \{1,m\}$) represents the $i^{th}$ input feature and $r$ ($r \in \{1,q\}$) represents the $r^{th}$ RF neuron. The firing strength $\phi_i^r$ determines the presynaptic spike time $s_i^r$ \Big($s_i^r=T.\big(1-\phi_i^r\big)$\Big) for the $r^{th}$ RF neuron corresponding to the $i^{th}$ input feature. A higher RF neuron firing strength ($\phi_i^r$ closer to $1$) corresponds to an early presynaptic firing ($s_i^r$ closer to $0s$) and a lower RF neuron firing strength ($\phi_i^r$ closer to $0$) correspond to a late presynaptic firing ($s_i^r$ closer to $Ts$). 

A generic population encoding scheme is denoted by $G(.)$ and given by,

\begin{equation}
G(\textbf{x})=\textbf{s} ~
\label{equation:population_encoding}
\end{equation}
where $\textbf{s}\in[0,T]^{m\times q}$. For an $i^{th}$ input feature $x_i$, an encoded presynaptic spike pattern $\textbf{s}_i$ is obtained by,
\begin{equation}
G(x_i)=\textbf{s}_i=\big\{s_i^1,s_i^2,.....,s_i^q\big\}
\end{equation} 

It may be noted that in this scheme each presynaptic neuron will fire only one spike for a given input pattern. Hence, there are $m\times q$ number of presynaptic neurons in the classifier.

\subsection{MC-SEFRON's learning algorithm}
\label{section:Multi_Sefron}

In a MC-SEFRON classifier, $j^{th}$ output neuron fires a postsynaptic spike $\hat{t}_j$ when the postsynaptic potential $v_j(t)$ crosses its firing threshold $\theta_j$. 
\begin{equation}
\hat{t}_j=\big\{t|v_j(t)=\theta_j\big\}
\end{equation}
Here, the postsynaptic potential $v_j(t)$ of $j^{th}$ output neuron is determined as,
\begin{equation}
v_j(t)=\sum_{i\in \{1,m\}}^{}\sum_{r \in \{1,q\}} w_i^r\big(s_i^r,j\big). \epsilon\big(t-s_i^r\big). H\big(t-s_i^r\big)
\end{equation}
where, $\epsilon(t)$ is the spike response function as in~\cite{Abeegithan} $\Big(\epsilon(t)=\frac{t}{\tau}.exp\big(1-\frac{t}{\tau}\big)\Big)$. $H(t)$ represents a Heaviside step function.   $w_i^r\big(s_i^r,j\big)$ is the momentary weight at the presynaptic spike time $s_i^r$ of the synapse (connection) between the $j^{th}$ output neuron and the input neuron corresponding to the $r^{th}$ RF neuron of the $i^{th}$ input feature. $w_i^r\big(s_i^r,j\big)$ is obtained by sampling the time-varying weight function $w_i^r\big(t,j\big)$ at the time instant $s_i^r$ as shown in Fig\ref{fig:MC_SEFRON2}.

In MC-SEFRON, synaptic efficacy functions are initialized using the first sample from the each classes. For a $j^{th}$ output neuron, synaptic efficacy function $w^r_i(t,j)$ of the $r^{th}~\big(r \in \{1,q\} \big)$ RF neuron of the $i^{th}$ input feature is initialized as, 
\begin{equation}
w^r_i(t,j)_\text{initial} = u^r_i\big(\hat{t}_d\big).exp\bigg(\frac{-(t-s^r_i)^2}{2\hat{\sigma}^2}\bigg)
\label{equation:Initial_W}
\end{equation}

Here, $u^r_i\big(\hat{t}_d\big)$ is the normalized STDP with respect to the desired postsynaptic firing time $\hat{t}_d$. A generic $u^r_i(\hat{t})$ with respect to a given postsynaptic spike time $\hat{t}$ is calculated as,
\begin{equation}
u^r_i(\hat{t})=\begin{dcases}
\frac{exp(-|\hat{t}-s^r_i|/\tau) }{\sum_{i=1}^{m}\sum_{r=1}^{q}exp(-|\hat{t}-s^r_i|/\tau)} & \text{for all $s^r_i \leq \hat{t}$ } \\ 
\\
\frac{-exp(-|\hat{t}-s^r_i|/\tau) }{\sum_{i=1}^{m}\sum_{r=1}^{q}exp(-|\hat{t}-s^r_i|/\tau)}& \text{for all $s^r_i >\hat{t}$ }
\end{dcases}
\label{equation:UTI}
\end{equation}
where, the sum of $u^r_i(\hat{t})$ corresponding to presynaptic spikes fired before and after $\hat{t}$ are equal to $1$ and $-1$ respectively. The firing threshold $\theta_j$ of $j^{th}$ output neuron is initialized as,

\begin{equation}
\theta_j= \sum_{i\in \{1,m\}}^{}\sum_{r \in \{1,q\}}u^r_i\big(\hat{t}_d\big).\epsilon\big(\hat{t}_d-s^r_i\big)
\label{equation:Initial_theta}
\end{equation}

During training, if the correct class output neuron fires earlier than the other class output neurons with a marginal time $t_m$, then the sample is not used to update any of the $w^r_i(t,j)$. Synaptic efficacy functions are only updated if a wrong class output neuron fires within the marginal time during a correct classification or if there is a misclassification. The weight update of the connection between the $j^{th}$ output neuron and the input neuron corresponding to the $r^{th}$ RF neuron of the $i^{th}$ input feature  is denoted by $\triangle w^r_i(s^r_i,j)$ and determined as,
\begin{equation}
\begin{aligned}
\triangle w^r_i\big(s^r_i,j\big)=\lambda. u^r_i\big(\hat{t}^\text{rf}_j\big). \theta_j. \Bigg(\frac{1}{\sum_{i=1}^{m}\sum_{r=1}^{q}u^r_i\big(\hat{t}^\text{rf}_j\big).\epsilon\big(\hat{t}_d-s^r_i\big)} - \frac{1}{\sum_{i=1}^{m}\sum_{r=1}^{q}u^r_i\big(\hat{t}_j\big).\epsilon\big(\hat{t}_j-s^r_i\big)}\Bigg)
\end{aligned}
\end{equation}
Here, $\lambda$ is the learning rate, $\hat{t}_j$ is the actual postsynaptic firing time of the $j^{th}$ output neuron and $\hat{t}^\text{rf}_j$ is the reference postsynaptic spike time (coded output signal for supervised learning) . Please refer to~\cite{Abeegithan} for a detailed discussion on deriving the weight update rule. For an input belonging to $j^{th}$ output class, during a wrong classification, $\hat{t}^\text{rf}_j$ is set to $\hat{t}_d$ and $\hat{t}^\text{rf}_h$ ($h\neq j$) is set to $min\big(\hat{t}_j+t_m, T+\delta T\big)$. Algorithm~\ref{Algo:Pseudo_code} presents the pseudocode for determining $\hat{t}^\text{rf}_j$ and also training the MC-SEFRON classifier. 

The change in weight $\triangle w^r_i\big(s^r_i,j\big)$ is a single value and it has been embedded in a time-varying function $g^r_i(t,j)$ as,
\begin{equation}
g^r_i(t,j)=\triangle w^r_i\big(s^r_i,j\big).exp\bigg(\frac{-(t-s^r_i)^2}{2\sigma^2}\bigg)
\end{equation}
where $\sigma$ is the efficacy update range. More information on a single synapse can be stored by setting a smaller value to $\sigma$ whereas an infinite value for $\sigma$ results in a single-weight model instead of a time-varying weight model. 

For the $j^{th}$ output neuron, the synaptic efficacy function of the $r^{th}$ RF neuron for the $i^{th}$ input feature is updated as, 
\begin{equation}
w^r_i(t,j)_\text{new} = w^r_i(t,j)_\text{old} + g^r_i(t,j)
\label{equation:Synaptic_efficacy_update}
\end{equation}

\begin{algorithm}[htb!]
	\caption{Pseudo code to train a MC-SEFRON classifier}
	\label{Algo:Pseudo_code}

	\For{all samples}{
		\eIf{first sample from class $j$ }{
			Initialize $w^r_i(t,j)$ for all $i$ and $r$ and firing threshold $\theta_j$ (using equation ~\ref{equation:Initial_W} and ~\ref{equation:Initial_theta} respectively)\;
		}
		{
			
			\If{sample belongs to class $j$}{			
				$\hat{t}_j \gets$  Postsynaptic spike time of class $j$ output neuron\;
				$\hat{t}_h \gets$ Postsynaptic spike times of class $h$ $(h\neq j)$ output neurons\;

				{\eIf{$\hat{t}_j+t_m\leq min(\hat{t}_{h})$}{   
						The sample is not used to update the network parameters.\;
					}
					{
						\If{$\hat{t}_j > \hat{t}_d$}
						{Update $w^r_i(t,j)$ for all $i$ and $r$ with $\hat{t}^\text{rf}_j=\hat{t}_d$. \;}
						\For{all output class $h$ $(h\neq j)$ }{
						\If{$\hat{t}_h<\hat{t}_j+t_m$}{
								Update $w^r_i(t,h)$ for all $i$ and $r$ with $\hat{t}^\text{rf}_h=min\big(\hat{t}_j+t_m$, $T+\delta T\big)$ \;}
						}			
					}
				}
			}	
			
		}	
	}
\end{algorithm}

\section{Knowledge encoding method to extract interpretable knowledge from MC-SEFRON}
\label{section:Knowledge_Extraction}

The previous section laid the groundwork for building and training an MC-SEFRON classifier. A well-trained classifier may have a better representation of knowledge on the dataset compared to a weak classifier. In this section, a framework is derived to extract the knowledge encoded in a trained MC-SEFRON classifier. 

The classification rule in equation~\ref{equation:classification_rule}  for an ideal MC-SEFRON classifier can be re-written as,

\begin{equation}
	\begin{aligned}
	\hat{c}&=\argmin_j \{\textbf{y} \}\\
	&=\argmin_j \min\{ t~|~ \frac{1}{\theta_j} v_j(t) > 1 \}\\
	&=\argmin_j  \min\{ t~|~ \frac{1}{\theta_j} \sum_{i=1}^{m}\sum_{r=1}^{q} w^r_i(s_i^r,j). \epsilon(t-s_i^r).H(t-s_i^r) > 1 \}
	\end{aligned}
	\label{equation:classification_rule_modified}
\end{equation}

For a given input ,an ideal classifier would fire the postsynaptic spikes at the desired firing times. Therefore, at the time of classification, the correct output neuron would have fired a postsynaptic spike at $\hat{t}_d$ and the remaining output neurons would fire postsynaptic spikes later at $\hat{t}_d + t_m$ (refer to equation~\ref{equation:output_y}). Hence, the term $\frac{1}{\theta_j} \sum_{i=1}^{m}\sum_{r=1}^{q} w_i(s_i^r,j) \epsilon(t-s_i^r). H(t-s_i^r)$ is maximum at $t=\hat{t}_d$ for the correct output class. Therefore, the classification rule can be modified as, 

\begin{equation}
\begin{aligned}
\hat{c}&=\argmax_j \frac{1}{\theta_j} \sum_{i=1}^{m}\sum_{r=1}^{q} w_i^r(s_i^r,j). \epsilon(\hat{t}_d-s_i^r). H(\hat{t}_d-s_i^r)\\
&= \argmax_j \sum_{i=1}^{m} \frac{1}{\theta_j} \sum_{r=1}^{q} w_i^r(s_i^r,j).\epsilon(\hat{t}_d-s_i^r).H(\hat{t}_d-s_i^r). 
\end{aligned}
\label{equation:classification_rule_max}
\end{equation}

The classification rule in~\ref{equation:classification_rule_max} is only applicable for an ideal classifier as $\hat{t}_d$ is always the earliest firing time in an ideal classifier. However a trained MC-SEFRON classifier \big($\hat{S}(.)$\big) produces an output $\hat{\textbf{y}}$ that only approximates the ideal output $\textbf{y}$ (produced by $S(.)$). Hence, the earliest firing time may not be same as the desired firing time due to the differences in the input patterns. Therefore, the earliest firing time has to be approximated to avoid this loss in performance. Here $\hat{t}_o$ \big($\hat{t}_o=\hat{t}_d+\varepsilon$\big) denotes the approximated earliest firing time. The equation~\ref{equation:classification_rule_max} can be rewritten after the approximation as,

\begin{equation}
\begin{aligned}
\hat{c}^*=\argmax_j \sum_{i=1}^{m} \frac{1}{\theta_j} \mathlarger{\sum}_{r=1}^{q} w_i^r\Big(s_i^r,j\Big). \epsilon\Big(\hat{t}_o-s_i^r\Big). H\Big(\hat{t}_o-s_i^r\Big)
\end{aligned}
\label{equation:classification_rule_max_modified}
\end{equation}

The approximated classification rule in equation~\ref{equation:classification_rule_max_modified} is preserved during the extraction of the interpretable knowledge. In the earlier section, population encoding scheme was used to convert the real valued input to spike times (refer to section~\ref{section:Population_coding}). Here, an inverse of the population encoding scheme is used to determine the preimage of the spike time under the map of $G(.)$. Multiple Gaussian functions are used as RF neurons in the population encoding scheme. Hence, in the inverse population encoding scheme, preimage of each presynaptic spike under the map of corresponding RF neuron will have two solutions. However, the collective solutions for the set of presynaptic spikes $\big\{s_i^1,s_i^2,.....,s_i^q\big\}$ have only one common solution $\textbf{x}_i$. Hence, in an inverse population encoding scheme, preimage of the encoded spike pattern $\textbf{s}_i$ $(\textbf{s}_i=\big\{s_i^1,s_i^2,.....,s_i^q\big\})$ has a unique solution $\textbf{x}_i$ . The inverse population encoding scheme is given by,
\begin{equation}
G^{-1}(\textbf{s}_i)=\big\{\textbf{x}_i~|~G(\textbf{x}_i)=\textbf{s}_i\big\}
\end{equation}

Due to the unique solutions in the inverse mapping it can be written as,
\begin{equation}
\begin{aligned}
\textbf{s}_i&=G\Big(G^{-1}\big(\textbf{s}_i\big)\Big)\\
&=G(\textbf{x}_i)
\end{aligned}
\end{equation}

Using this inverse population encoding scheme, the feature strength function $\psi_i(x_i,j)$ of the $i^{th}$ input feature for the $j^{th}$ output class is extracted from the weighted postsynaptic potential as,

\begin{equation}
\begin{aligned}
\psi_i(x_i,j)&=\frac{1}{\theta_j} \mathlarger{\sum}_{r=1}^{q} w_i^r\Big(s_i^r,j\Big). \epsilon\Big(\hat{t}_o-s_i^r\Big). H\Big(\hat{t}_o-s_i^r\Big)\\&=\frac{1}{\theta_j}\mathlarger{\sum}_{r=1}^{q} w_i^r\Big(G(x_i)^r,j\Big). \epsilon\Big(\hat{t}_o-G(x_i)^r\Big). H\Big(\hat{t}_o-G(x_i)^r\Big)
\end{aligned}
\label{equation:FSF}
\end{equation} 

\begin{figure}[htb!]	
	\centering
	\includegraphics[width=0.6\linewidth]{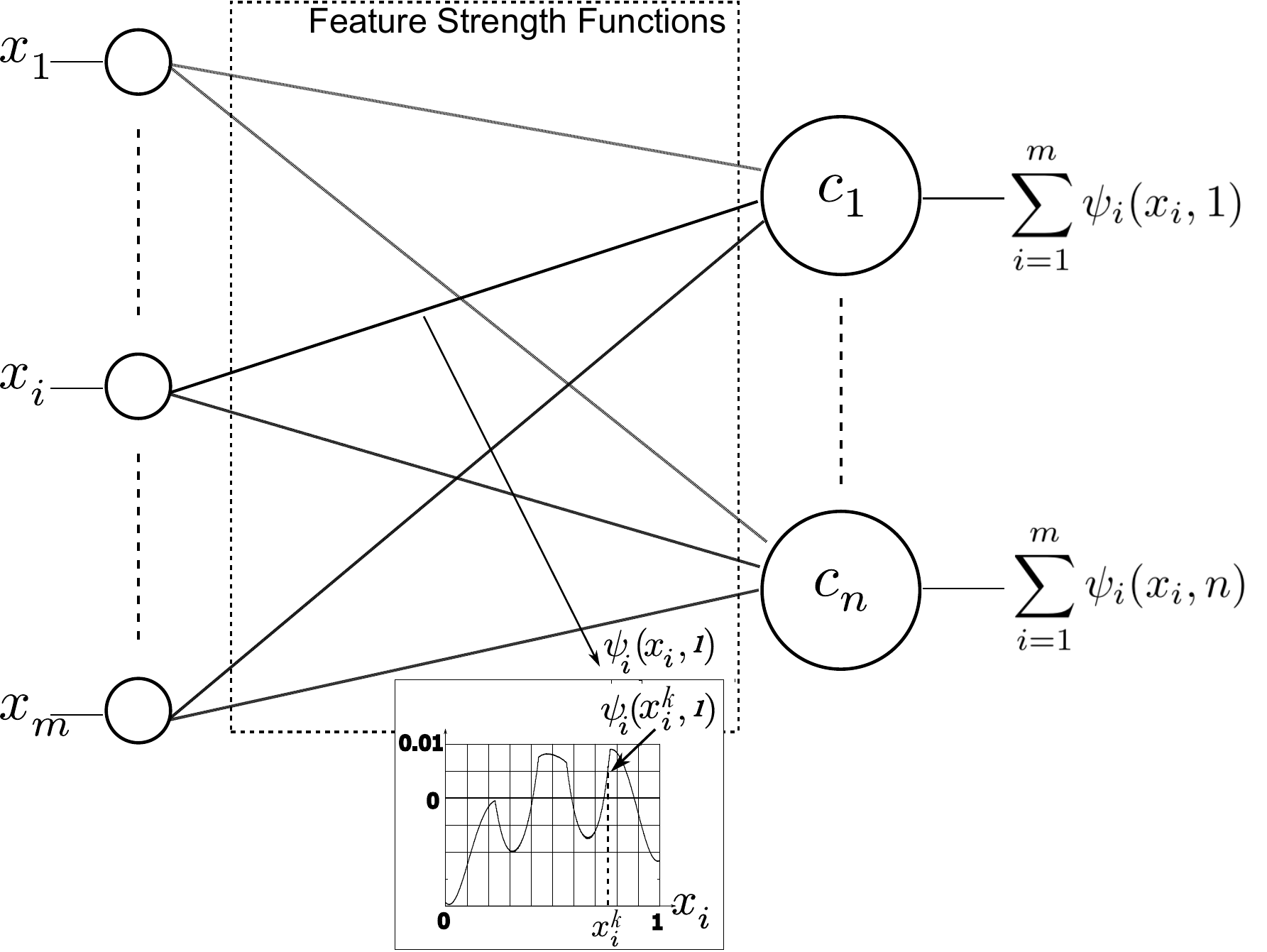}
	\caption{Architecture for using FSFs for classification tasks. Zoomed in view of the Feature Strength Function $\psi_i(x_i,j)$ shows the sampling of feature strength value $\psi_i(x^k_i,j)$ for an input feature value $x^k_i$.}
	\label{fig:FSF_classifier}
\end{figure}

This FSF imitates the input-output function learned by MC-SEFRON as templates of feature functions. Feature strength for a given feature value is sampled from the corresponding FSF. For a given input, the collective sampled feature strength values are used directly for classification and are also used to explain why a particular classification is made for those input values. For a classification task, RHS of the equation~\ref{equation:classification_rule_max_modified} is replaced by the FSF in the equation~\ref{equation:FSF}. Using FSFs, the output class label $\hat{c}^*$ for a given input sample $\textbf{x}^k=\{x^k_1,x^k_2.....,x^k_i,...,x^k_m\}$ is predicted as,
\begin{equation}
\hat{c}^*=\argmax_j\sum_{i=1}^{m}\psi_i(x^k_i,j)
\label{equation:classification_FSF}
\end{equation}
here $x^k_i \in[0,1]$ is the $i^{th}$ feature value. For classification tasks, FSF can be considered as a single input-output layer classifier. Fig~\ref{fig:FSF_classifier} shows the architecture for using FSFs for classification tasks directly that relies on equation~\ref{equation:classification_FSF}. It can be seen from Fig~\ref{fig:FSF_classifier}, that for a classification task FSF values are sampled and the feature strength values are summed at the output neuron. The output class label is predicted by that output neuron corresponding to the highest aggregated feature strength value. 

\section{Performance evaluation of MC-SEFRON and FSF using UCI datasets and MNIST dataset}
\label{section:experiment}
\subsection{Performance evaluation based on UCI datasets}
Performance of MC-SEFRON and the accuracy of the extracted knowledge are evaluated using ten benchmark datasets from the UCI machine learning repository for classification tasks. Performance of MC-SEFRON classifier is also compared with other existing SNN classifiers. The population encoding scheme is used to convert the real-valued input data into spike patterns. In the population encoding scheme, the number of RF neurons, overlap constant and the presynaptic spike interval ($T$) are set to $6$, $0.7$ and $3ms$ respectively as given in~\cite{TMM-SNN}. For all the experiments, $\delta T$, $t_m$ and $\hat{t}_d$ are set to $1ms$, $0.05ms$ and $2ms$ respectively. For each dataset, the efficacy update range $\sigma$ and the time constant $\tau$ of the STDP rule are chosen using cross validation. To enable fairness in evaluation, for the seven datasets, 10-random fold cross-validation is conducted and for the remaining three datasets a single fold validation is conducted as stated in~\cite{TMM-SNN}. Cross-validation is used to choose $\hat{t}_o$ to extract FSFs from the trained MC-SEFRON classifier. Description of the ten UCI machine learning datasets and the chosen values for $\sigma$, $\tau$-STDP and $\hat{t}_o$ are given in table~\ref{table:datasets}.

\begin{table}[htb]
	\centering
	\caption{Description of Dataset used for validation}
	\label{table:datasets}
	\resizebox{\columnwidth}{!}{
		\begin{tabular}{|c|c|c|c|c|c|c|c|}
			\hline
			\multicolumn{1}{|c|}{\multirow{2}{*}{Dataset}} & \multicolumn{1}{c|}{\multirow{2}{*}{\# Features}} & \multicolumn{1}{c|}{\multirow{2}{*}{\# Classes}} & \multicolumn{2}{c|}{\# Samples} &
			{\multirow{2}{*}{$\sigma$ (ms)}}  &	{\multirow{2}{*}{$\tau$-STDP (ms)}} &	{\multirow{2}{*}{$\hat{t}_o$ (ms)}}\\  \cline{4-5} 
		                         & \multicolumn{1}{c|}{}                             & \multicolumn{1}{c|}{}  & Training        & Testing   & \multicolumn{1}{c|}{}  & \multicolumn{1}{c|}{}  & \multicolumn{1}{c|}{}       \\ \hline
		                         \multicolumn{8}{|c|}{10 fold cross-validation}\\ \hline
			
			Iris            & 4               & 3          & 75            & 75   &0.55  &1.6 &2.38        \\ \hline
			Wine       & 13              & 3          & 60            & 118        &0.85  &4.9   &1.54  \\ \hline
			Acoustic emission           & 5               & 4          & 62            & 137      &0.9  &3.7 &   2.99    \\ \hline
					Liver          & 6              & 2          & 170            & 175     &0.25  &7.35  &2.37      \\ \hline				
				Breast Cancer          & 9              & 2          & 350            & 333 &0.4  &3.35  &2.03         \\ \hline
				Ionosphere         & 34              & 2          & 175            & 176    &0.3 &4.5  &2.64      \\ \hline
				PIMA          & 8              & 2          & 384            & 384     &0.35 &3.7 &1.85      \\ \hline

			\multicolumn{8}{|c|}{Single fold validation}\\ \hline
			Image Segmentation         & 18             & 7          & 210            &2100    &0.45  &4.55   &2.41     \\ \hline
			EEG eye state       & 14             & 2          & 9990            &4990       &0.3  &3.6  &2.11 \\ \hline
			Yeast       & 8             & 10         & 990            &494      &0.35  &2.05  &1.85 \\ \hline
	
		\end{tabular}
	}
\end{table}

Experimental results with the 10-random fold cross-validations are compared with SpikeProp~\cite{SpikeProp}, SWAT~\cite{SWAT}, SRESN~\cite{SRESN}, and TMM-SNN~\cite{TMM-SNN}. Single fold experimental results are compared with SpikeTemp~\cite{SpikeTemp}, eSNN~\cite{eSNN} and TMM-SNN~\cite{TMM-SNN}. All the results used for comparison are reproduced from~\cite{TMM-SNN}. Table~\ref{table:results_UCI} shows the performance comparison of MC-SEFRON with other algorithms and the performance of FSFs.

\begin{table*}[htb!]
	\centering
	\caption{Performance comparison on UCI dataset }
	\resizebox{0.6\totalheight}{!}{	

	\begin{tabular}{|c|c|c|c|}
		\hline
		Dataset & Method & \begin{tabular}[c]{@{}c@{}}Classifier \\ Training/Testing  accuracy (\%)\end{tabular} & \begin{tabular}[c]{@{}c@{}}Interpretable knowledge \\ Training/Testing accuracy (\%)\end{tabular} \\ \hline
		\multicolumn{4}{|c|}{10 fold cross-validation}\\ \hline
		Iris & \begin{tabular}[c]{@{}c@{}}SpikeProp\\ SWAT\\ SRESN\\ TMM-SNN\\ MC-SEFRON\end{tabular} & \begin{tabular}[c]{@{}c@{}}97.2/96.7\\ 96.7/92.4\\ 96.9/97.3\\ 97.5/97.2\\ 98.4/97.1\end{tabular} & \begin{tabular}[c]{@{}c@{}}-\\ -\\ -\\ -\\ 97.3/96.7\end{tabular} \\ \hline
		Wine & \begin{tabular}[c]{@{}c@{}}SpikeProp\\ SWAT\\ SRESN\\ TMM-SNN\\ MC-SEFRON\end{tabular} & \begin{tabular}[c]{@{}c@{}}99.2/96.8\\ 98.6/92.3\\ 96.9/91.0\\ 100/97.5\\ 98.8/94.6\end{tabular} & \begin{tabular}[c]{@{}c@{}}-\\ -\\ -\\ -\\ 97.5/94.5\end{tabular} \\ \hline
		Acoustic emission & \begin{tabular}[c]{@{}c@{}}SpikeProp\\ SWAT\\ SRESN\\ TMM-SNN\\ MC-SEFRON\end{tabular} & \begin{tabular}[c]{@{}c@{}}98.5/97.2\\ 93.1/91.5\\ 93.9/94.2\\ 97.6/97.5\\ 98.2/97.7\end{tabular} & \begin{tabular}[c]{@{}c@{}}-\\ -\\ -\\ -\\ 97.3/96.0\end{tabular} \\ \hline 
		Liver & \begin{tabular}[c]{@{}c@{}}SpikeProp\\ SWAT\\ SRESN\\ TMM-SNN\\ MC-SEFRON\end{tabular} & \begin{tabular}[c]{@{}c@{}}71.5/65.1\\74.8/60.9\\60.4/59.7\\74.2/70.4\\77.3/69.6\end{tabular} & \begin{tabular}[c]{@{}c@{}}-\\ -\\ -\\ -\\ 75.1/67.2\end{tabular} \\ \hline 
		Breast Cancer & \begin{tabular}[c]{@{}c@{}}SpikeProp\\ SWAT\\ SRESN\\ TMM-SNN\\ MC-SEFRON\end{tabular} & \begin{tabular}[c]{@{}c@{}}97.3/97.2\\96.5/95.8\\97.7/97.2\\97.4/97.2\\98.4/97.4\end{tabular} & \begin{tabular}[c]{@{}c@{}}-\\ -\\ -\\ -\\ 98.0/97.4\end{tabular} \\ \hline 
		Ionosphere & \begin{tabular}[c]{@{}c@{}}SpikeProp\\ SWAT\\ SRESN\\ TMM-SNN\\ MC-SEFRON\end{tabular} & \begin{tabular}[c]{@{}c@{}}89.0/86.5\\86.5/90.0\\91.9/88.6\\98.7/92.4\\94.2/89.7\end{tabular} & \begin{tabular}[c]{@{}c@{}}-\\ -\\ -\\ -\\ 92.0/87.2\end{tabular} \\ \hline 
		PIMA & \begin{tabular}[c]{@{}c@{}}SpikeProp\\ SWAT\\ SRESN\\ TMM-SNN\\ MC-SEFRON\end{tabular} & \begin{tabular}[c]{@{}c@{}}78.6/76.2\\77.0/72.1\\70.5/69.9\\79.7/78.1\\77.5/75.4\end{tabular} & \begin{tabular}[c]{@{}c@{}}-\\ -\\ -\\ -\\ 76.2/74.8\end{tabular} \\ \hline 
	
		\multicolumn{4}{|c|}{Single fold validation}\\ \hline
		Image Segmentation & \begin{tabular}[c]{@{}c@{}}SpikeTemp\\ eSNN\\ TMM-SNN\\ MC-SEFRON\end{tabular} & \begin{tabular}[c]{@{}c@{}}89.1/82.0\\ 71.9/70.9\\ 96.2/88.9\\ 98.1/90.3\end{tabular} & \begin{tabular}[c]{@{}c@{}}-\\ -\\ -\\ 97.1/90.2\end{tabular} \\ \hline
		EEG eye state & \begin{tabular}[c]{@{}c@{}}SpikeTemp\\ eSNN\\ TMM-SNN\\ MC-SEFRON\end{tabular} & \begin{tabular}[c]{@{}c@{}}55.4/54.6\\ 55.4/54.6\\ 55.1/55.2\\ 71.4/70.8\end{tabular} & \begin{tabular}[c]{@{}c@{}}-\\ -\\ -\\ 71.3/70.8\end{tabular} \\ \hline
		Yeast & \begin{tabular}[c]{@{}c@{}}SpikeTemp\\ eSNN\\ TMM-SNN\\ MC-SEFRON\end{tabular} & \begin{tabular}[c]{@{}c@{}}56.7/31.6\\50.5/31.4\\59.3/62.4\\56.2/55.5\end{tabular} & \begin{tabular}[c]{@{}c@{}}-\\ -\\ -\\ 56.2/55.5\end{tabular} \\ \hline
	\end{tabular}
}
\label{table:results_UCI}
\end{table*}

From the table~\ref{table:results_UCI}, it can be seen that the classification performance of MC-SEFRON is on par with other classifiers on the 10-fold cross-validation and outperforms all the classifiers on single fold validation. For the EEG eye state dataset, MC-SEFRON performs 15\% better than any other classifier. This also highlights that the performance of MC-SEFRON is better than other SNN classifiers. However, the main focus of this work is not to outperform other algorithms in the classification task but to emphasise on the quality of the knowledge extracted from MC-SEFRON. It can be seen from table~\ref{table:results_UCI}, that the performance loss is very minimal ($0.0-2.5\%$) when FSF is used directly for the classification task. This implies that the knowledge represented by FSFs is reliable. Next, the Iris dataset is used to illustrate the classification by using FSFs that directly represents the knowledge stored in an MC-SEFRON classifier.

\subsubsection{Illustration of direct classification using FSFs for the Iris dataset}

\begin{figure*}[htb!]
	\begin{subfigure}{0.5\textwidth }
		\centering
		\includegraphics[height=0.18\textheight]{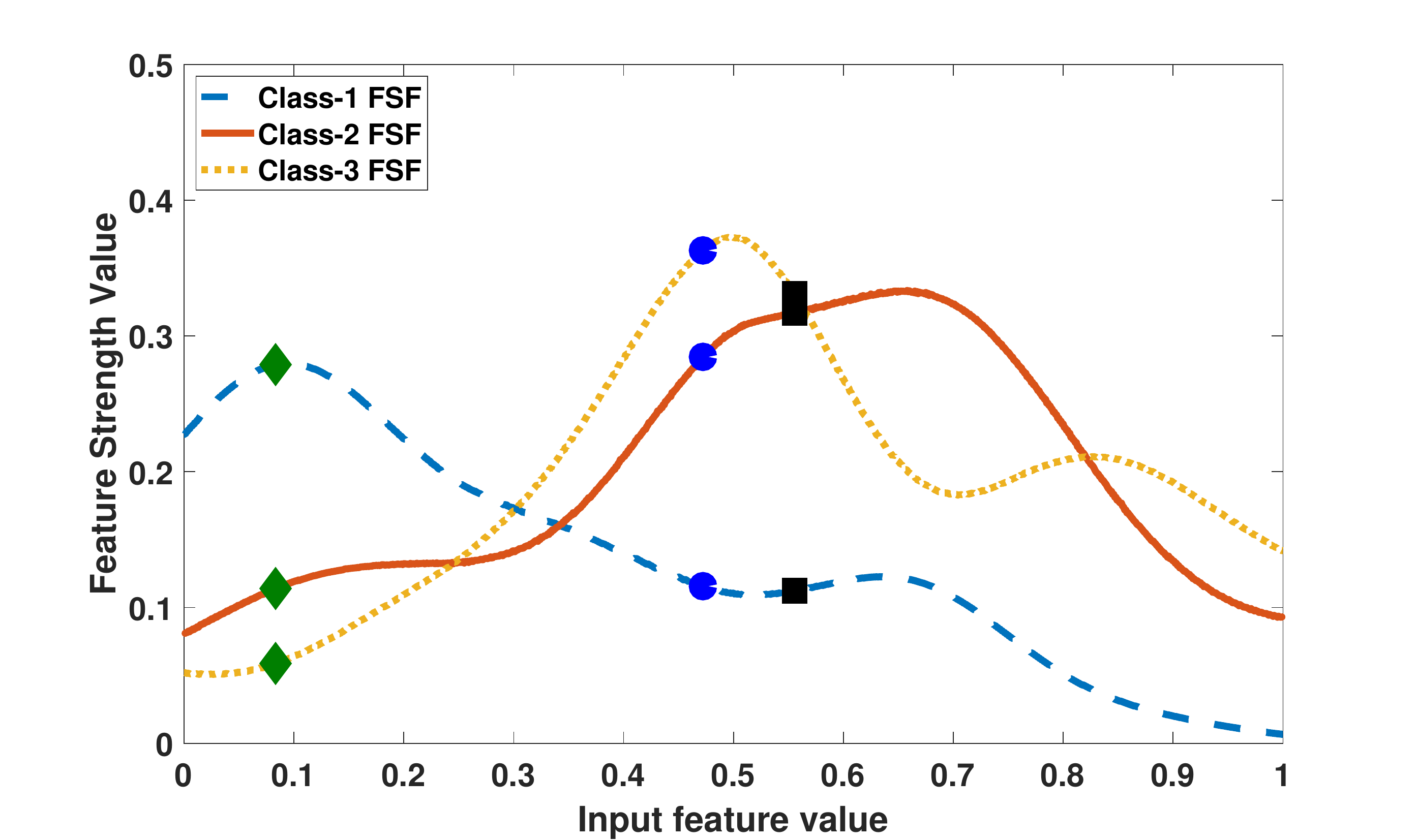}
		\caption{Sampling FSF for input feature 1}
		\label{fig:f1}
	\end{subfigure}
	\begin{subfigure}{.5\textwidth}
		\centering
		\includegraphics[height=0.18\textheight]{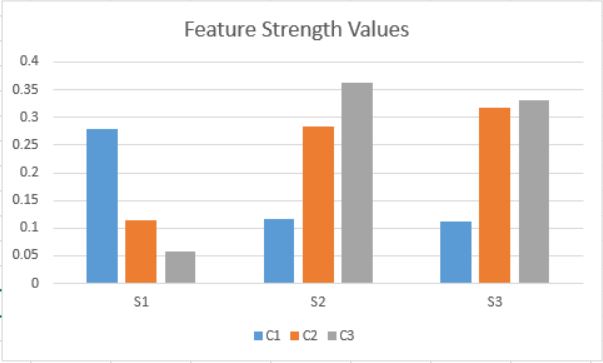}
		\caption{Sampled feature strength values for feature 1 }   \label{fig:fs1}
	\end{subfigure}
	\begin{subfigure}{.5\textwidth}
		\centering
		\includegraphics[height=0.18\textheight]{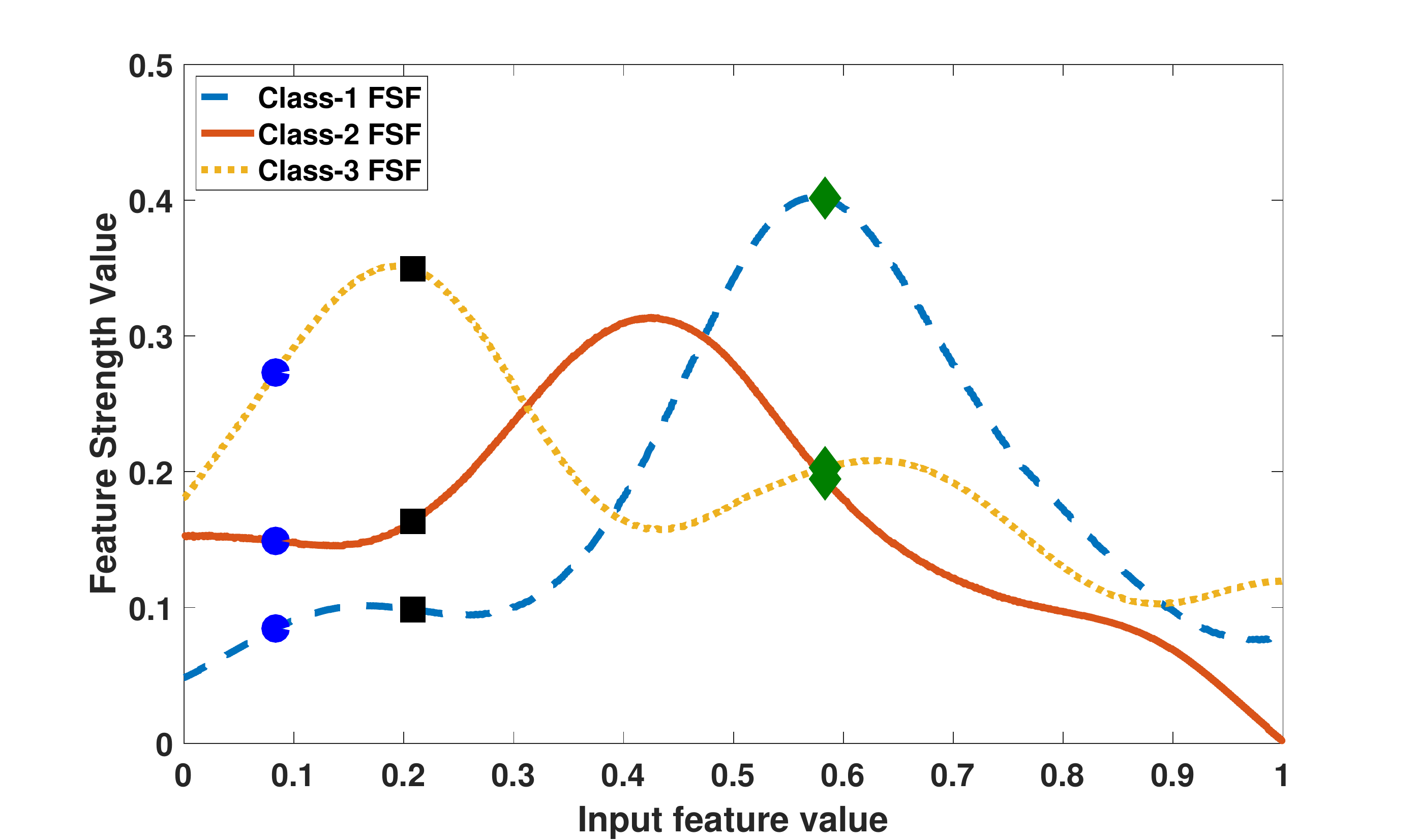}
		\caption{Sampling FSF for input feature 2}
		\label{fig:f2}
	\end{subfigure}
	\begin{subfigure}{.5\textwidth}
		\centering
		\includegraphics[height=0.18\textheight]{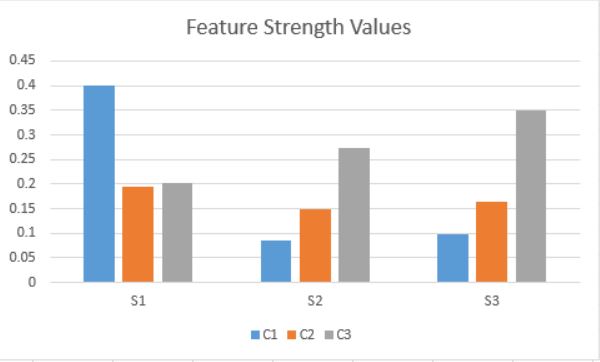}
		\caption{Sampled feature strength values for feature 2}   \label{fig:fs2}
	\end{subfigure}
	\begin{subfigure}{.5\textwidth}
		\centering
		\includegraphics[height=0.18\textheight]{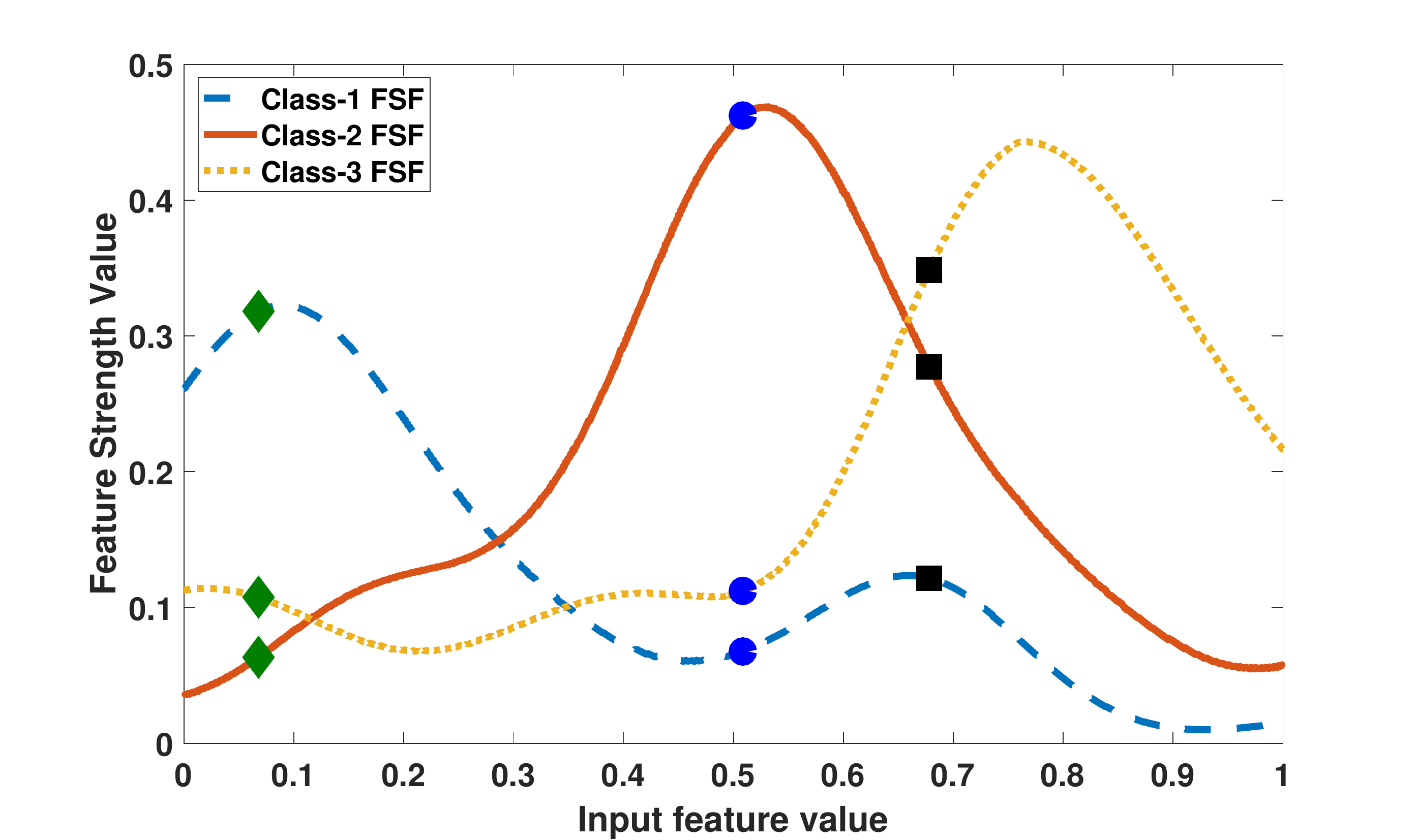}
		\caption{Sampling FSF for input feature 3}
		\label{fig:f3}
	\end{subfigure}
	\begin{subfigure}{.5\textwidth}
		\centering
		\includegraphics[height=0.18\textheight]{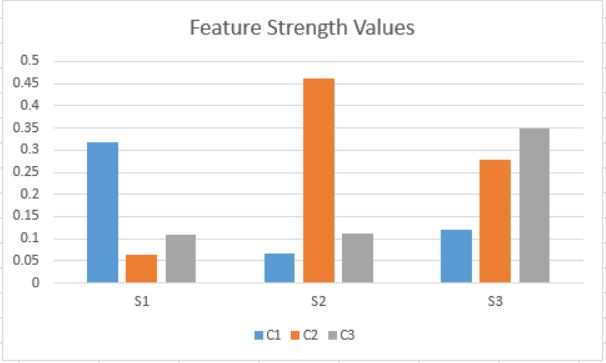}
		\caption{Sampled feature strength values for feature 3}   \label{fig:fs3}
	\end{subfigure}
	\begin{subfigure}{.5\textwidth}
		\centering
		\includegraphics[height=0.18\textheight]{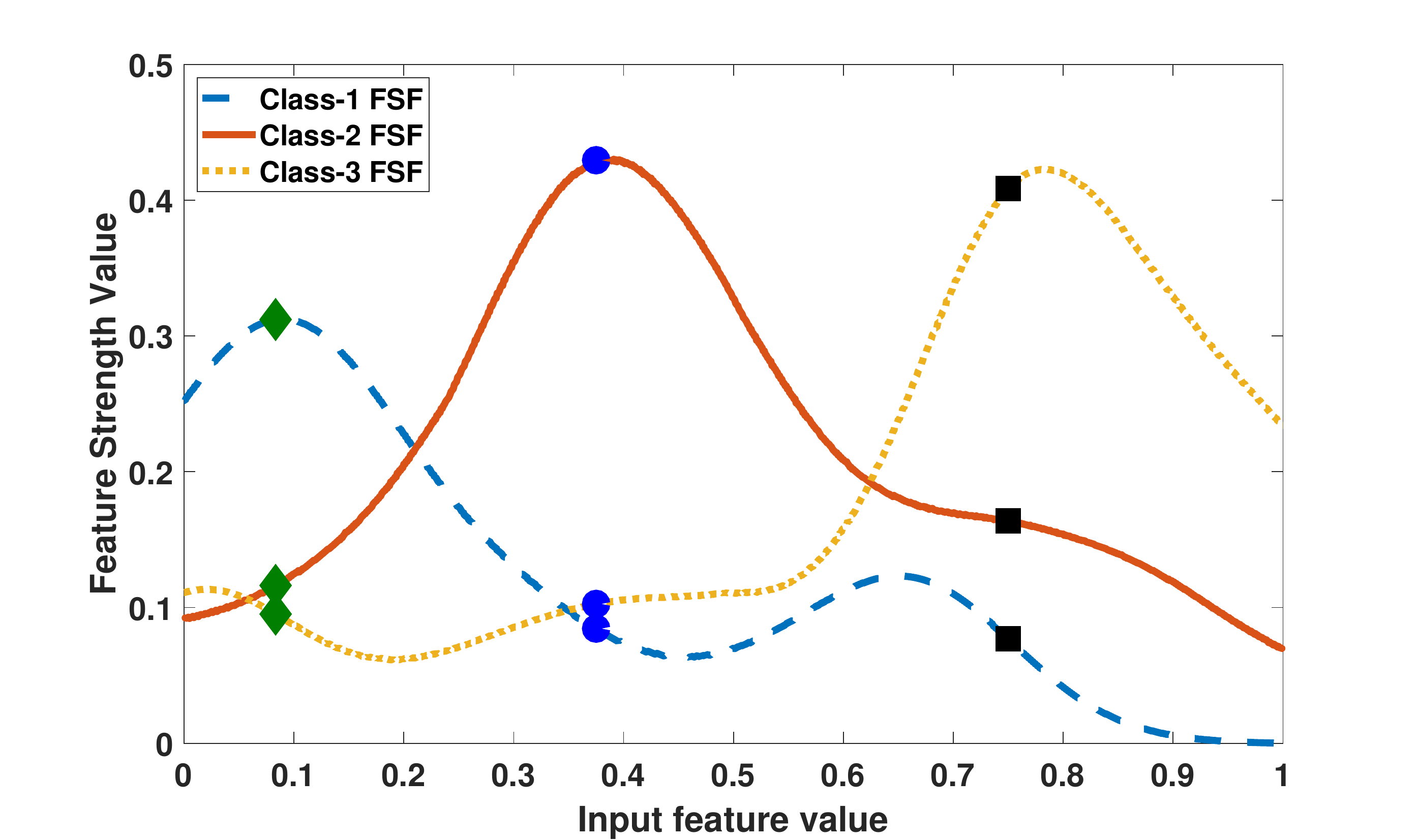}
		\caption{Sampling FSF for input feature 4}
		\label{fig:f4}
	\end{subfigure}
	\begin{subfigure}{.5\textwidth}
		\centering
		\includegraphics[height=0.18\textheight]{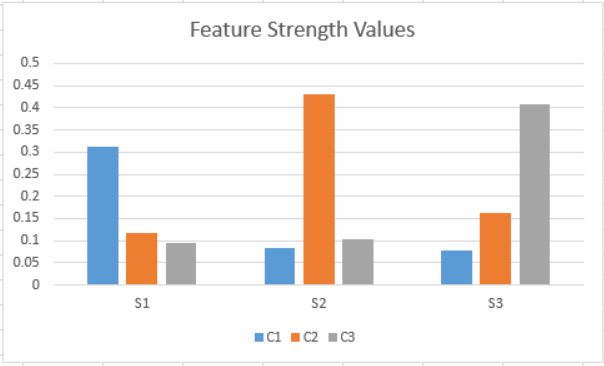}
		\caption{Sampled feature strength values for feature 4}  
		\label{fig:fs4}
	\end{subfigure}	
	\caption{FSF is used for classification task for three input samples $S1=[0.083,0.583,0.068,0.083]$ (diamond), $S2=[0.472,0.083,0.508,0.375]$ (circle) and $S3=[0.556,0.208,0.678,0.75]$ (square) from class-1,class-2 and class-3 respectively. FSF is sampled at input feature values to obtain the corresponding feature strength values.}
	\label{fig:IRIS}
\end{figure*}

Iris dataset contains $3$ classes, with each input data having $4$ features (attributes). A trained MC-SEFRON classifier with accuracies of $98.67\%$ and $97.33\%$ for training and testing datasets respectively is used to extract the FSFs. Fig~\ref{fig:f1},~\ref{fig:f2},~\ref{fig:f3} and~\ref{fig:f4} show the FSFs of input feature 1, 2, 3 and 4 respectively for all the classes. Three input data $S1=[0.083,~0.583,~0.068,~0.083]$, $S2=[0.472,~0.083,~0.508,~0.375]$ and $S3=[0.556,~0.208,~0.678,~0.75]$ from class-1 ($C1$), class-2 ($C2$) and class-3 ($C3$) respectively are used to show the sampling of feature strength values from FSFs (refer to Fig~\ref{fig:FSF_classifier} for the architecture).  In Fig~\ref{fig:f1},~\ref{fig:f2},~\ref{fig:f3} and~\ref{fig:f4}, sampling feature strength values from FSFs for $S1$, $S2$ and $S3$ are denoted by symbols that have shapes of 'diamond' (green), 'circle' (blue) and 'square'(black) respectively. Fig~\ref{fig:fs1},~\ref{fig:fs2},~\ref{fig:fs3} and~\ref{fig:fs4} show the sampled feature strength values for input feature 1, 2, 3 and 4 respectively. 

\begin{figure}[htb!]
	\centering
	\includegraphics[width=0.5\linewidth]{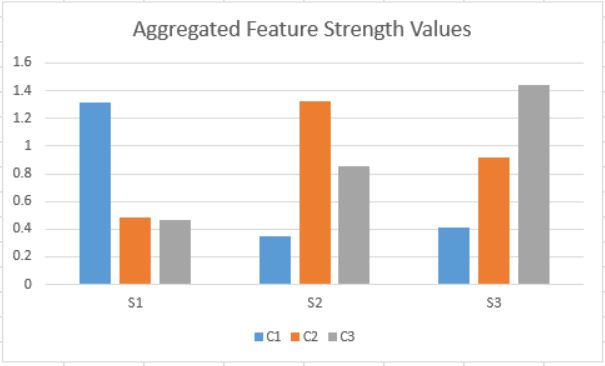}
	\caption{Sum of all the sampled feature strength values from Fig~\ref{fig:IRIS}. Class label for inputs are predicted using equation~\ref{equation:classification_FSF}. Here for all the inputs ($S1,S2,S3$), maximum aggregated feature strength value is obtained for the correct class. This implies correct classifications.}
	\label{fig:IRIS_sum}
\end{figure}

From Fig~\ref{fig:fs1},~\ref{fig:fs2},~\ref{fig:fs3} and~\ref{fig:fs4}, it can be seen that for $S1$, all the feature strength values are higher for $C1$ compared to other classes. Hence, it can be said that $S1$ is easily classifiable as $C1$. Similarly for $S3$, all the features are higher for $C3$ compared to other classes. However, for feature $1$ and $3$, feature strength values for $C2$ are very close to that of $C3$. This makes both the features $2$ and $4$ to be very significant for $S3$ during classification. For $S2$, feature strength values for feature $1$ and $2$ are higher for $C3$ and feature strength values for feature $3$ and $4$ are higher for $C2$. Individual feature strength values alone are not sufficient to interpret the classification for $S2$. By looking at the aggregated feature strength values for $S2$ in Fig~\ref{fig:IRIS_sum}, it can be observed that class label for $S2$ is predicted as $C2$. Hence, it can be said that the feature $3$ and $4$ are significant for $S2$ during classification.  

Fig~\ref{fig:IRIS_sum}, shows the aggregated feature strength values for $S1$, $S2$ and $S3$ for all the classes. For example, for $S1$, the aggregated feature strength value corresponding to $C1$ ($\sum_i^m\psi_i(S1_i,C1)$) is higher than that of other classes. Hence, using the equation~\ref{equation:classification_FSF}, the class label for $S1$ is predicted as $C1$, implying a correct classification. The same can be observed for $S2$ and $S3$. 

\subsection{Experimental results for MNIST dataset}

This section presents the classification performance of MC-SEFRON and FSFs for the MNIST dataset~\cite{MNIST} and shows the use of the extracted FSFs to interpret the predictions made by the MC-SEFRON classifier. Experiments on MNIST dataset are repeated 10 times to ensure that the performance is not affected by the first sample presented from each class to initialize the $w^r_i(t,j)$ and $\theta_j$. In the population encoding scheme, $5$ RF neurons are used in all the experiments. For the 10 experiments, average training and testing accuracies of $93.64\%$ and $92.30\%$ were obtained respectively. Cross-validation is used to choose $\hat{t}_o$ to extract FSFs from the 10 trained MC-SEFRON classifiers. Average accuracies of $93.47\%$ and $92.20\%$ were obtained for the training and testing dataset respectively by using FSFs for classification. Classification accuracies are compared in Table~\ref{table:results} with other SNN methods in the literature.

\begin{table*}[htb]
	\centering
	\caption{Performance comparison on MNIST dataset }
	\resizebox{\linewidth}{!}{
	\begin{threeparttable}
	\begin{tabular}{|c|c|c|c|c|}
		\hline
		Model & Architecture& Method&\begin{tabular}[c]{@{}c@{}}Classifier's\\ Accuracy (\%)\end{tabular} & \begin{tabular}[c]{@{}c@{}}Interpretable knowledge's \\ Accuracy (\%)\end{tabular} \\ \hline	
		\cite{Tavanaei2018} &Spiking CNN &STDP learning rule  &98.60 & - \\ \hline	
		\cite{Lee} &Spiking CNN &Backpropagation& 99.31 & - \\ \hline			
		\cite{zhao2015} &Spiking CNN & Tempotron learning rule &91.29 & - \\ \hline			
		\multicolumn{5}{|c|}{\rule{0pt}{15pt}Fully connected SNN models} \\ [0.25cm]\hline			
		\cite{EDBN} &784-500-500-10  &\begin{tabular}[c]{@{}c@{}}Spiking DBN\\ Converted from trained DBN\end{tabular}  & 94.09 & - \\ \hline	
		\cite{spike_rbm1} &(784+40)-500-10 &\begin{tabular}[c]{@{}c@{}}Spiking RBM\\ Contrastive divergence\end{tabular}  &91.90 & - \\ \hline	
		\cite{mostafa2018} &784-400-400-10 & \begin{tabular}[c]{@{}c@{}}3-layer SNN\\ Temporal backpropagation\end{tabular}  &97.14 & - \\ \hline				
		\cite{TAVANAEI2019} &784-500-150-10 & \begin{tabular}[c]{@{}c@{}}3-layer SNN\\ STDP-based backpropagation\end{tabular} &97.20 & - \\ \hline
		\cite{hussain2014}* &[25$\times$200]$_{10}$ - 10 & \begin{tabular}[c]{@{}c@{}}Spiking Cells\\ Morphology learning rule\end{tabular} &90.26 & - \\ \hline
		MC-SEFRON (This work) &784$\times$5 - 10 &\begin{tabular}[c]{@{}c@{}}Spiking neuron with time-varying weight\\ Modified STDP rule\end{tabular}& 92.30 & 92.20 \\ \hline		
	\end{tabular}
\begin{tablenotes}
	\item \textit{*- Full dataset is not used}
\end{tablenotes}
\end{threeparttable}
}
\label{table:results}
\end{table*}

From Table~\ref{table:results}, it can be seen that spiking Convolutional Neural Networks (spiking CNNs) achieve higher accuracies which are comparable to other non-spiking deep learning methods. It can be noted that MC-SEFRON is the only classifier with the simplest architecture (single input-output layer). The accuracy of MC-SEFRON is better than the spiking Restricted Boltzmann Machines (Spiking RBM) and Spiking cells. Here the parameters for the MC-SEFRON classifier were optimized to produce a testing accuracy higher than $91\%$ and this do not purport to be the best performance of MC-SEFRON. However, the main focus of this work is not to achieve the state-of-the-art performance on (MNIST) handwritten digit recognition but to highlight the quality of the knowledge that can be extracted from an SNN with time-varying weight model and also the interpretability of those extracted knowledge. 

Except for MC-SEFRON, all the other methods do not have a framework to extract interpretable knowledge from the trained network. From Table~\ref{table:results}, it can be seen that the classification accuracy for the MNIST dataset using interpretable knowledge (FSF) is $92.20\%$ and the loss in accuracy is minimal ($0.1\%$). This highlights that the extracted knowledge is a better representation of the classifier's learned knowledge.
 
\subsubsection{Interpreting MC-SEFRON's prediction using FSF on MNIST dataset}

\begin{figure*}[htb!]	
	\centering
	\begin{subfigure}[b]{0.48\textwidth}
		\includegraphics[width=\linewidth]{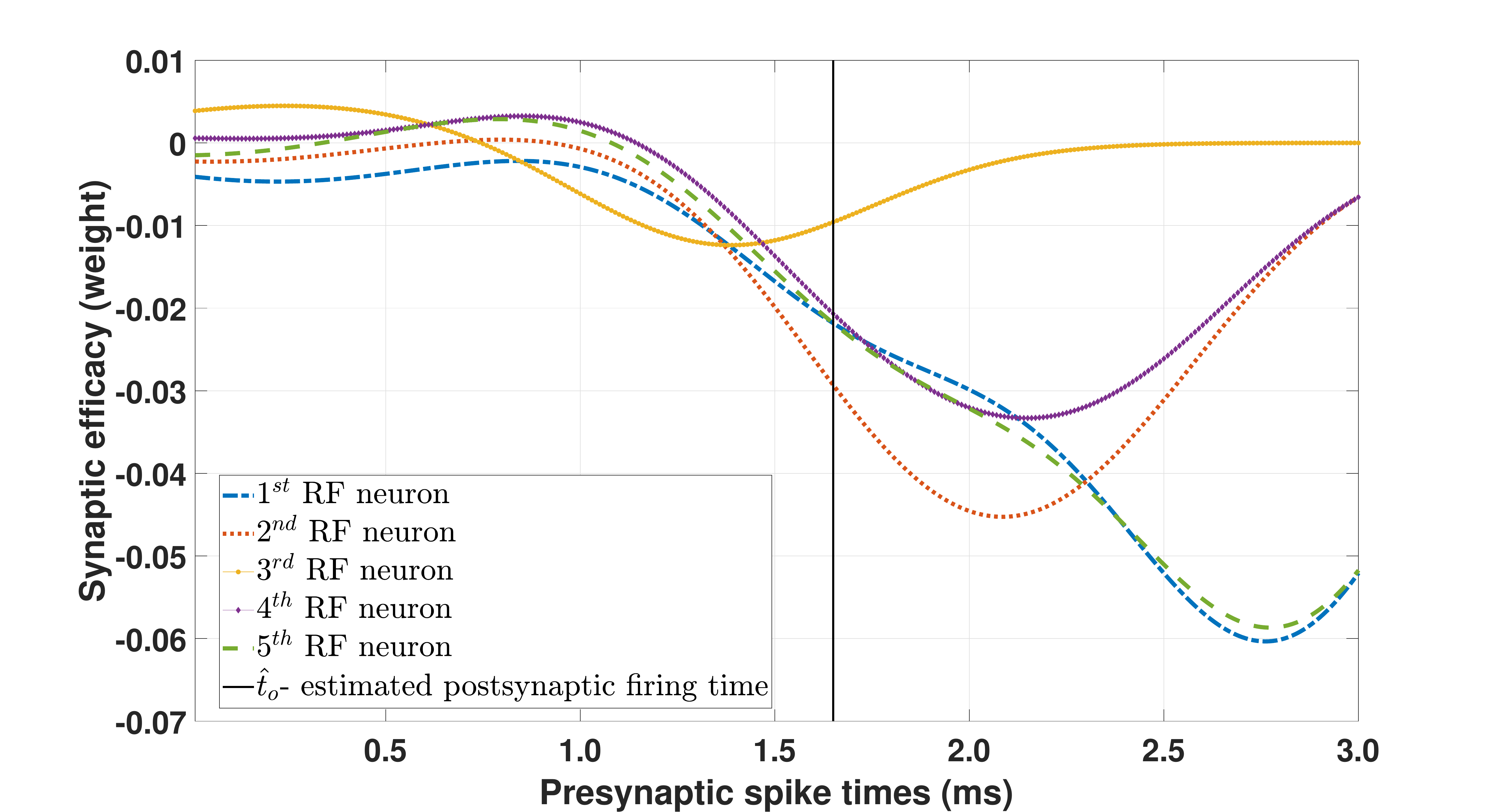}
		\caption{Synaptic efficacy functions $w^r_i(t,j)$ of all RF neurons}
		\label{fig:SEF}
	\end{subfigure}
	\begin{subfigure}[b]{0.48\textwidth}
		\includegraphics[width=\linewidth]{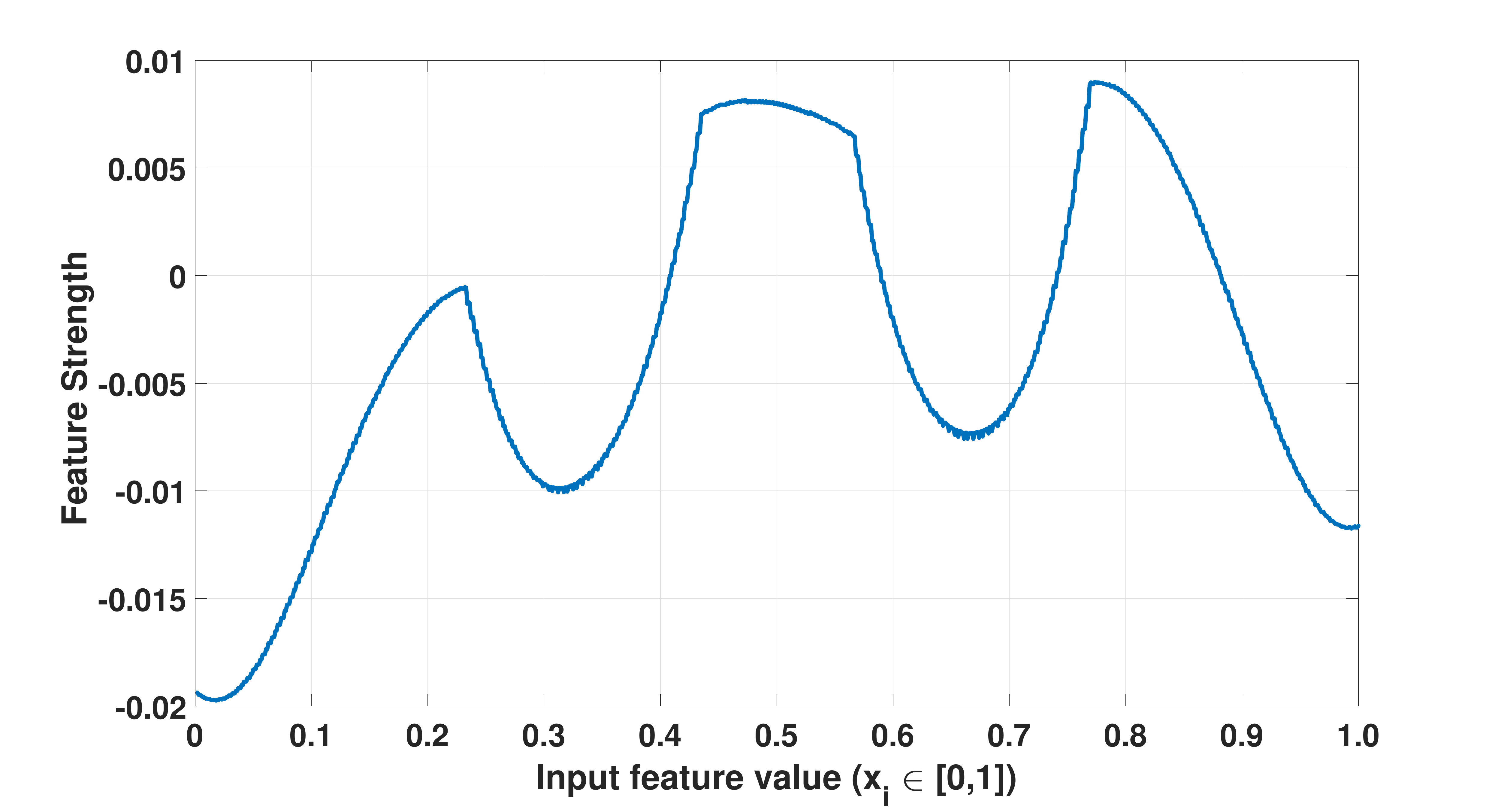}	
		\caption{Extracted Feature strength Function $\psi_i(x_i,j)$ from SEFs in Fig~\ref{fig:SEF}}
		\label{fig:FSF}
	\end{subfigure}
	\caption{Synaptic efficacy functions of all the RF neurons for $i^{th}$ feature and the extracted Feature strength function  for $i^{th}$ feature from all the Synaptic efficacy functions}
	\label{fig:SEF_to_FSF}
\end{figure*}

\begin{figure*}[htb!]
	\centering\setlength\tabcolsep{0.05em}
	\resizebox{0.8\totalheight}{!}{
	\begin{tabular}[t]{|c|c|c|c|c|c|c|c|c|c|c|}
		\hline
		\multirow{2}{*}{Input} & \multicolumn{10}{c|}{Feature strength heatmap and aggregated value \big($\sum_{i=1}^{m}\psi_i(x^k_i,j)$\big) for each output class} \\ \cline{2-11} 
		& `0'  & `1'  & `2'  & `3' & `4' & `5' & `6' & `7' & `8' & `9' \\ \hline
		
		\subcaptionbox*{}{\includegraphics[width=0.09\textwidth]{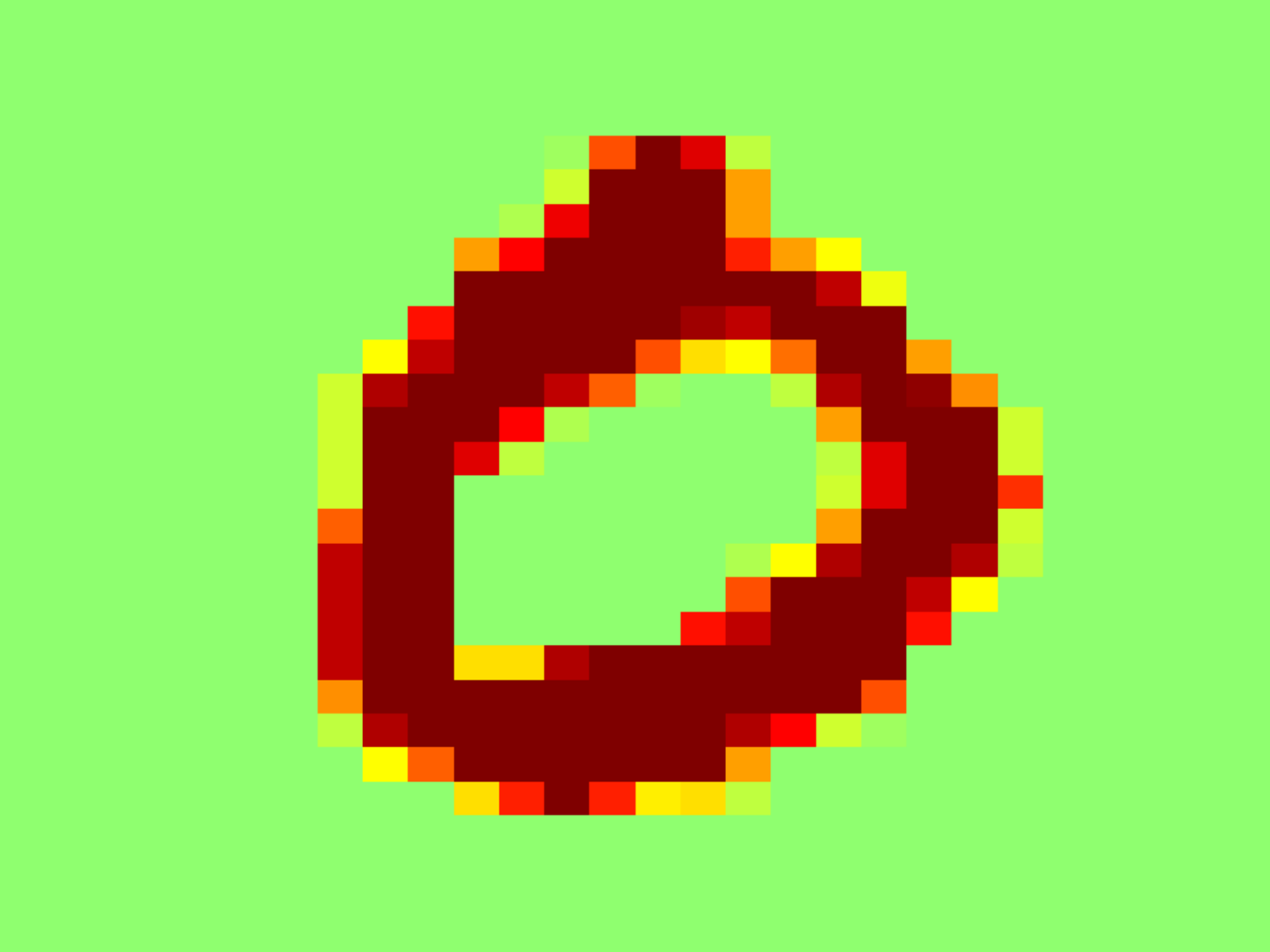}}&
		\subcaptionbox*{}{\includegraphics[width=0.09\textwidth]{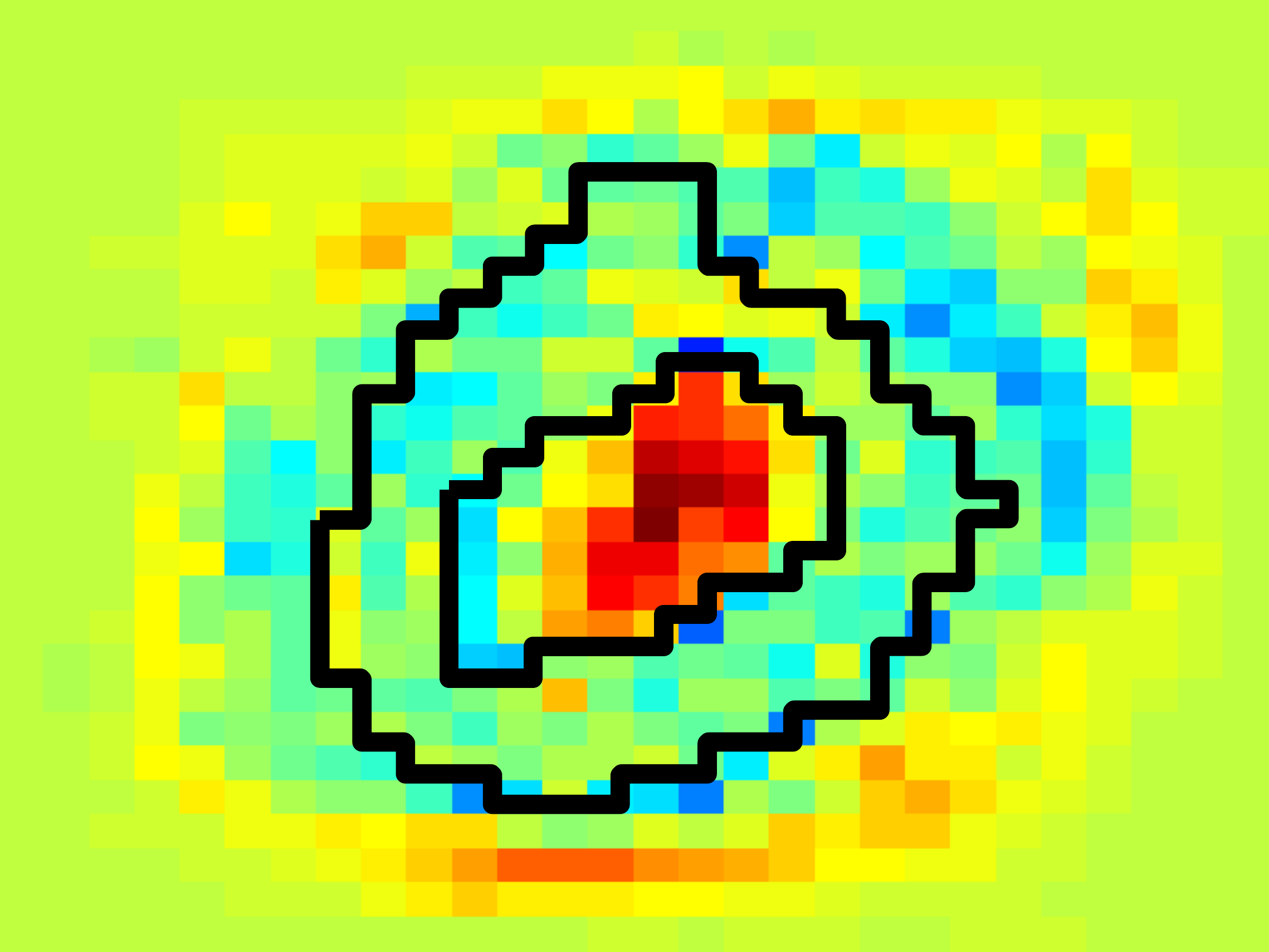}}&
		\subcaptionbox*{}{\includegraphics[width=0.09\textwidth]{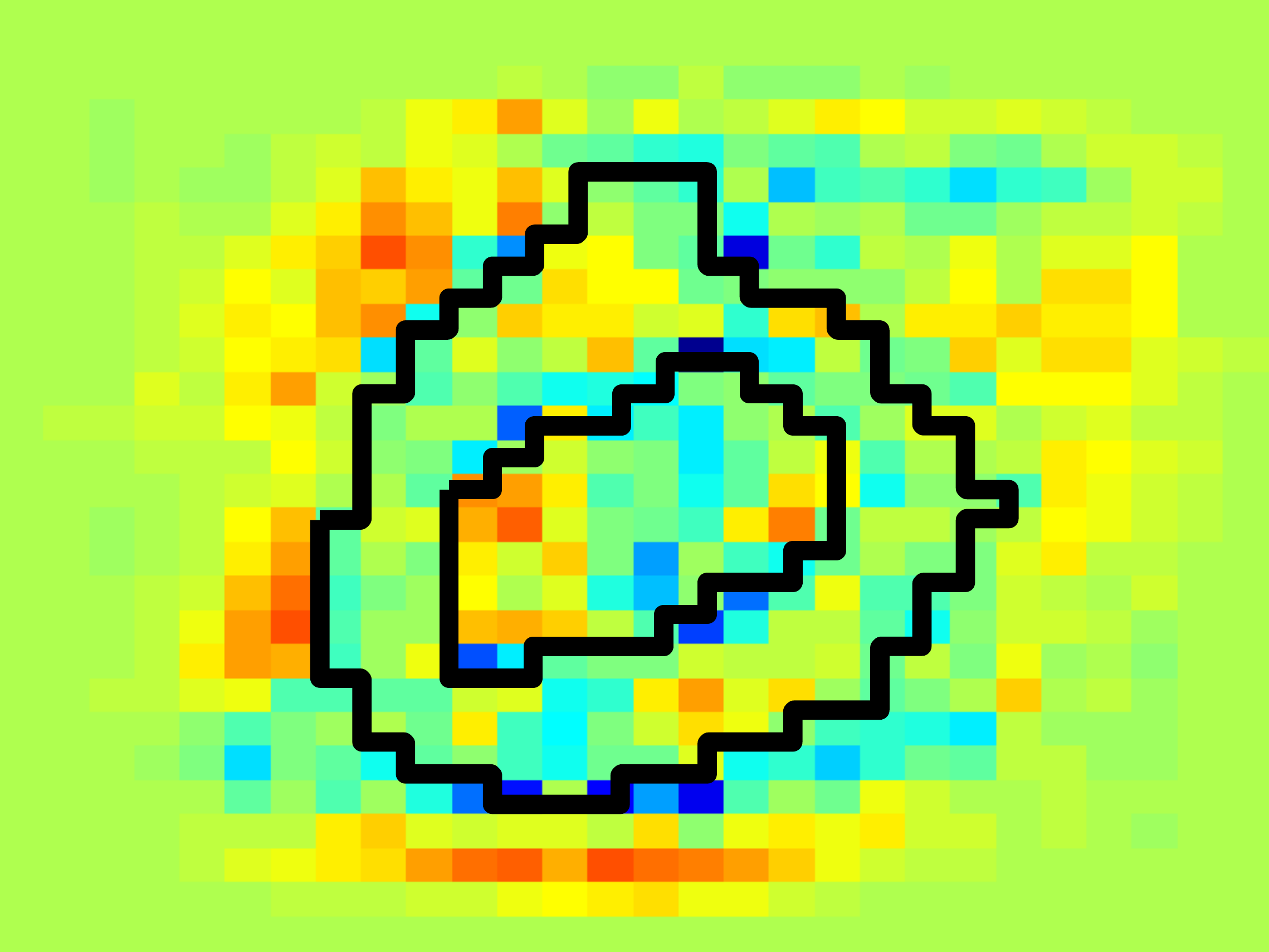}}&
		\subcaptionbox*{}{\includegraphics[width=0.09\textwidth]{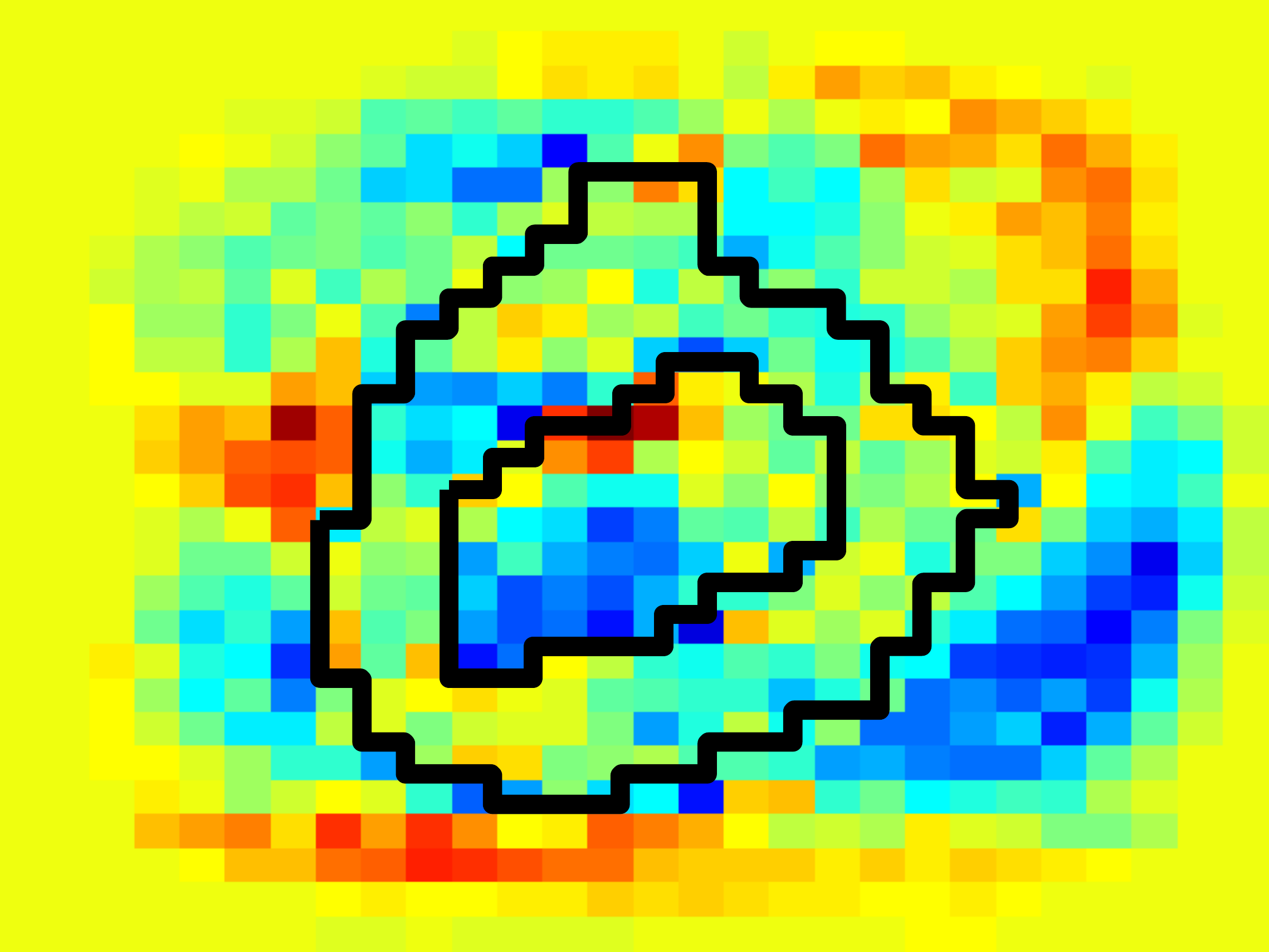}}&
		\subcaptionbox*{}{\includegraphics[width=0.09\textwidth]{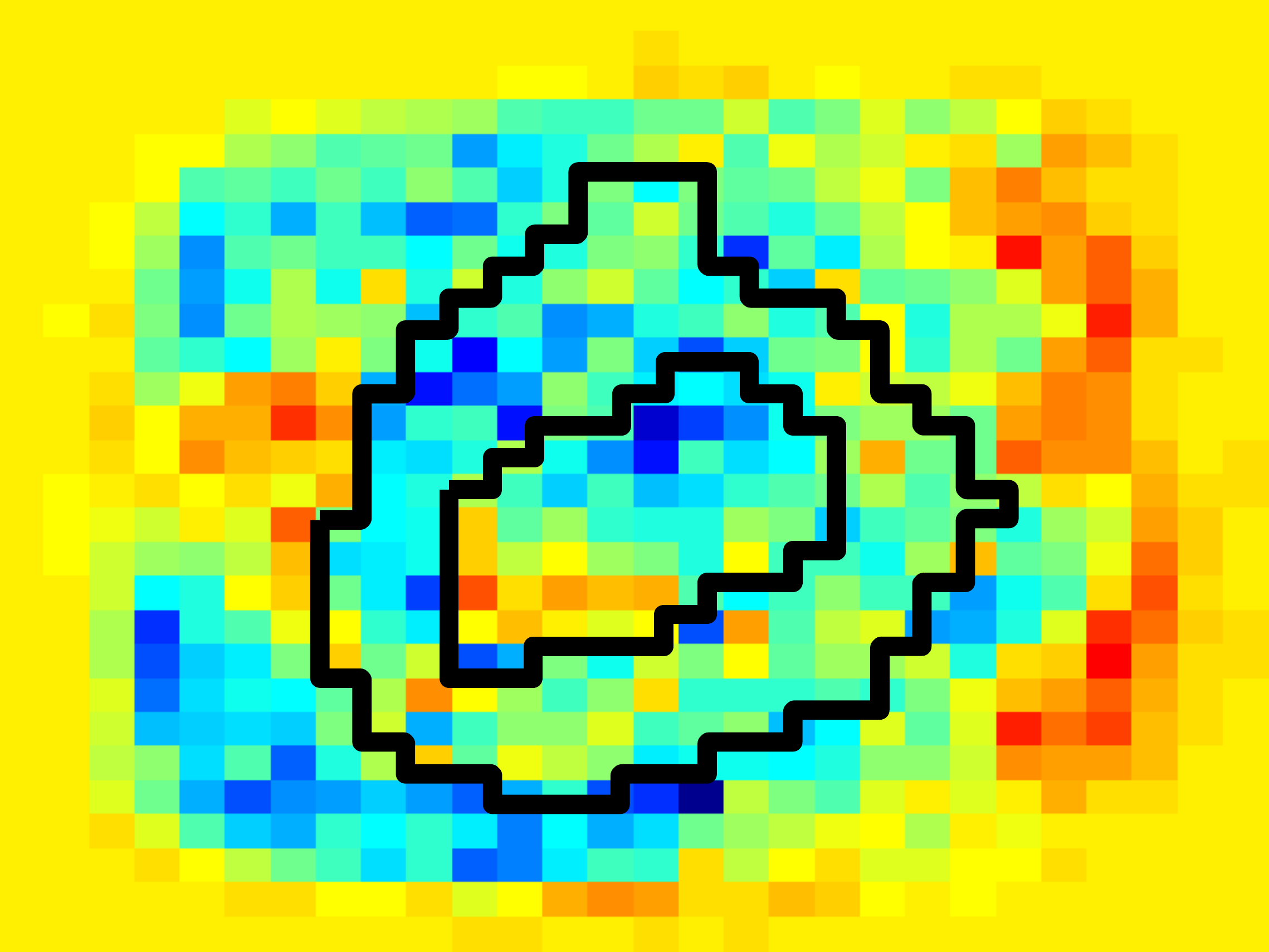}}&
		\subcaptionbox*{}{\includegraphics[width=0.09\textwidth]{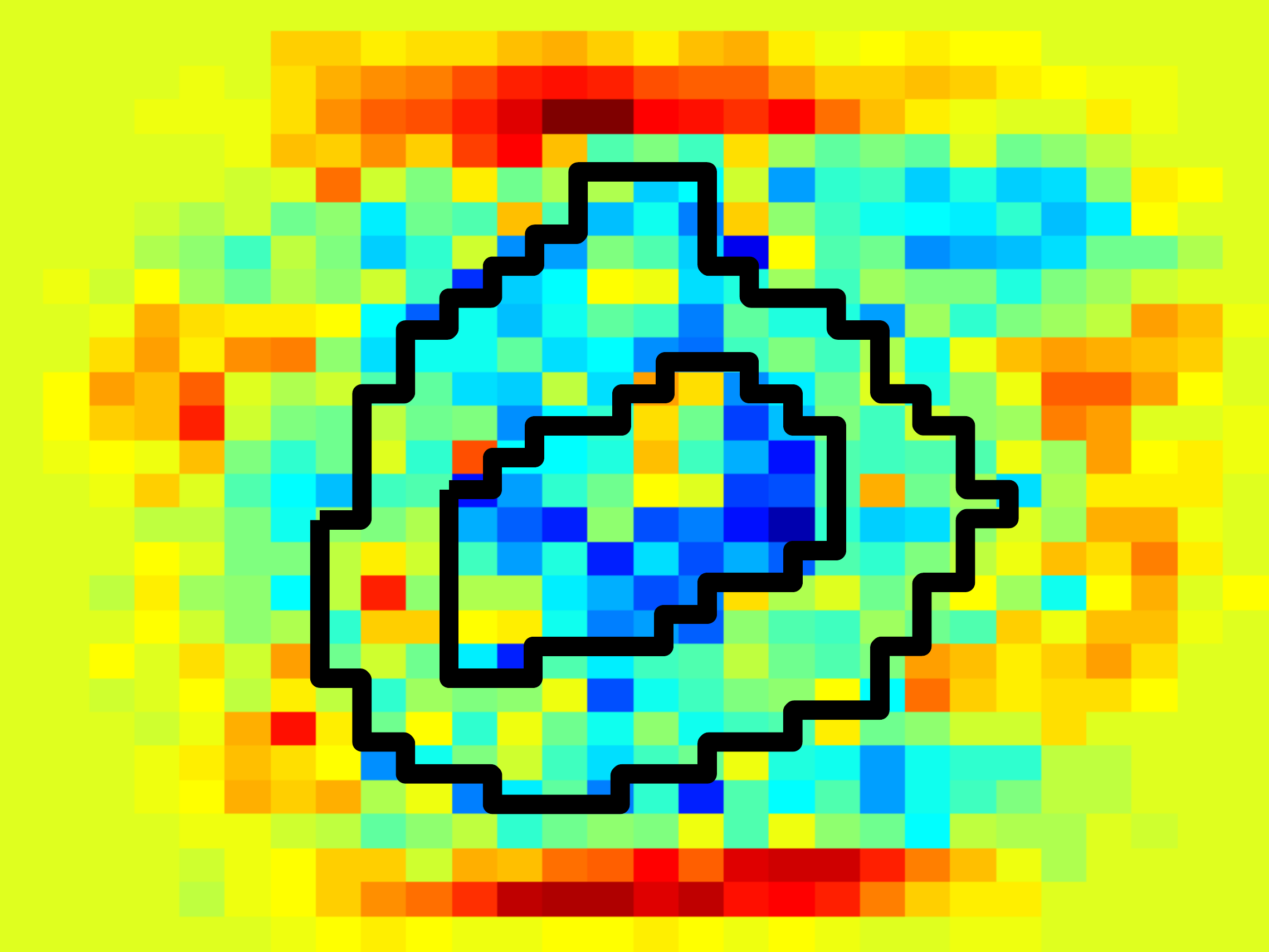}}&
		\subcaptionbox*{}{\includegraphics[width=0.09\textwidth]{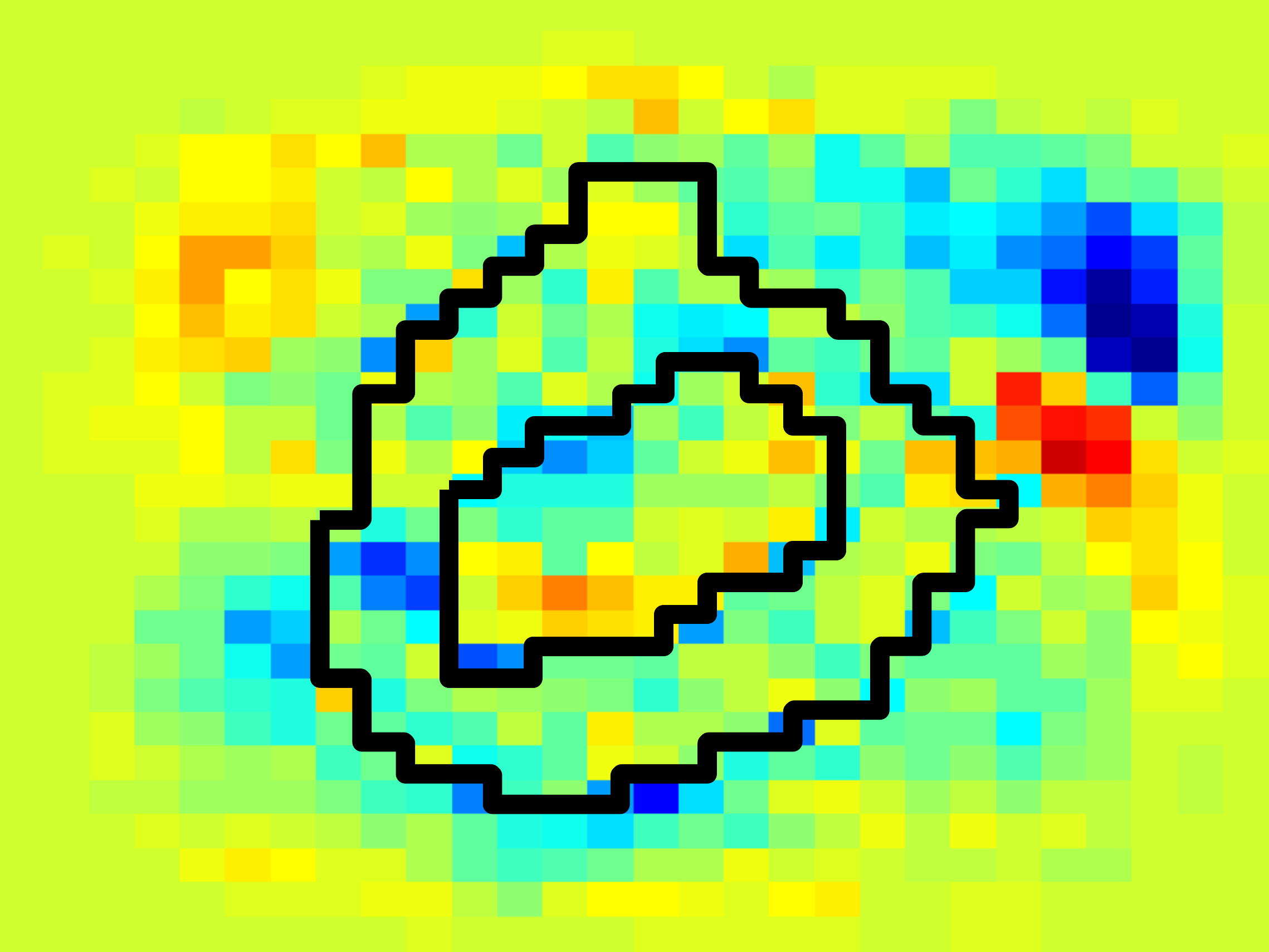}}&
		\subcaptionbox*{}{\includegraphics[width=0.09\textwidth]{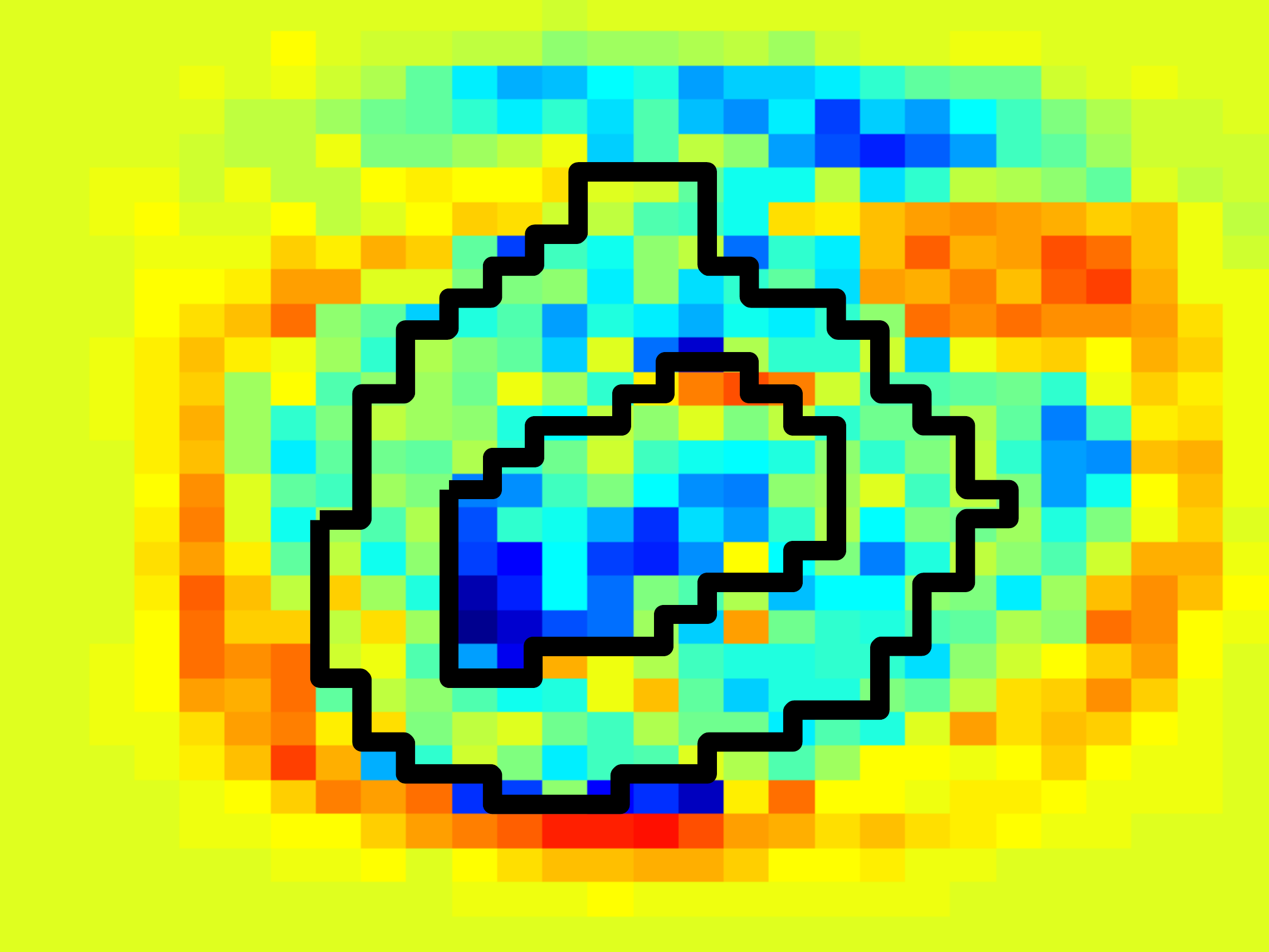}}&
		\subcaptionbox*{}{\includegraphics[width=0.09\textwidth]{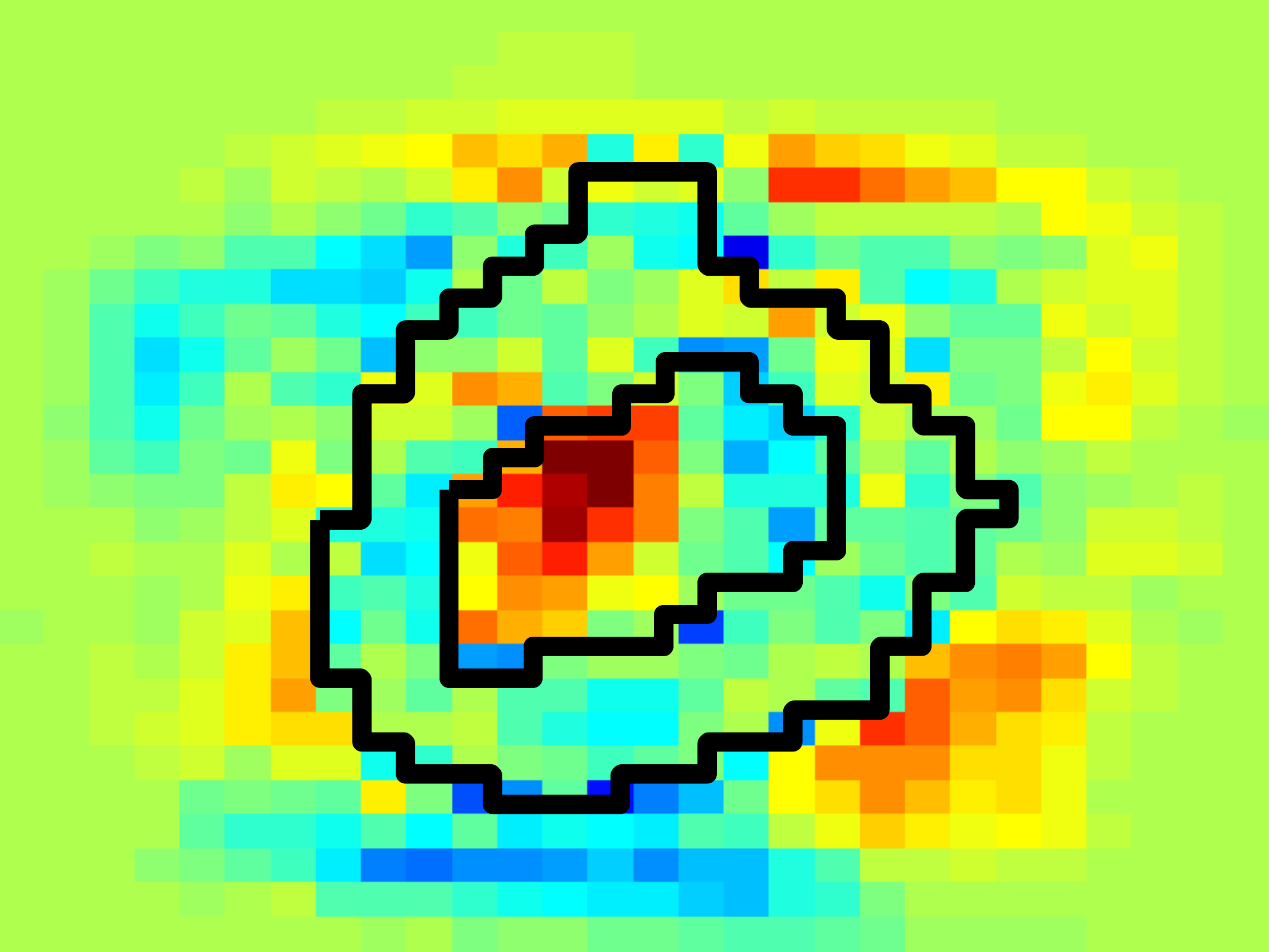}}&
		\subcaptionbox*{}{\includegraphics[width=0.09\textwidth]{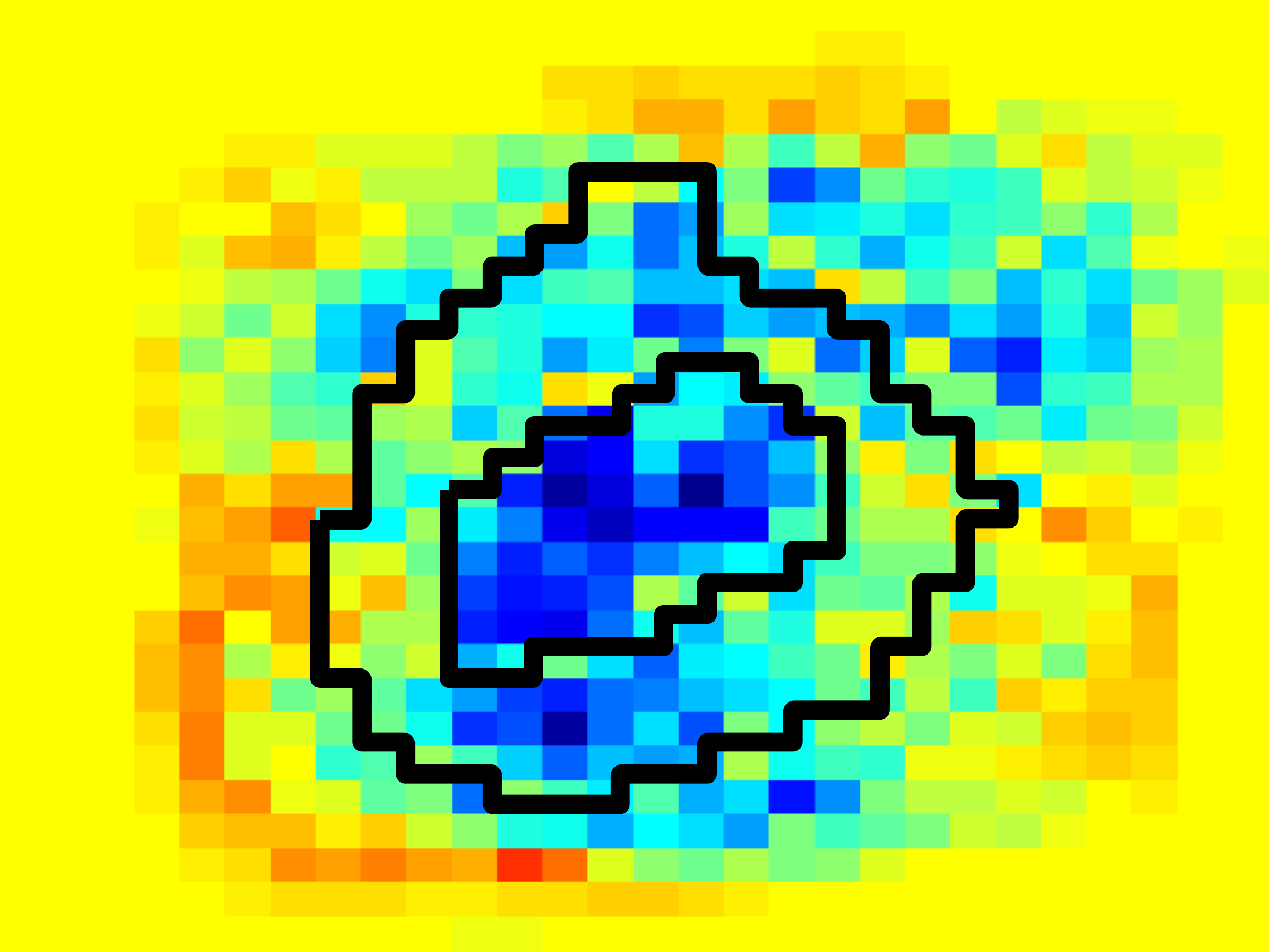}}&
		\subcaptionbox*{}{\includegraphics[width=0.09\textwidth]{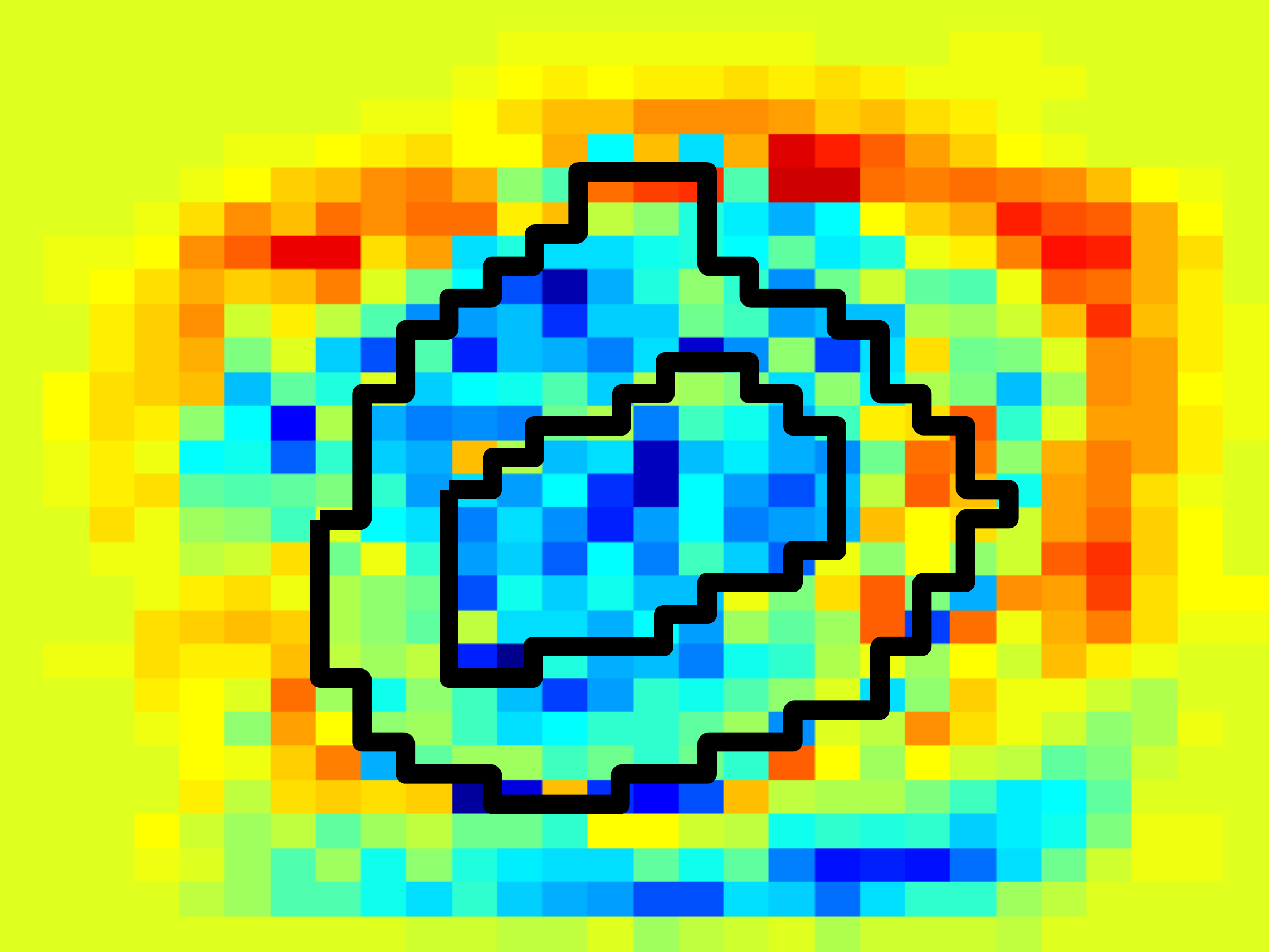}}
		\\[-1em]
		&1.19&0.69&0.84&0.85&0.83&0.86&0.84&0.77&0.69&0.73\\[0.5em]\hline
		
		\subcaptionbox*{}{\includegraphics[width=0.09\textwidth]{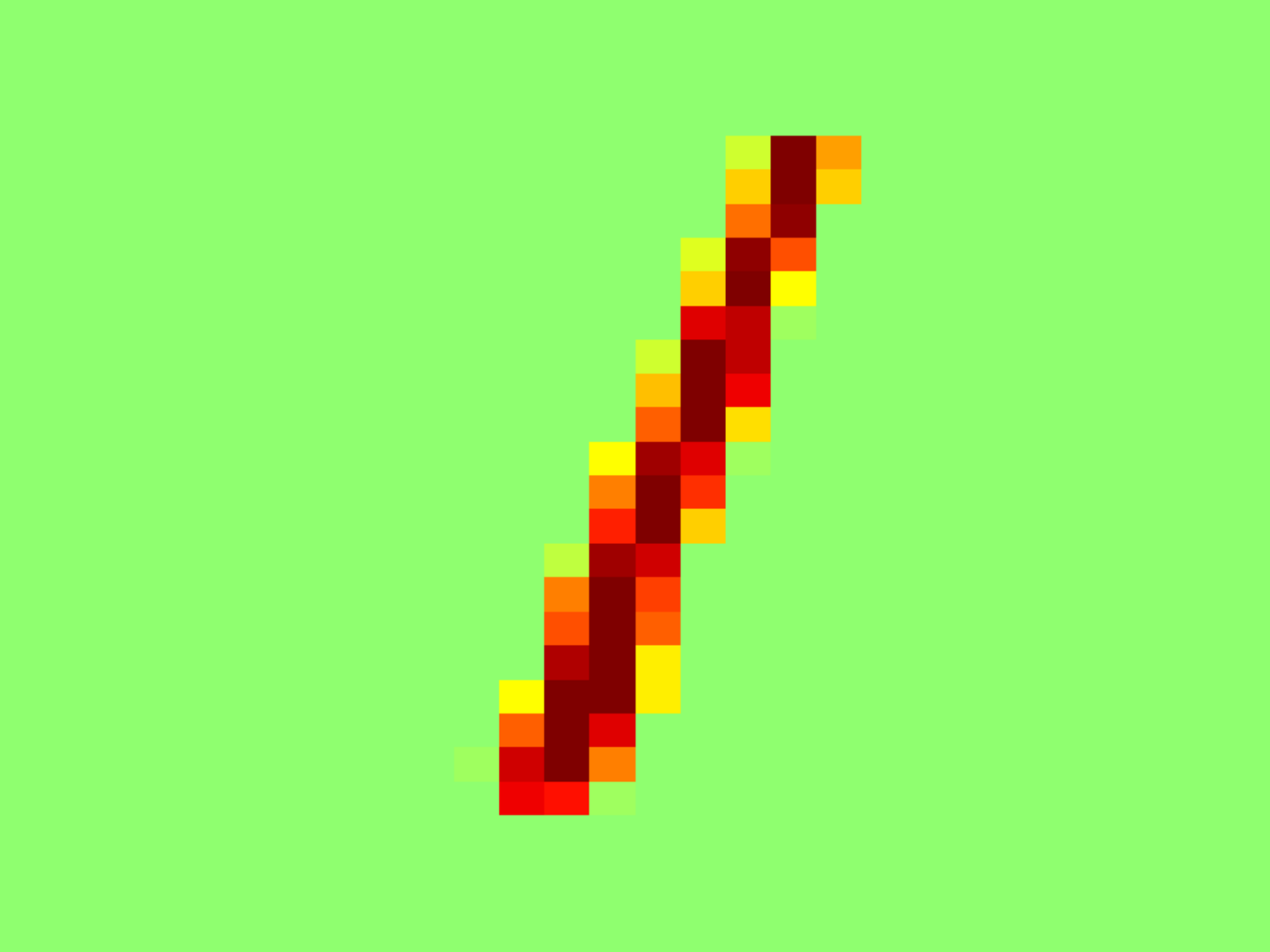}}&
		\subcaptionbox*{}{\includegraphics[width=0.09\textwidth]{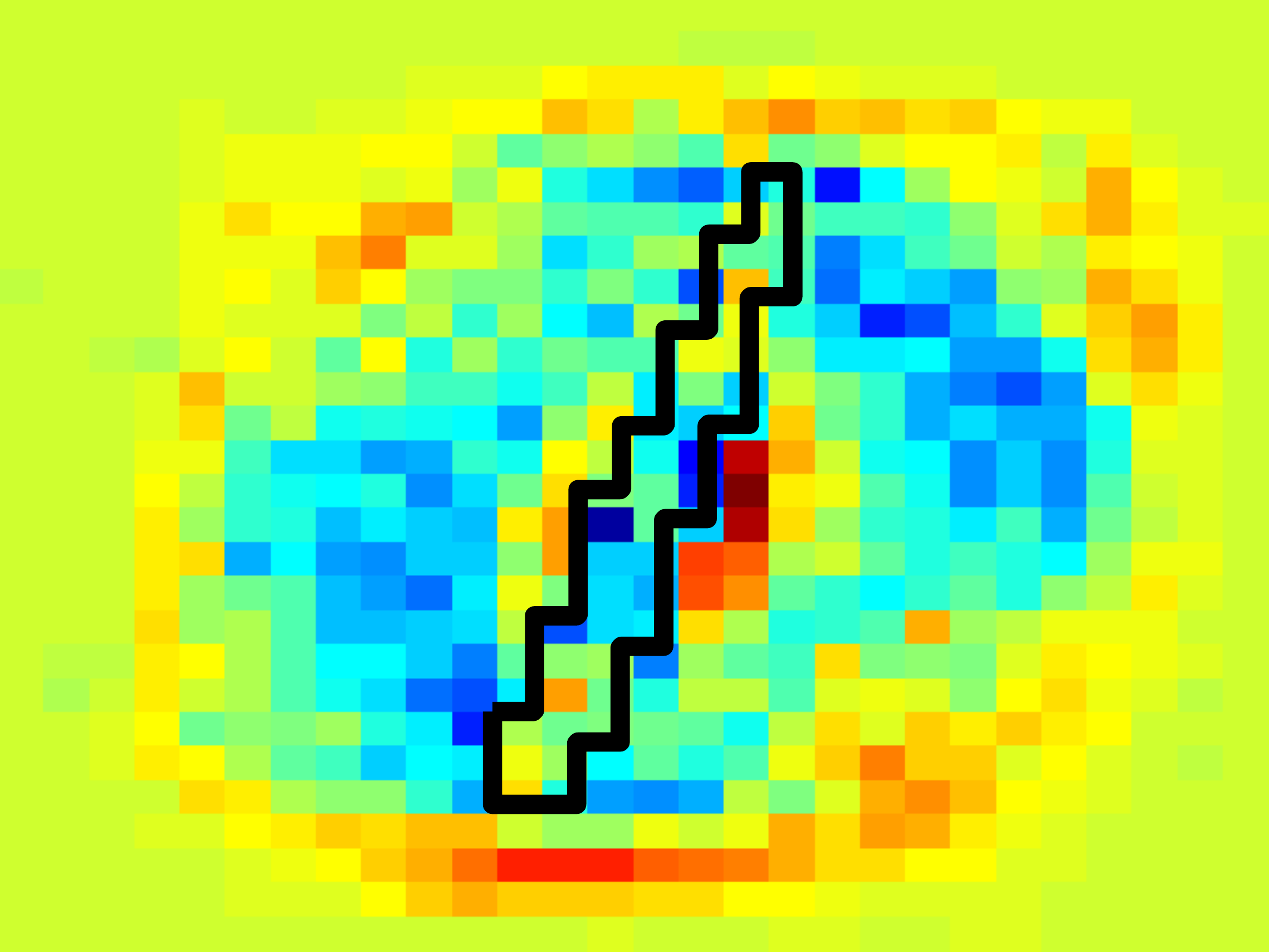}}&
		\subcaptionbox*{}{\includegraphics[width=0.09\textwidth]{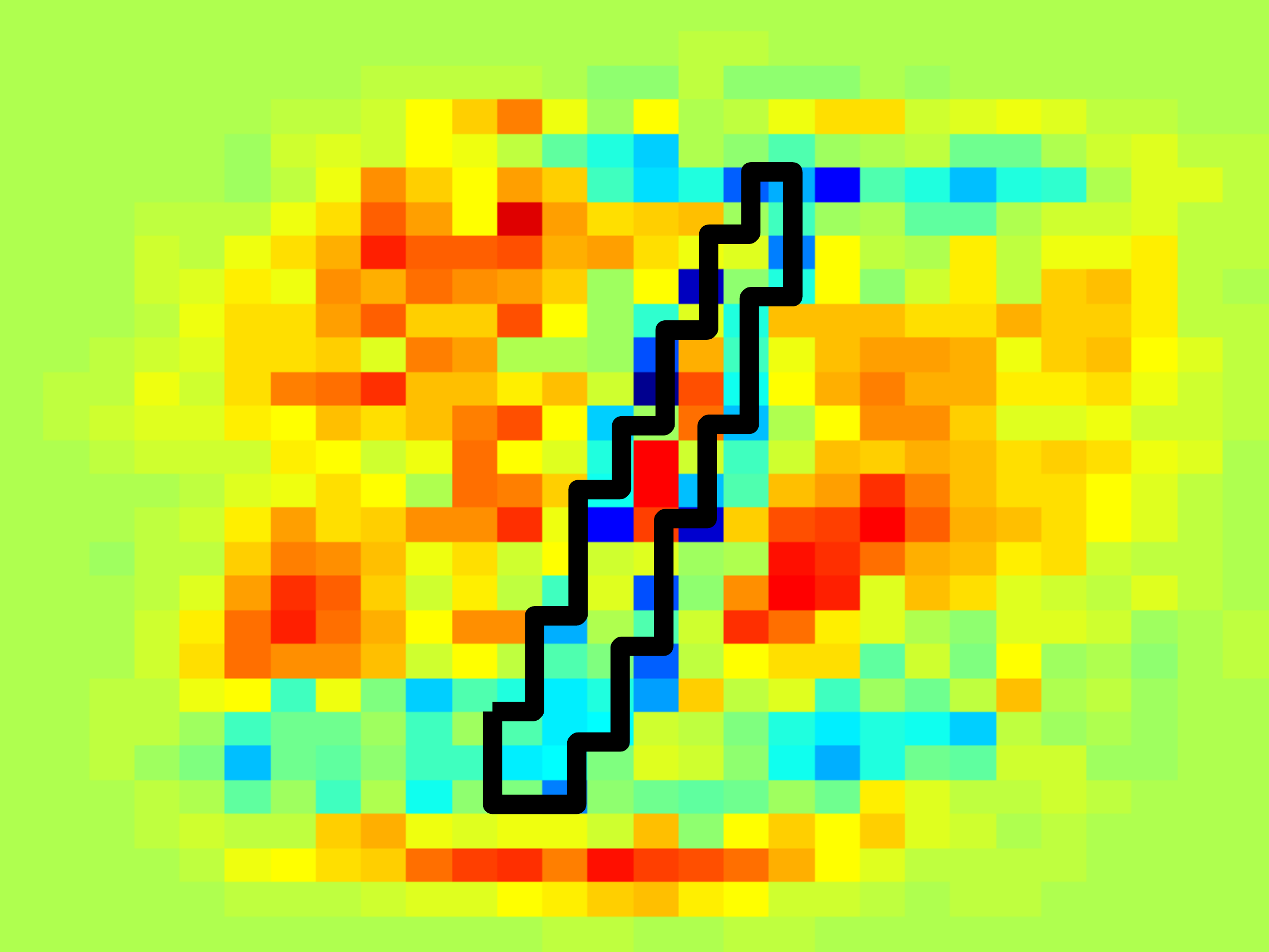}}&
		\subcaptionbox*{}{\includegraphics[width=0.09\textwidth]{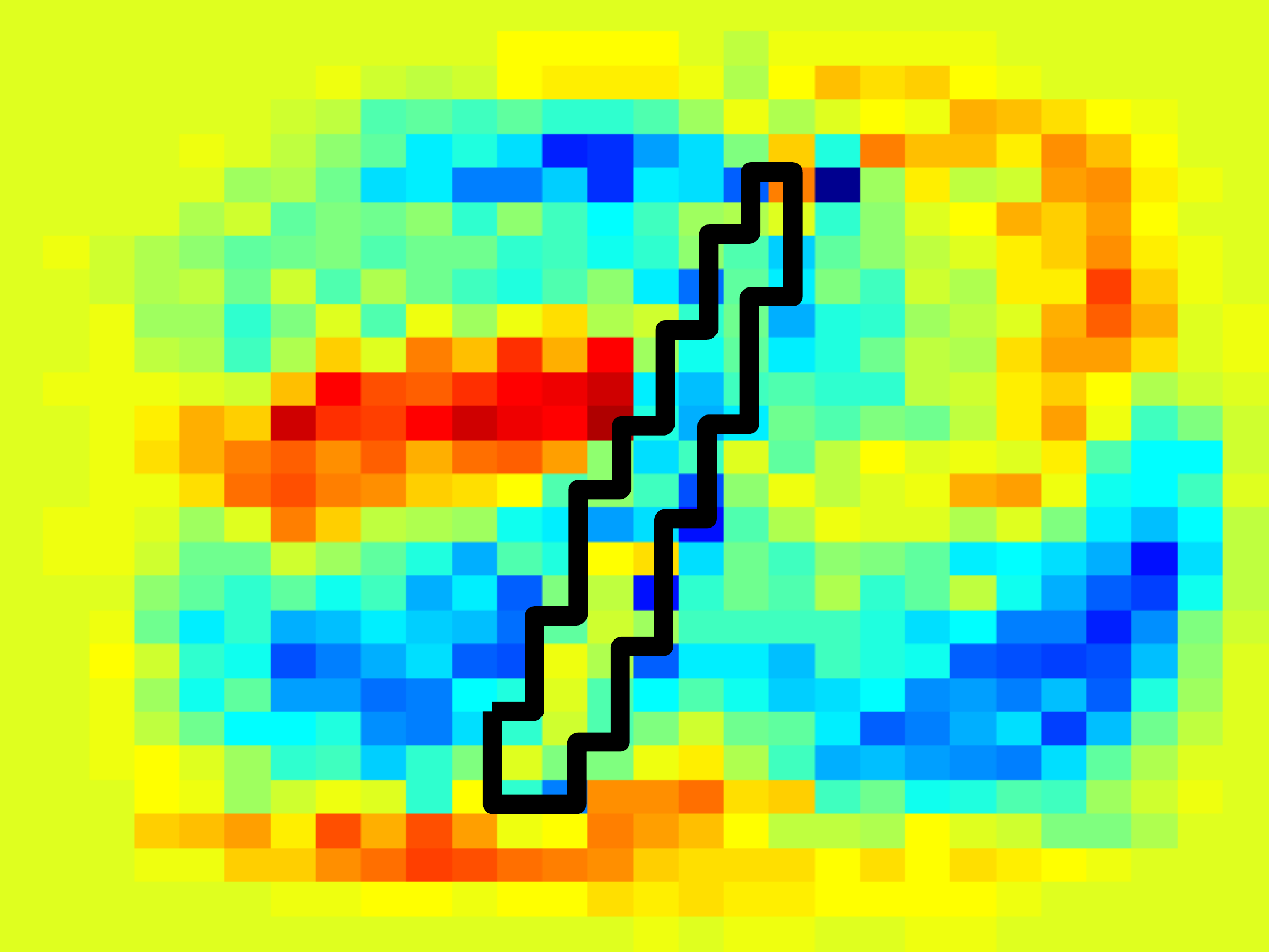}}&
		\subcaptionbox*{}{\includegraphics[width=0.09\textwidth]{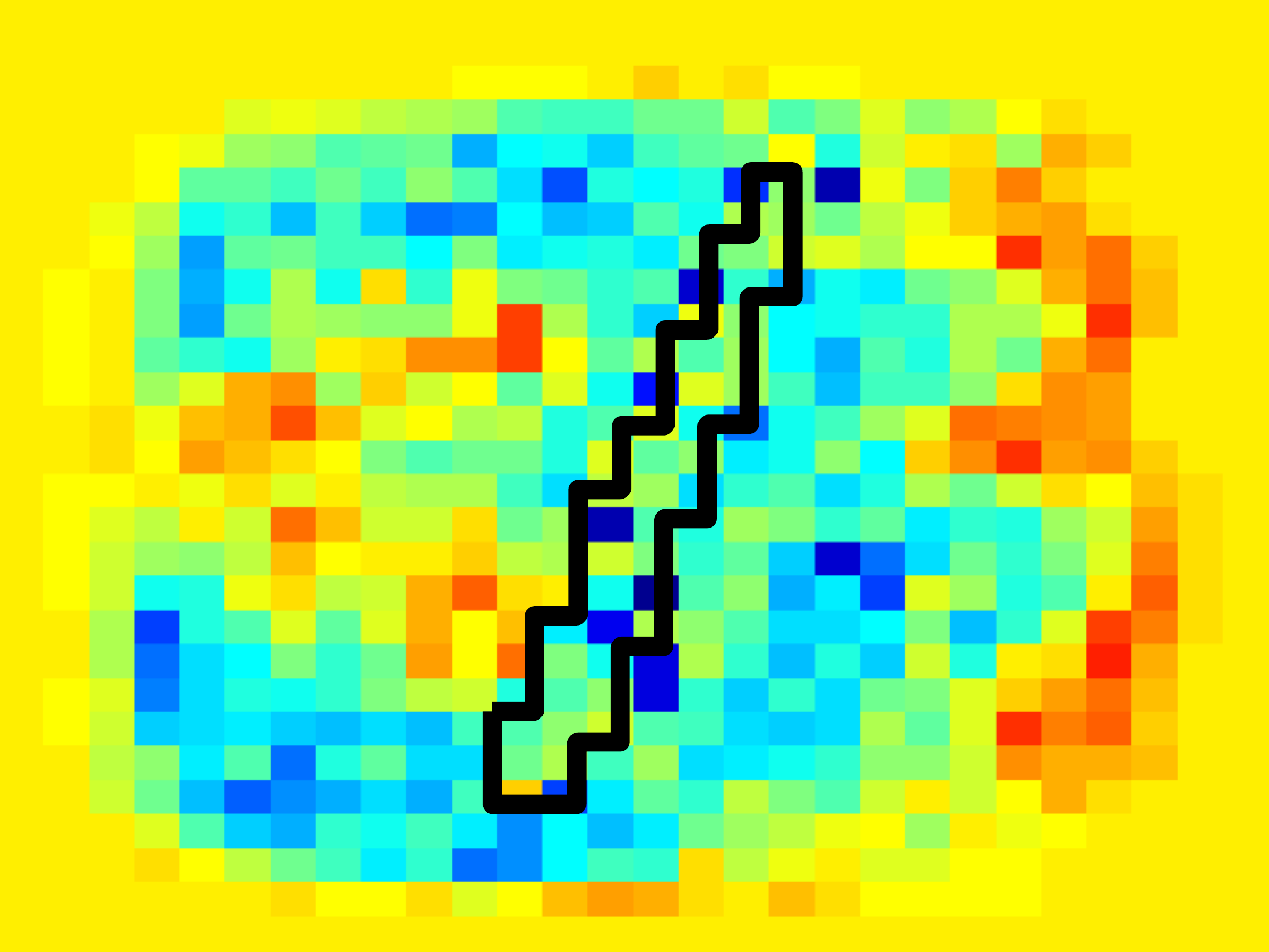}}&
		\subcaptionbox*{}{\includegraphics[width=0.09\textwidth]{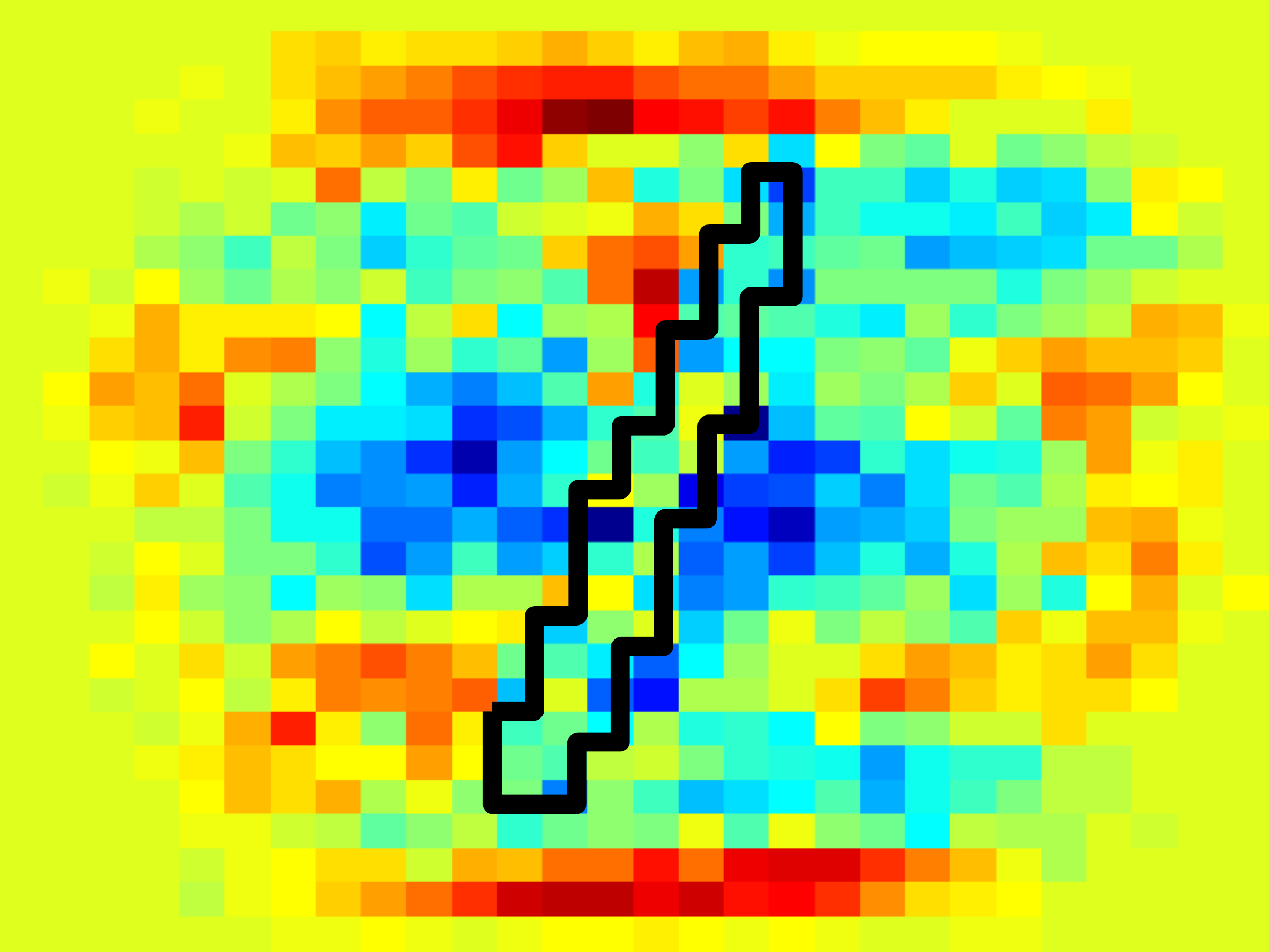}}&
		\subcaptionbox*{}{\includegraphics[width=0.09\textwidth]{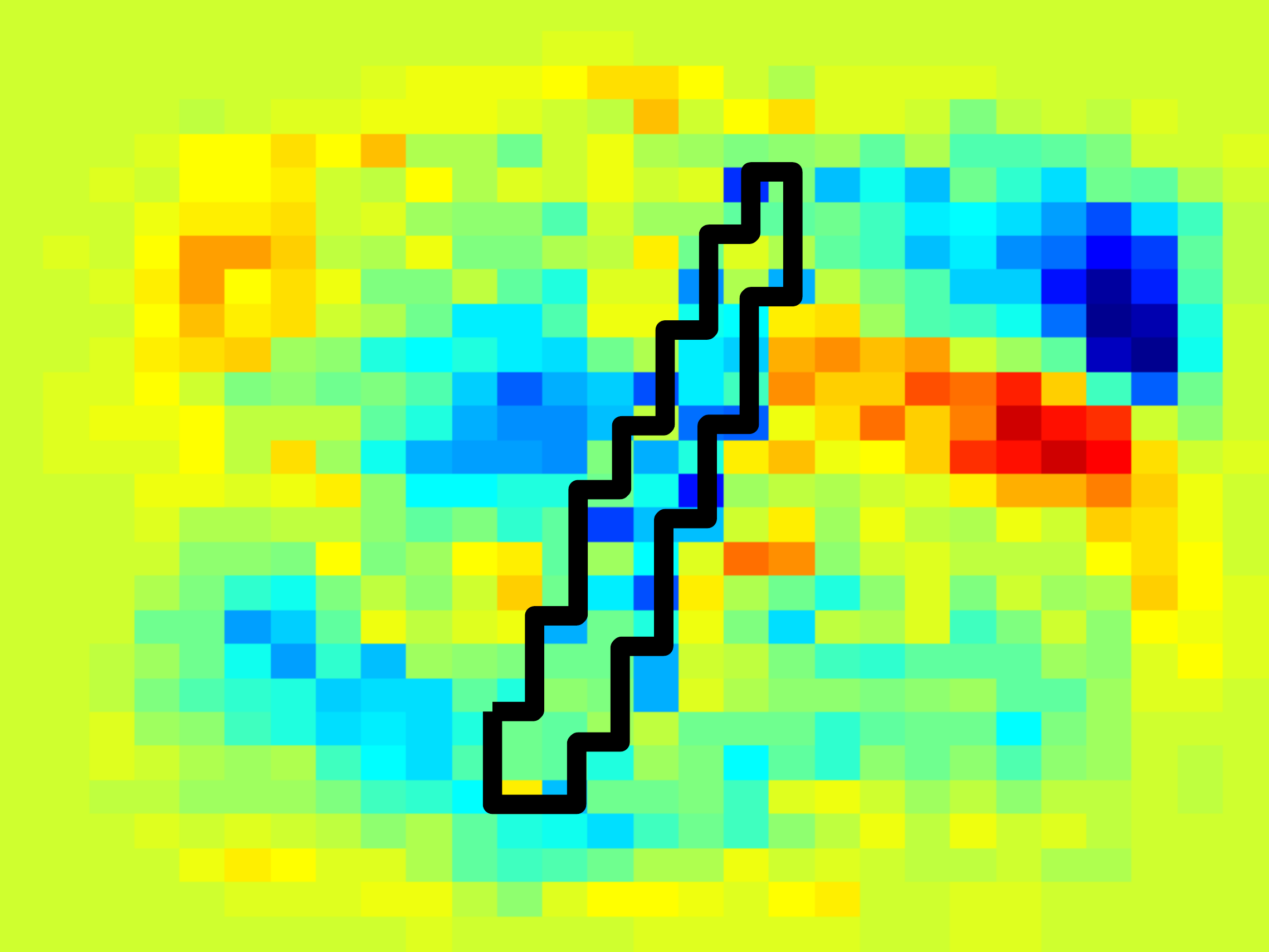}}&
		\subcaptionbox*{}{\includegraphics[width=0.09\textwidth]{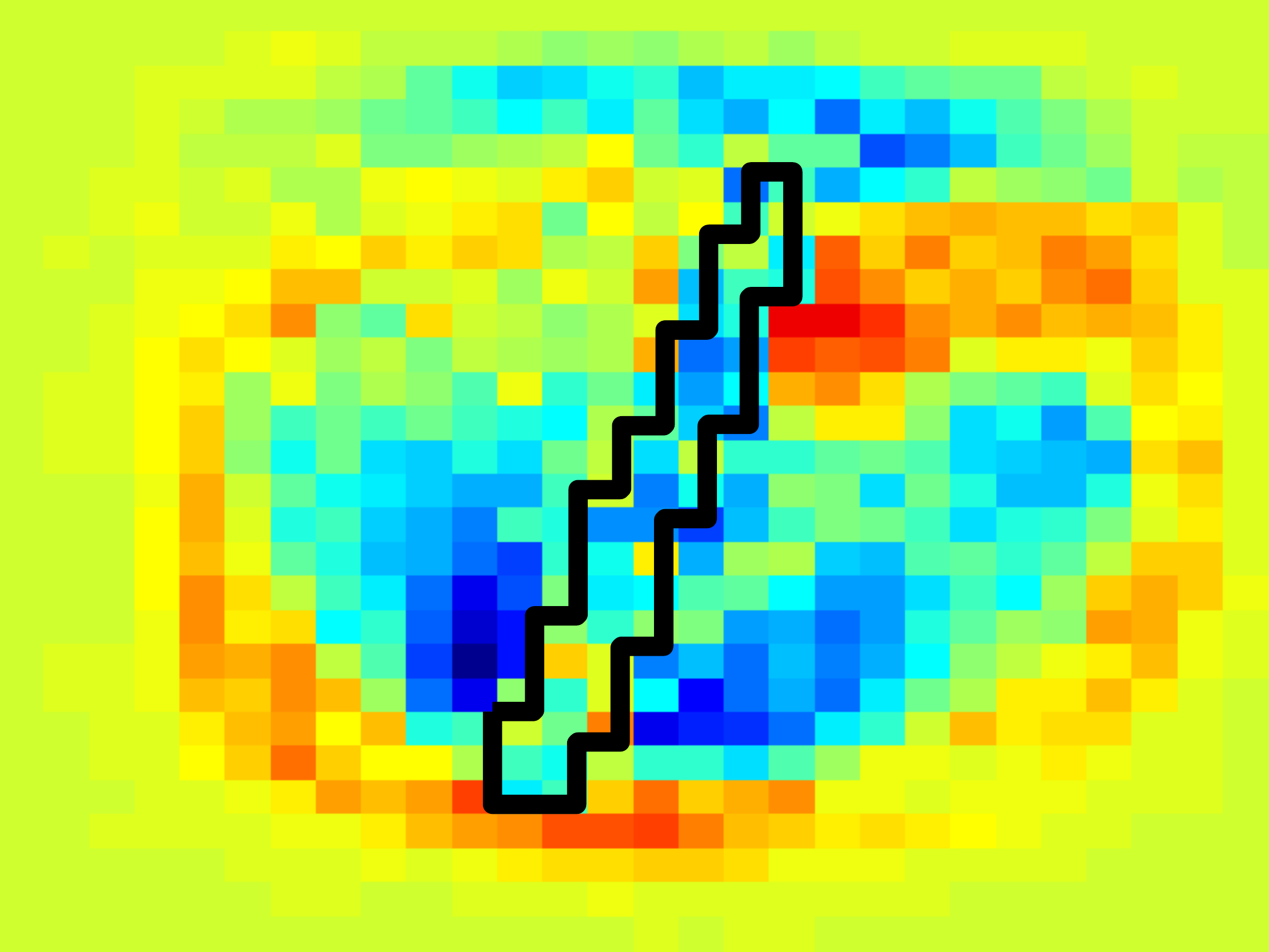}}&
		\subcaptionbox*{}{\includegraphics[width=0.09\textwidth]{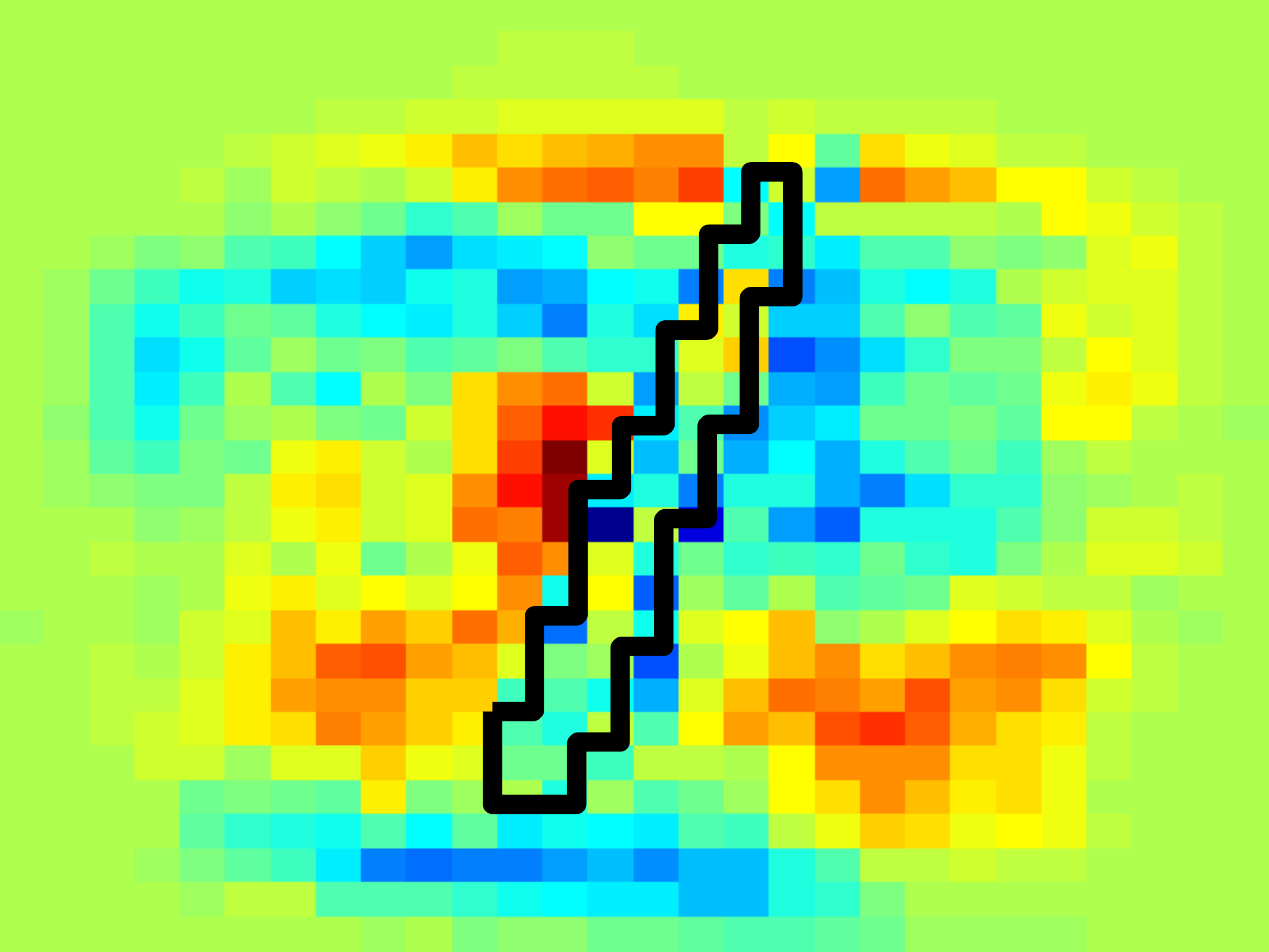}}&
		\subcaptionbox*{}{\includegraphics[width=0.09\textwidth]{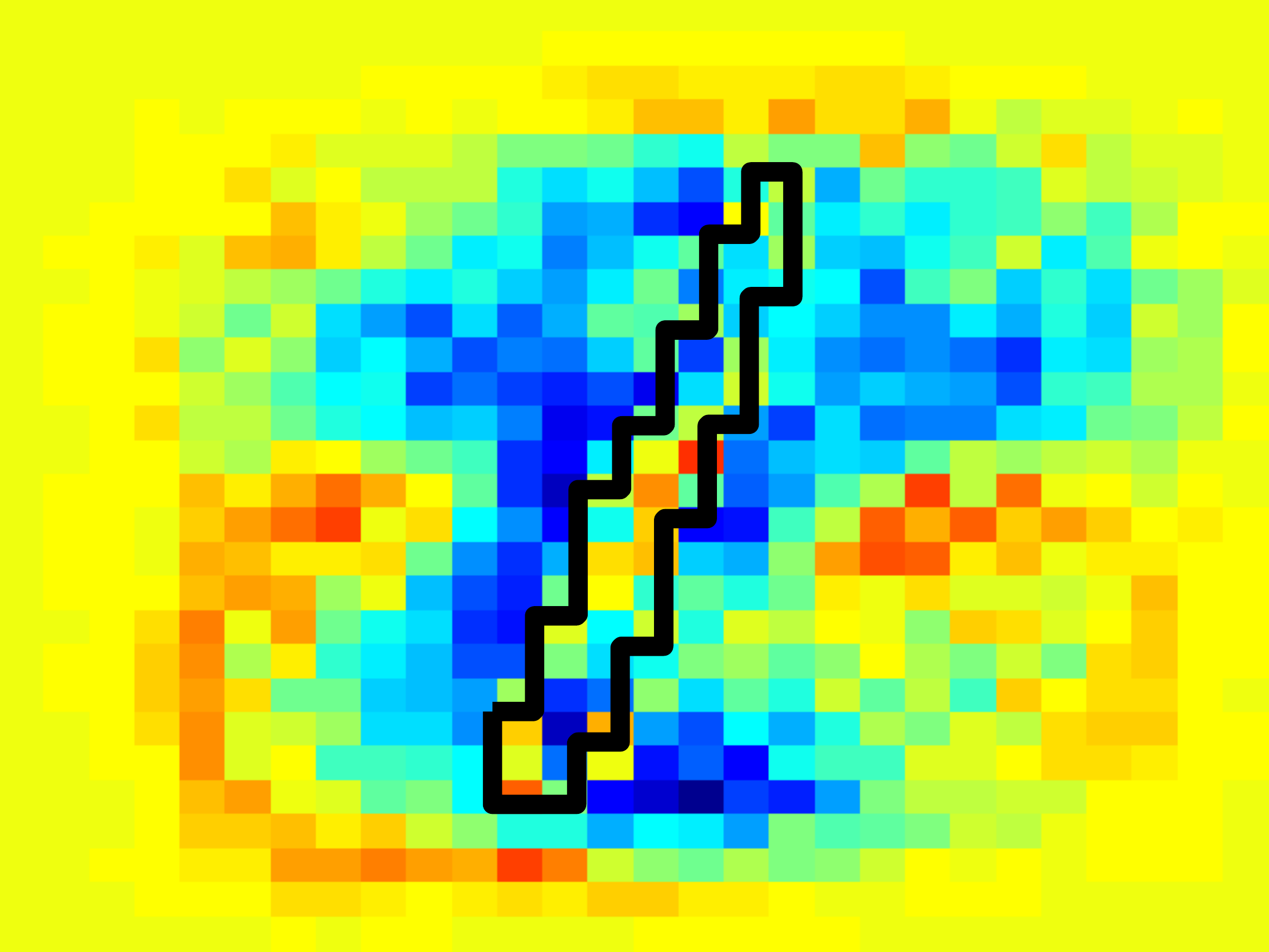}}&
		\subcaptionbox*{}{\includegraphics[width=0.09\textwidth]{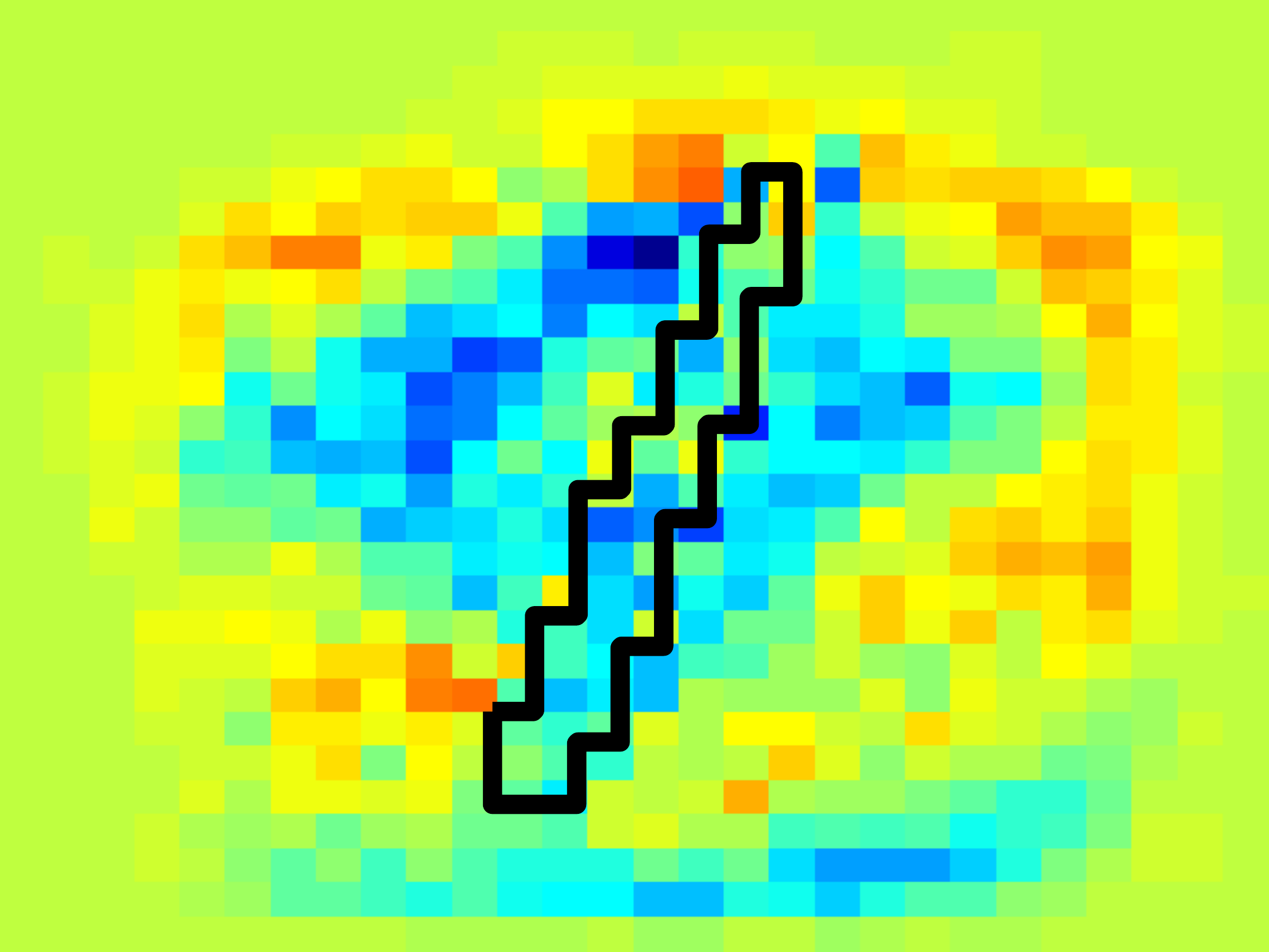}} 
		\\[-1em]
		&0.65&1.09&0.88&0.88&0.89&0.83&0.81&0.81&0.69&0.73\\[0.5em]\hline
		
		\subcaptionbox*{}{\includegraphics[width=0.09\textwidth]{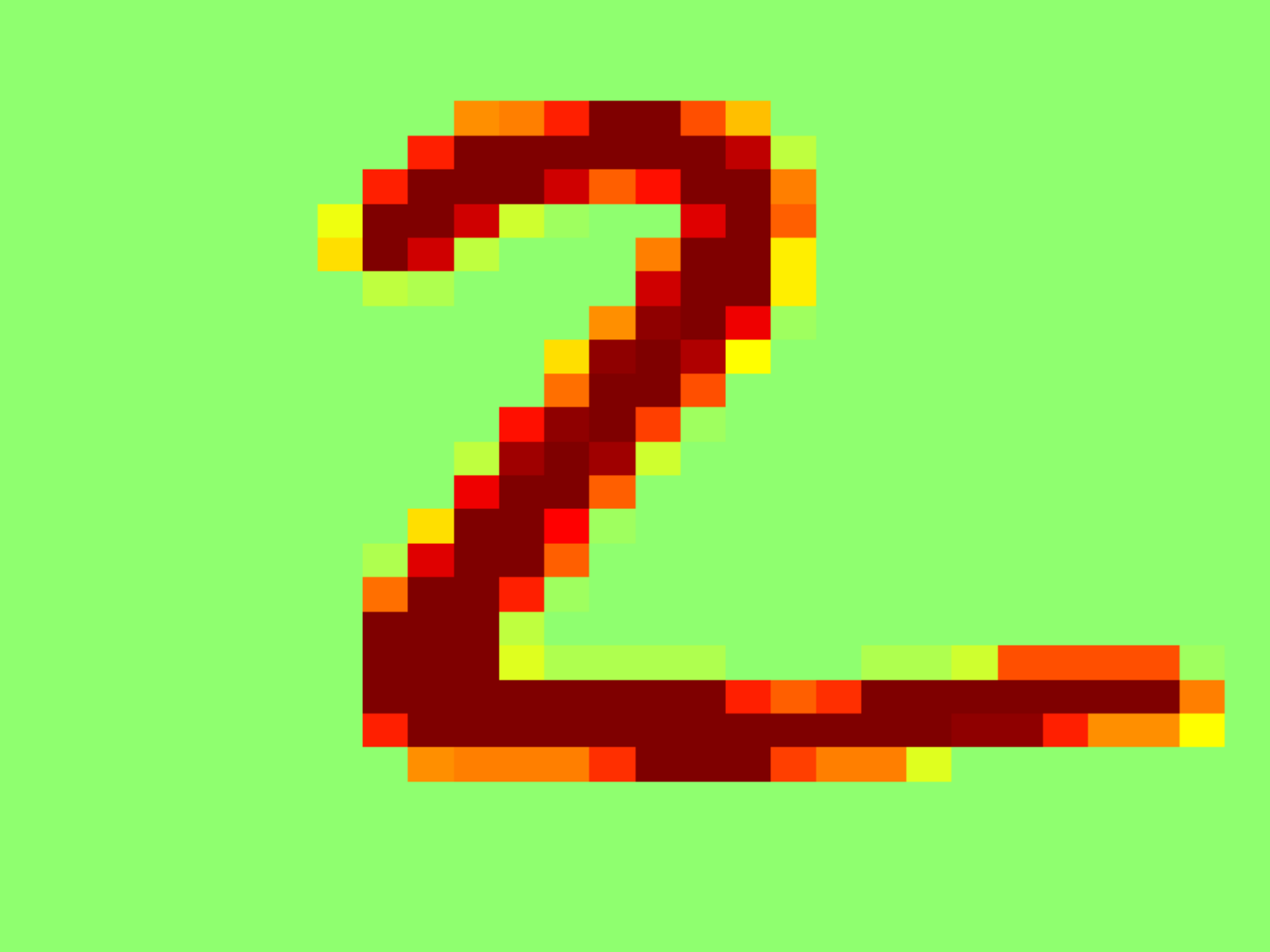}}&
		\subcaptionbox*{}{\includegraphics[width=0.09\textwidth]{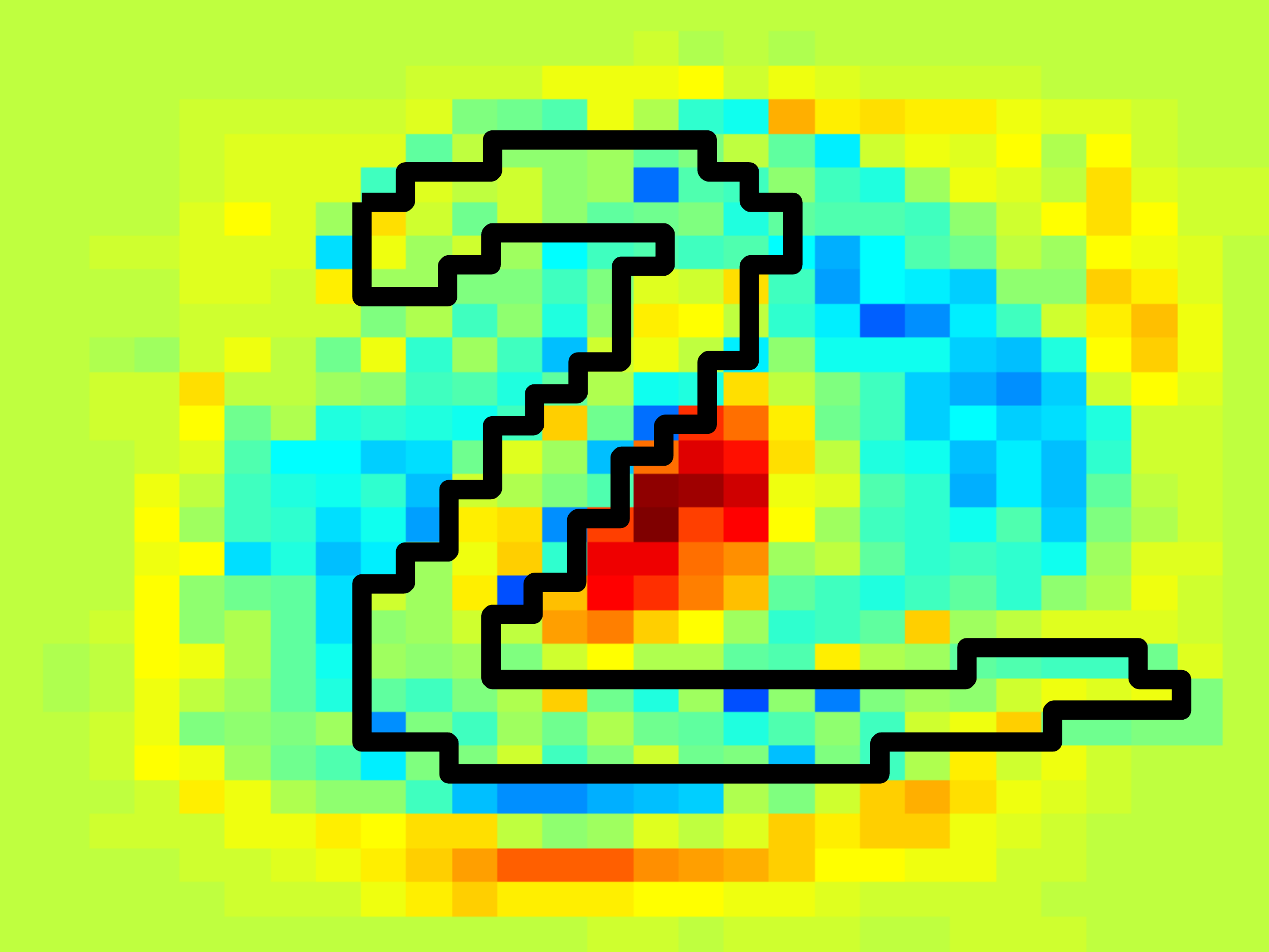}}&
		\subcaptionbox*{}{\includegraphics[width=0.09\textwidth]{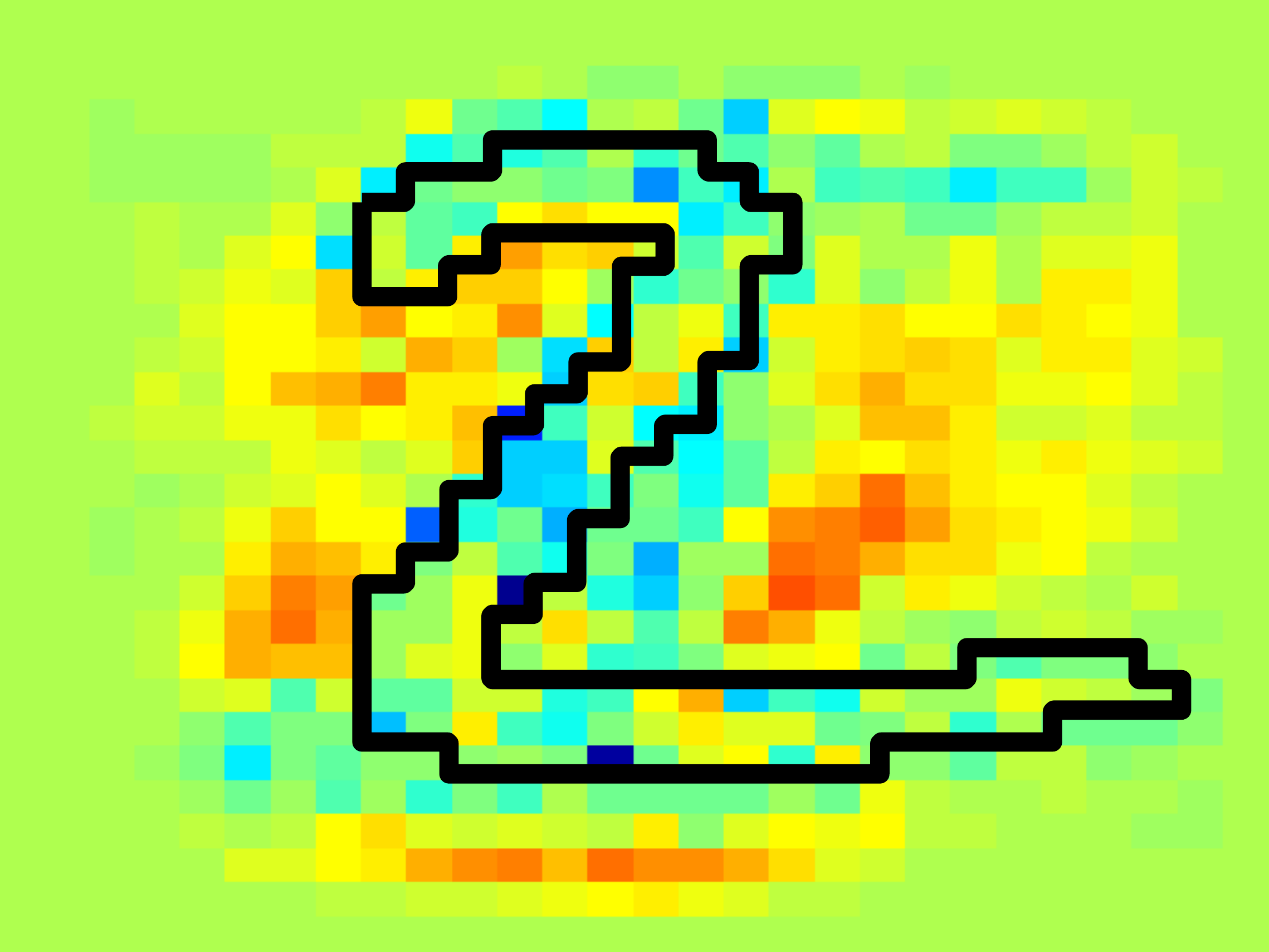}}&
		\subcaptionbox*{}{\includegraphics[width=0.09\textwidth]{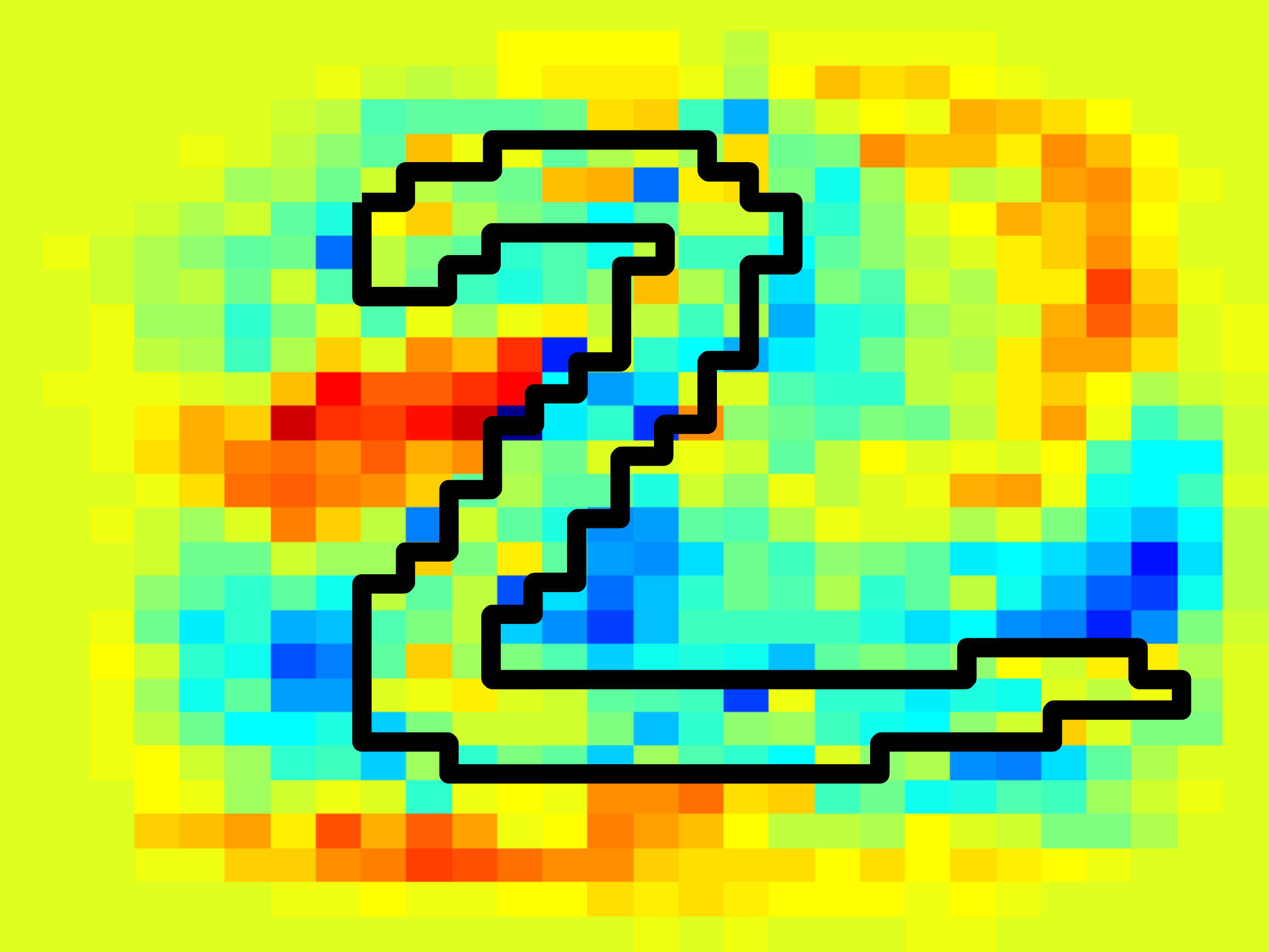}}&
		\subcaptionbox*{}{\includegraphics[width=0.09\textwidth]{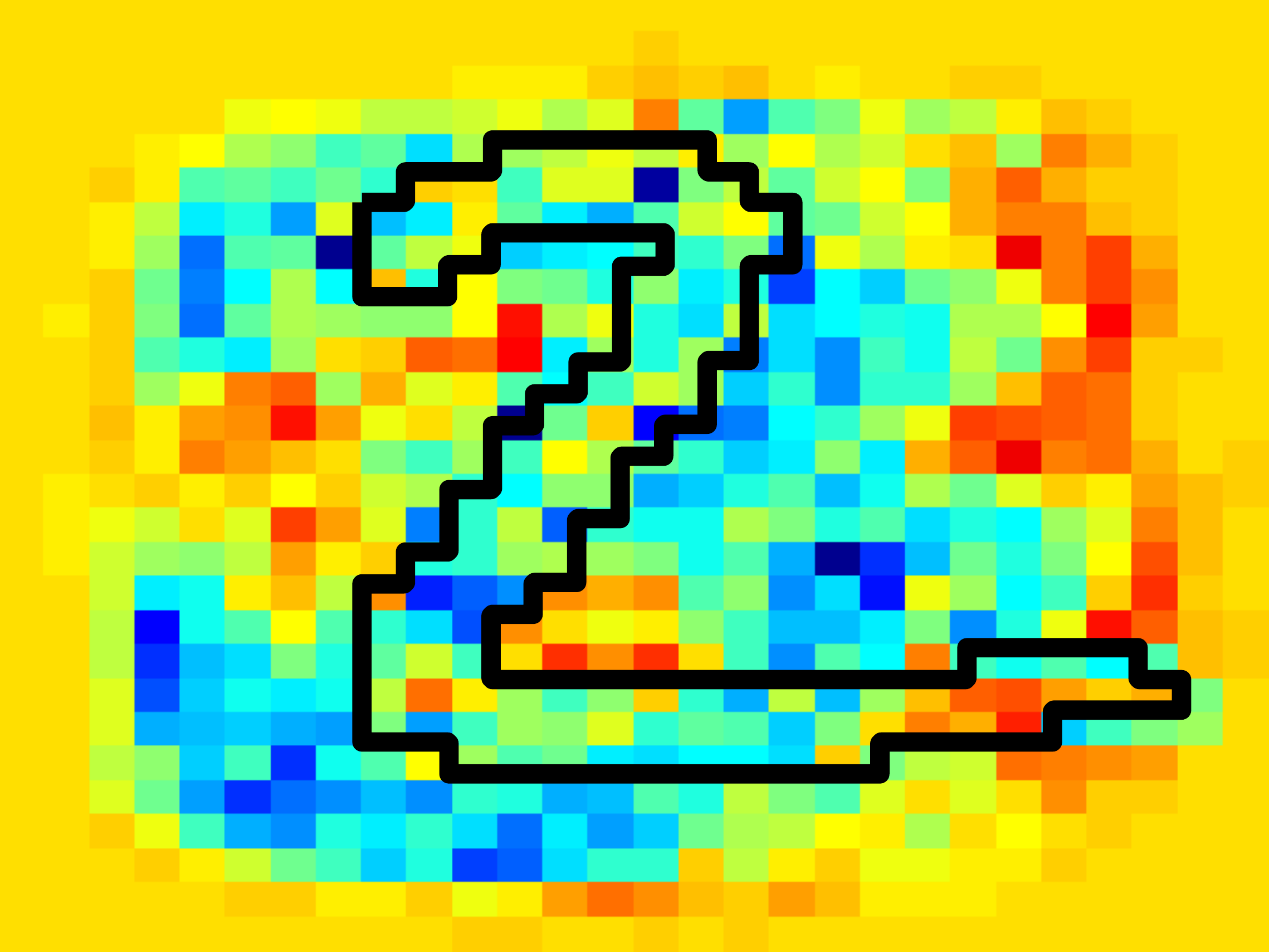}}&
		\subcaptionbox*{}{\includegraphics[width=0.09\textwidth]{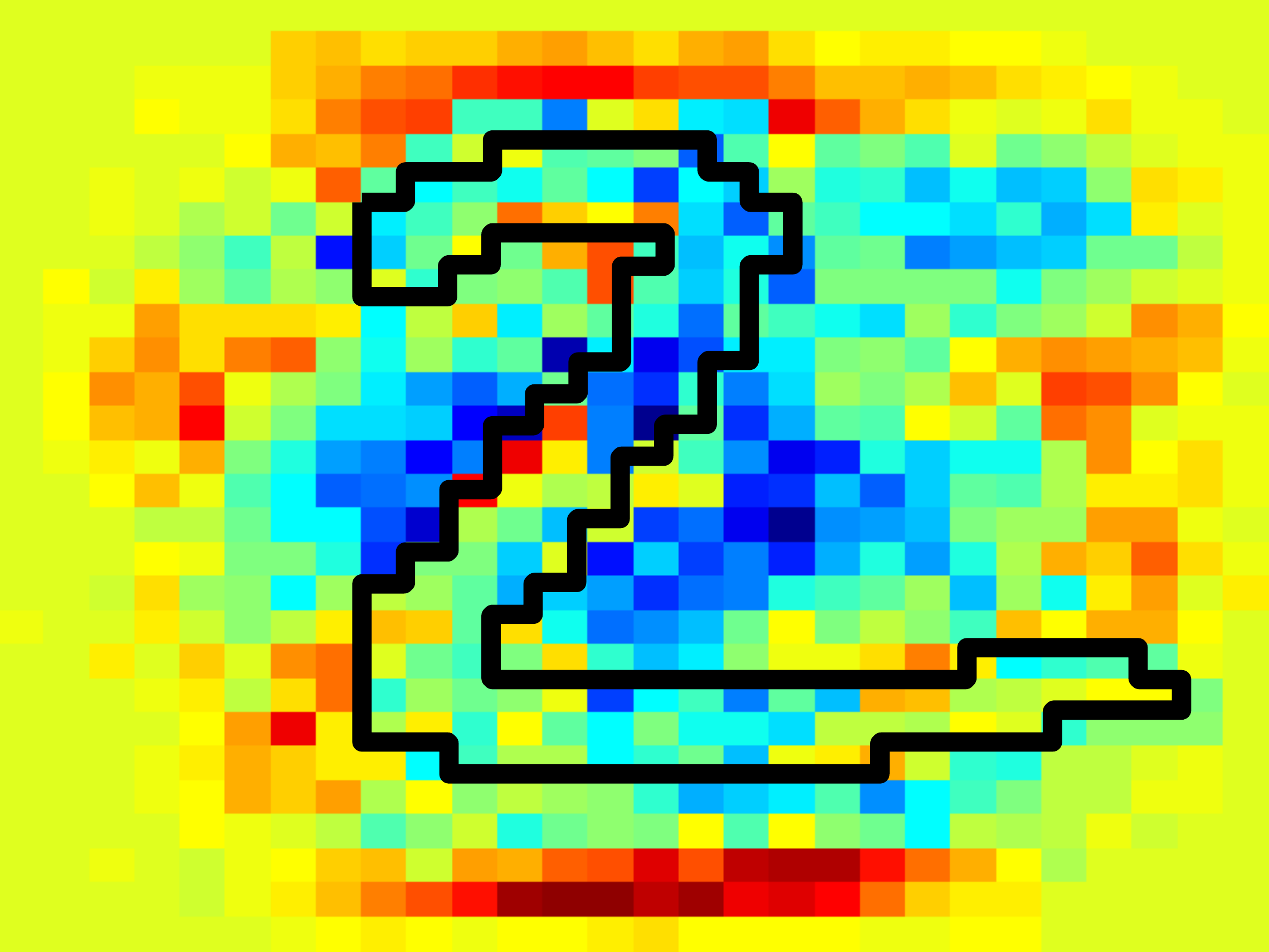}}&
		\subcaptionbox*{}{\includegraphics[width=0.09\textwidth]{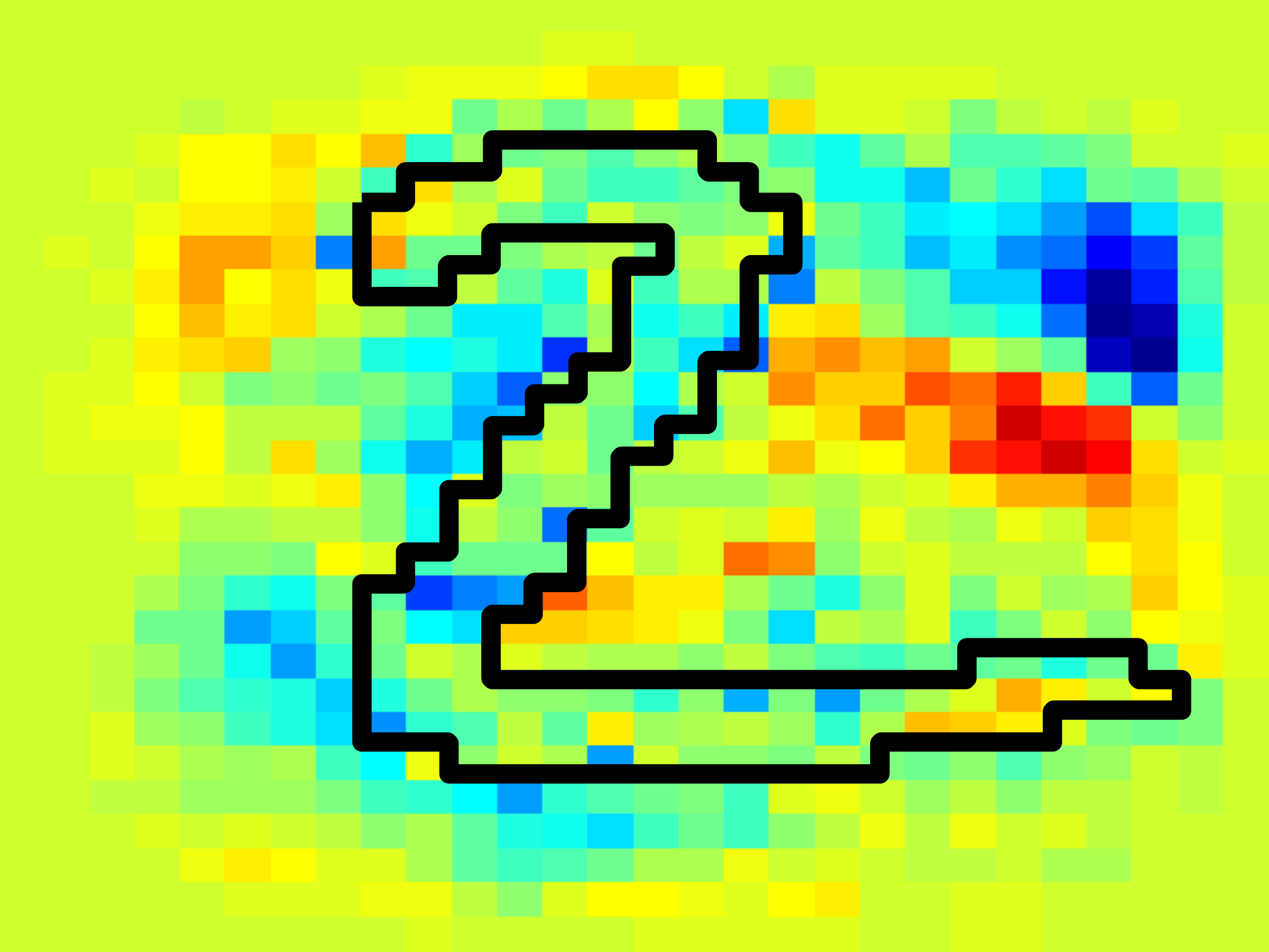}}&
		\subcaptionbox*{}{\includegraphics[width=0.09\textwidth]{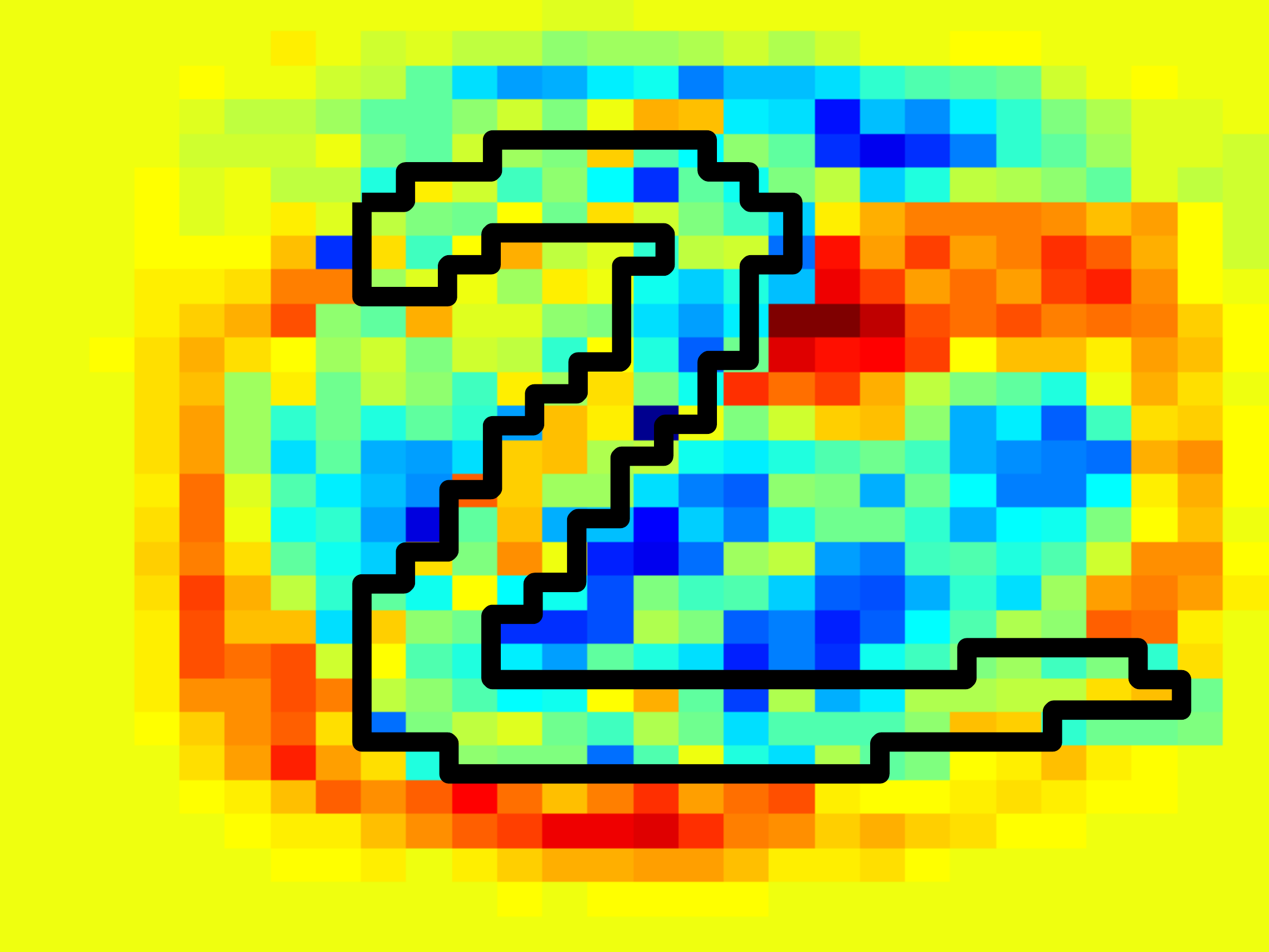}}&
		\subcaptionbox*{}{\includegraphics[width=0.09\textwidth]{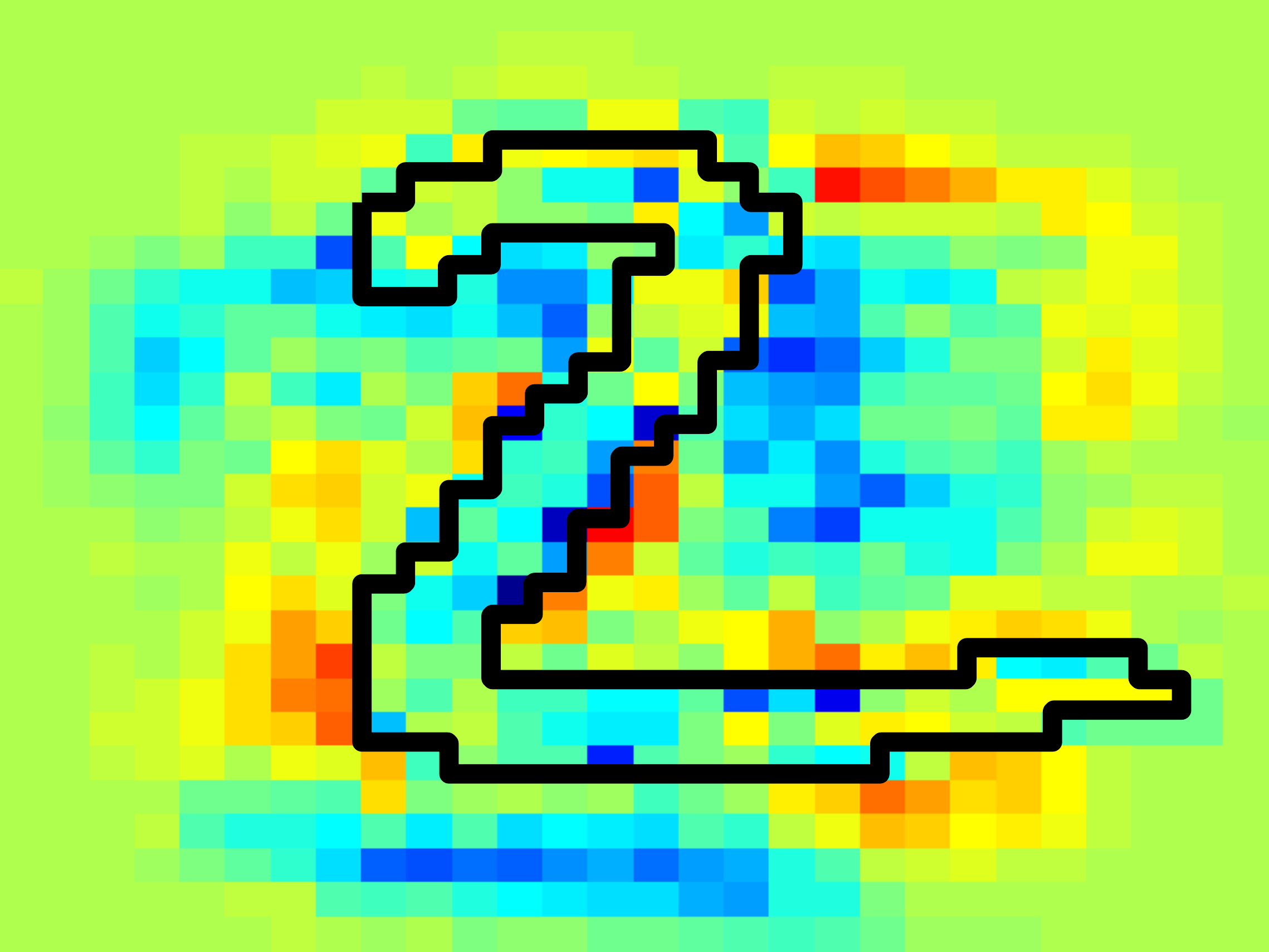}}&
		\subcaptionbox*{}{\includegraphics[width=0.09\textwidth]{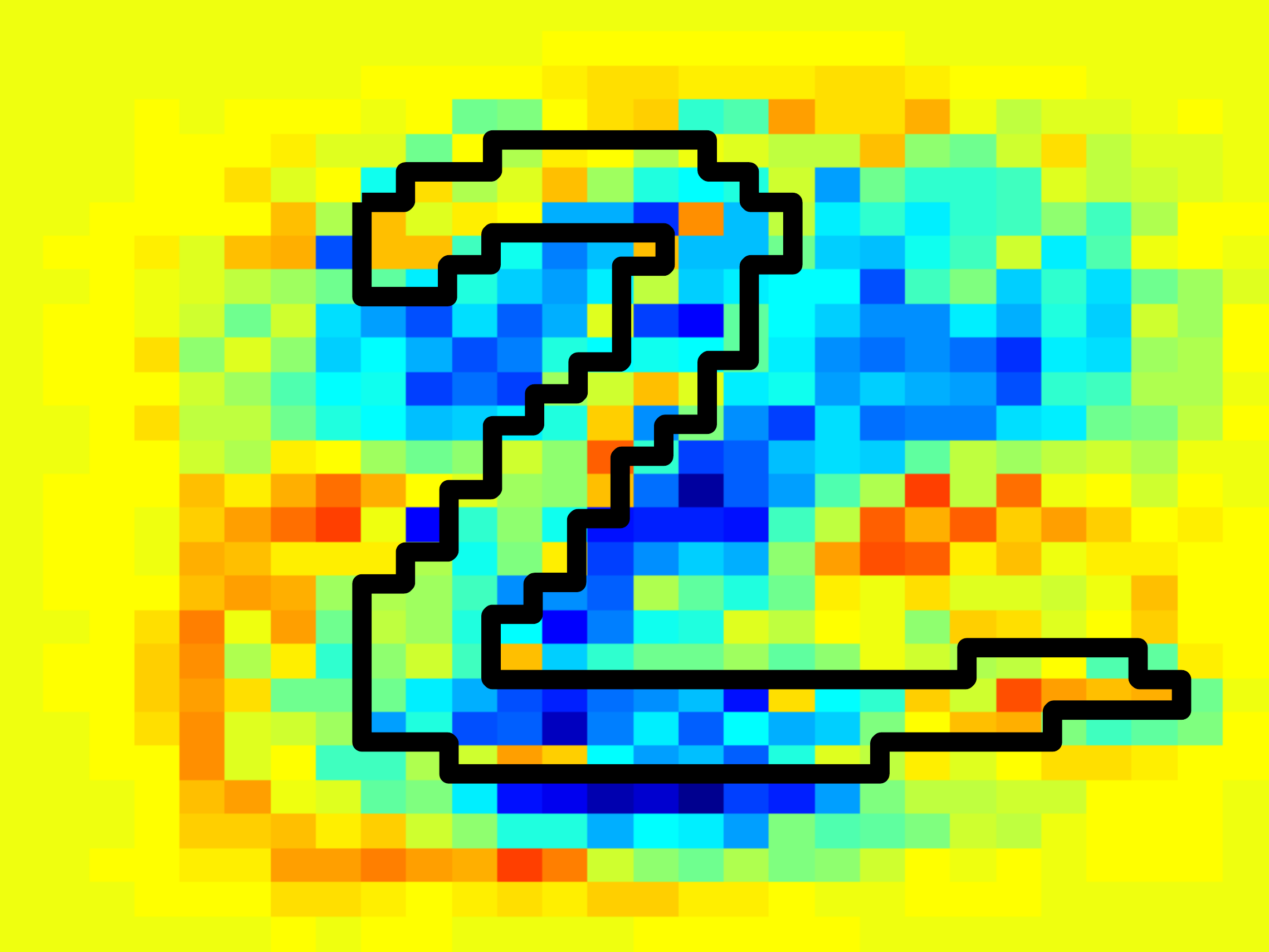}}&
		\subcaptionbox*{}{\includegraphics[width=0.09\textwidth]{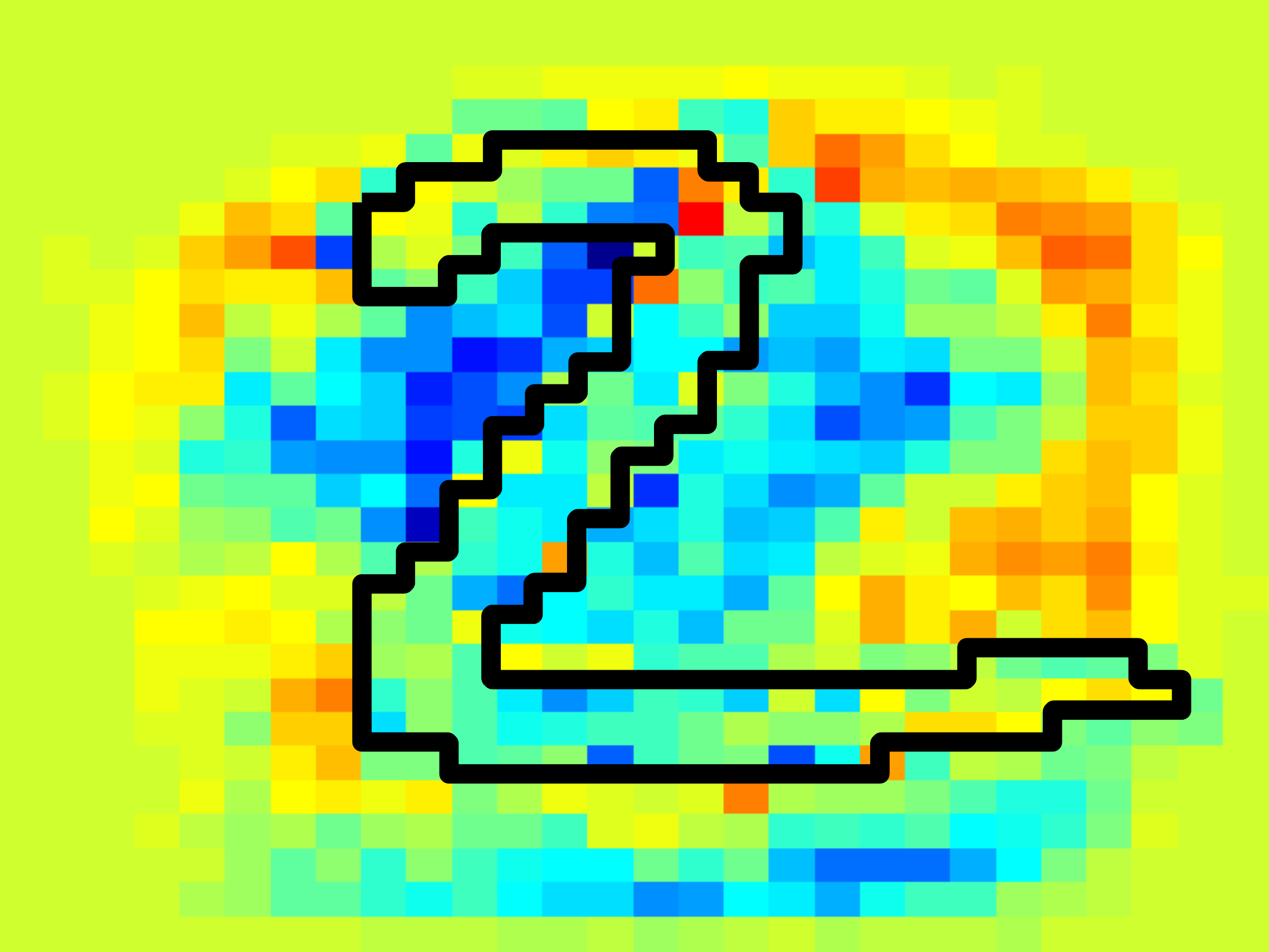}}
		\\[-1em]
		&0.87&0.90&1.25&0.97&0.64&1.01&0.97&0.21&0.80&0.52\\[0.5em]\hline
		
		\subcaptionbox*{}{\includegraphics[width=0.09\textwidth]{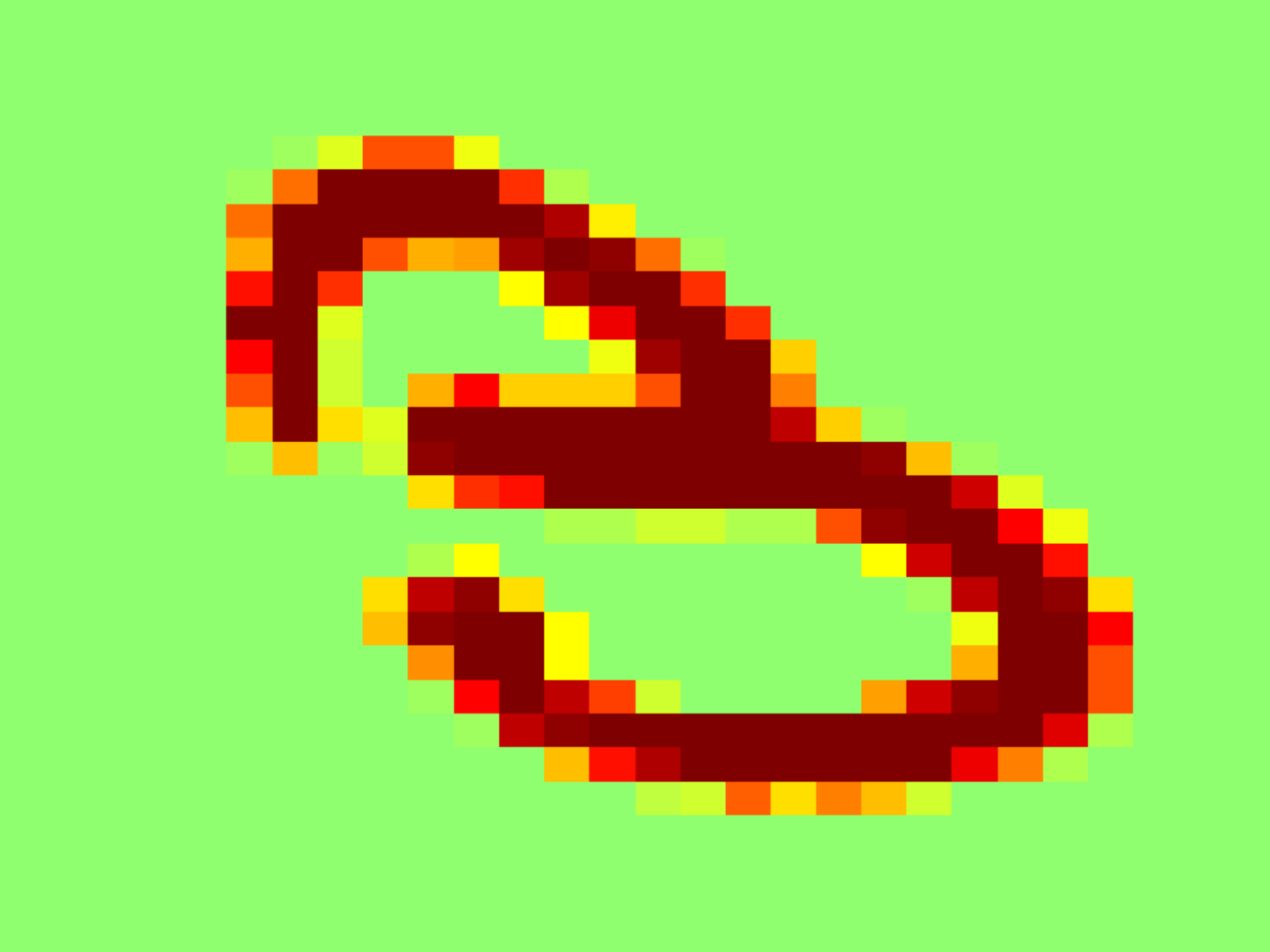}}&
		\subcaptionbox*{}{\includegraphics[width=0.09\textwidth]{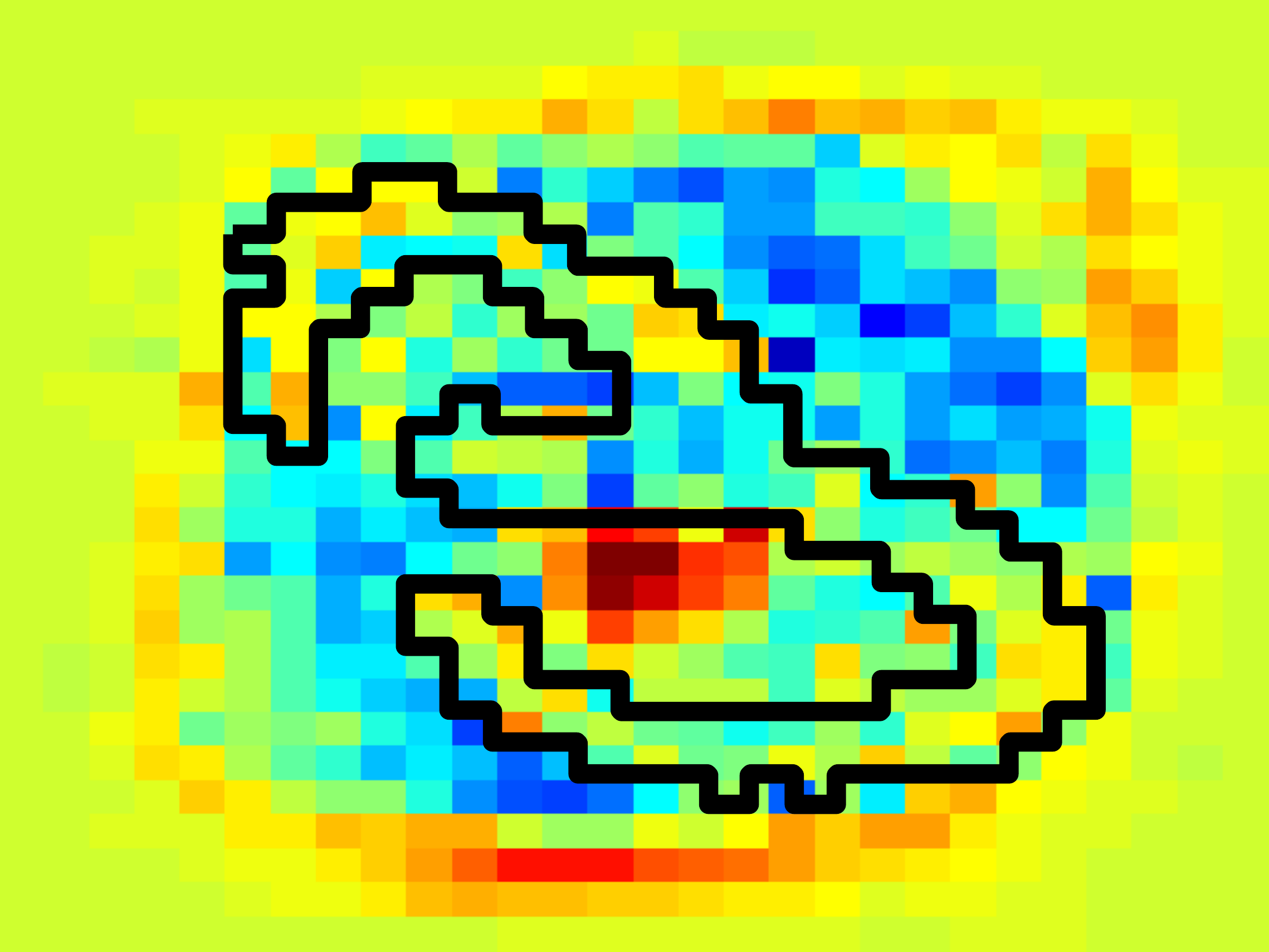}}&
		\subcaptionbox*{}{\includegraphics[width=0.09\textwidth]{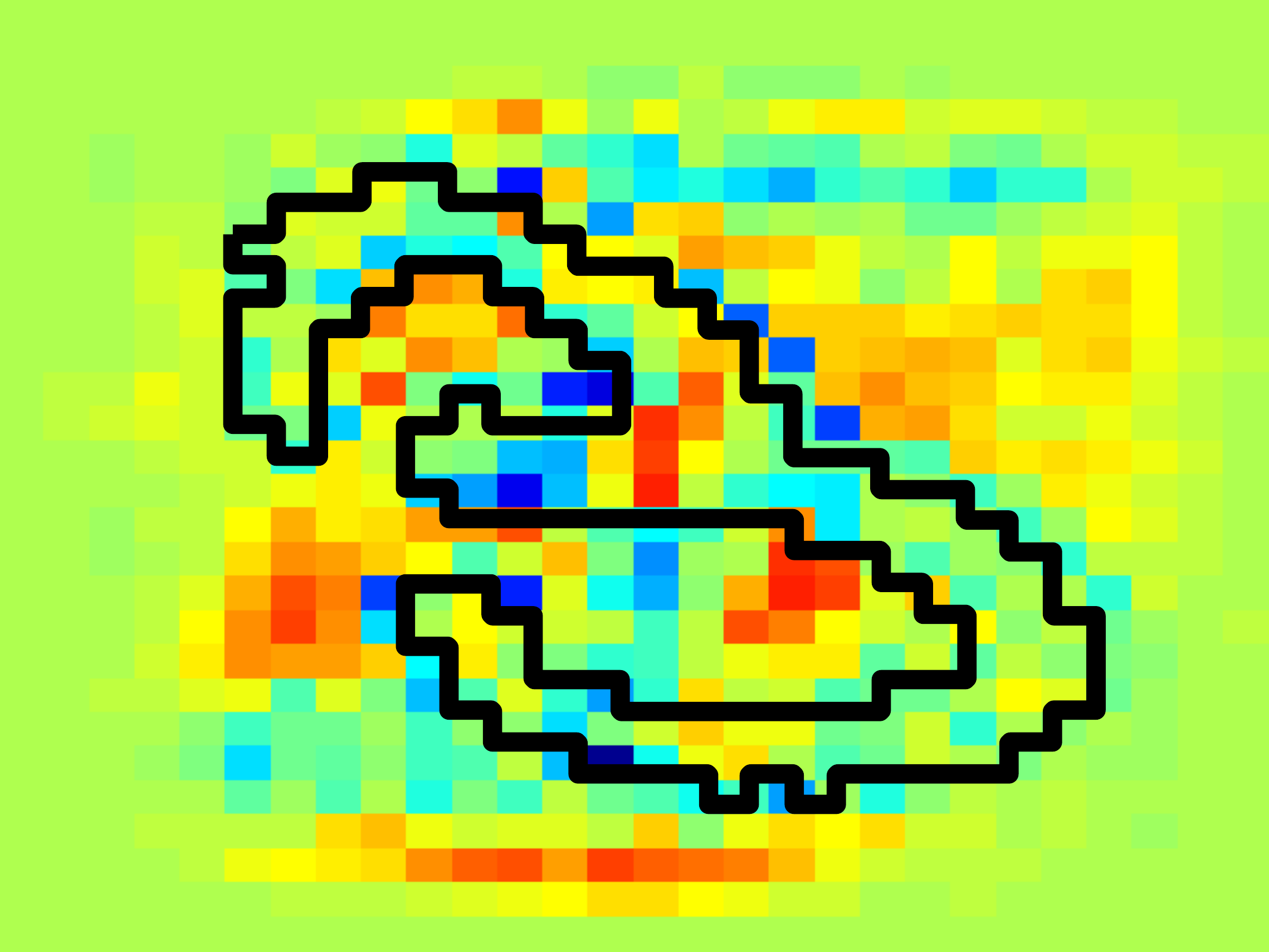}}&
		\subcaptionbox*{}{\includegraphics[width=0.09\textwidth]{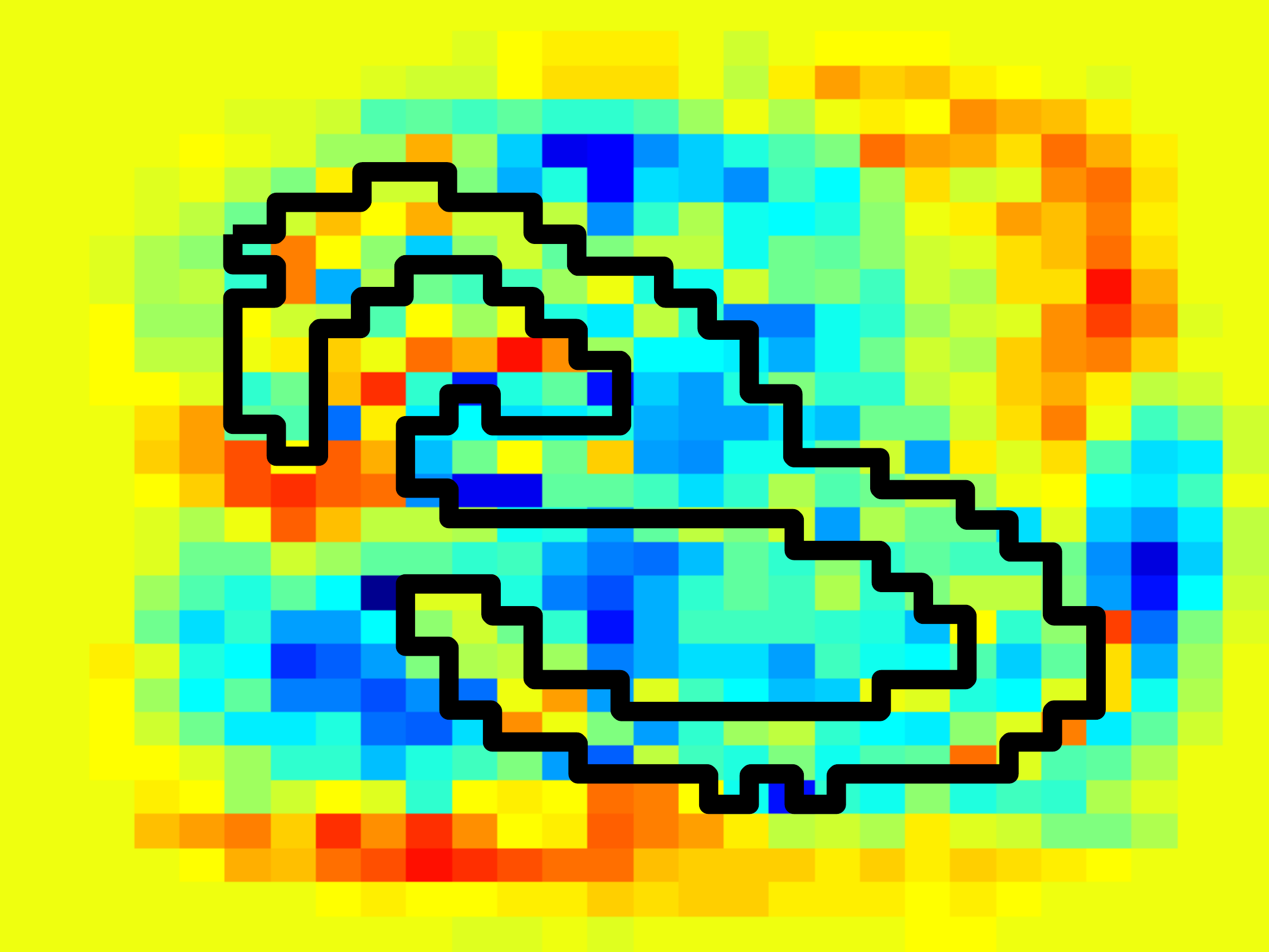}}&
		\subcaptionbox*{}{\includegraphics[width=0.09\textwidth]{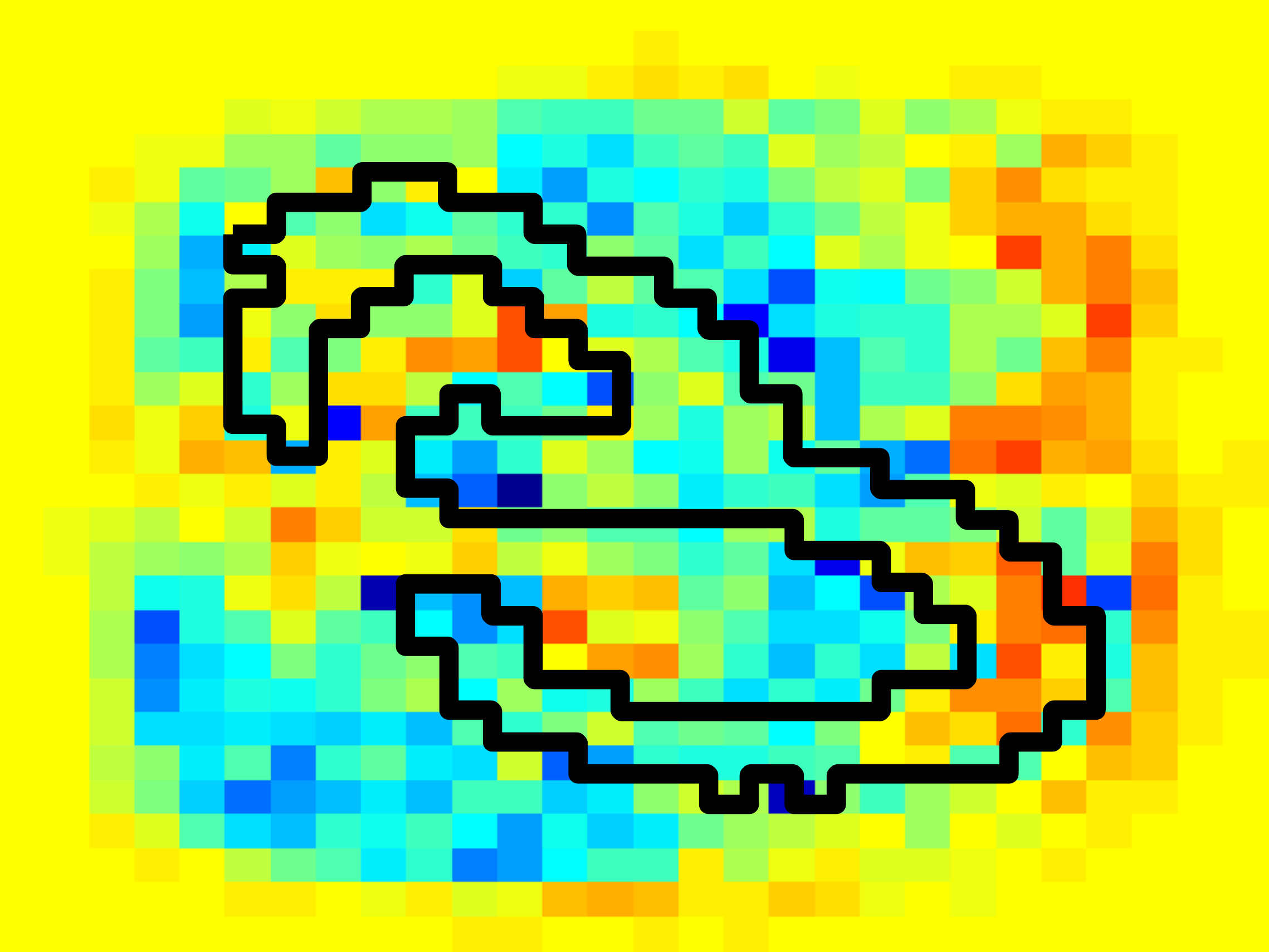}}&
		\subcaptionbox*{}{\includegraphics[width=0.09\textwidth]{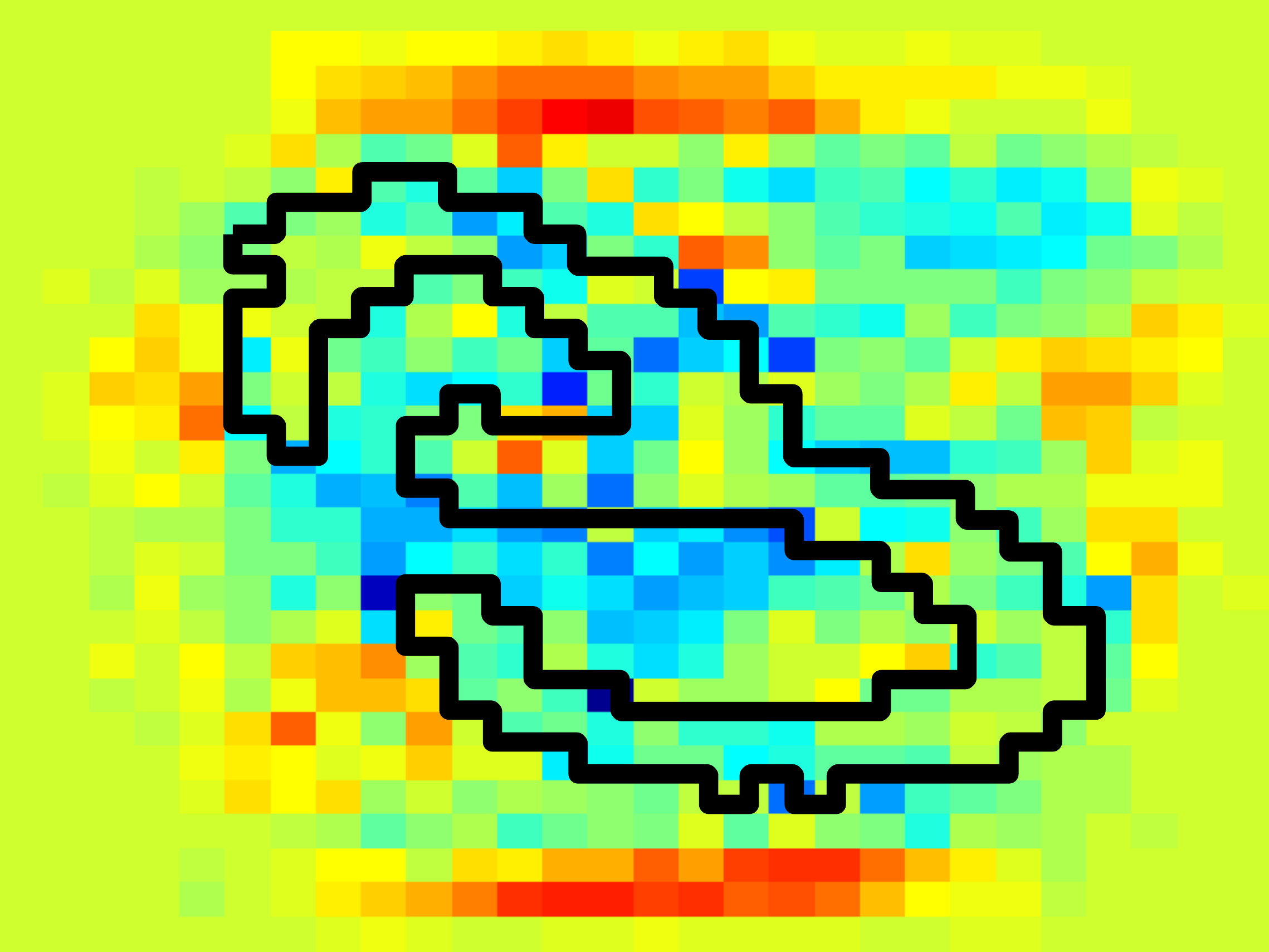}}&
		\subcaptionbox*{}{\includegraphics[width=0.09\textwidth]{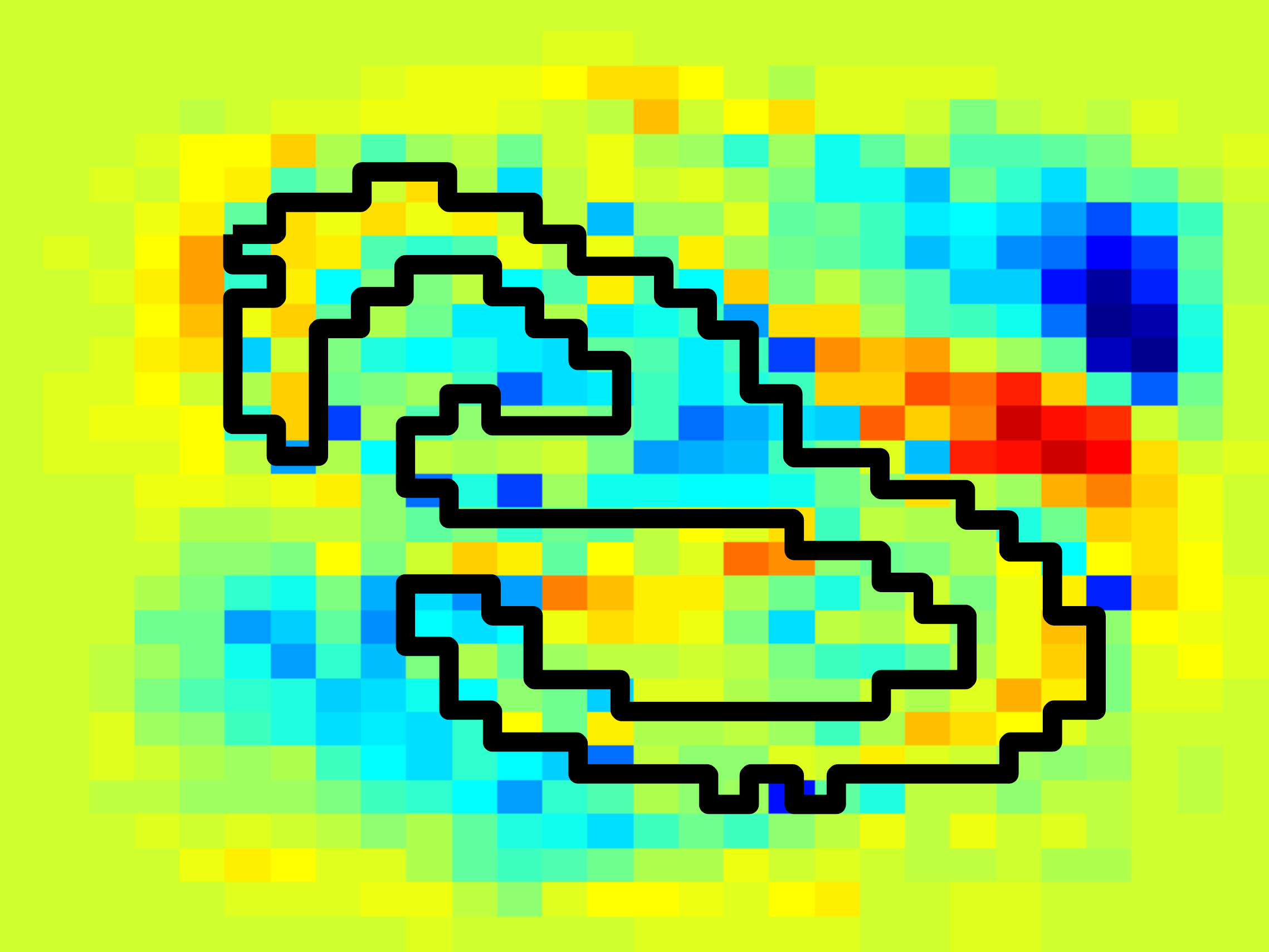}}&
		\subcaptionbox*{}{\includegraphics[width=0.09\textwidth]{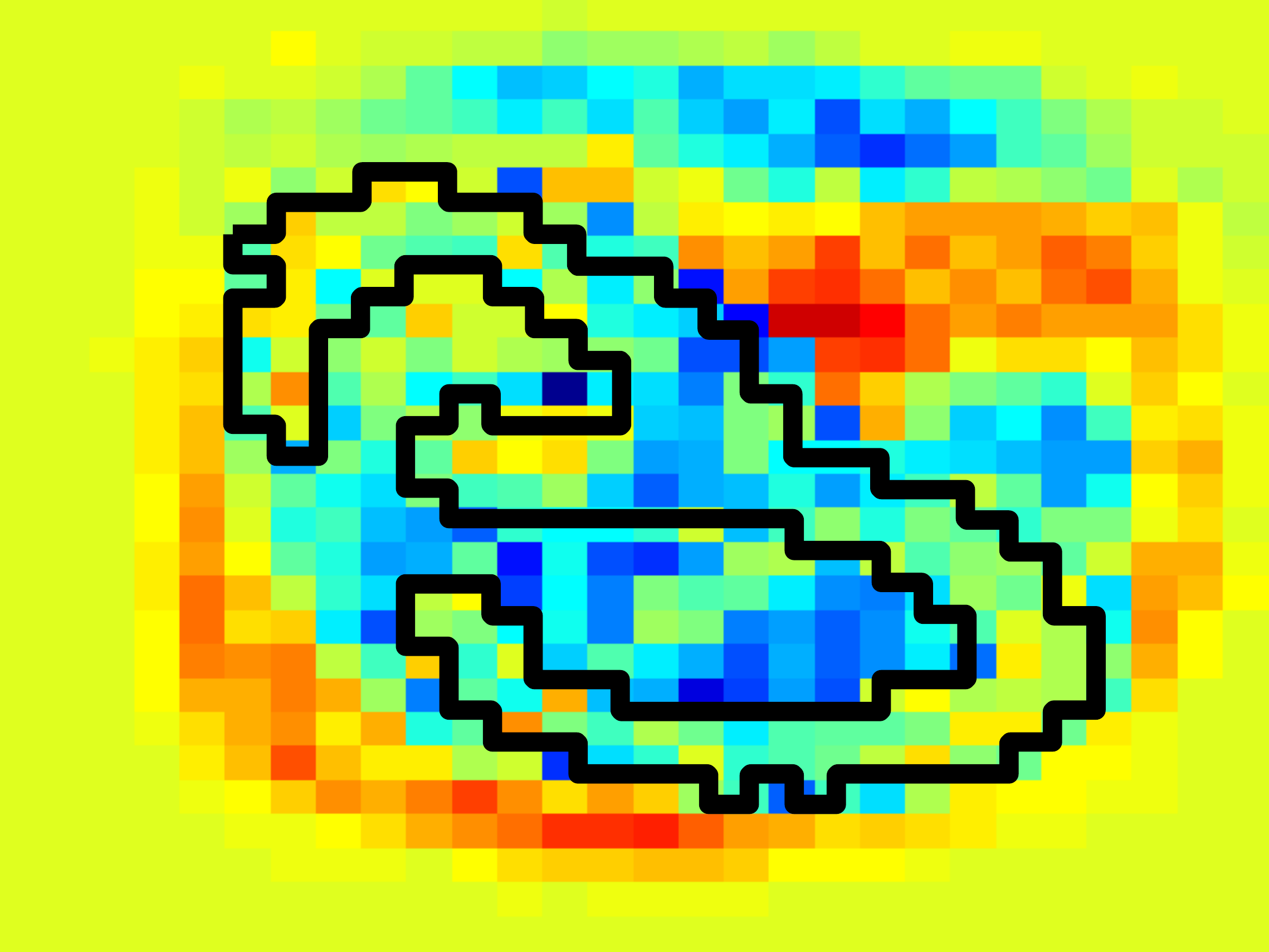}}&
		\subcaptionbox*{}{\includegraphics[width=0.09\textwidth]{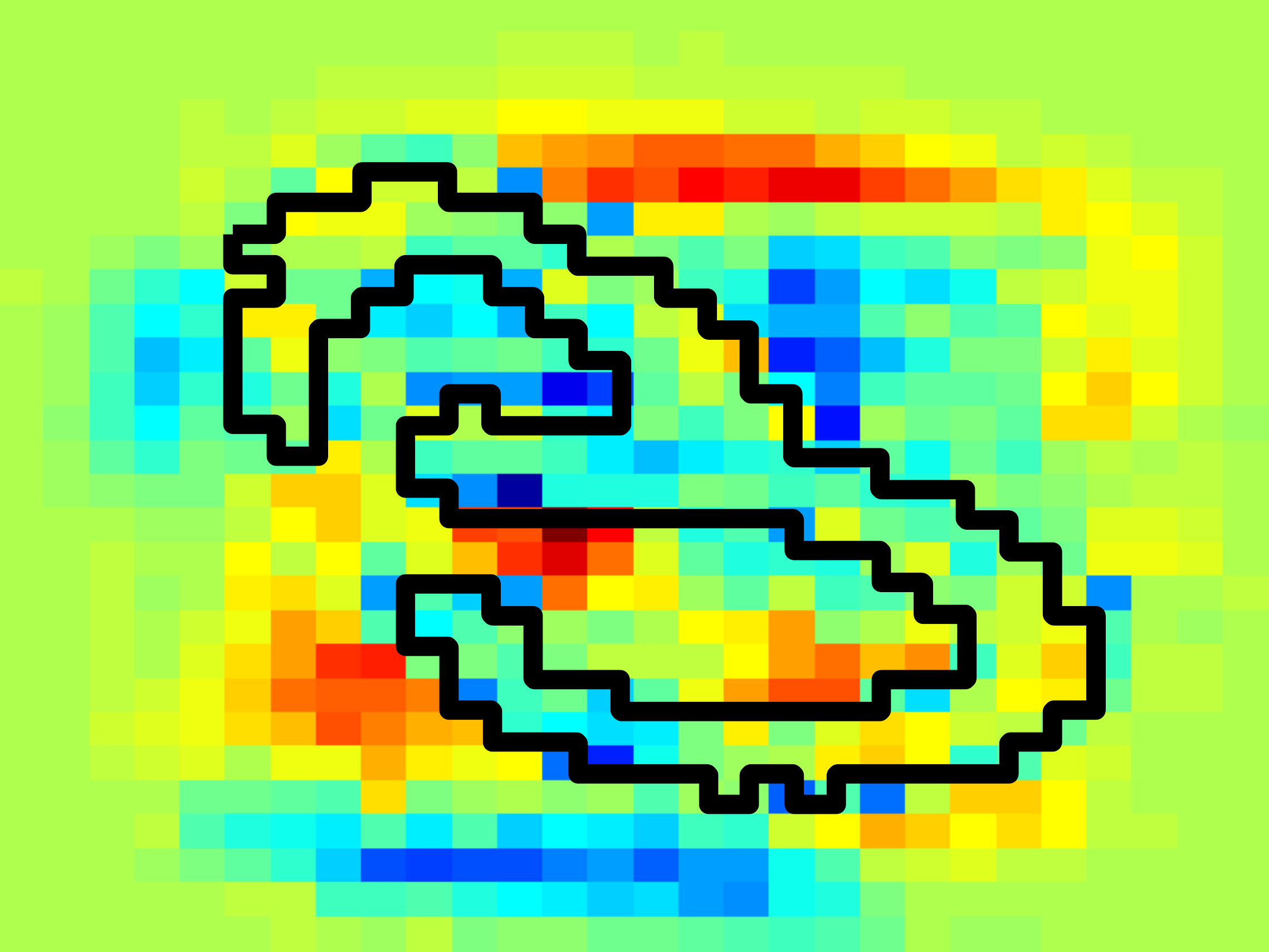}}&
		\subcaptionbox*{}{\includegraphics[width=0.09\textwidth]{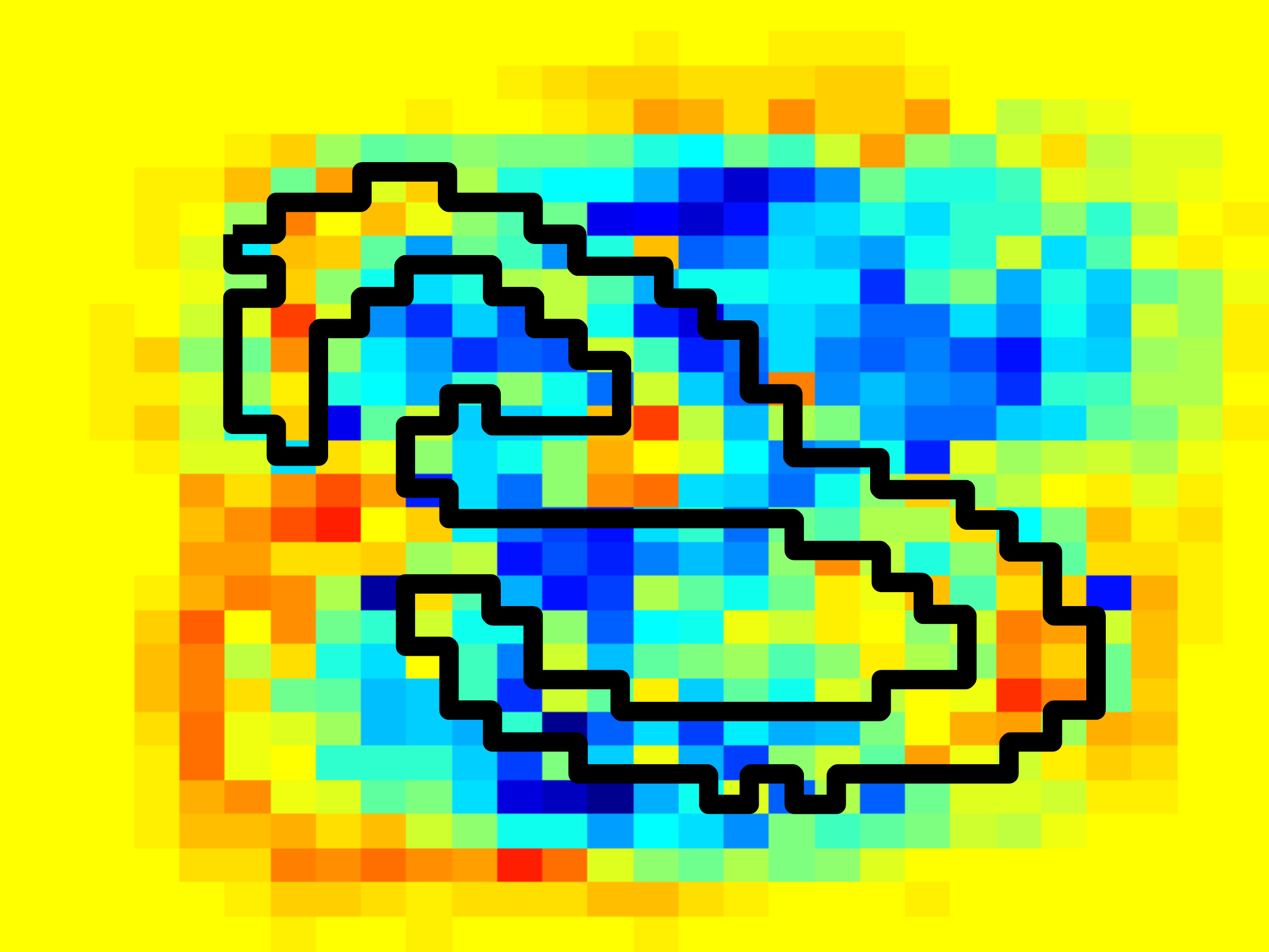}}&
		\subcaptionbox*{}{\includegraphics[width=0.09\textwidth]{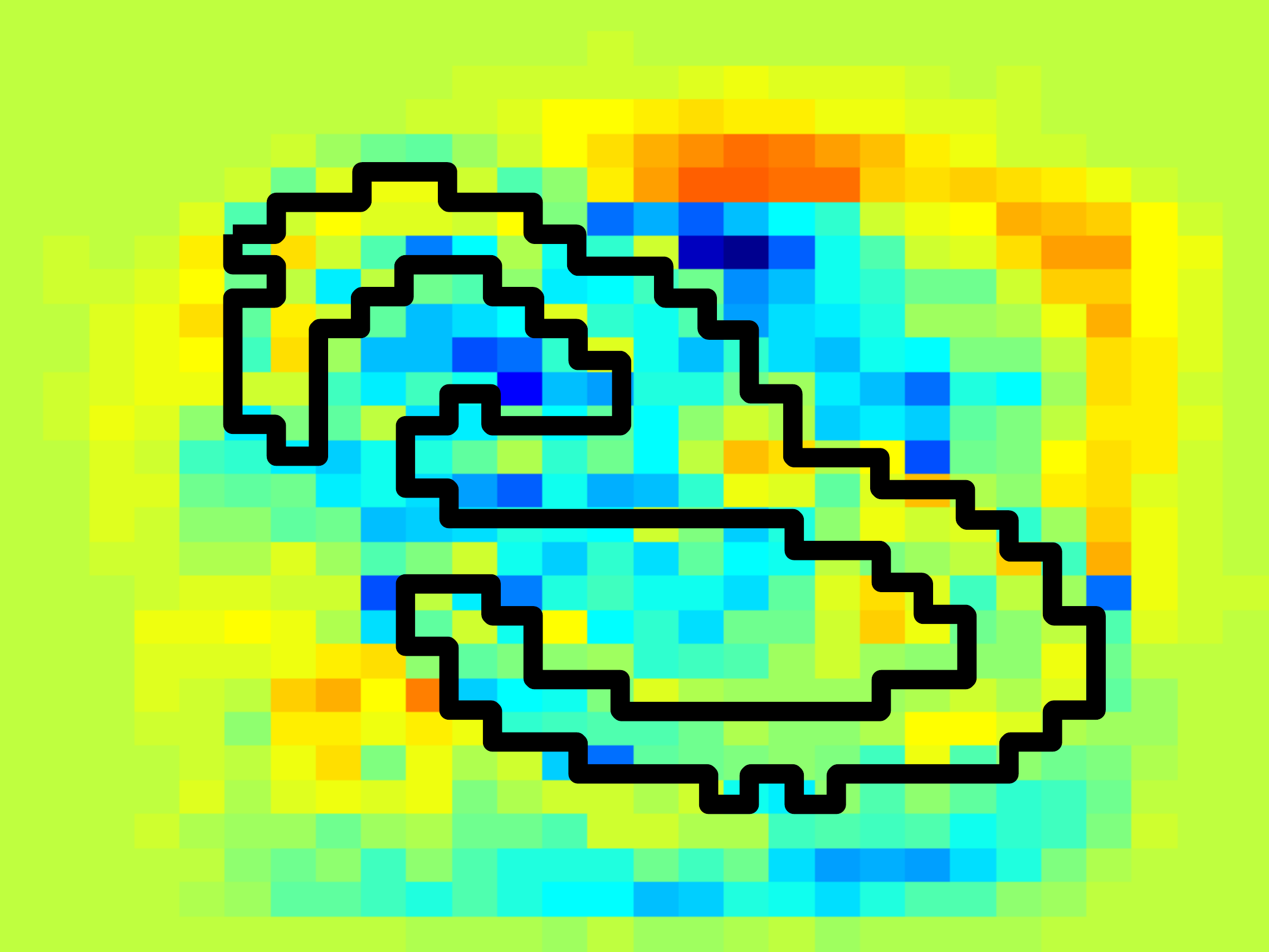}}
		\\[-1em]
		&0.65&0.75&0.81&0.89&0.80&0.80&0.78&0.55&0.74&0.61\\[0.5em]\hline
		
		\subcaptionbox*{}{\includegraphics[width=0.09\textwidth]{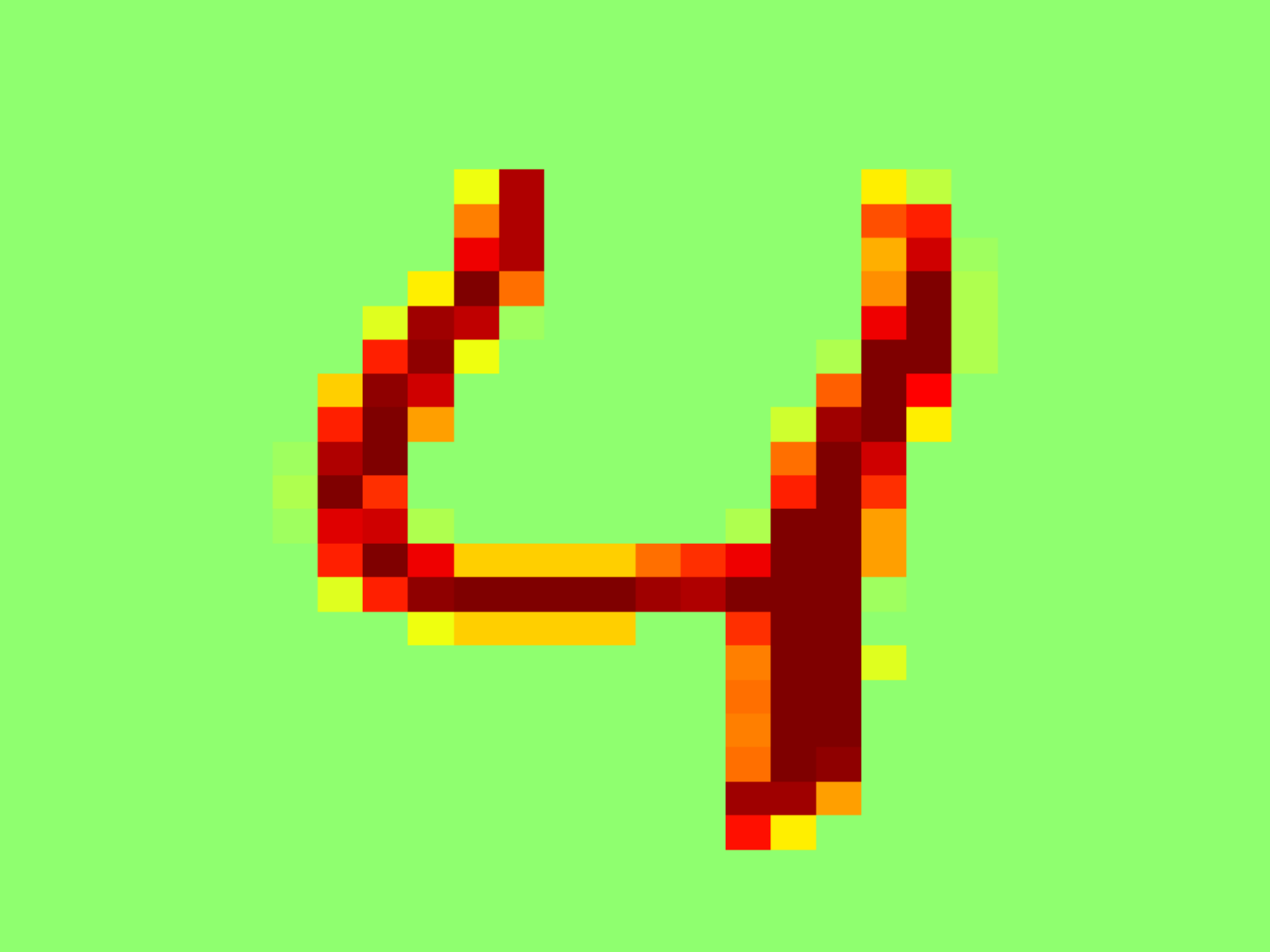}}&
		\subcaptionbox*{}{\includegraphics[width=0.09\textwidth]{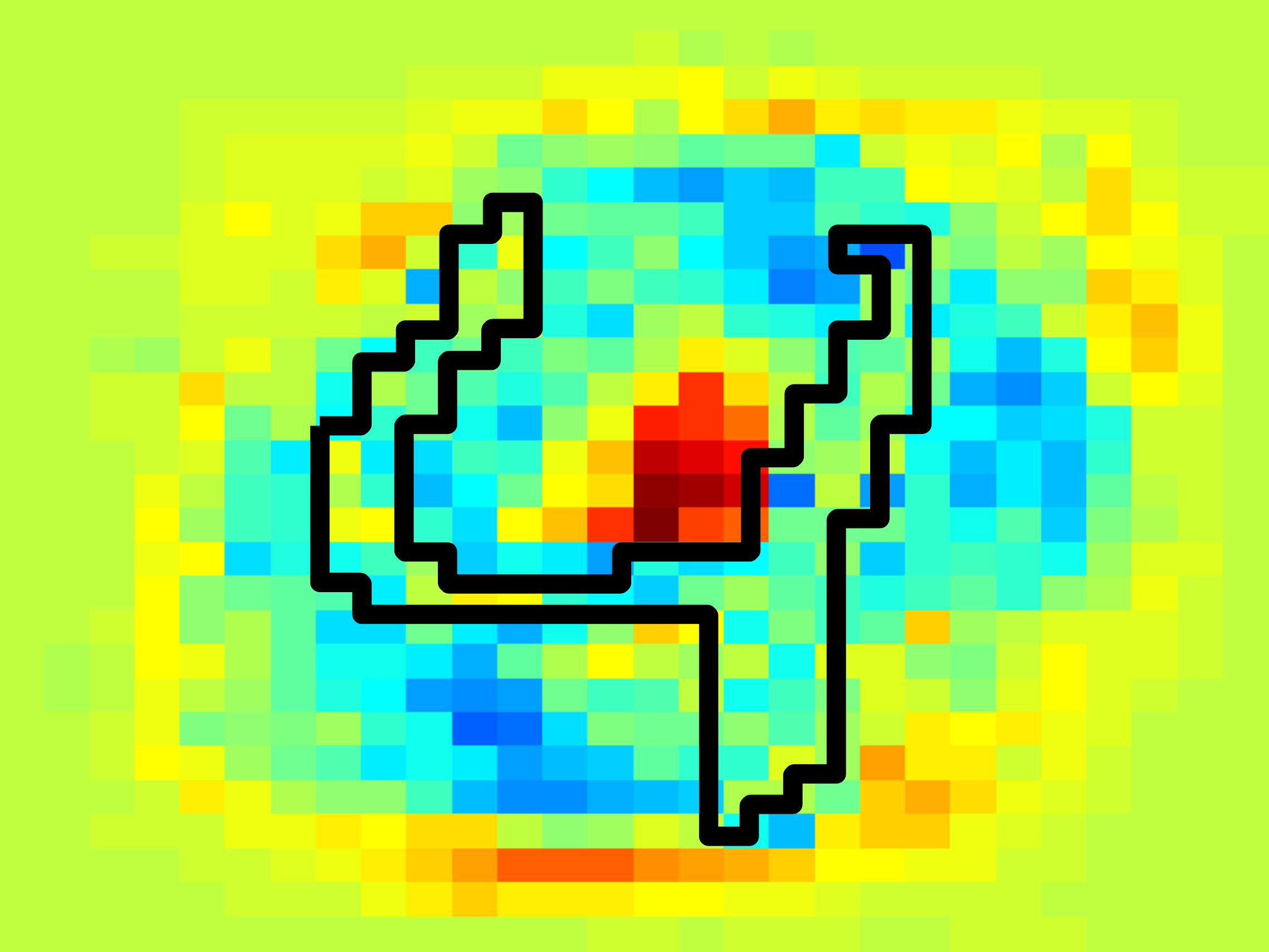}}&
		\subcaptionbox*{}{\includegraphics[width=0.09\textwidth]{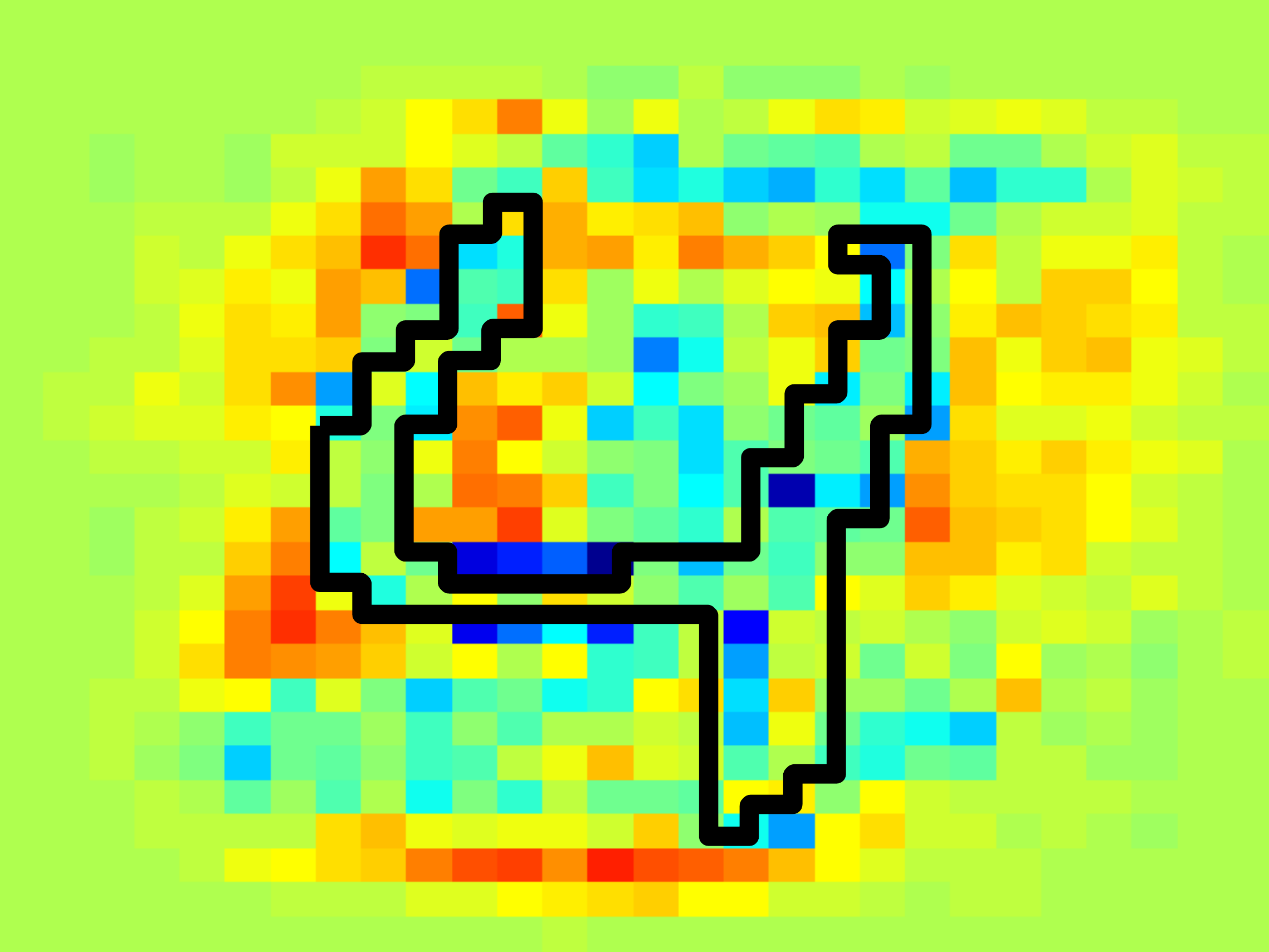}}&
		\subcaptionbox*{}{\includegraphics[width=0.09\textwidth]{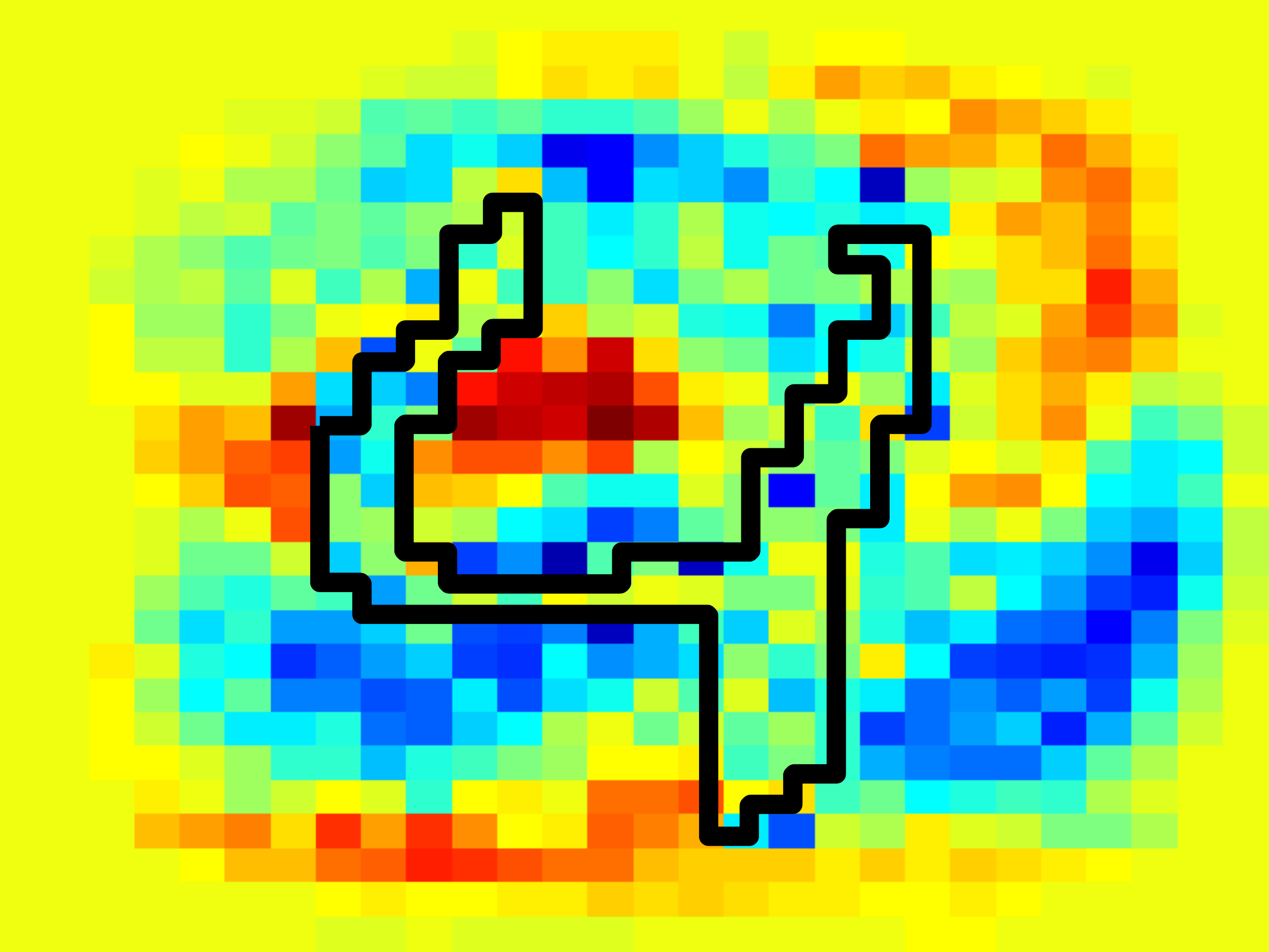}}&
		\subcaptionbox*{}{\includegraphics[width=0.09\textwidth]{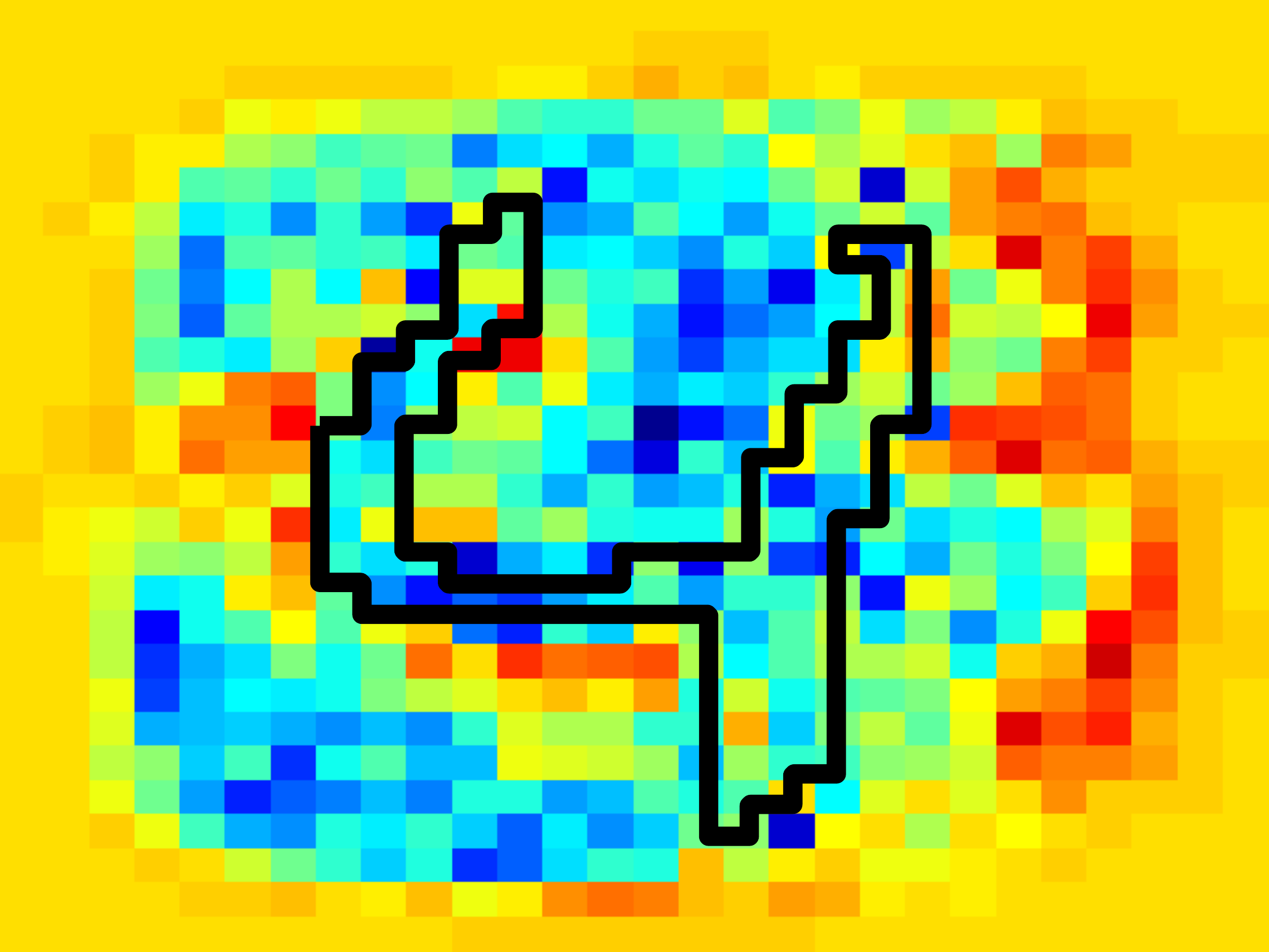}}&
		\subcaptionbox*{}{\includegraphics[width=0.09\textwidth]{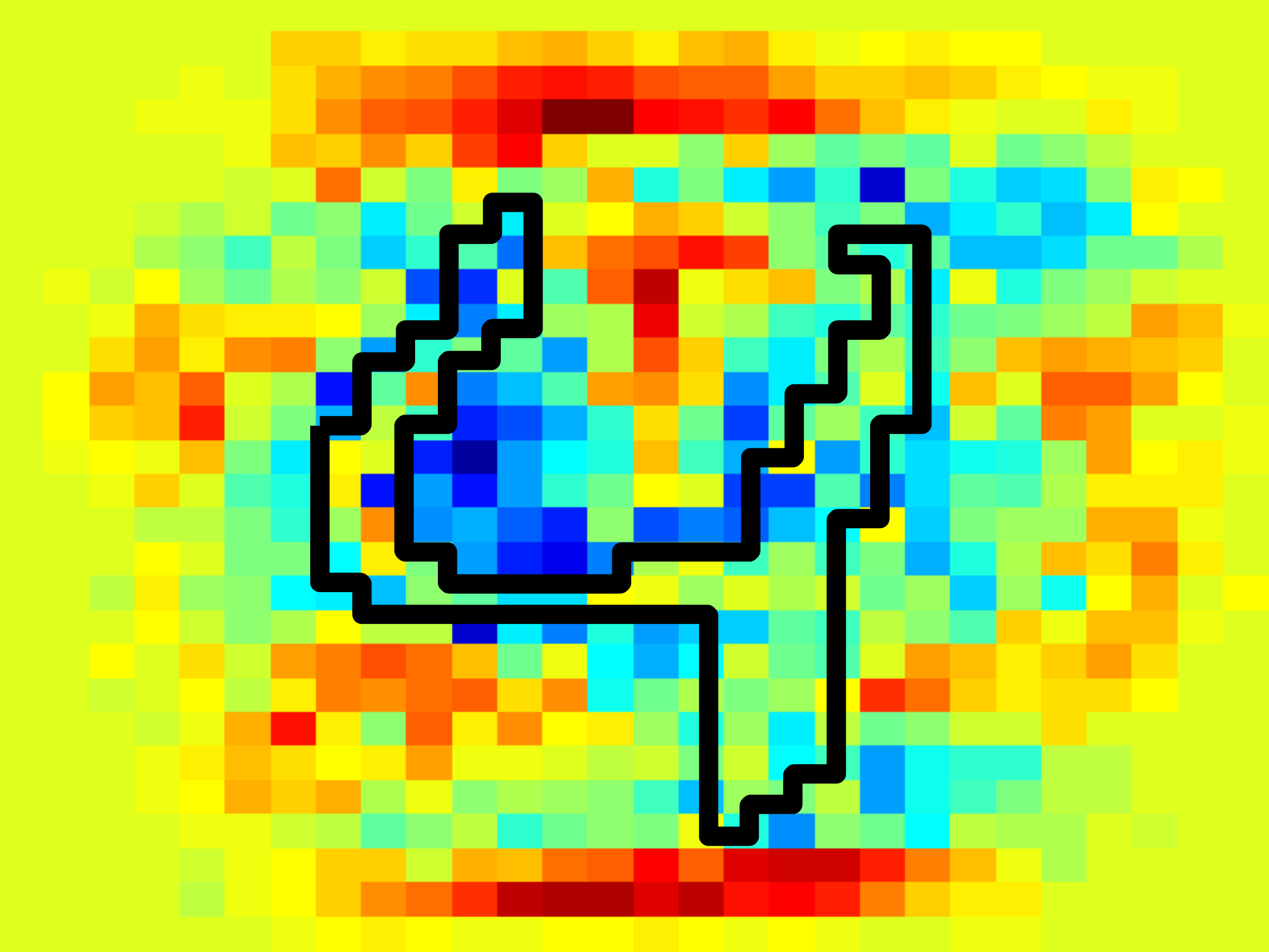}}&
		\subcaptionbox*{}{\includegraphics[width=0.09\textwidth]{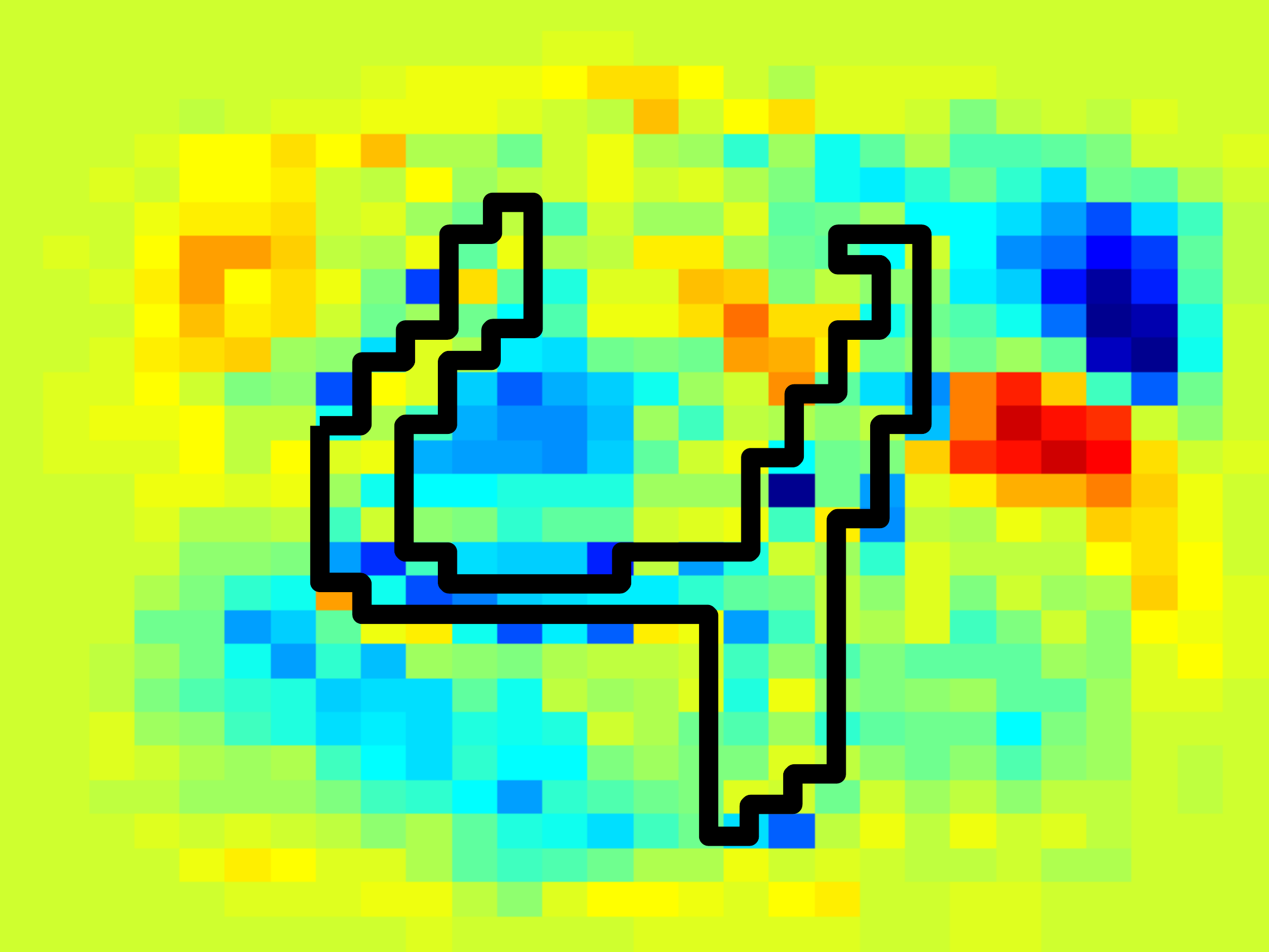}}&
		\subcaptionbox*{}{\includegraphics[width=0.09\textwidth]{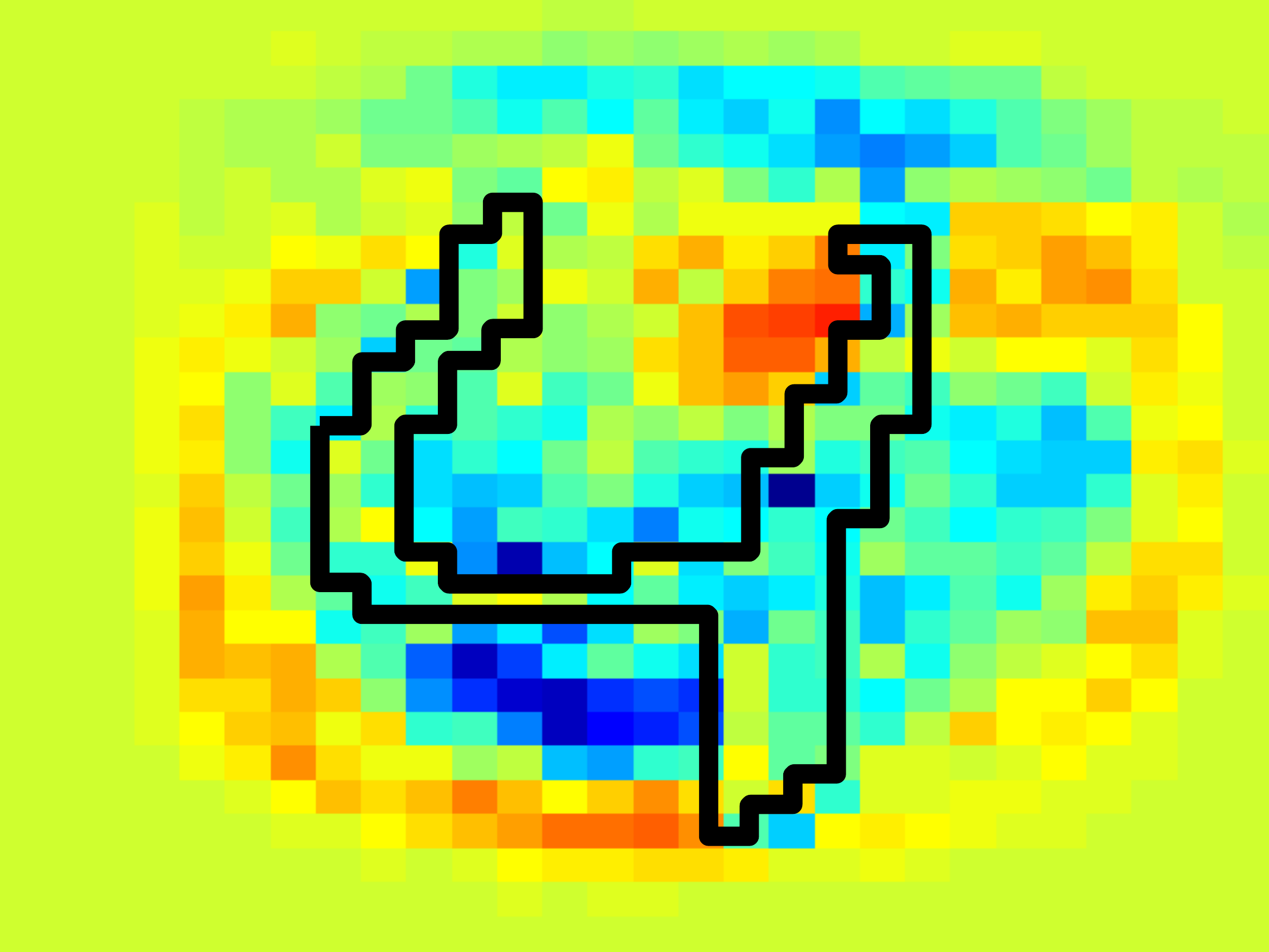}}&
		\subcaptionbox*{}{\includegraphics[width=0.09\textwidth]{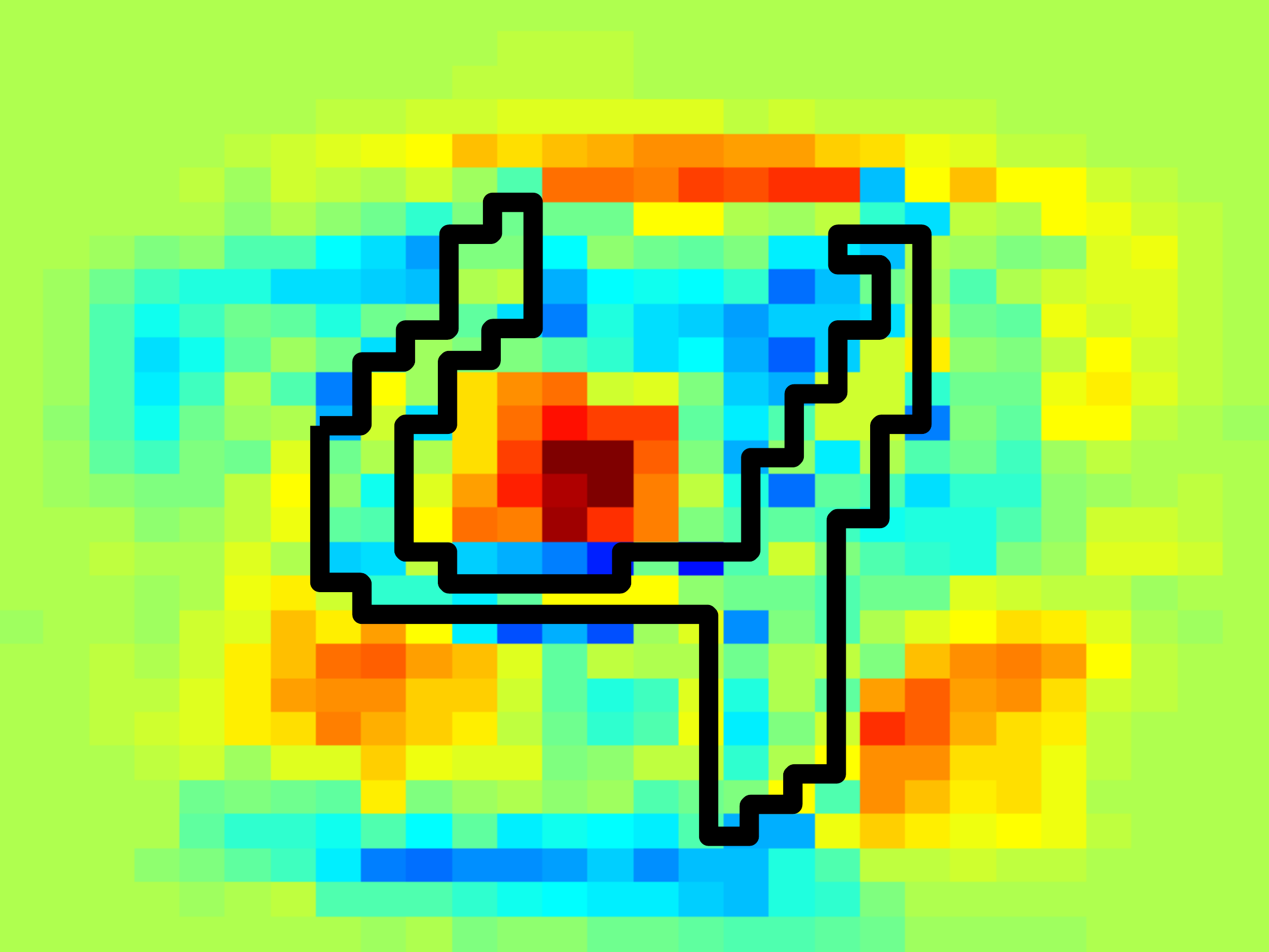}}&
		\subcaptionbox*{}{\includegraphics[width=0.09\textwidth]{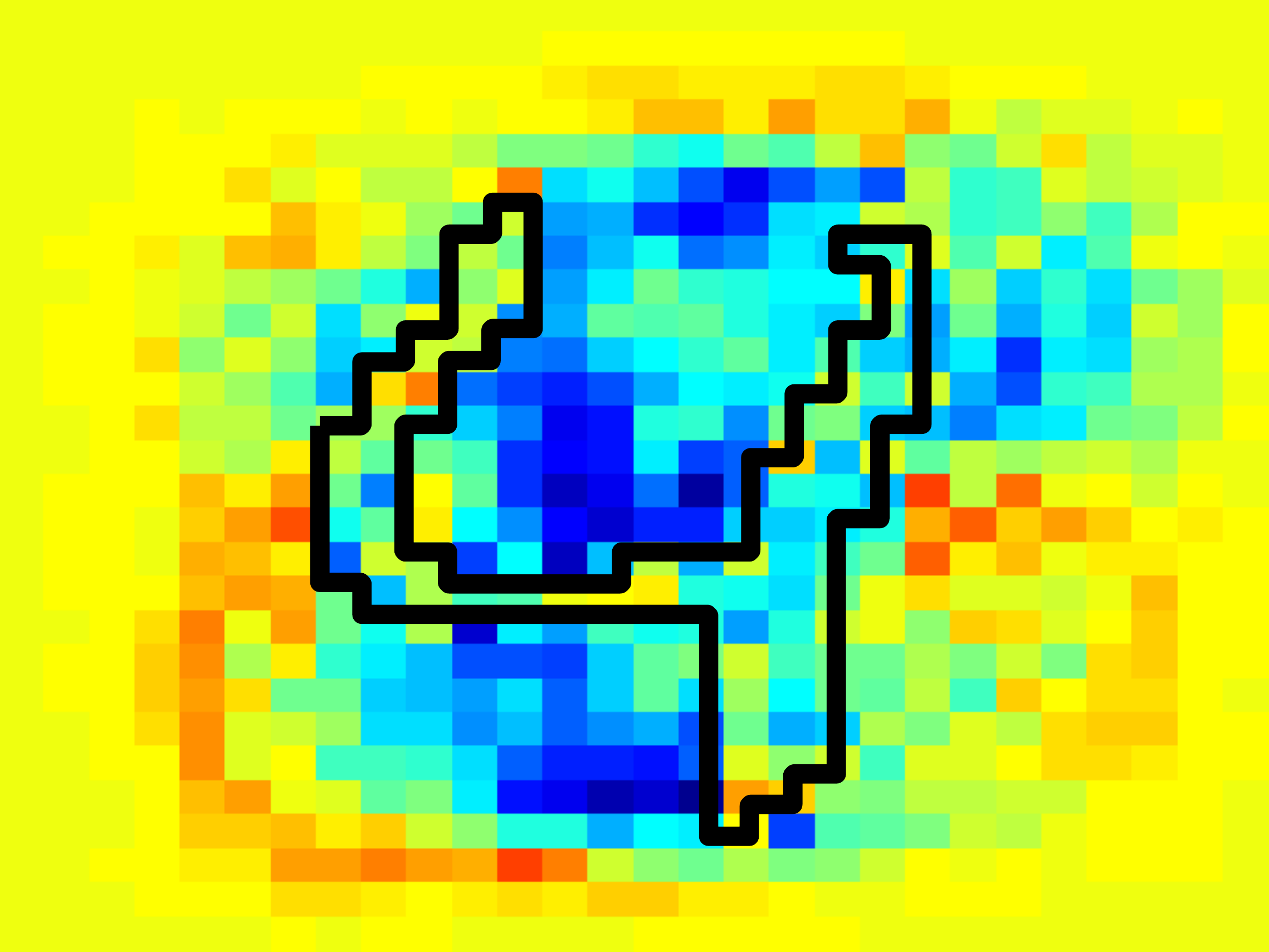}}&
		\subcaptionbox*{}{\includegraphics[width=0.09\textwidth]{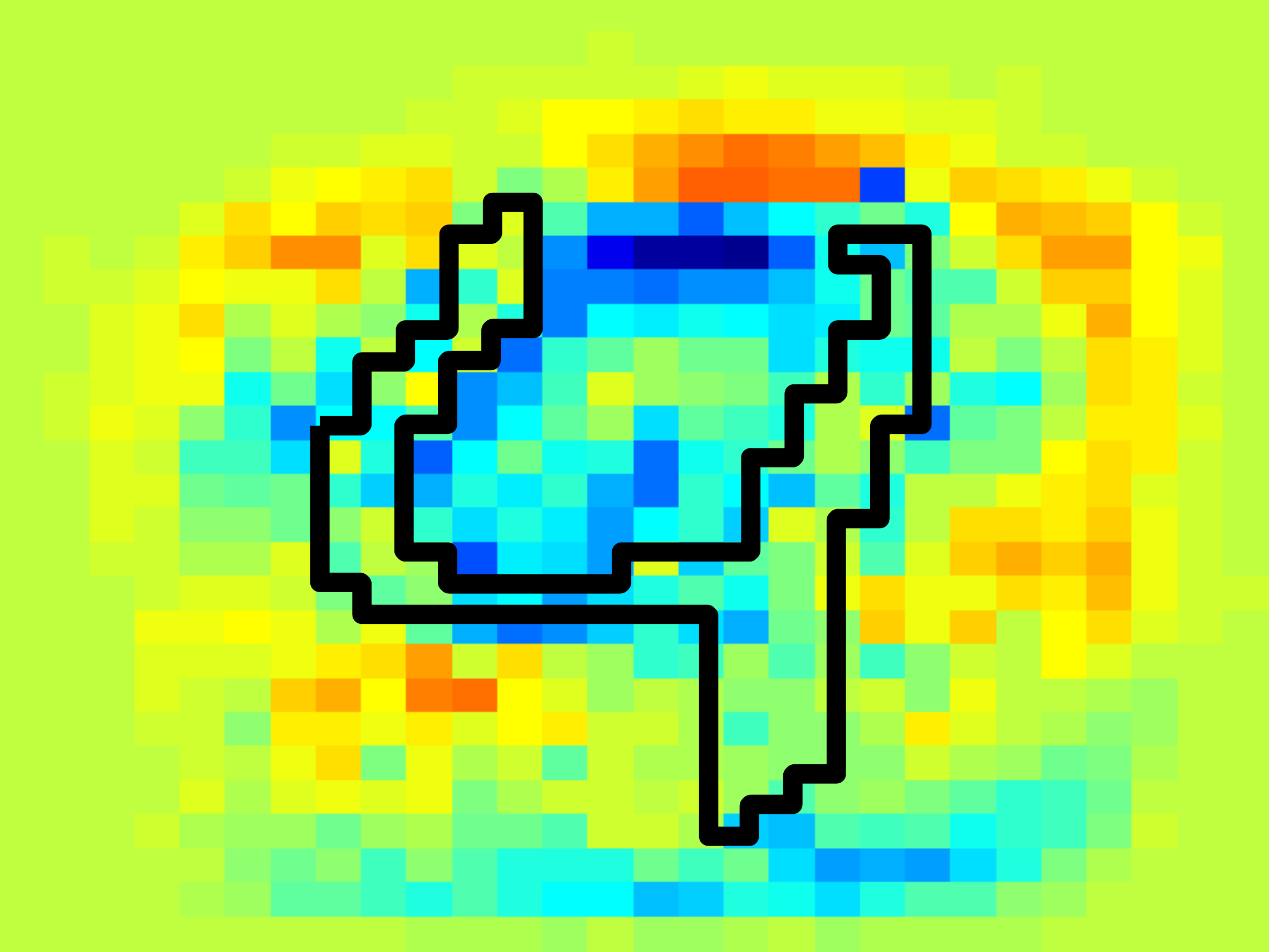}}
		\\[-1em]
		&0.74&0.74&0.79&0.76&1.06&0.71&0.79&0.77&0.60&0.82\\[0.5em]\hline

		\subcaptionbox*{}{\includegraphics[width=0.09\textwidth]{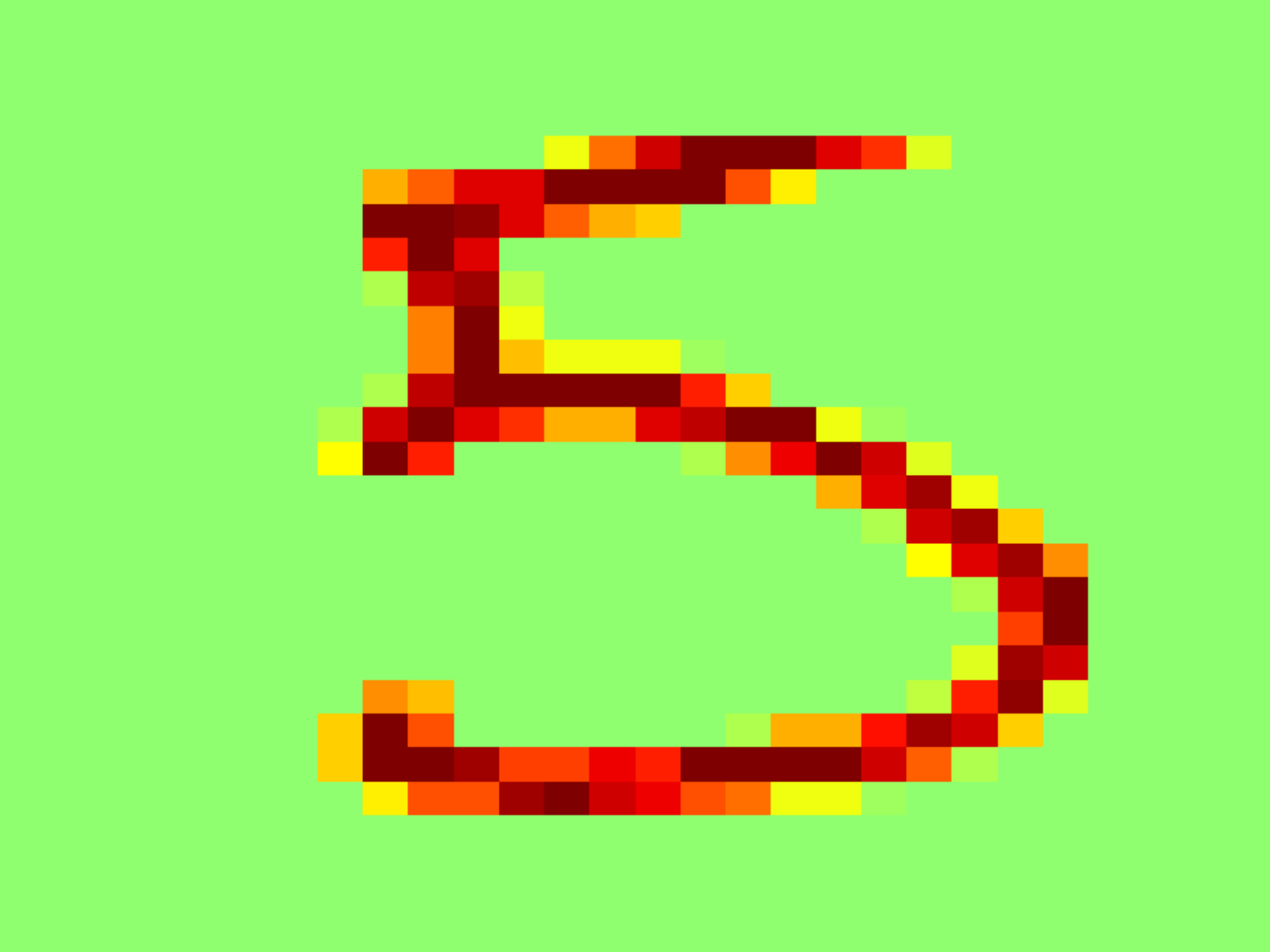}}&
		\subcaptionbox*{}{\includegraphics[width=0.09\textwidth]{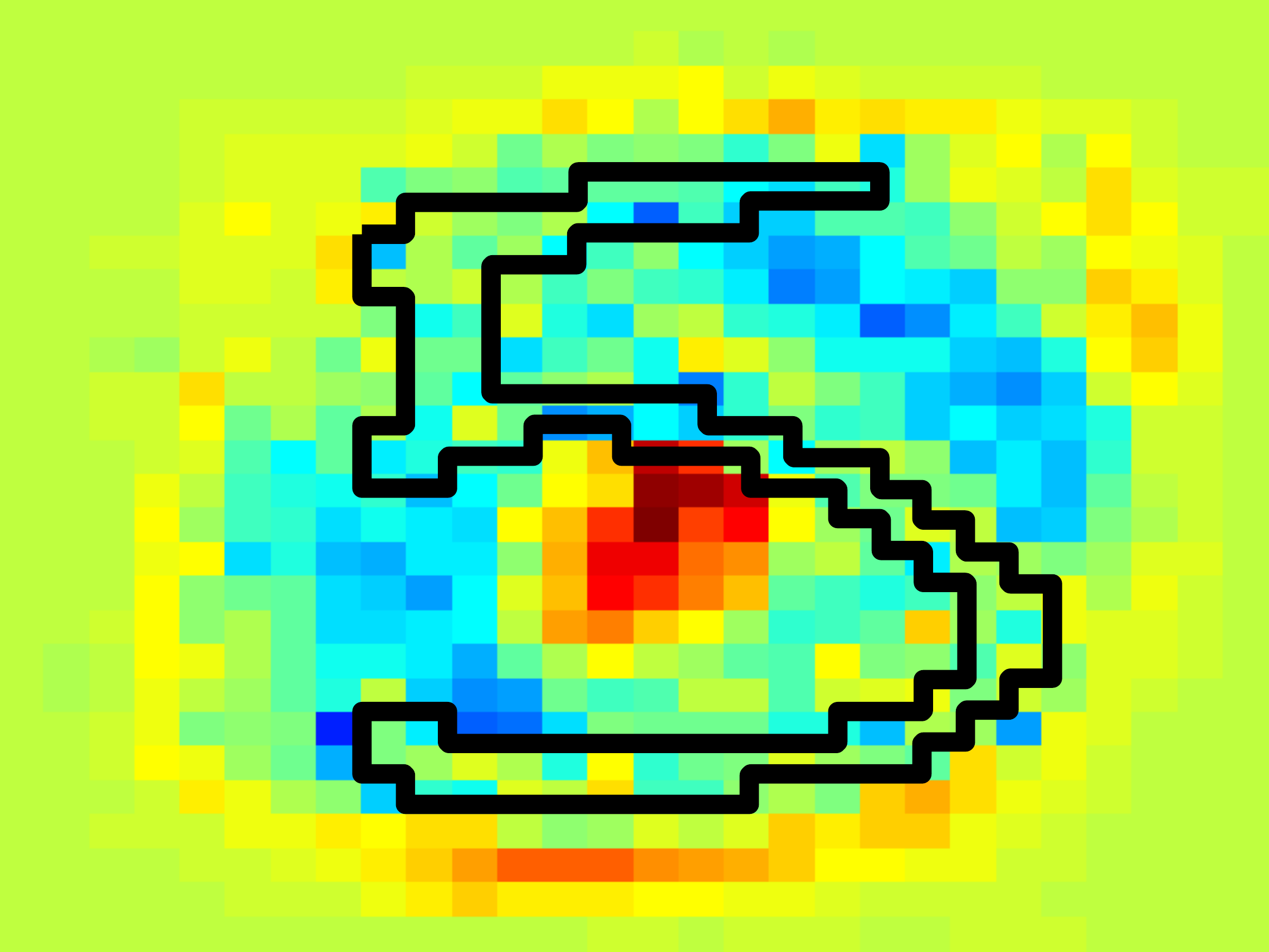}}&
		\subcaptionbox*{}{\includegraphics[width=0.09\textwidth]{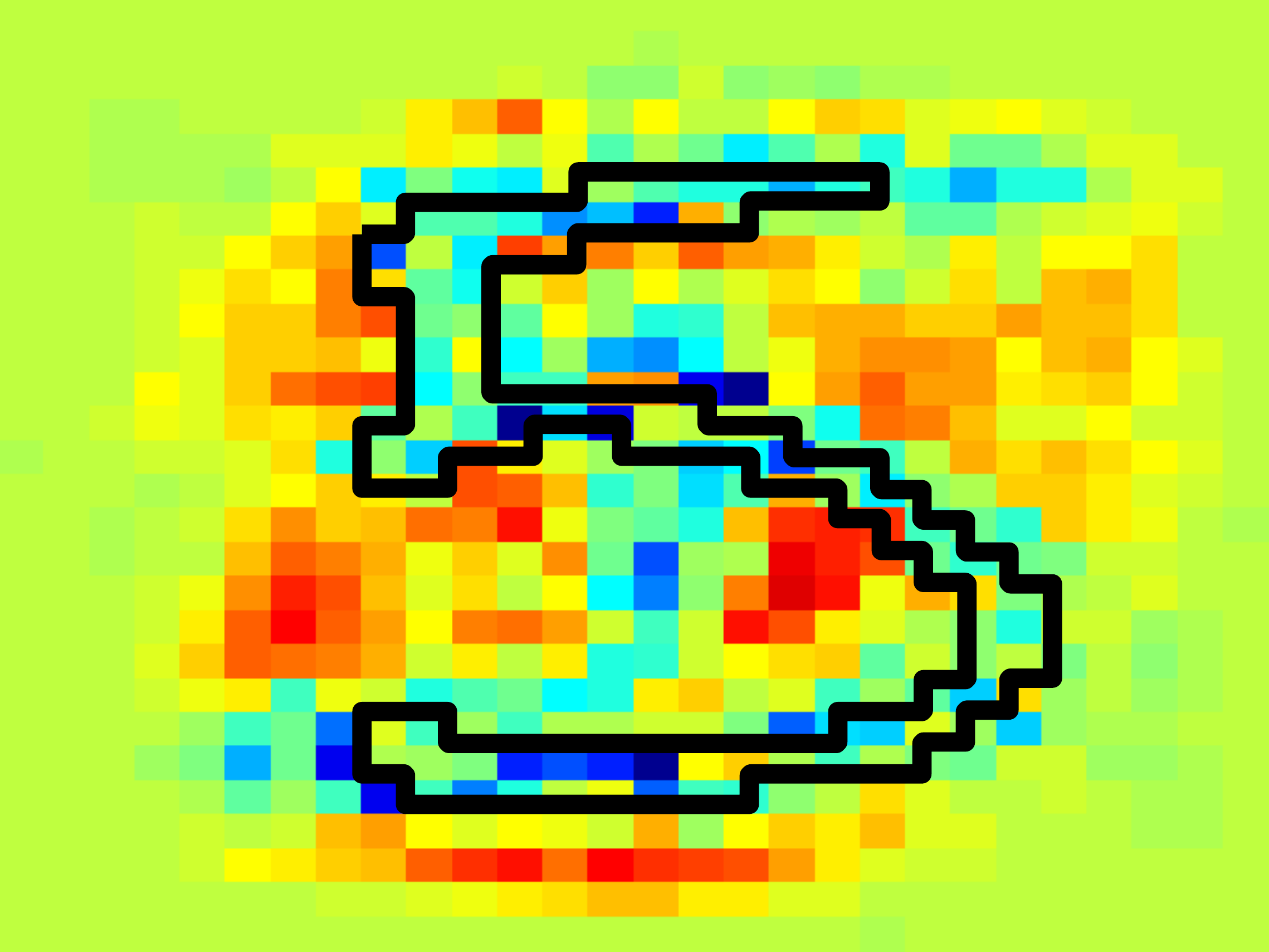}}&
		\subcaptionbox*{}{\includegraphics[width=0.09\textwidth]{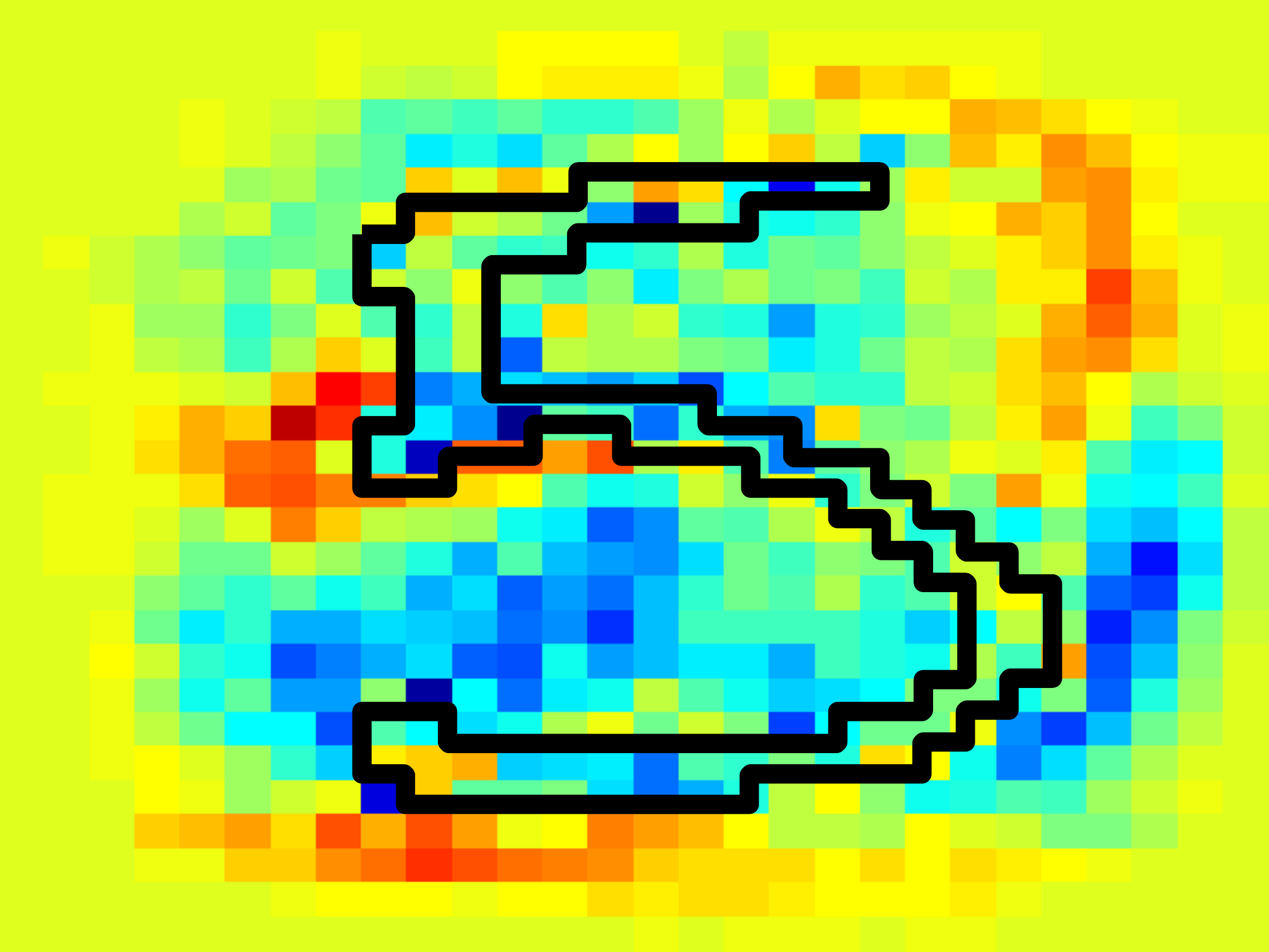}}&
		\subcaptionbox*{}{\includegraphics[width=0.09\textwidth]{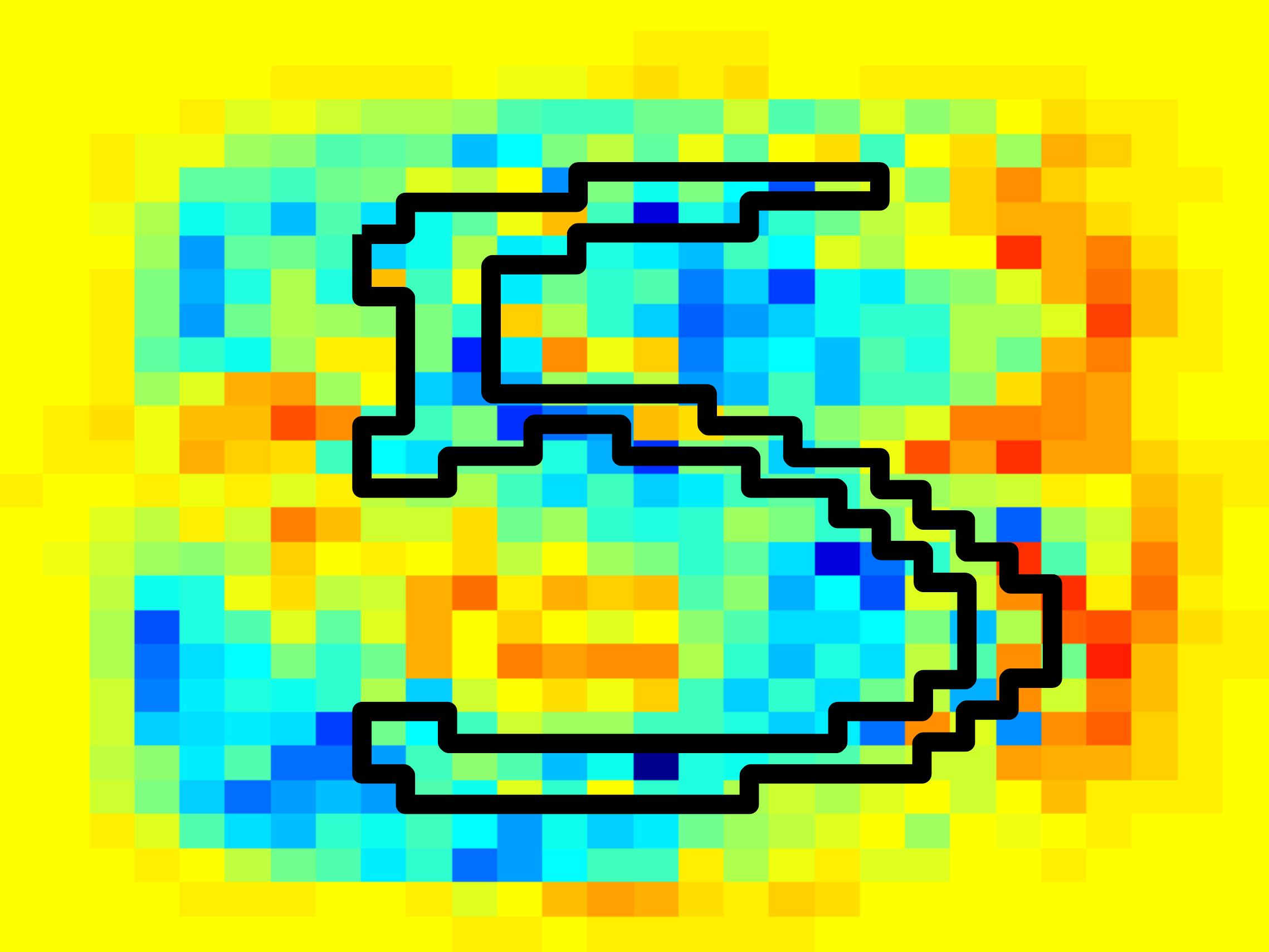}}&
		\subcaptionbox*{}{\includegraphics[width=0.09\textwidth]{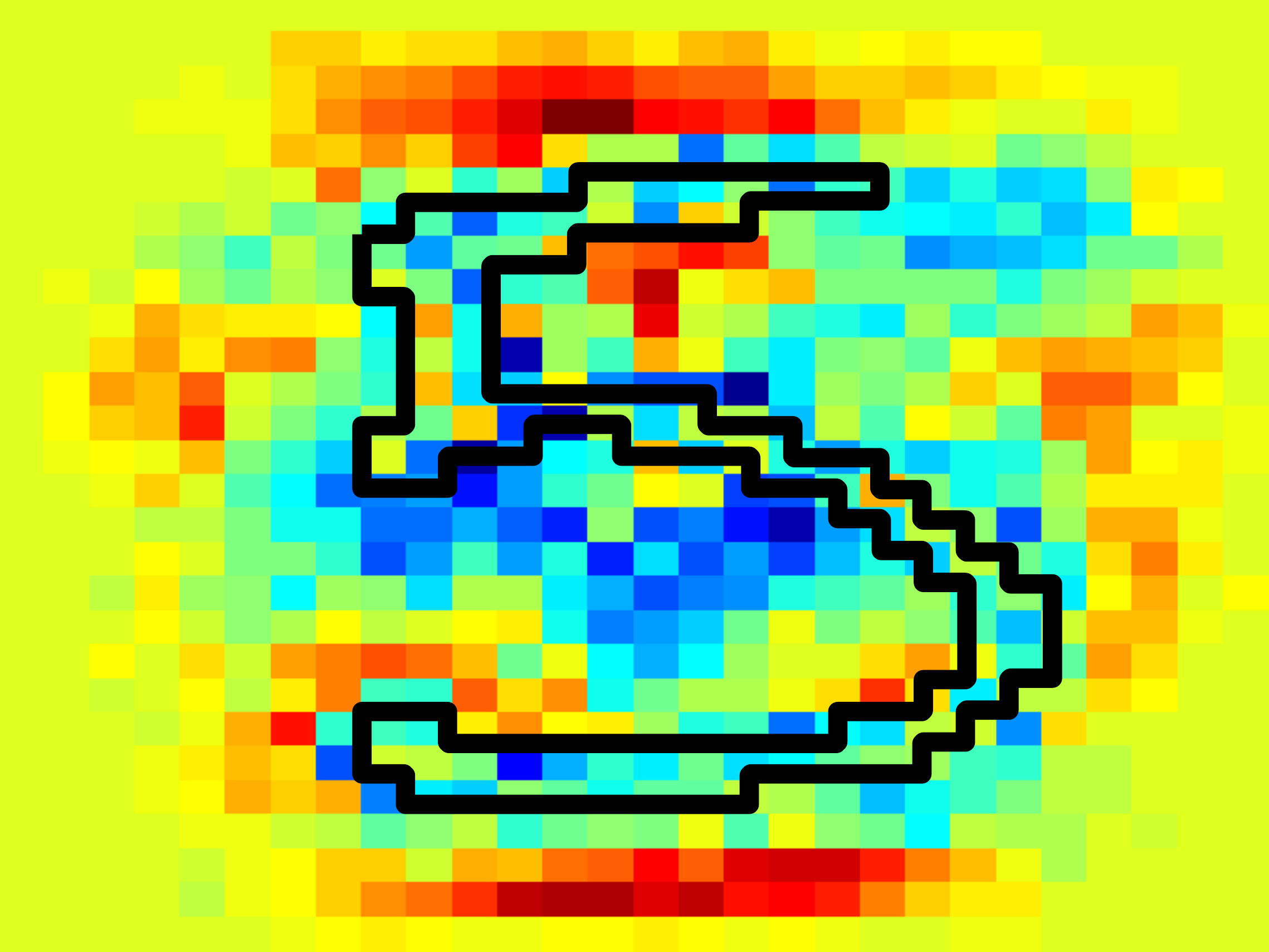}}&
		\subcaptionbox*{}{\includegraphics[width=0.09\textwidth]{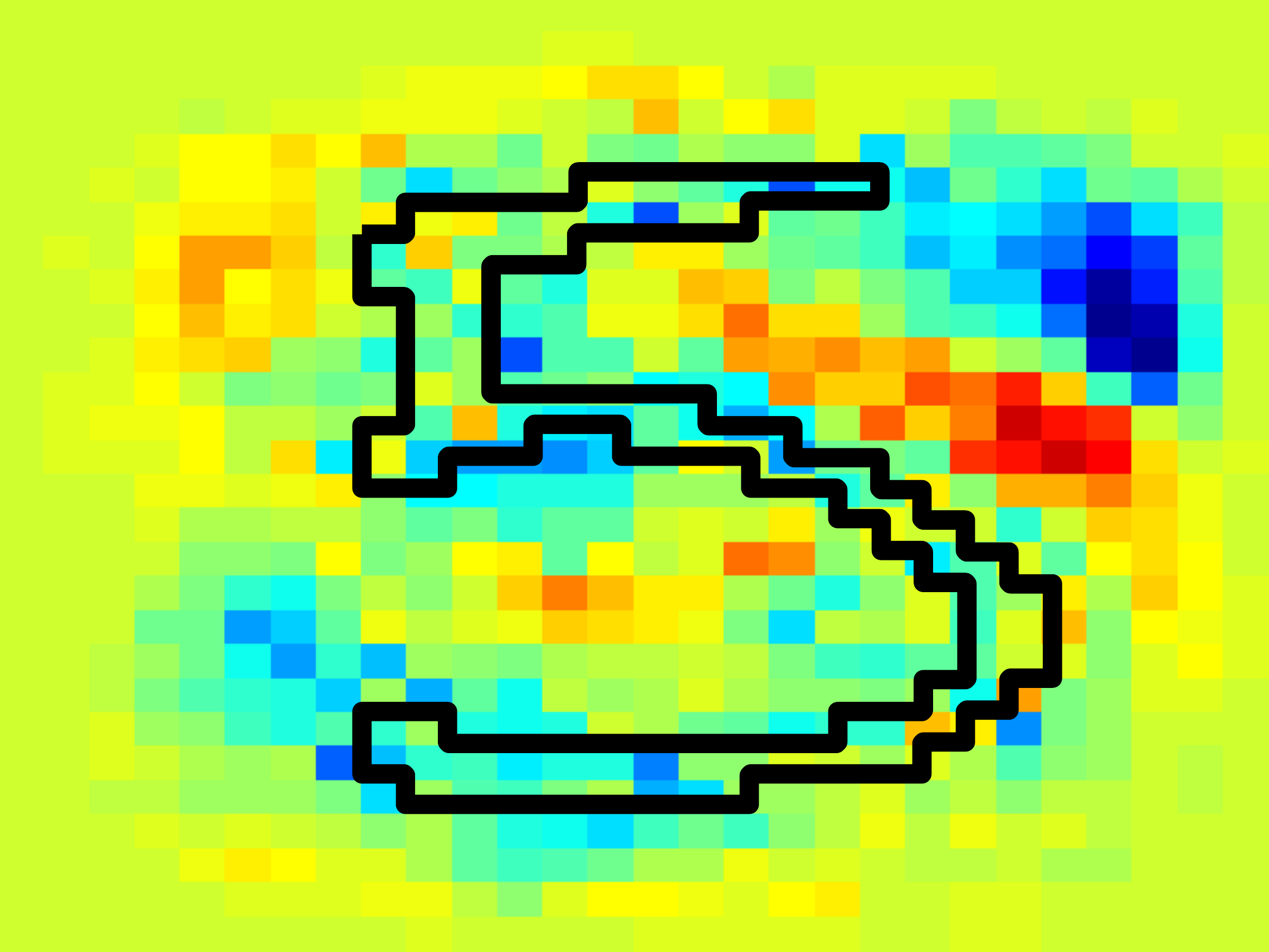}}&
		\subcaptionbox*{}{\includegraphics[width=0.09\textwidth]{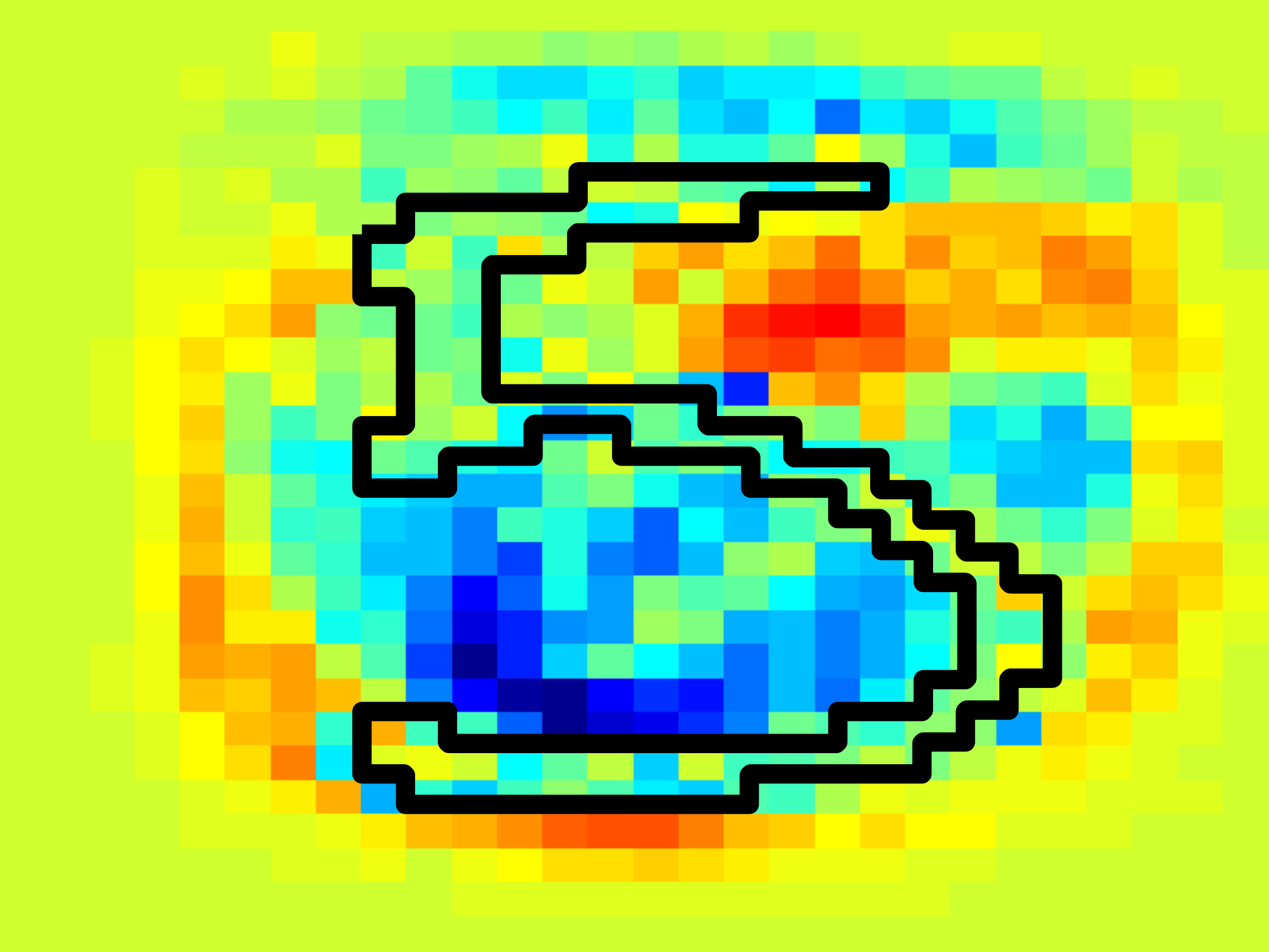}}&
		\subcaptionbox*{}{\includegraphics[width=0.09\textwidth]{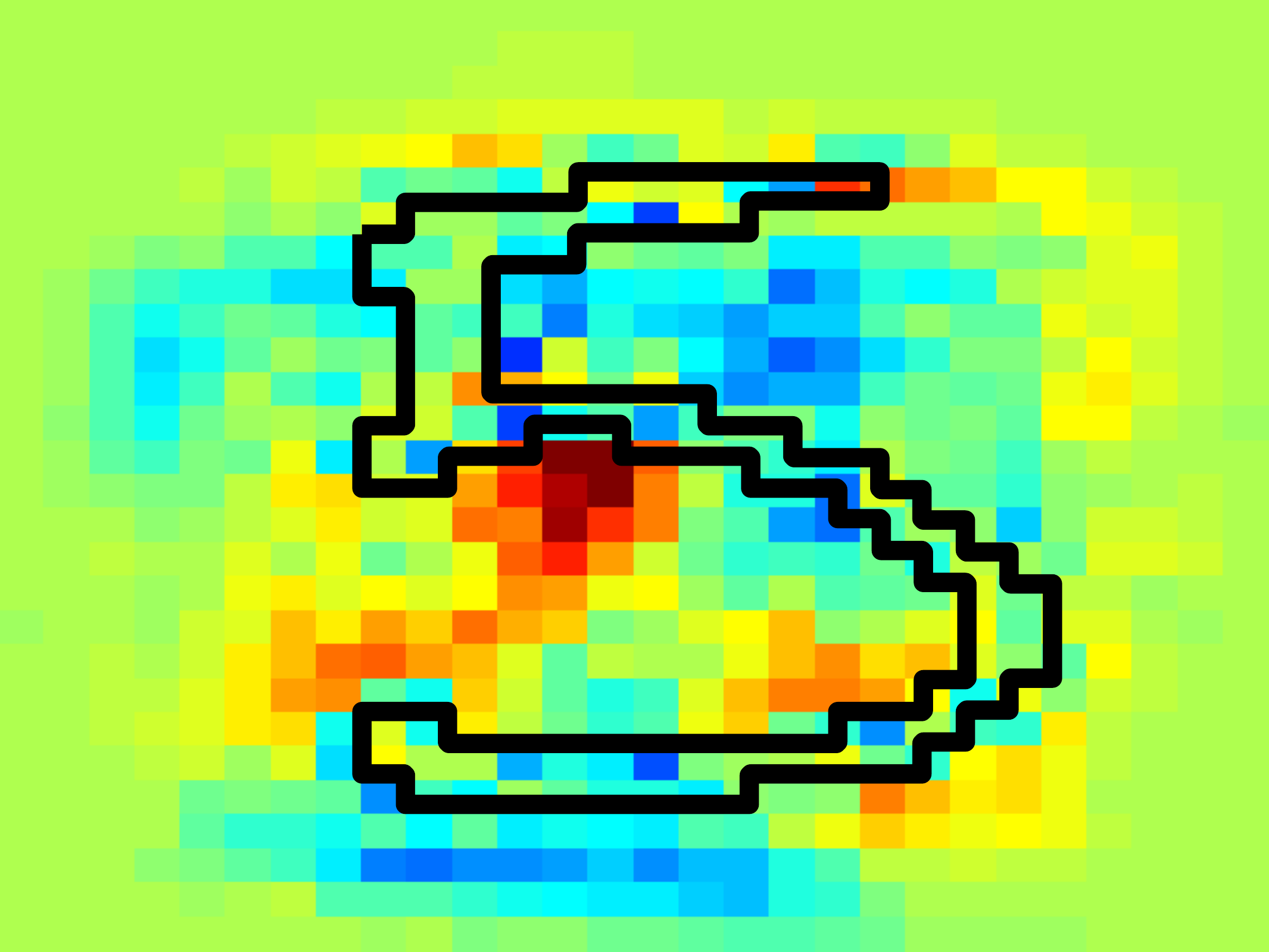}}&
		\subcaptionbox*{}{\includegraphics[width=0.09\textwidth]{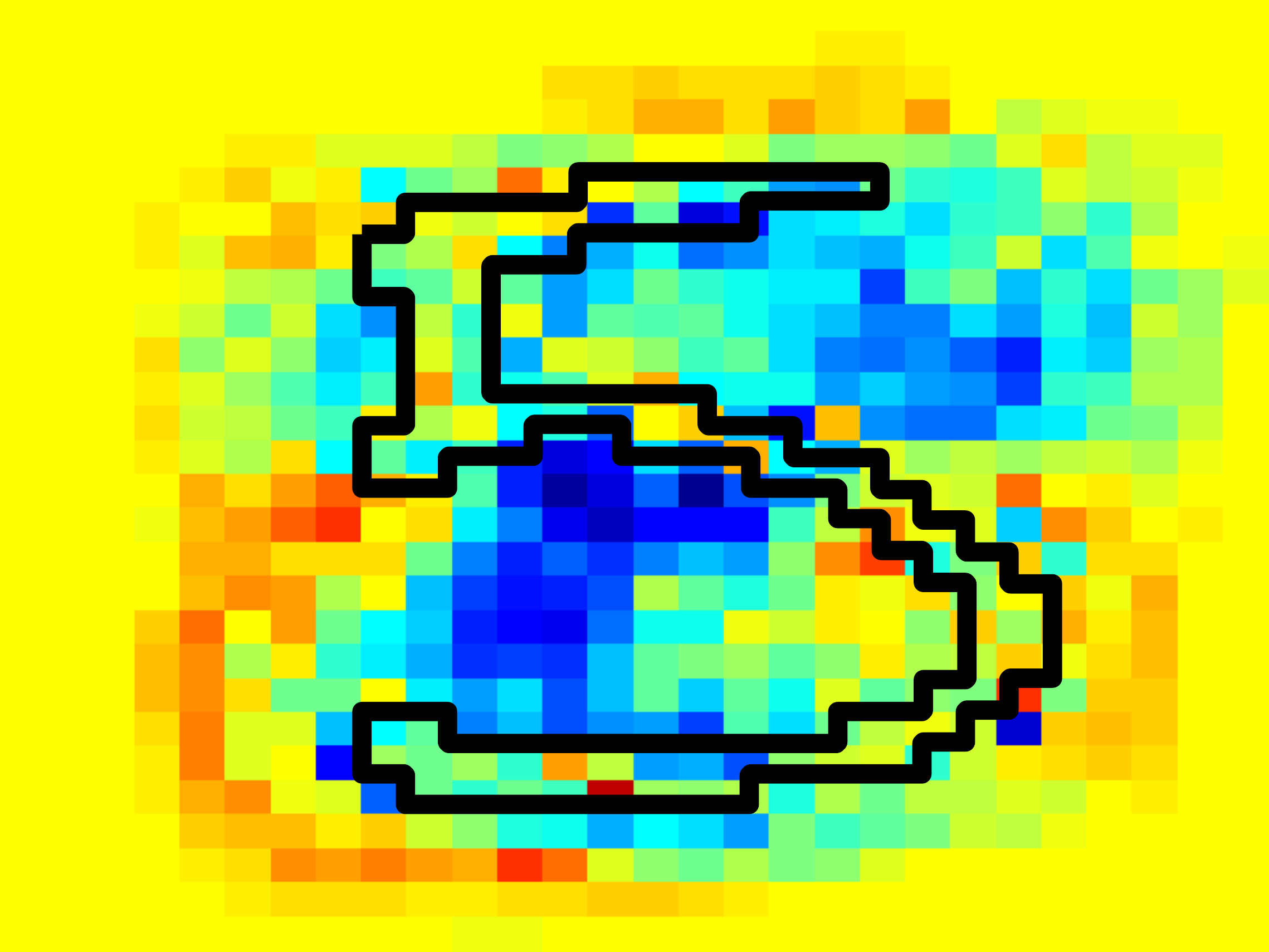}}&
		\subcaptionbox*{}{\includegraphics[width=0.09\textwidth]{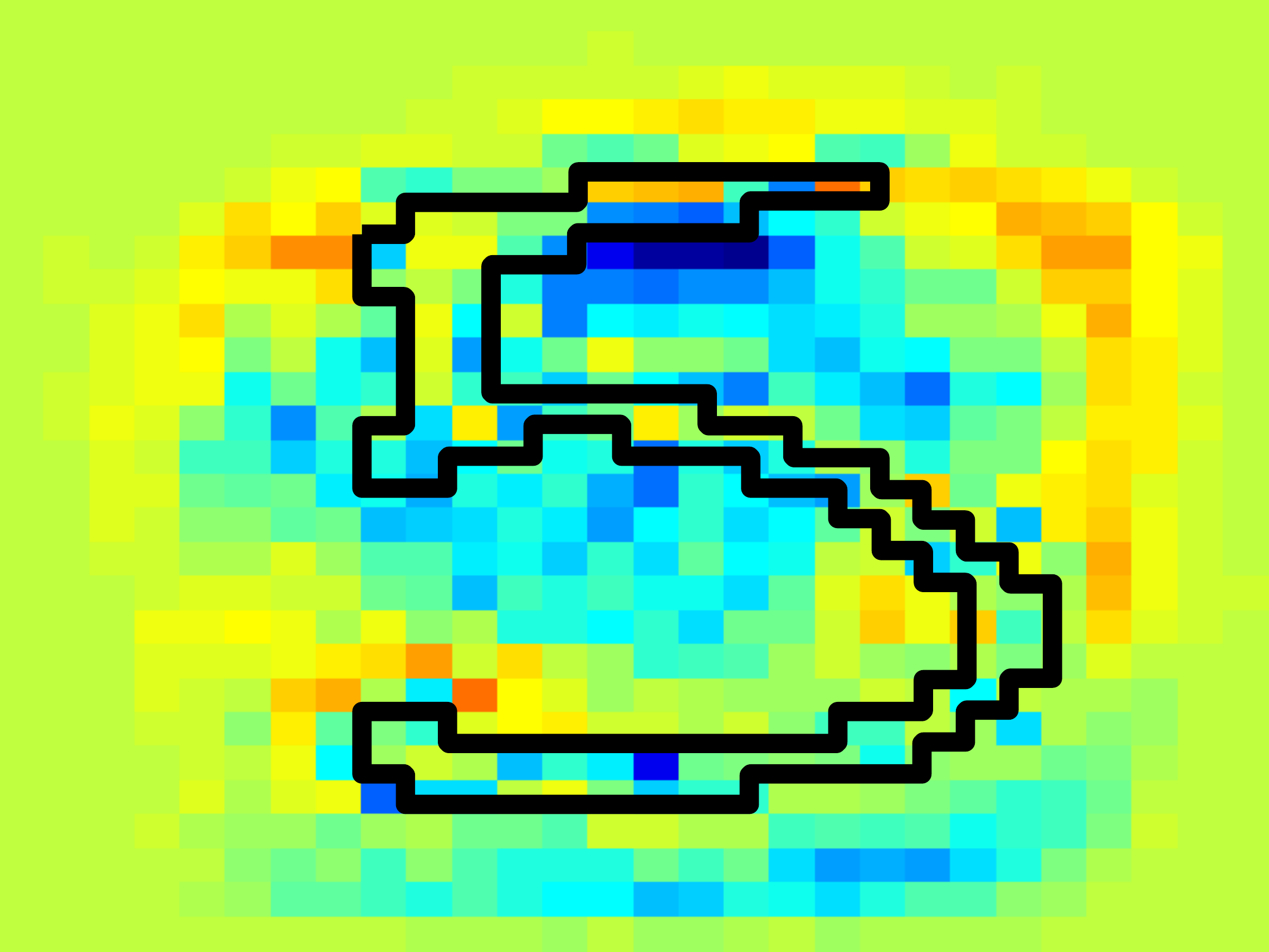}}
		\\[-1em]
		&0.77&0.84&0.69&0.94&0.84&1.08&0.67&0.59&0.82&0.43\\[0.5em]\hline

		\subcaptionbox*{}{\includegraphics[width=0.09\textwidth]{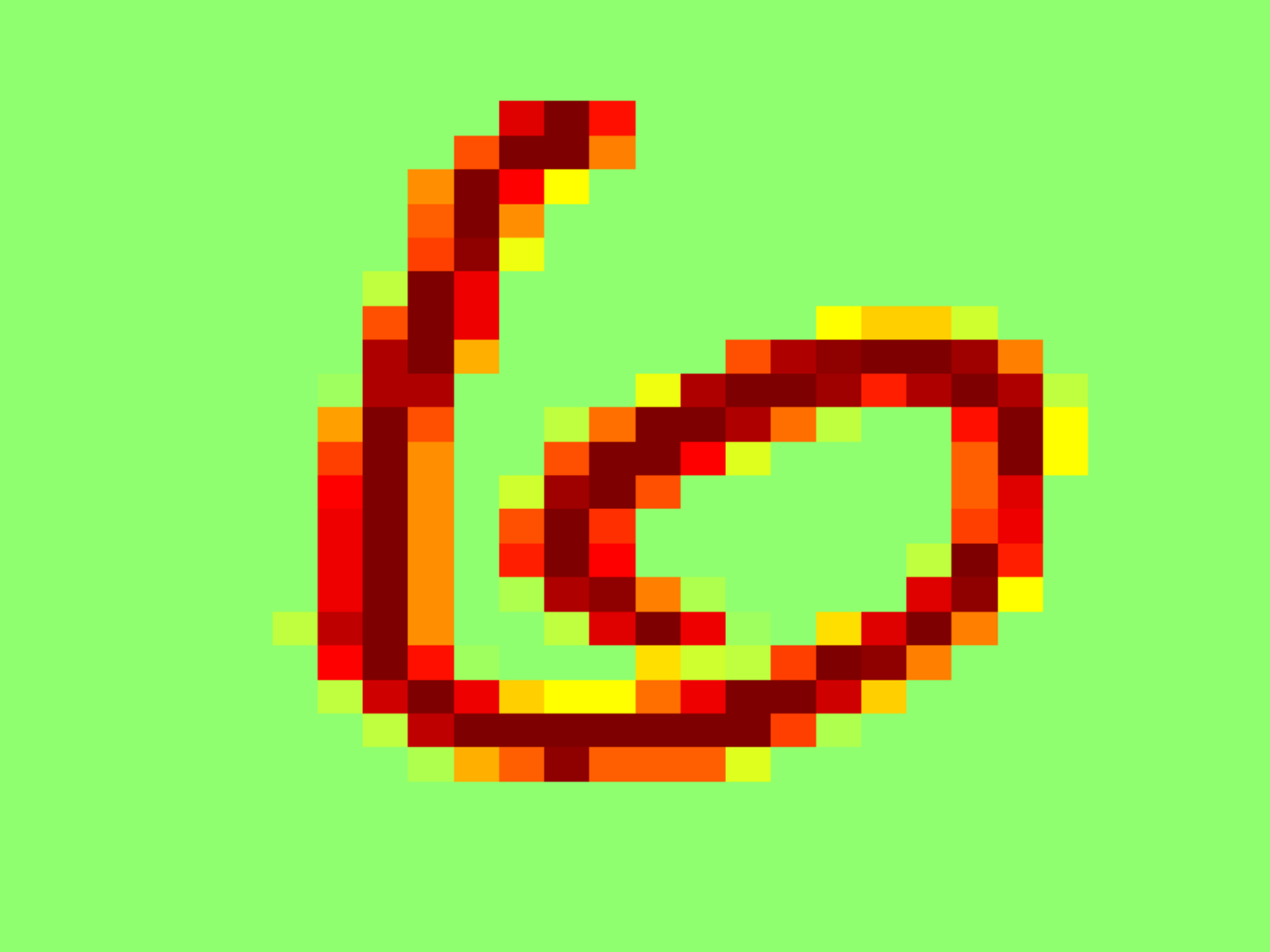}}&
		\subcaptionbox*{}{\includegraphics[width=0.09\textwidth]{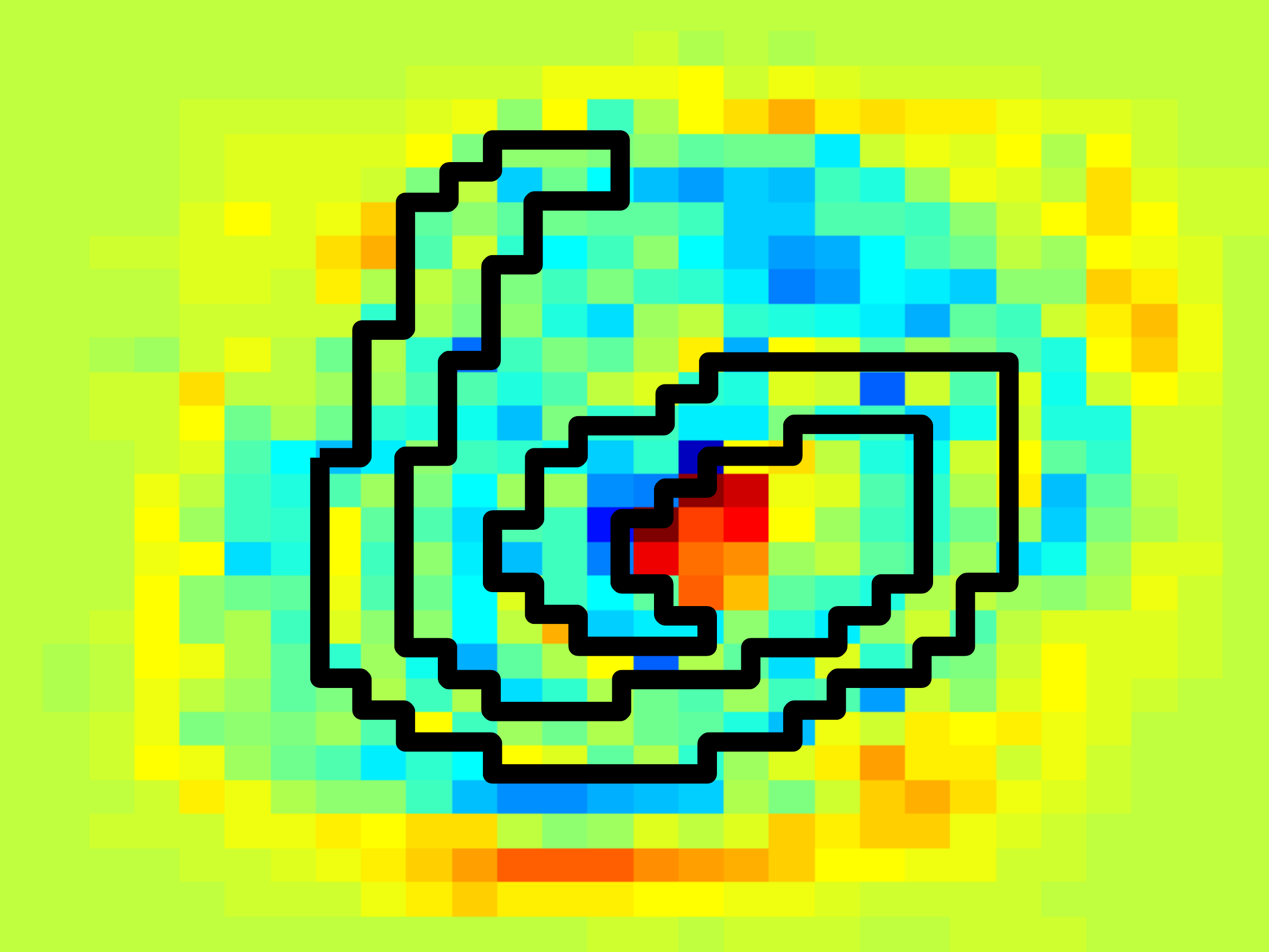}}&
		\subcaptionbox*{}{\includegraphics[width=0.09\textwidth]{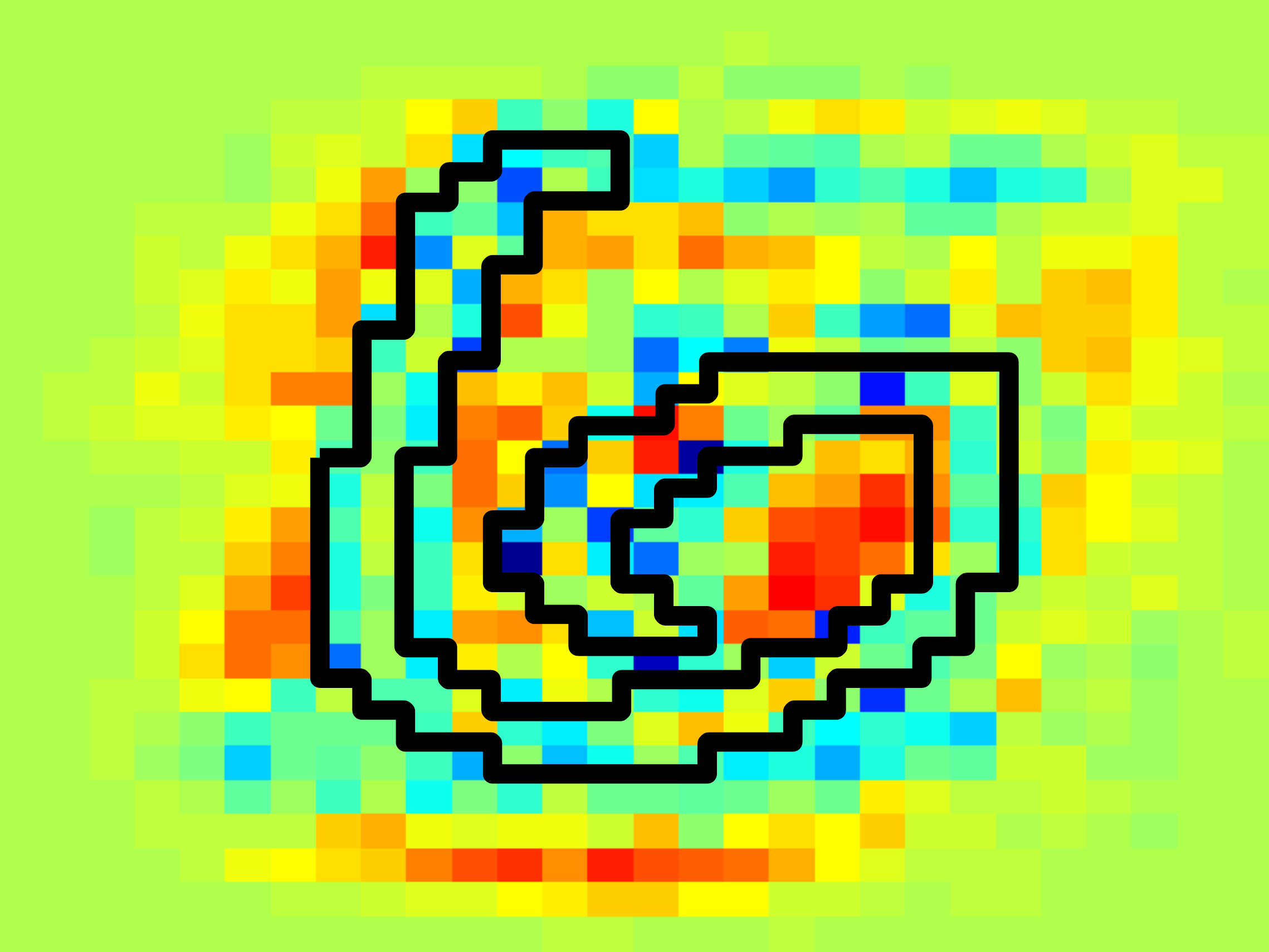}}&
		\subcaptionbox*{}{\includegraphics[width=0.09\textwidth]{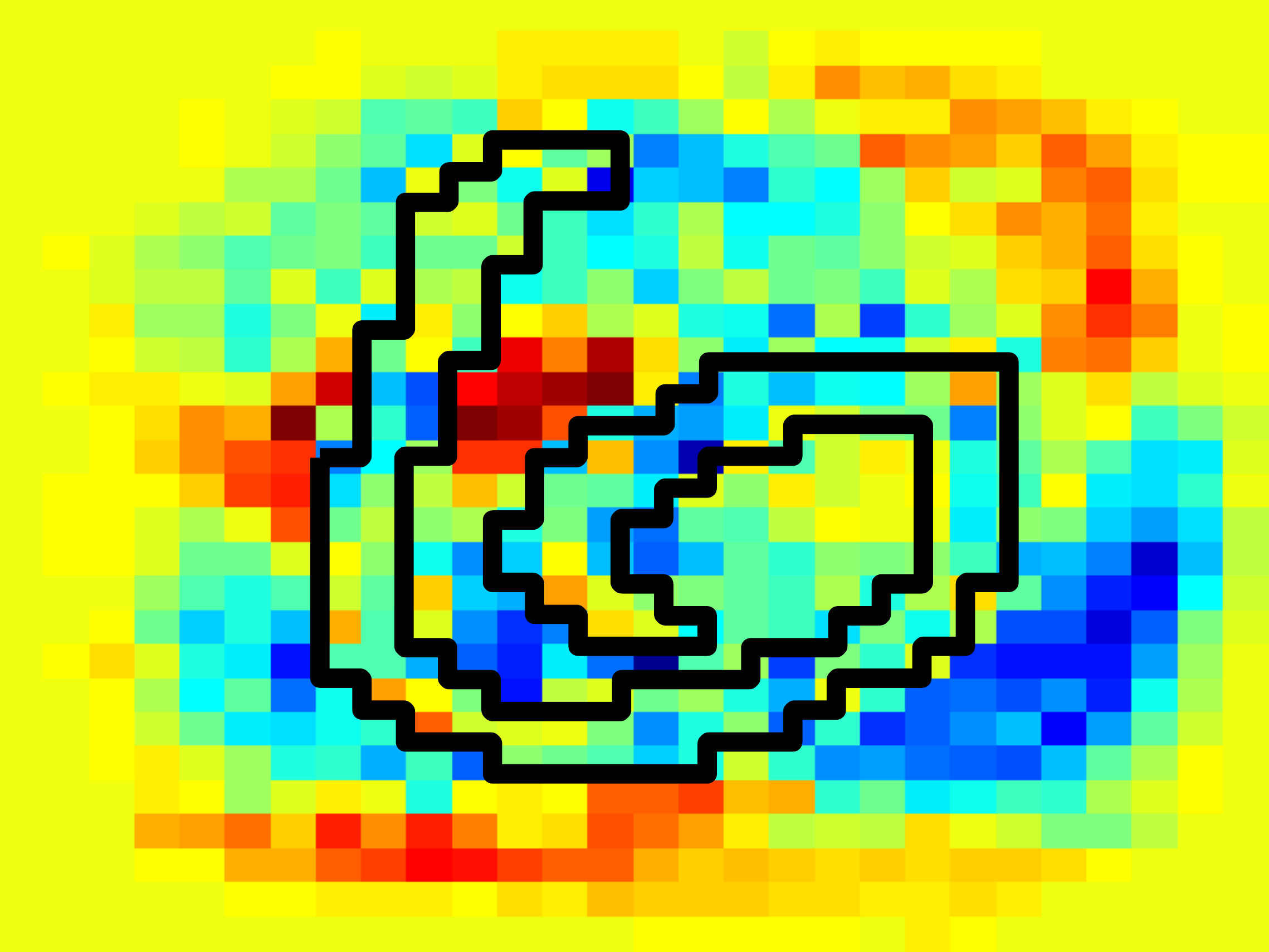}}&
		\subcaptionbox*{}{\includegraphics[width=0.09\textwidth]{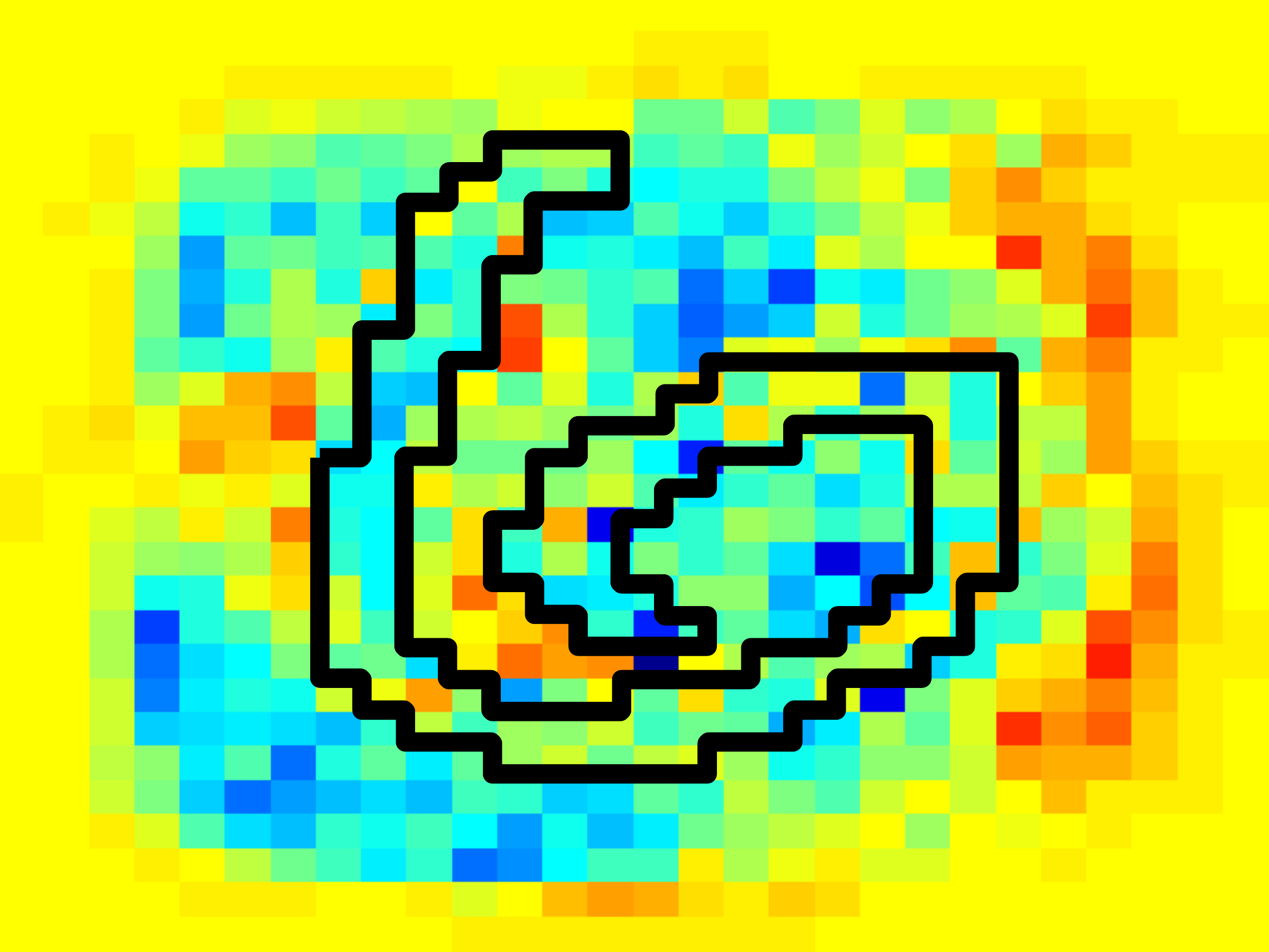}}&
		\subcaptionbox*{}{\includegraphics[width=0.09\textwidth]{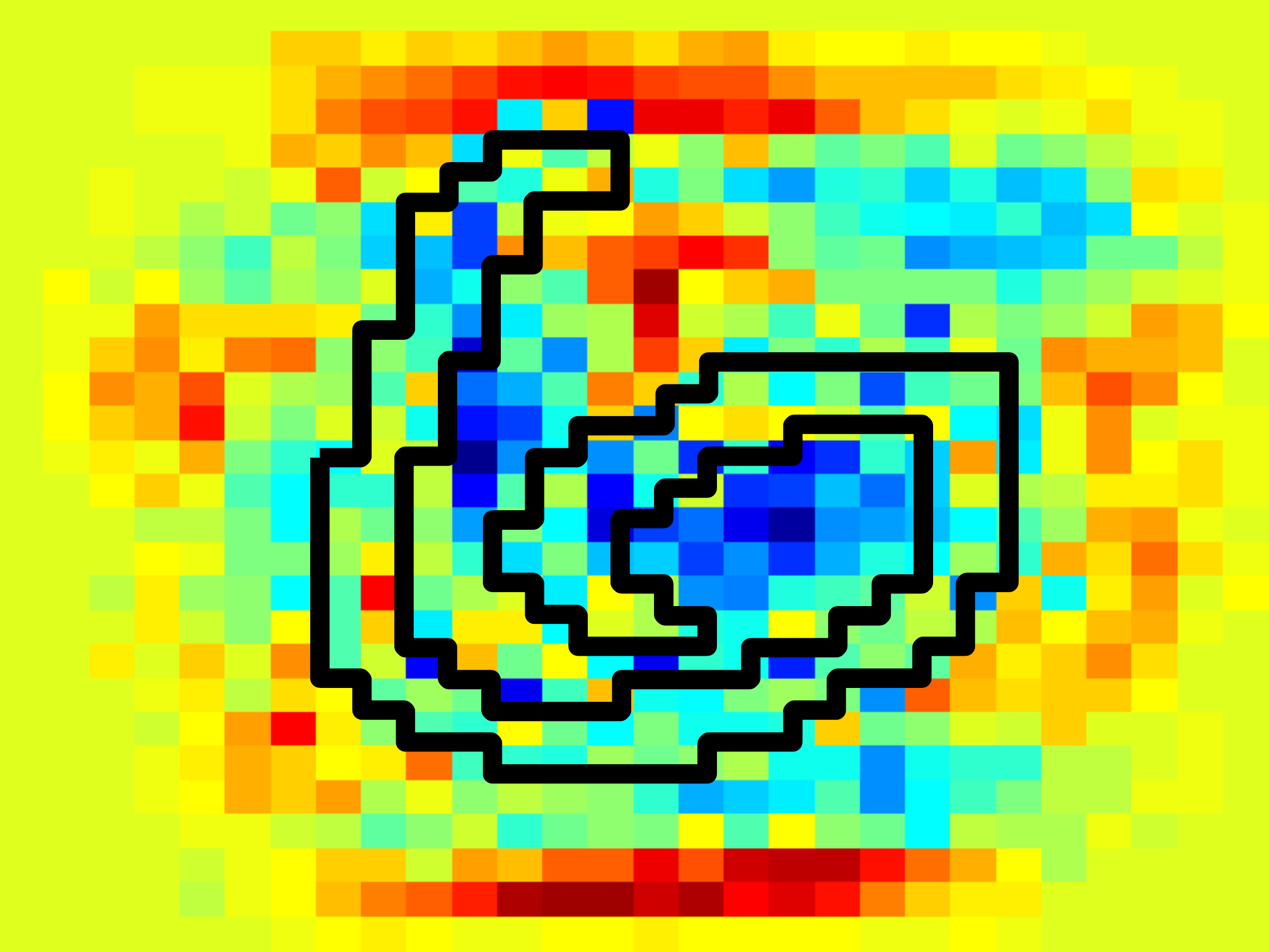}}&
		\subcaptionbox*{}{\includegraphics[width=0.09\textwidth]{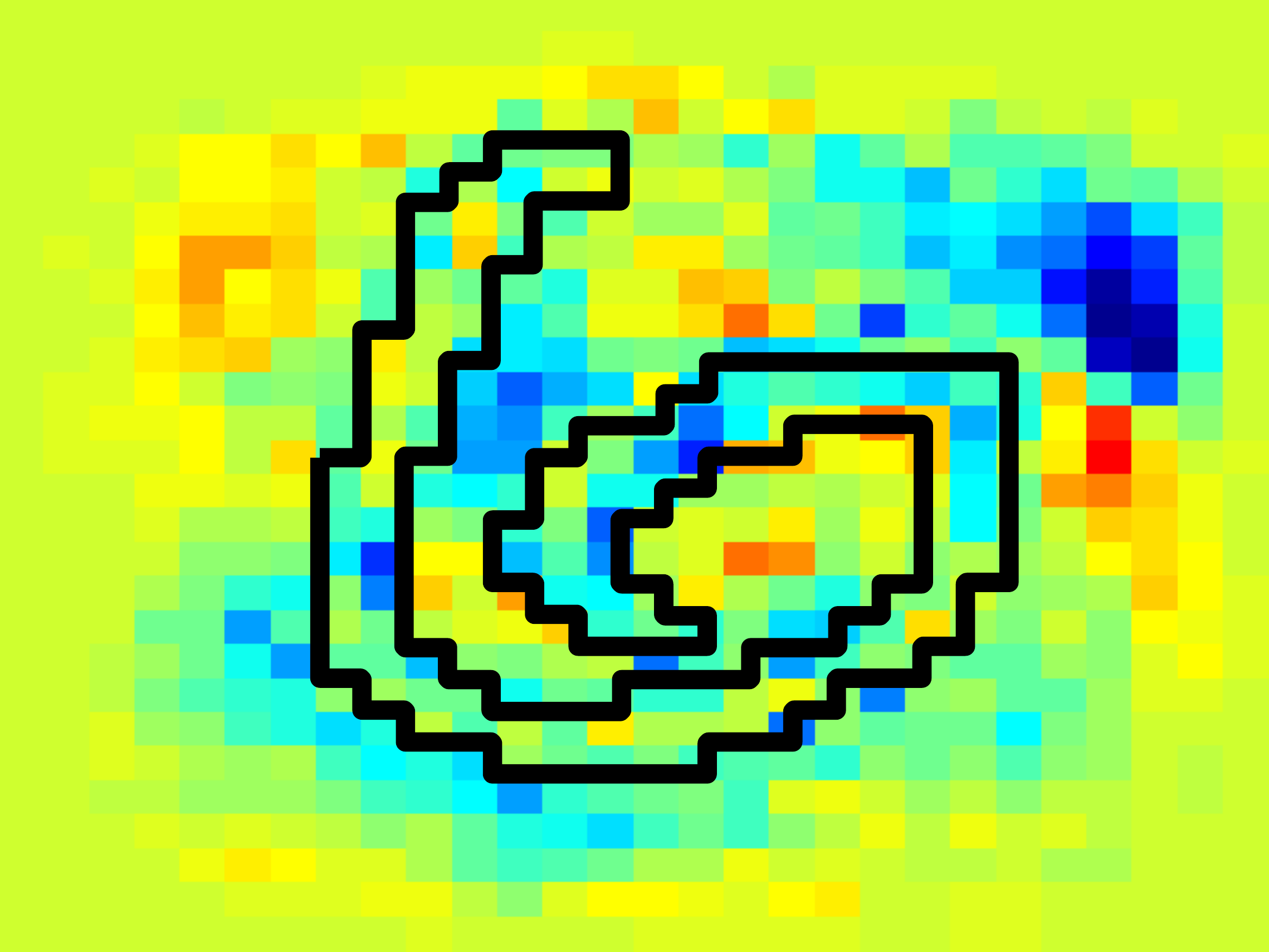}}&
		\subcaptionbox*{}{\includegraphics[width=0.09\textwidth]{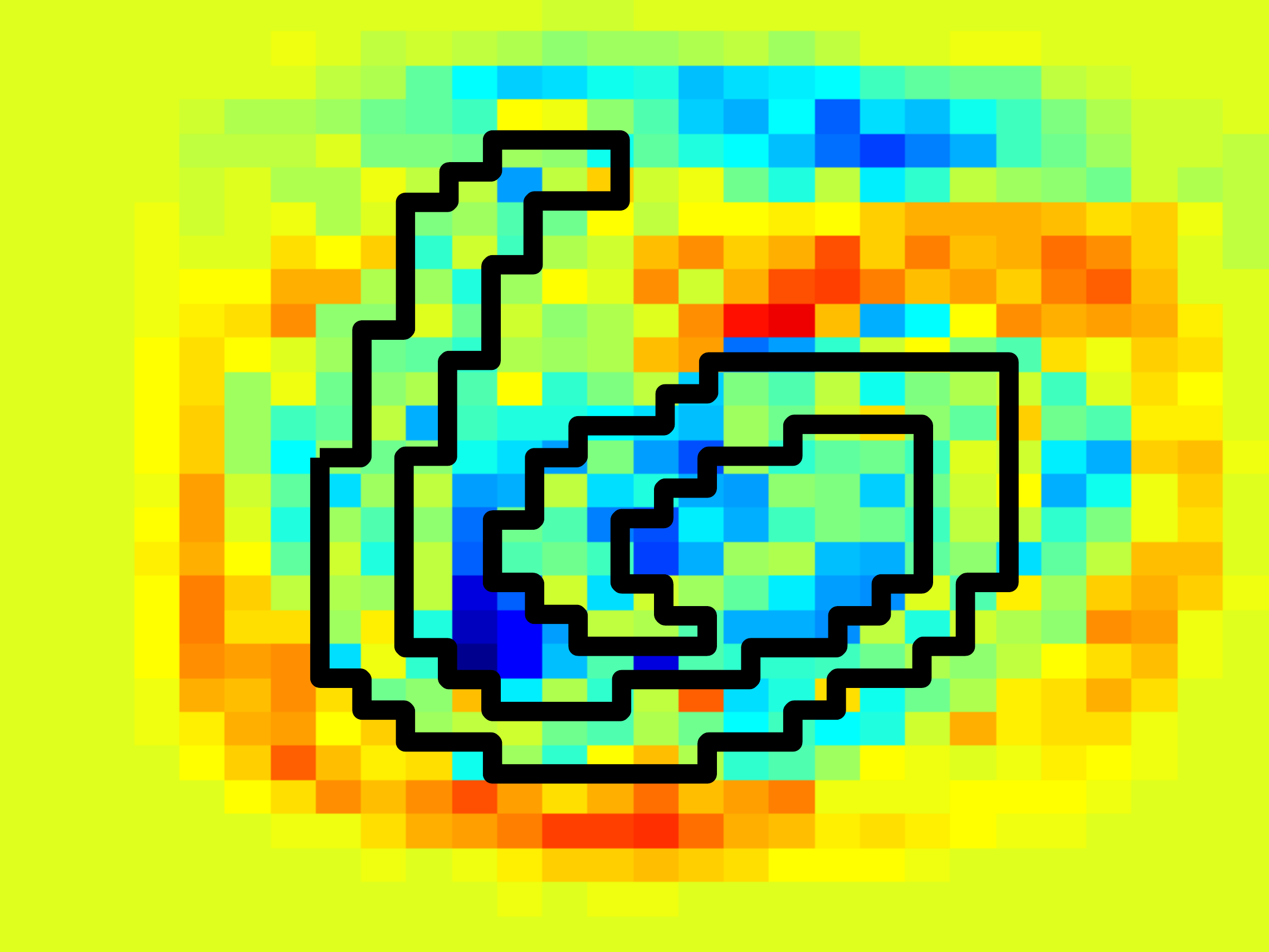}}&
		\subcaptionbox*{}{\includegraphics[width=0.09\textwidth]{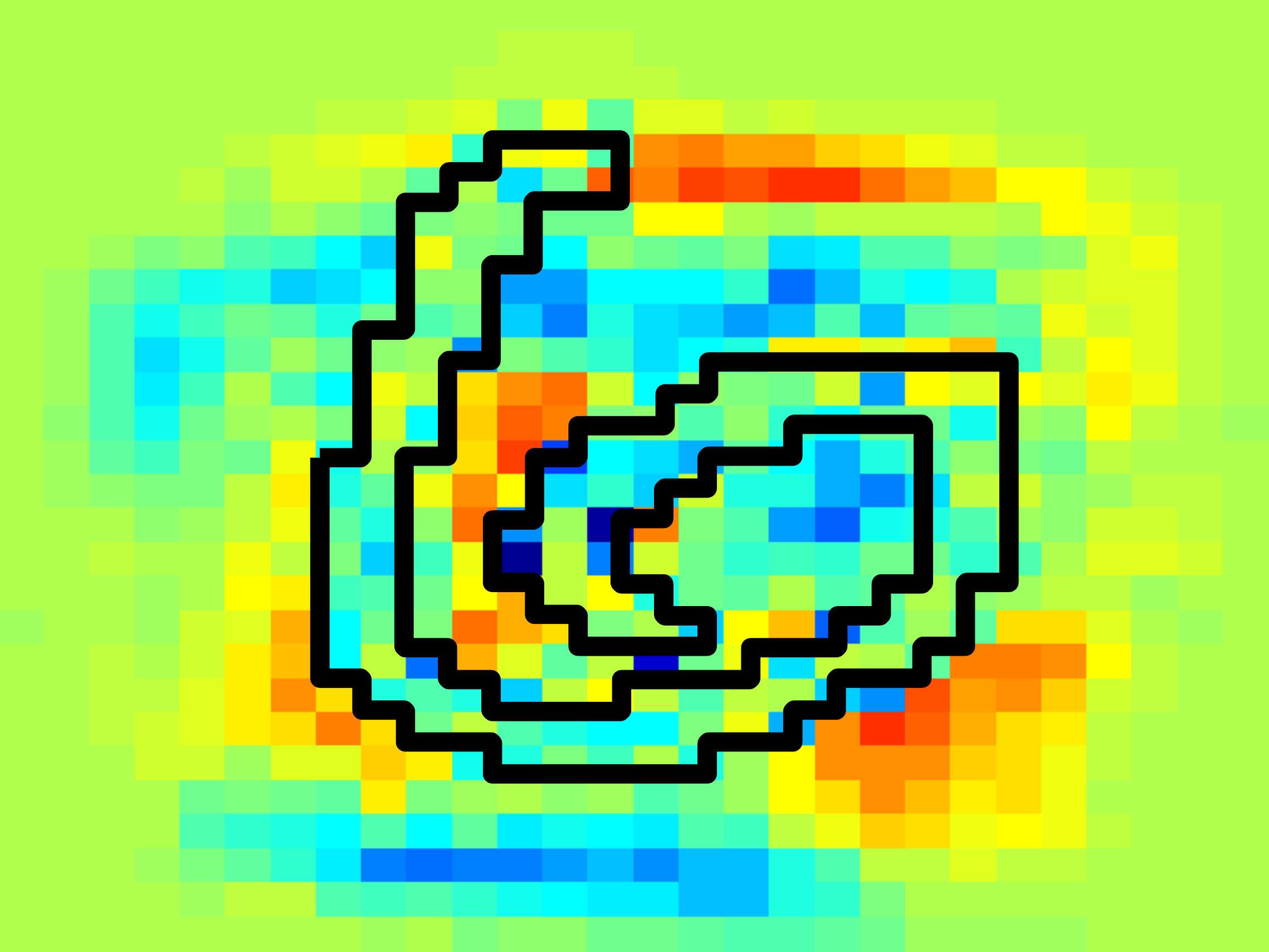}}&
		\subcaptionbox*{}{\includegraphics[width=0.09\textwidth]{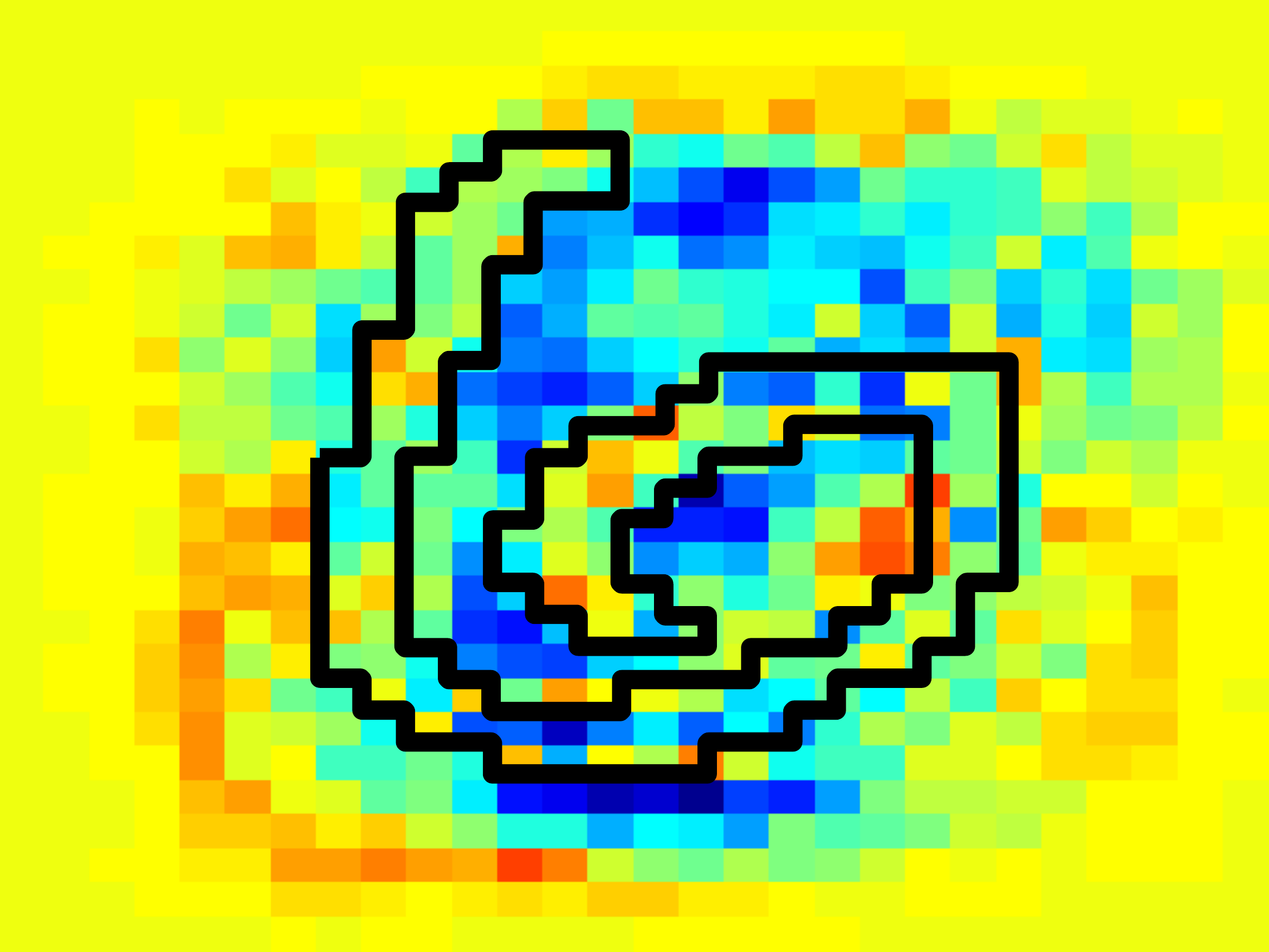}}&
		\subcaptionbox*{}{\includegraphics[width=0.09\textwidth]{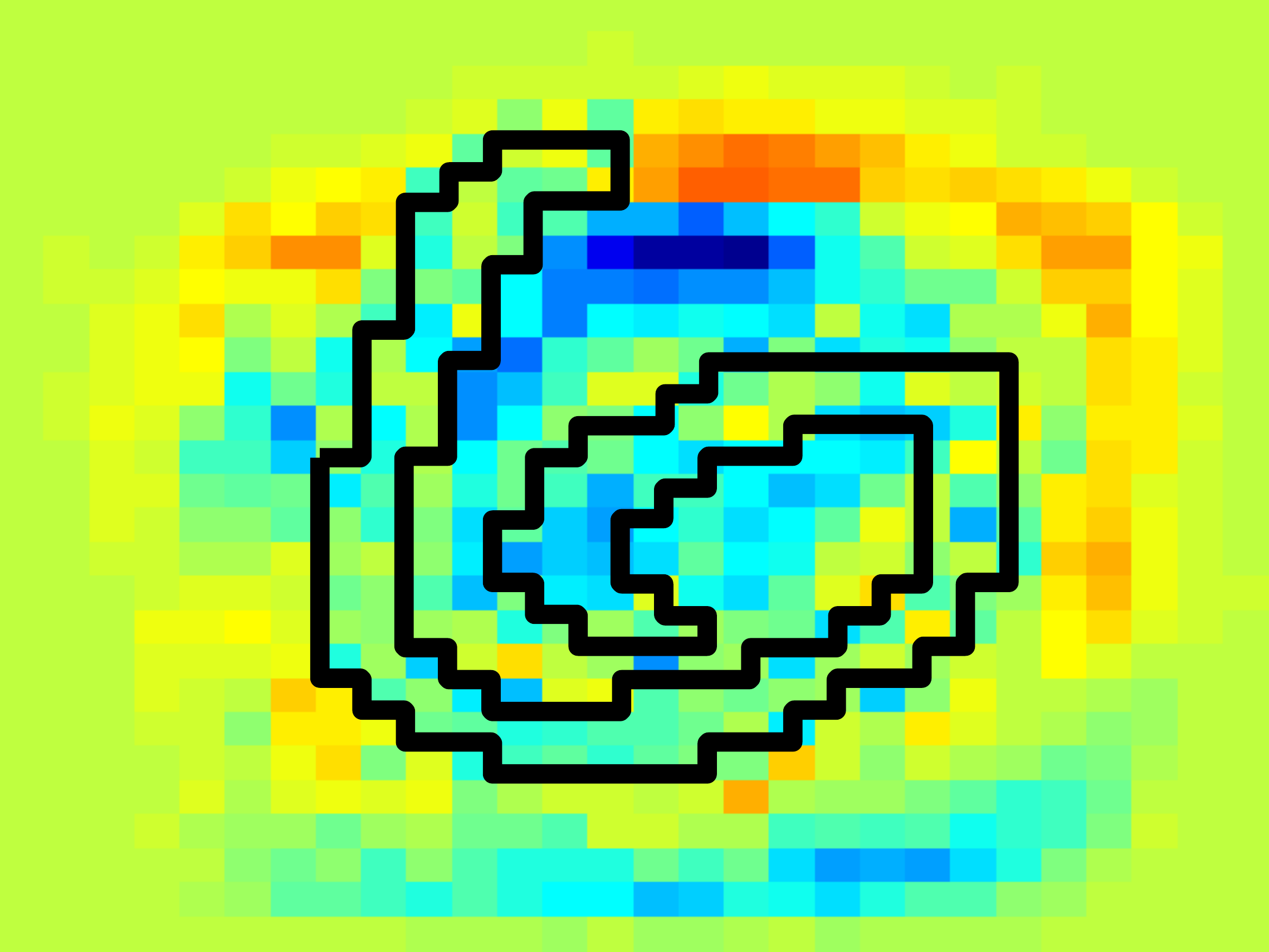}}
		\\[-1em]
		&0.74&0.71&0.89&0.93&0.88&0.69&1.06&0.48&0.90&0.65\\[0.5em]\hline

		\subcaptionbox*{}{\includegraphics[width=0.09\textwidth]{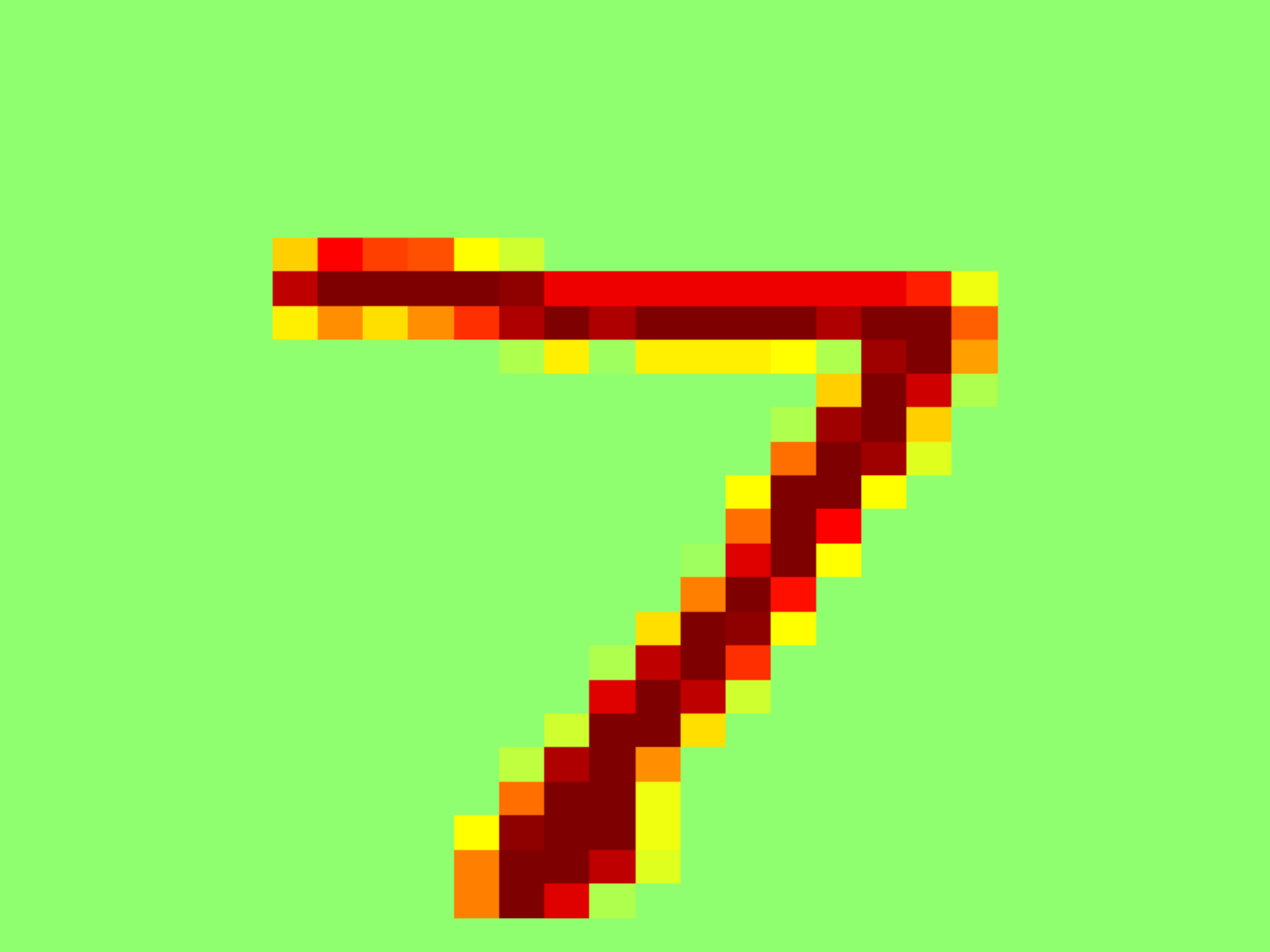}}&
		\subcaptionbox*{}{\includegraphics[width=0.09\textwidth]{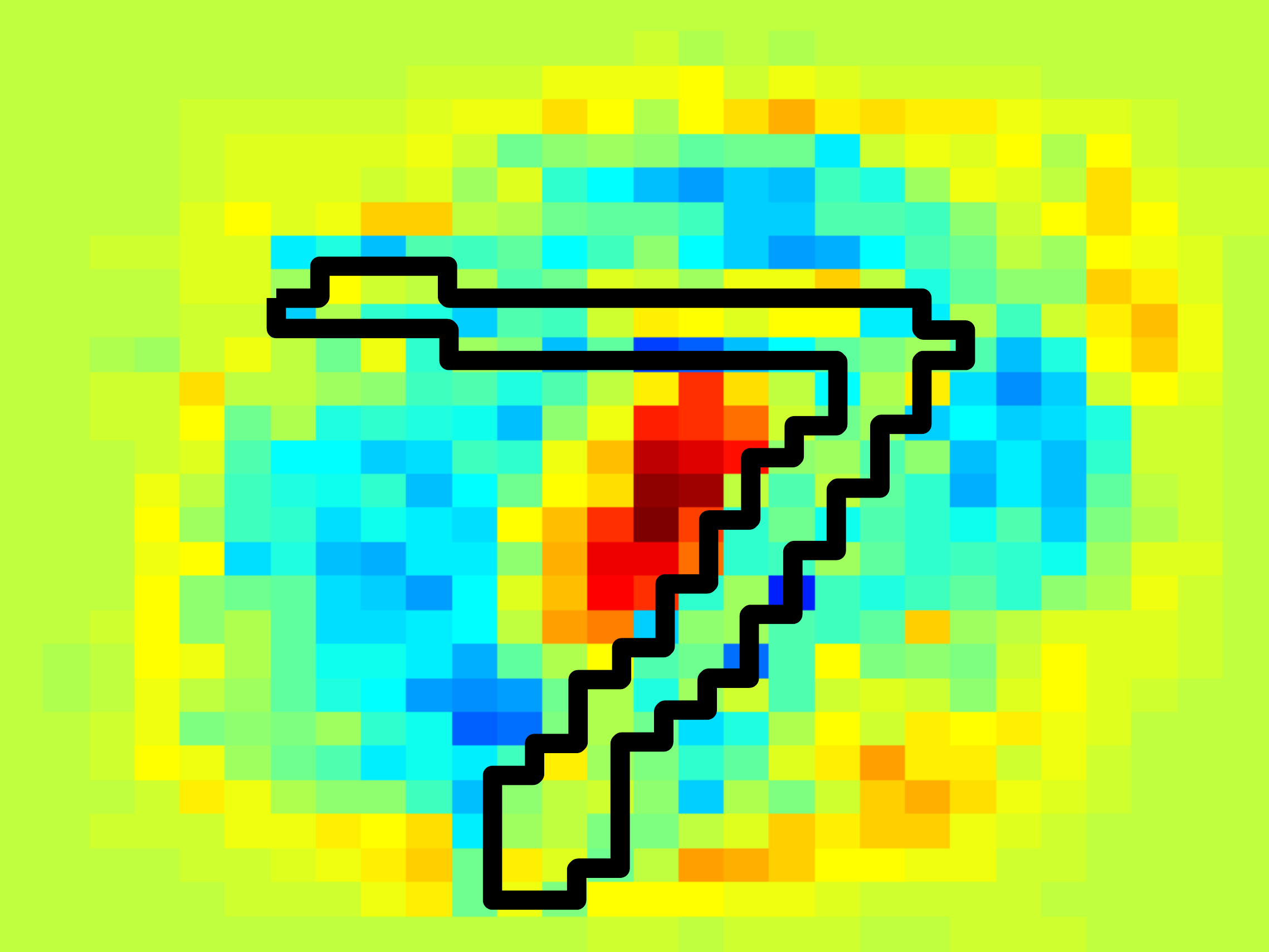}}&
		\subcaptionbox*{}{\includegraphics[width=0.09\textwidth]{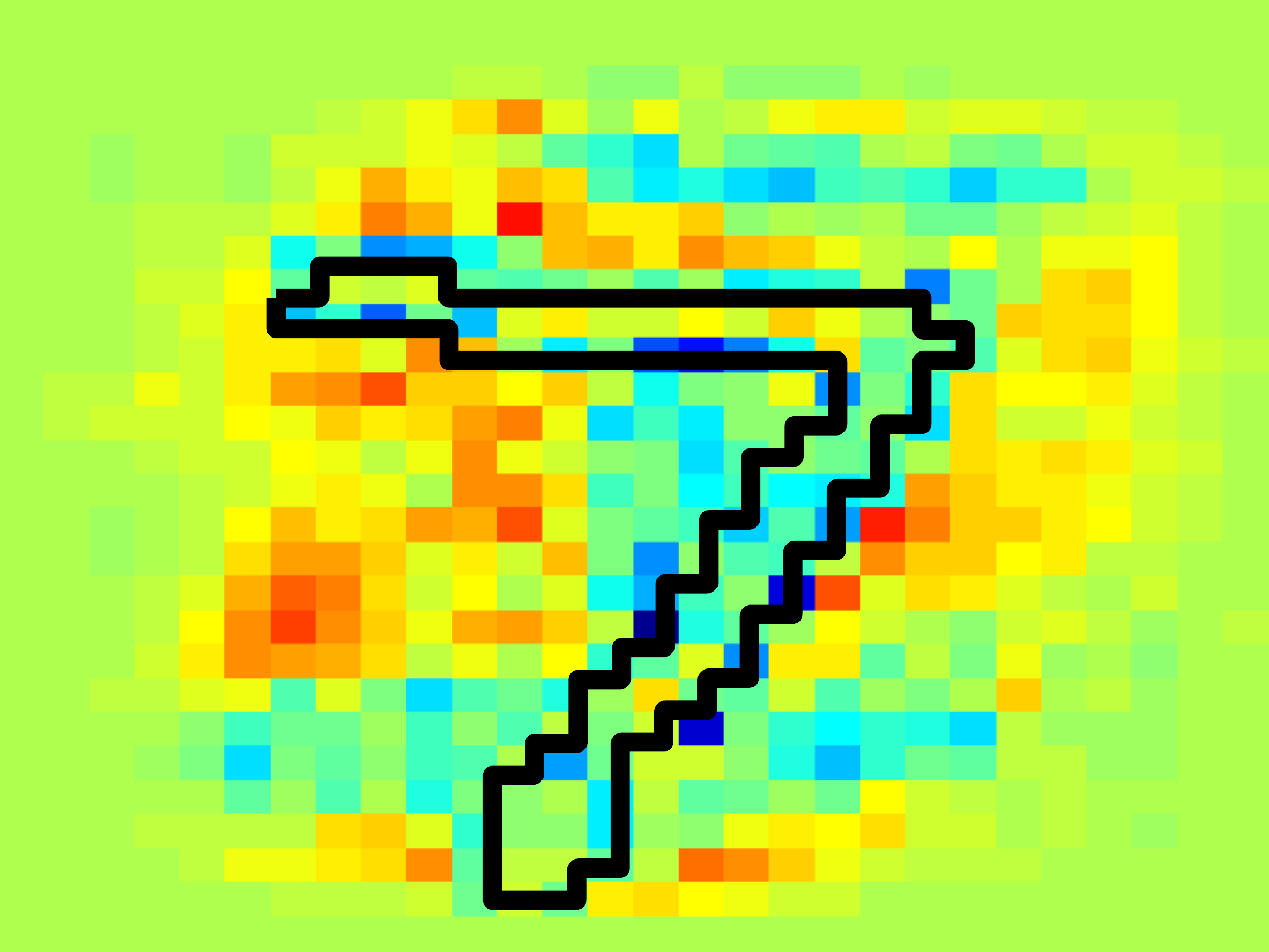}}&
		\subcaptionbox*{}{\includegraphics[width=0.09\textwidth]{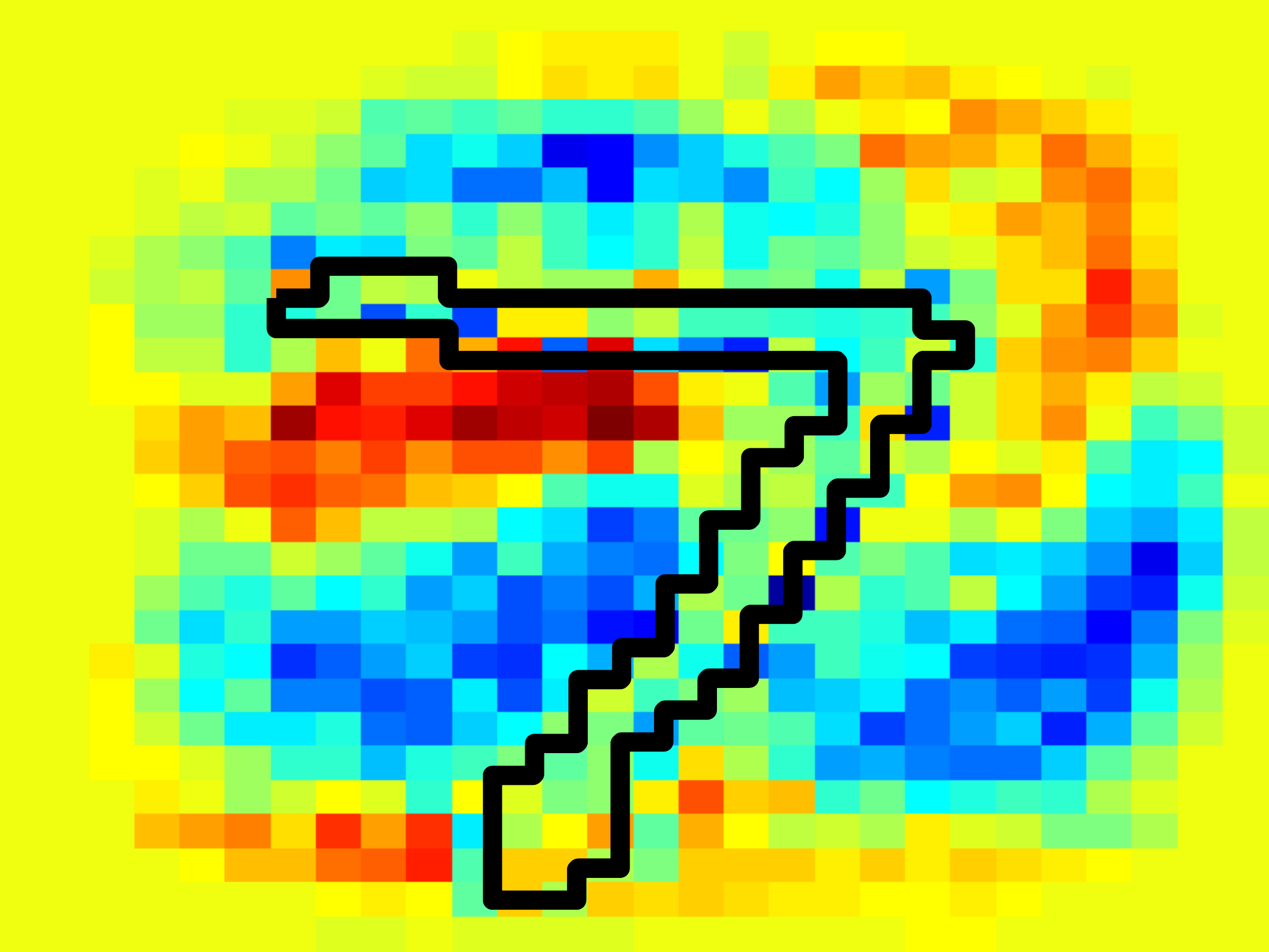}}&
		\subcaptionbox*{}{\includegraphics[width=0.09\textwidth]{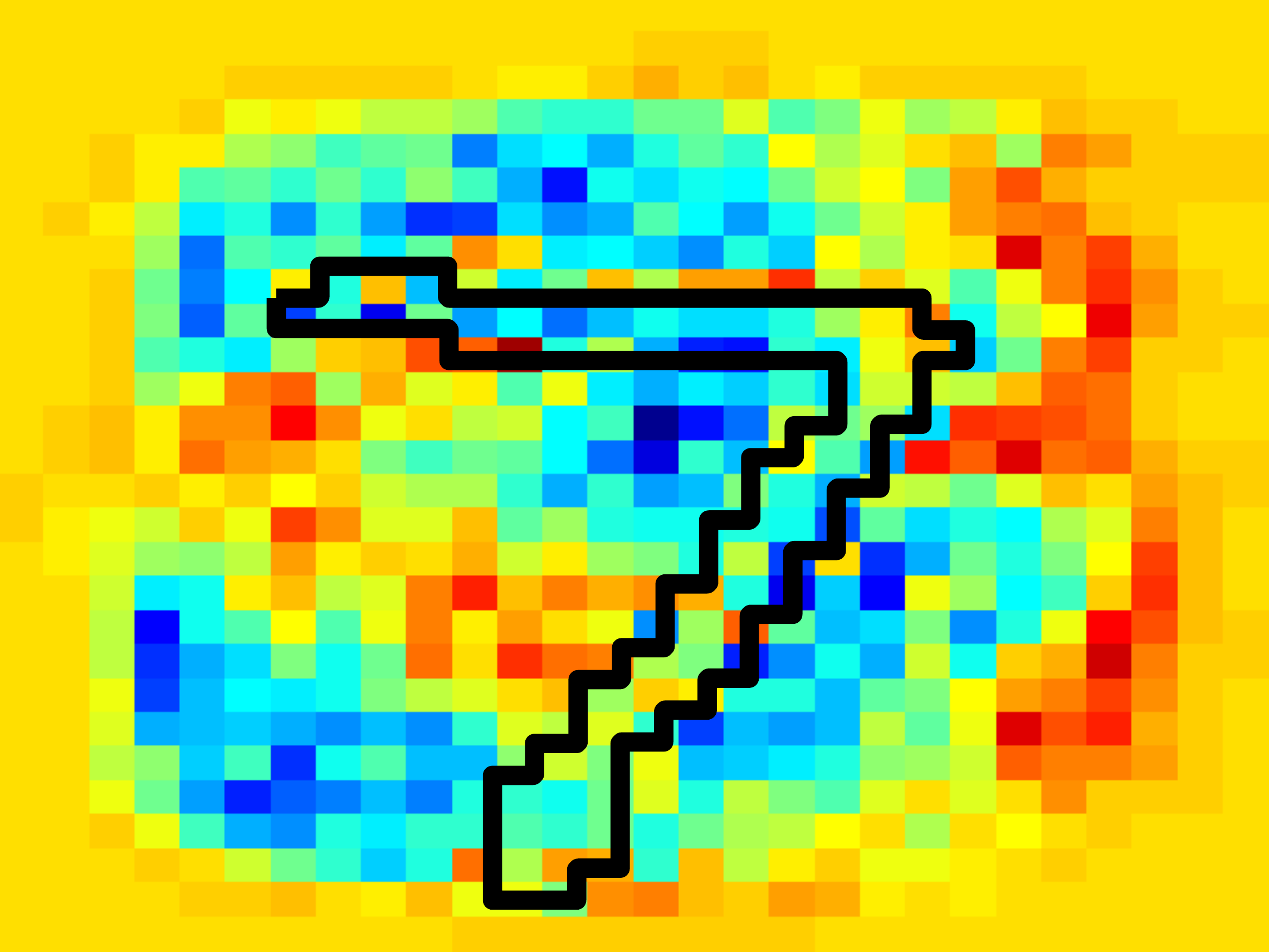}}&
		\subcaptionbox*{}{\includegraphics[width=0.09\textwidth]{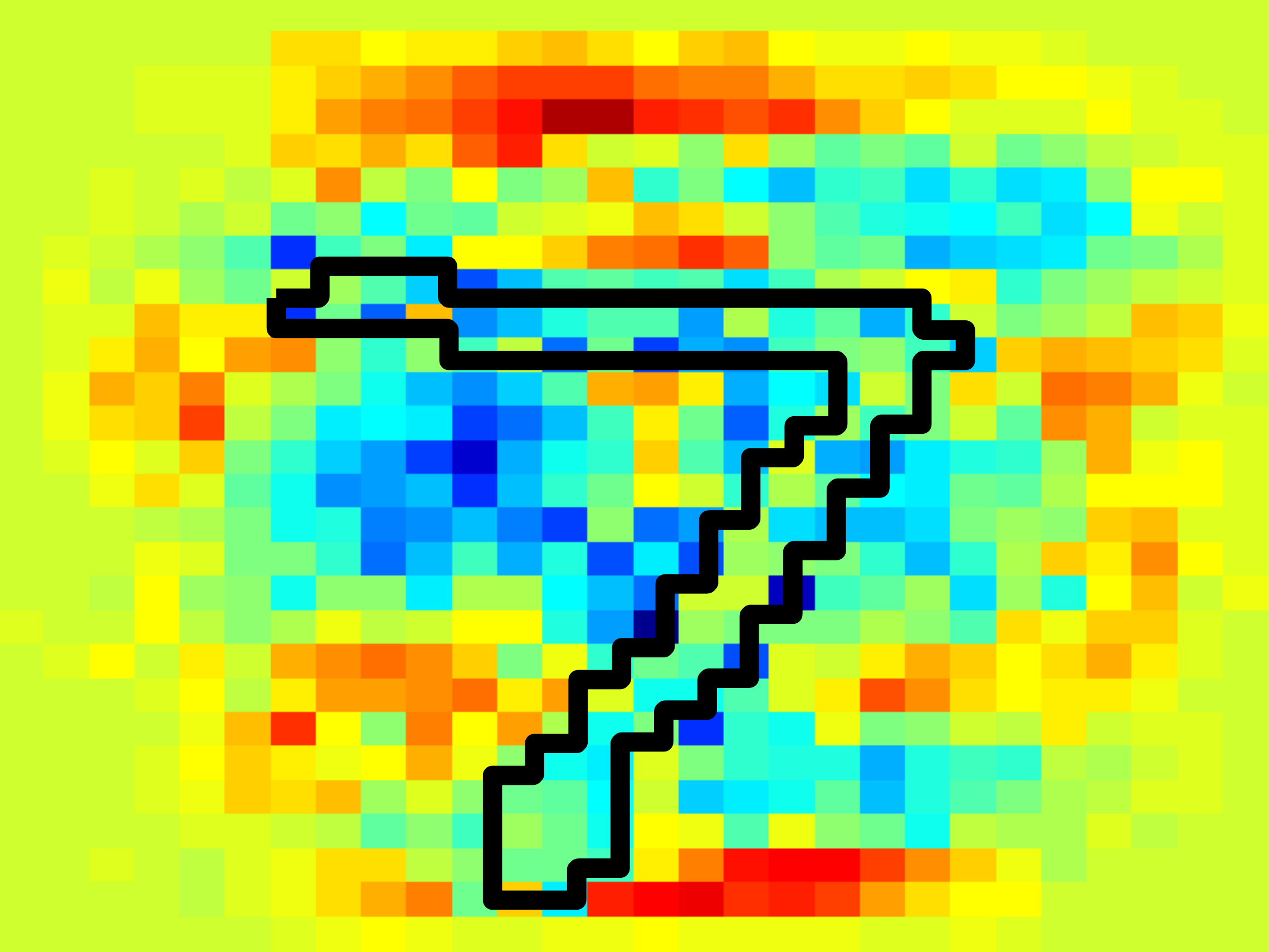}}&
		\subcaptionbox*{}{\includegraphics[width=0.09\textwidth]{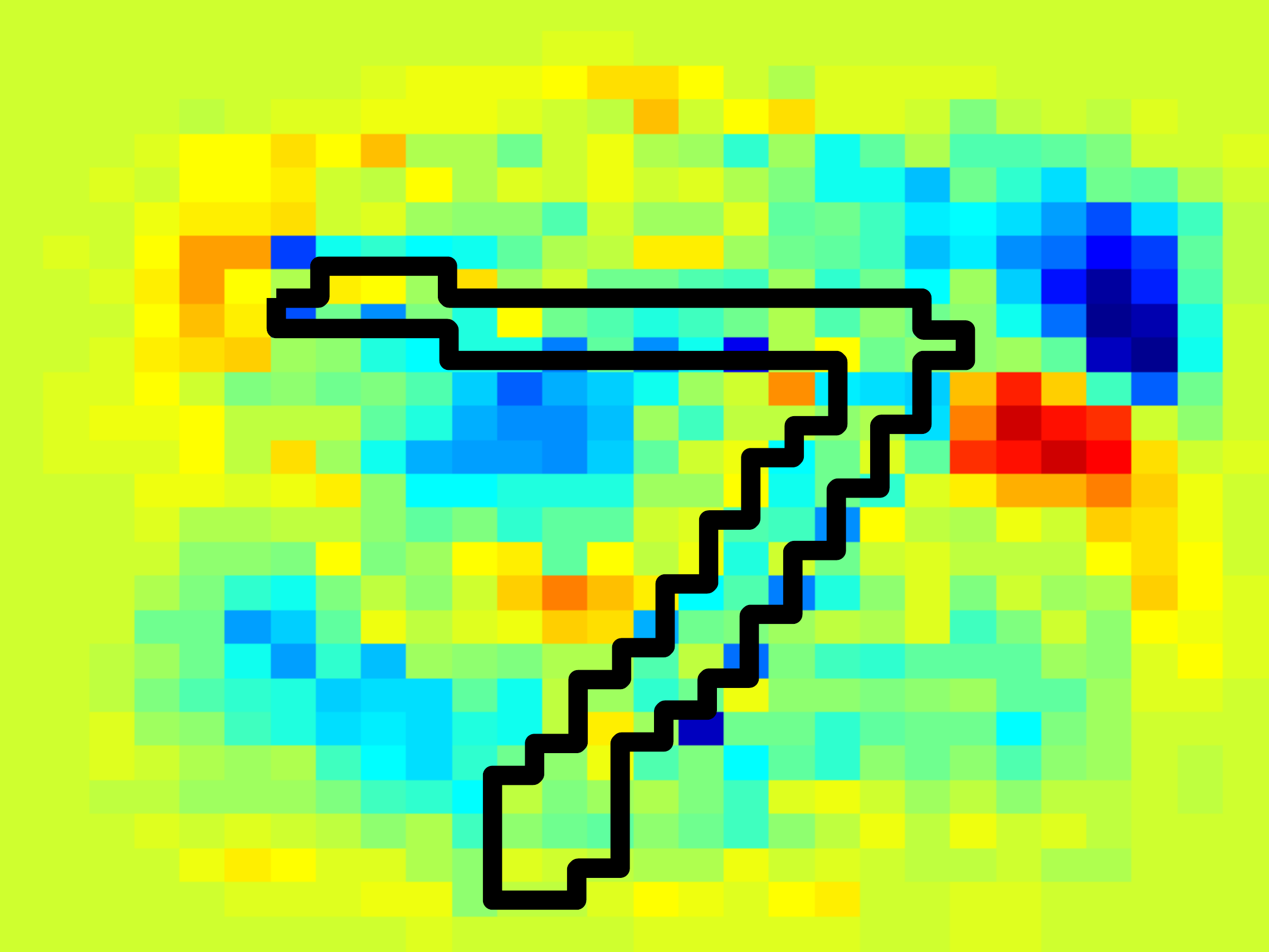}}&
		\subcaptionbox*{}{\includegraphics[width=0.09\textwidth]{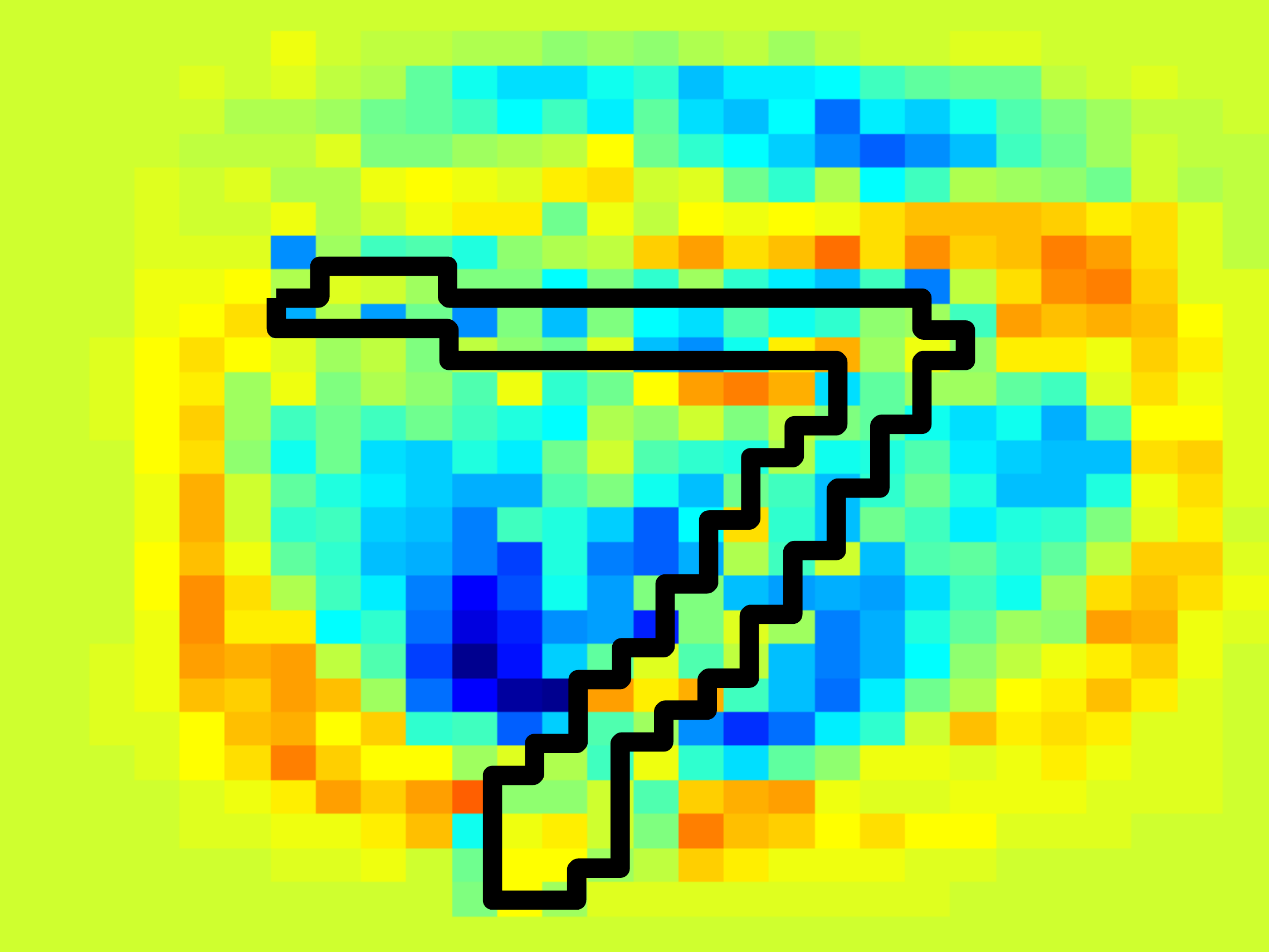}}&
		\subcaptionbox*{}{\includegraphics[width=0.09\textwidth]{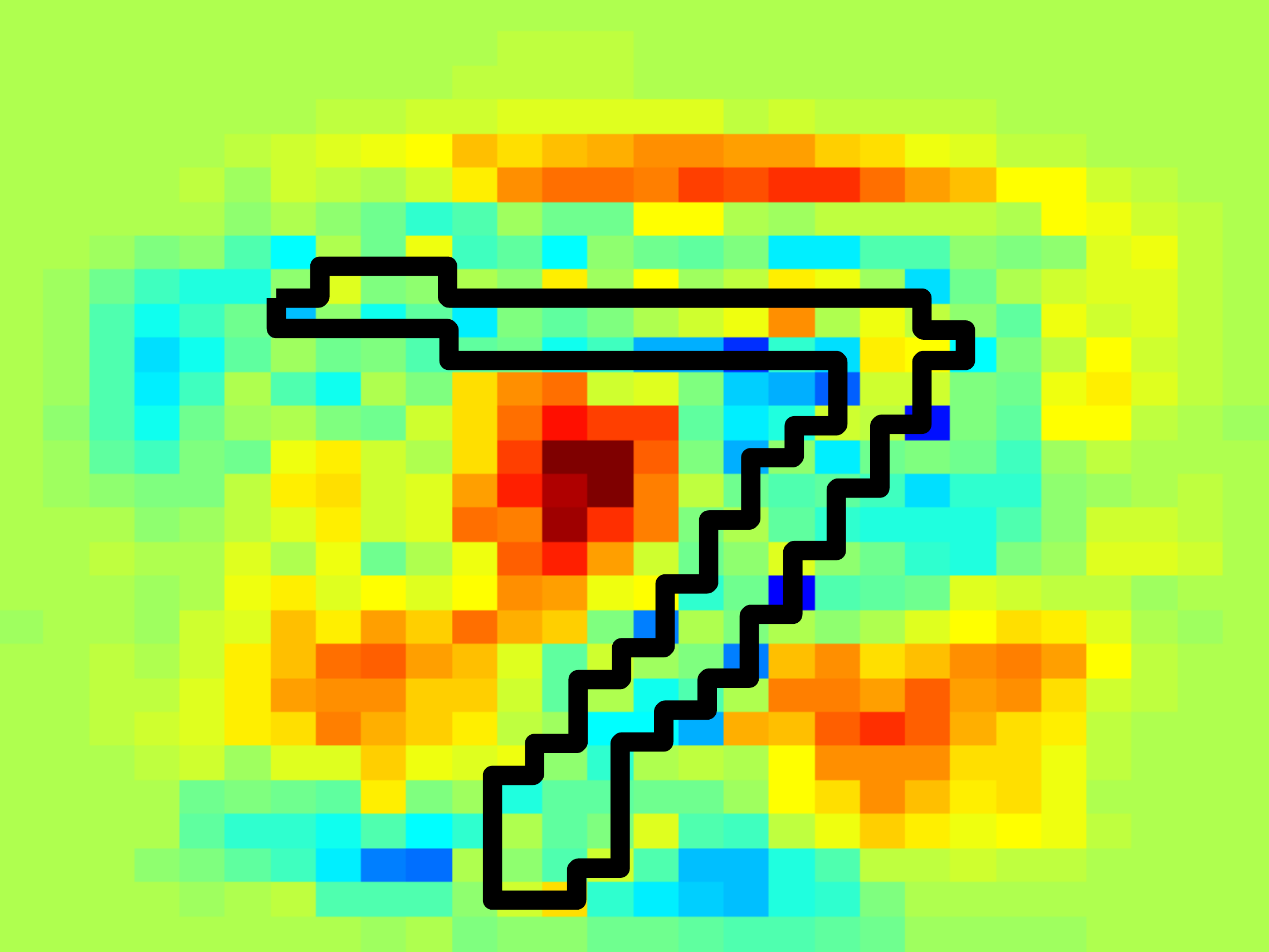}}&
		\subcaptionbox*{}{\includegraphics[width=0.09\textwidth]{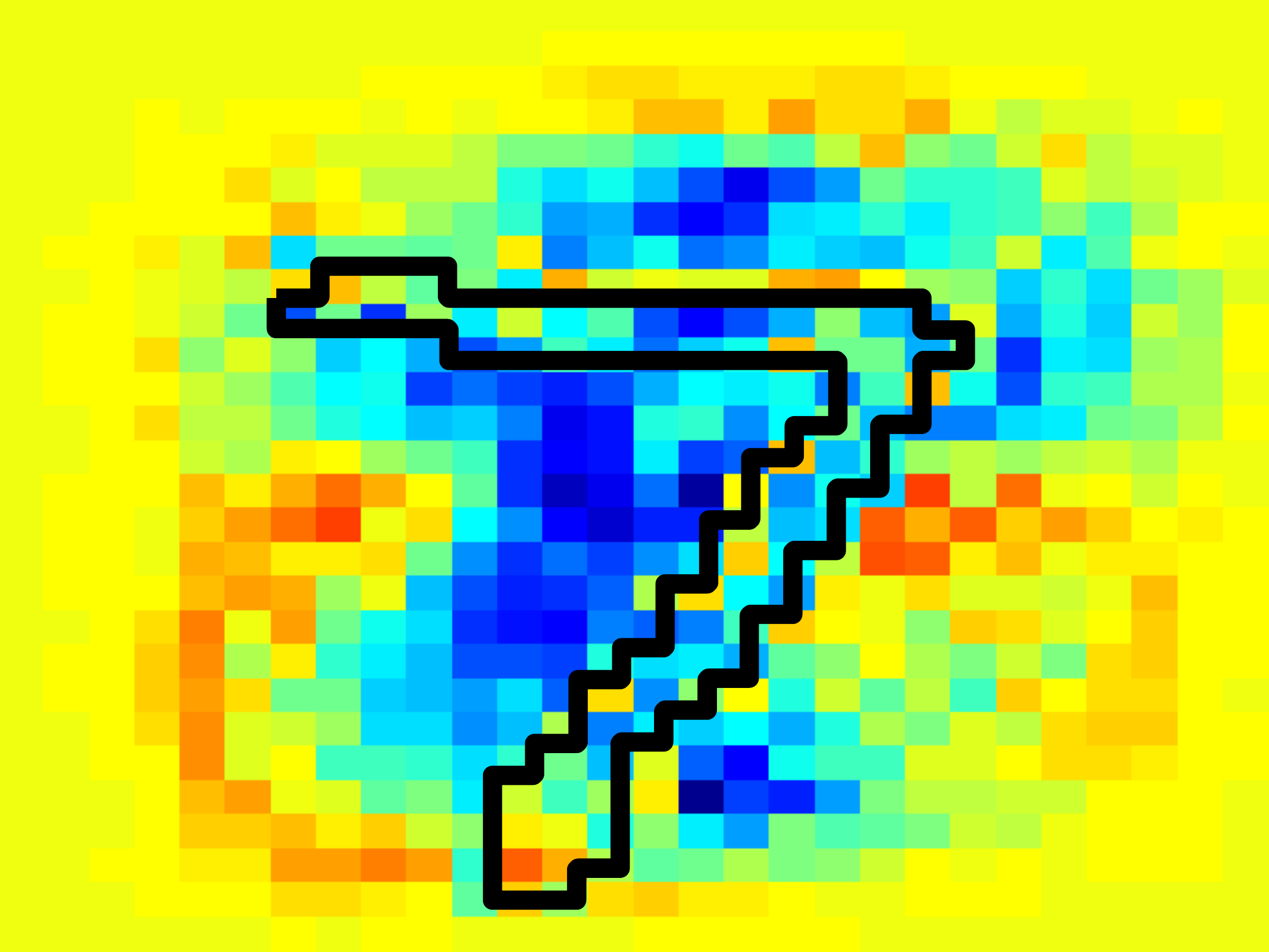}}&
		\subcaptionbox*{}{\includegraphics[width=0.09\textwidth]{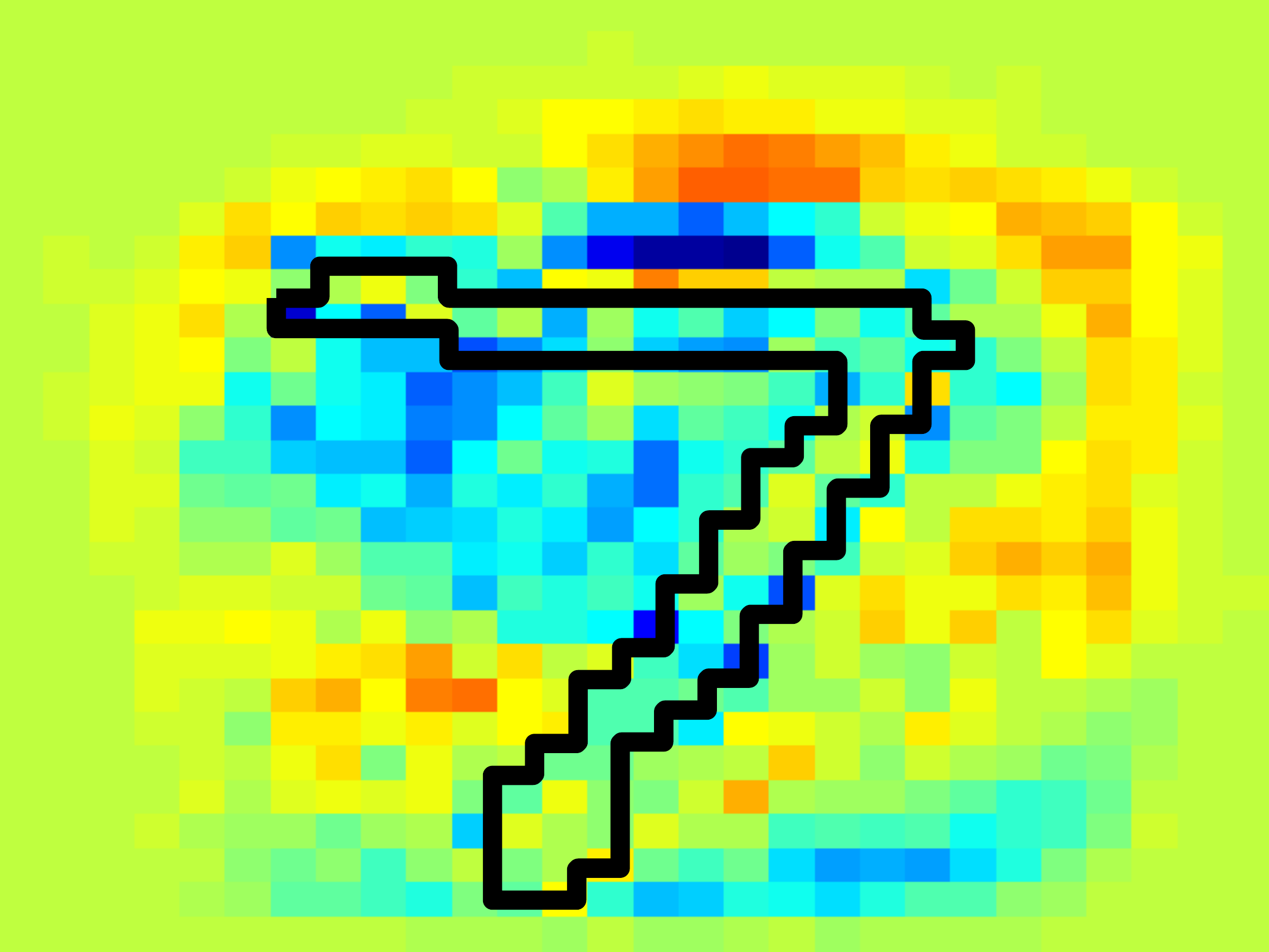}}
		\\[-1em]
		&0.79&0.75&0.74&1.04&0.83&0.75&0.57&1.40&0.63&0.88\\[0.5em]\hline
		
		\subcaptionbox*{}{\includegraphics[width=0.09\textwidth]{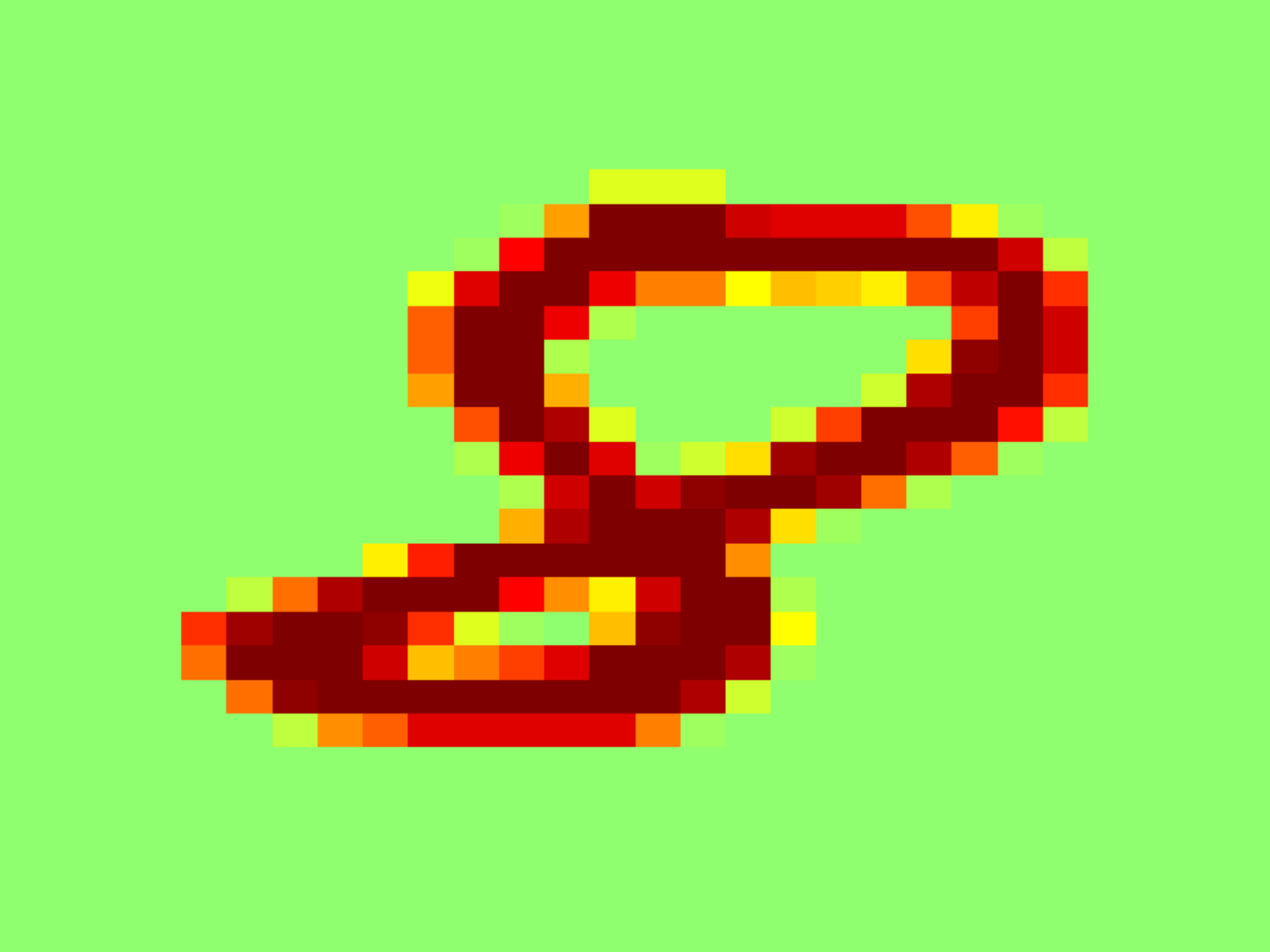}}&
		\subcaptionbox*{}{\includegraphics[width=0.09\textwidth]{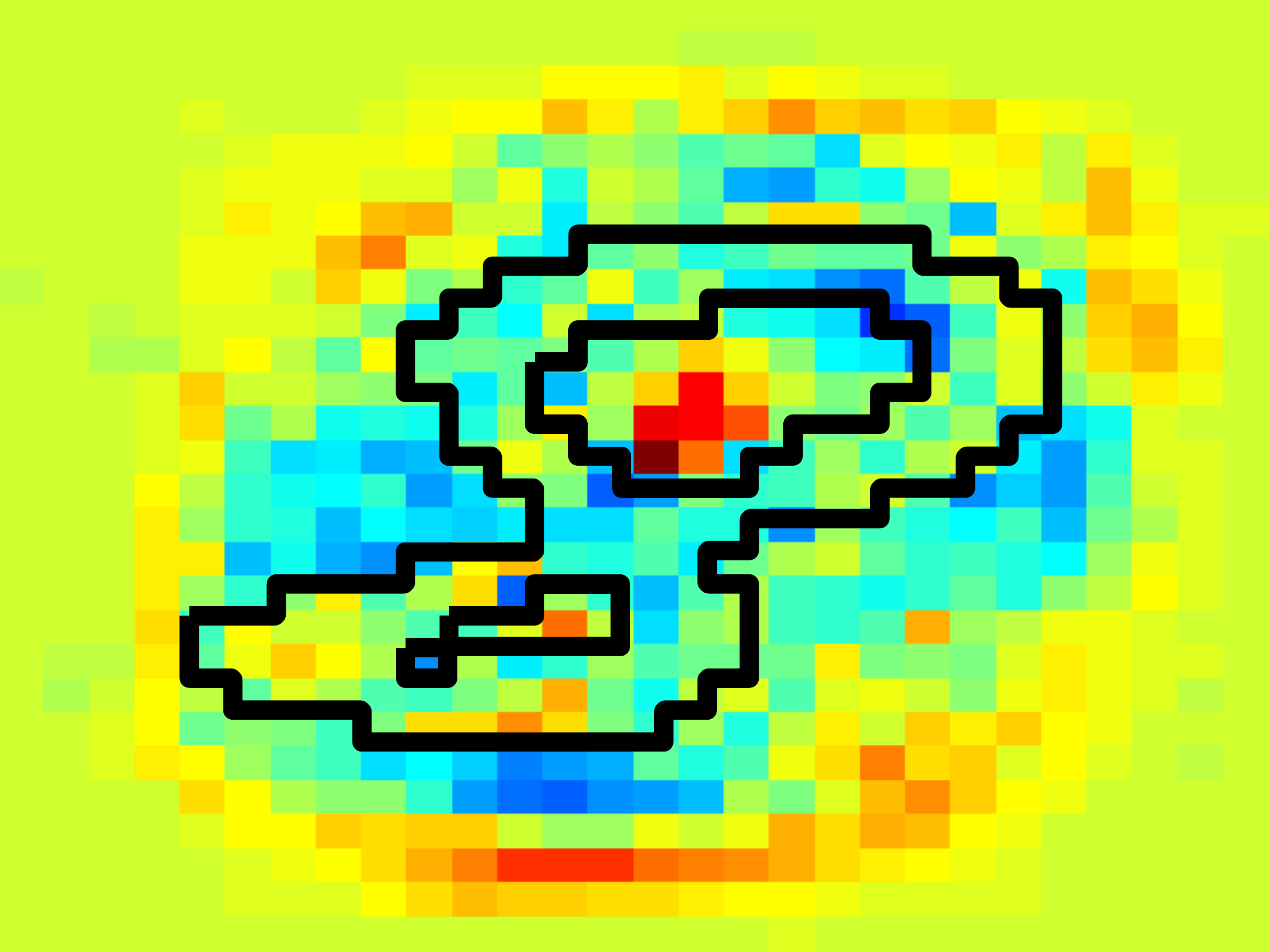}}&
		\subcaptionbox*{}{\includegraphics[width=0.09\textwidth]{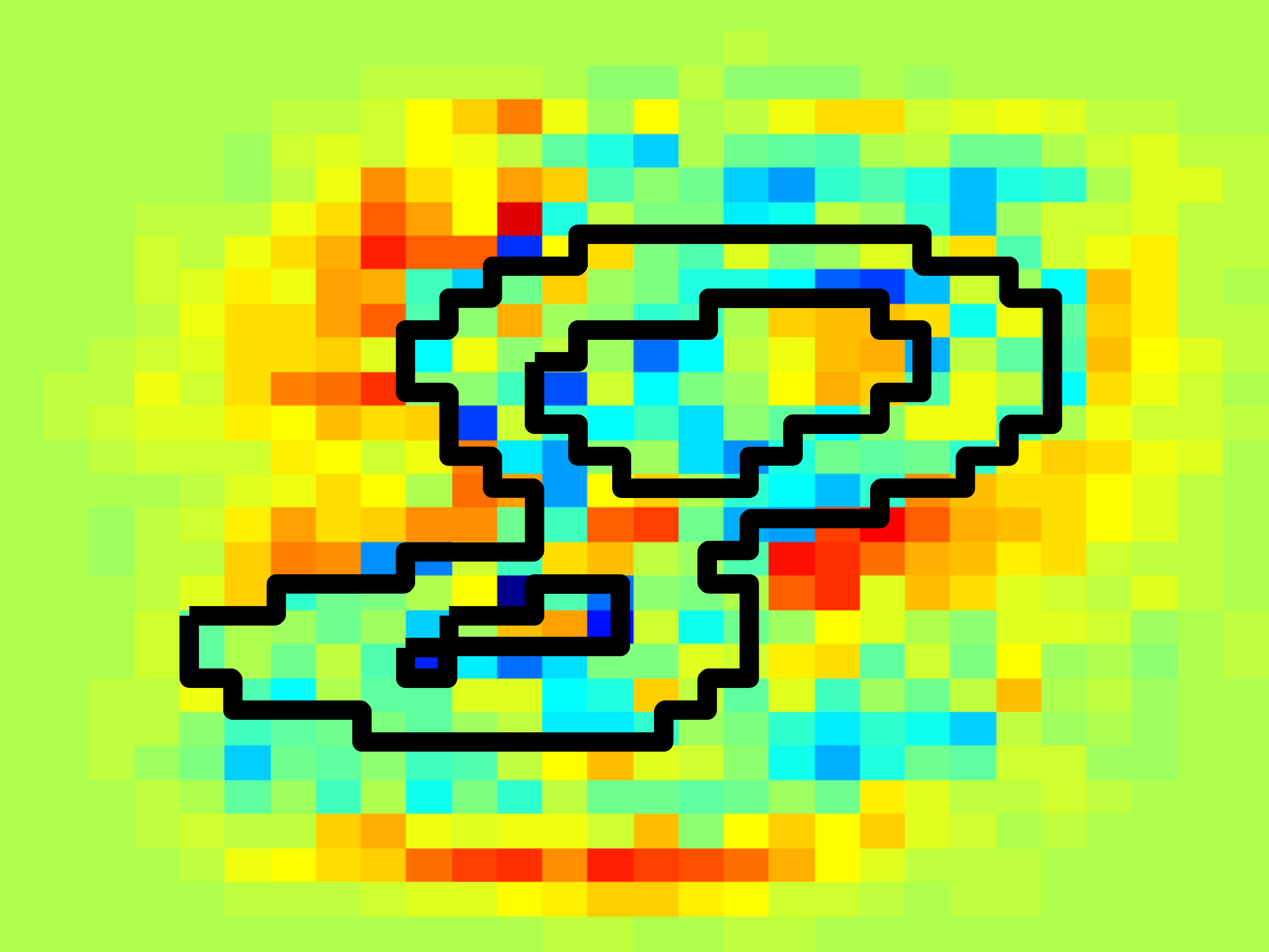}}&
		\subcaptionbox*{}{\includegraphics[width=0.09\textwidth]{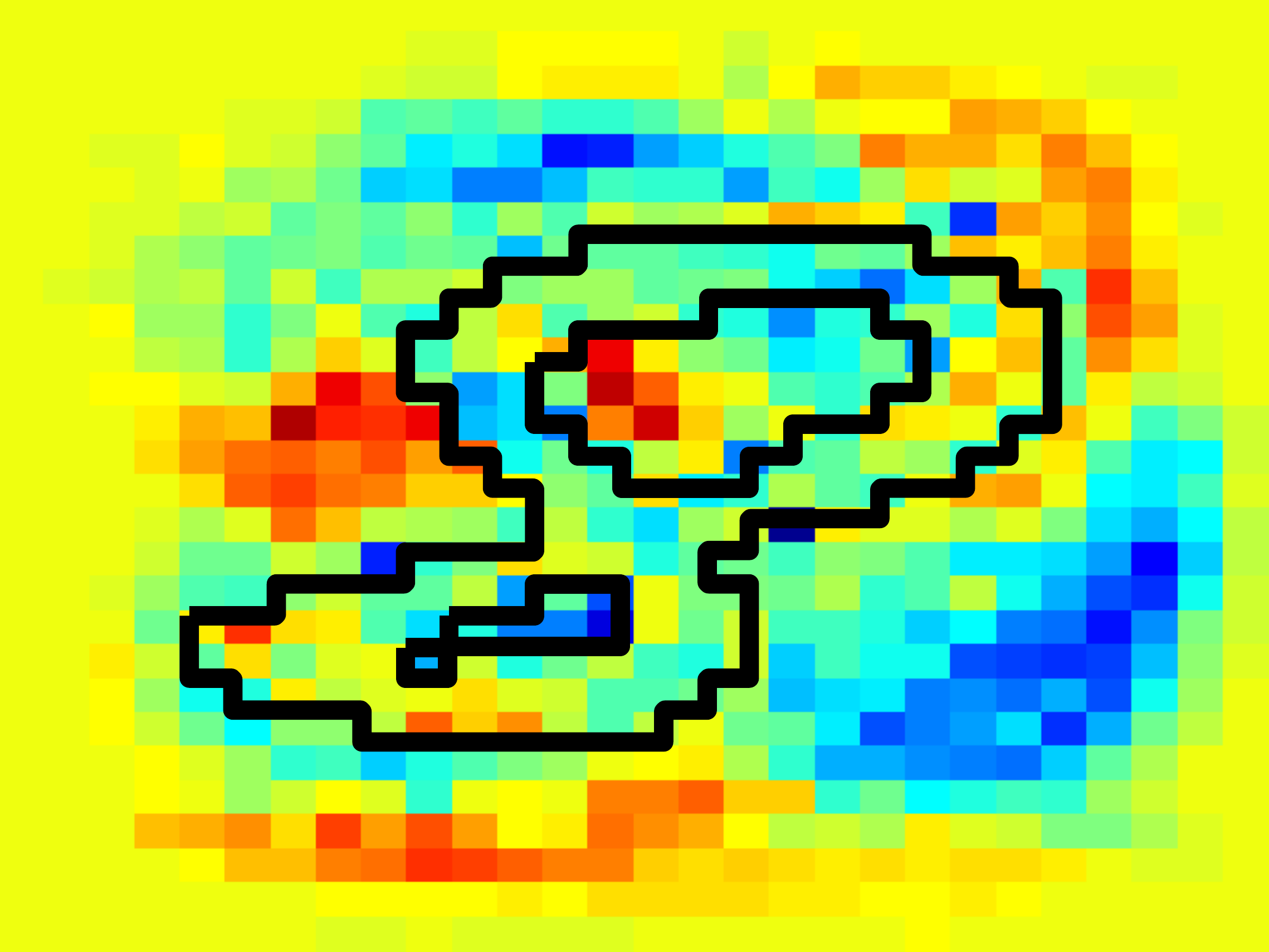}}&
		\subcaptionbox*{}{\includegraphics[width=0.09\textwidth]{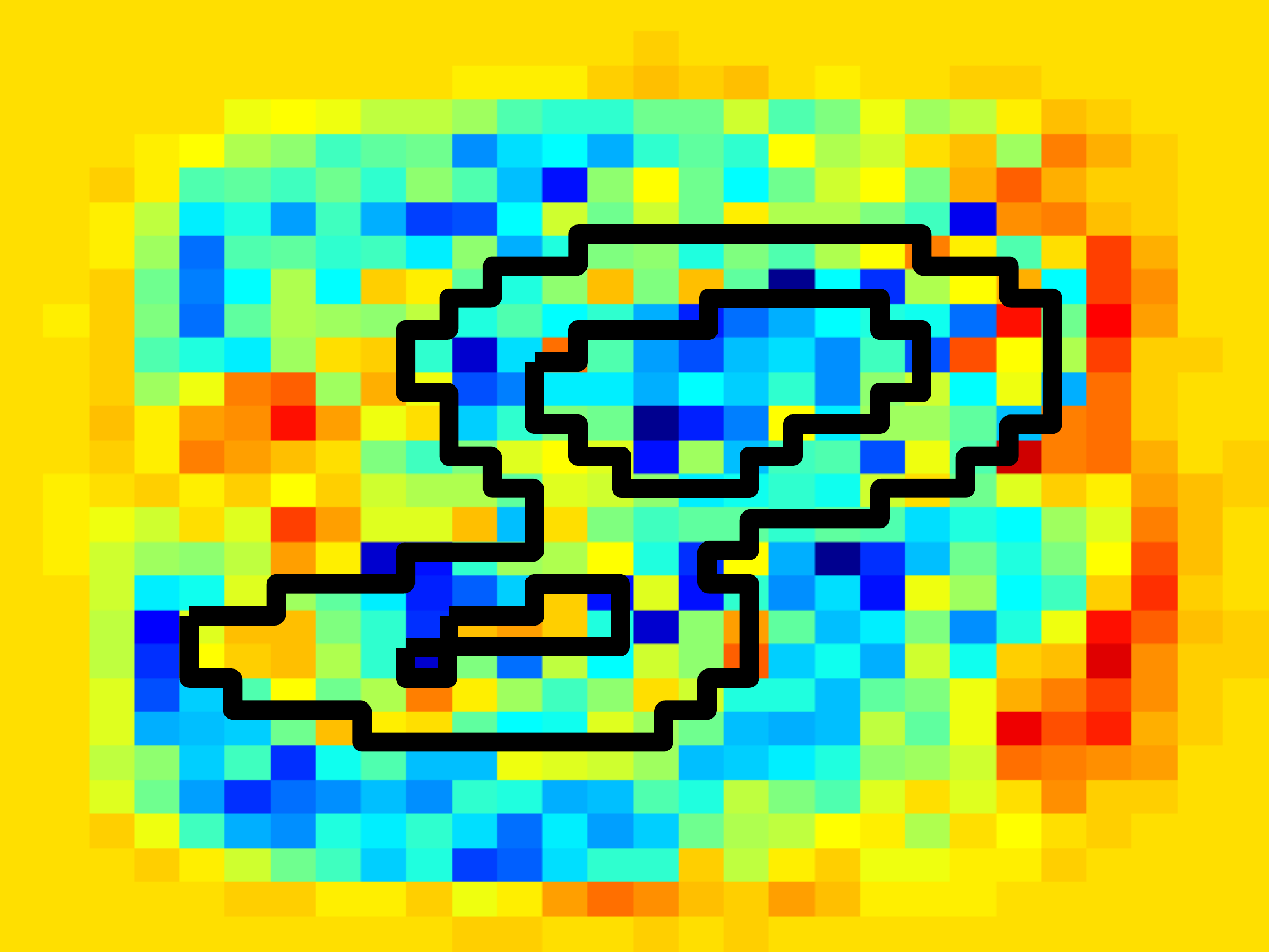}}&
		\subcaptionbox*{}{\includegraphics[width=0.09\textwidth]{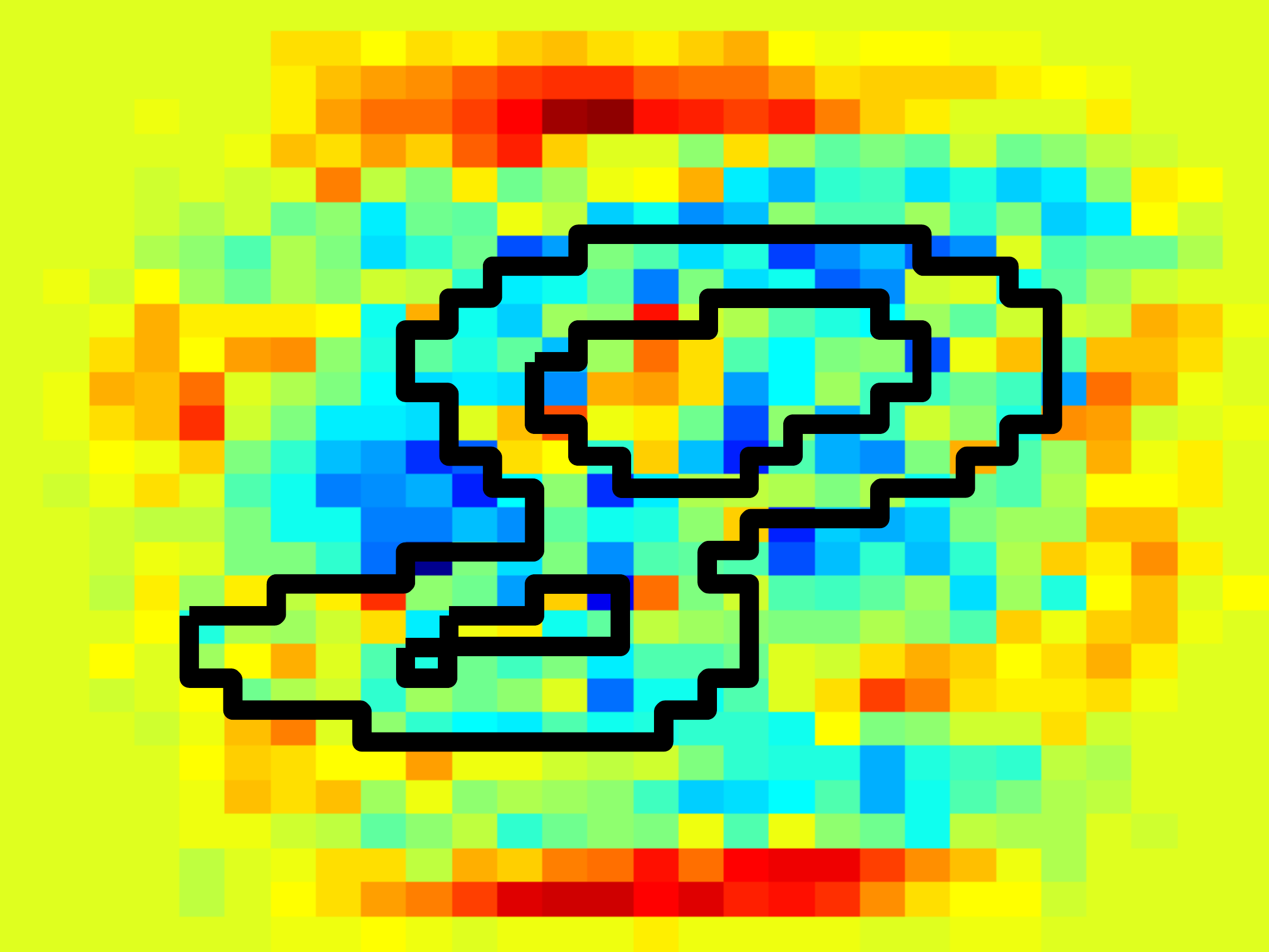}}&
		\subcaptionbox*{}{\includegraphics[width=0.09\textwidth]{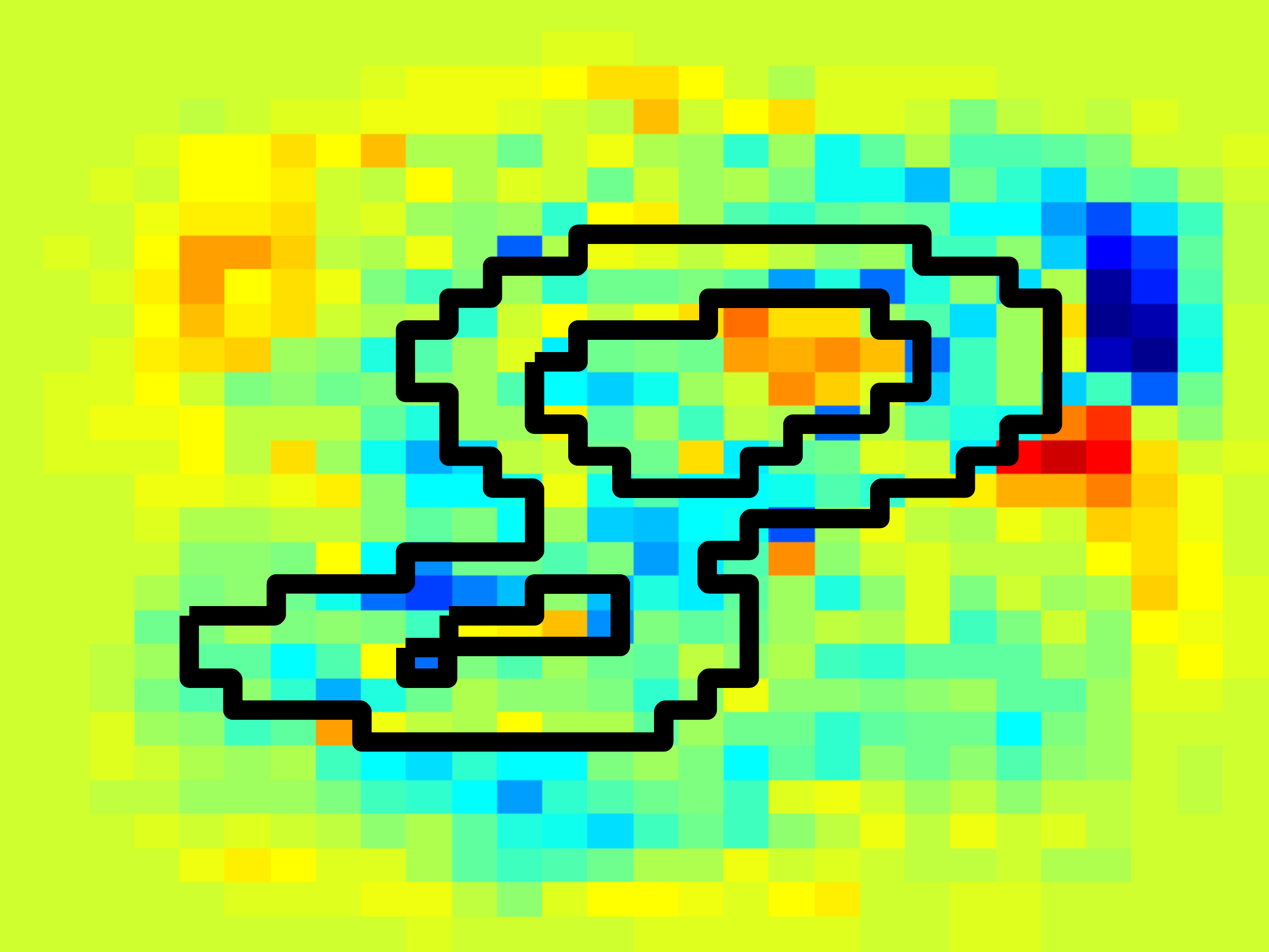}}&
		\subcaptionbox*{}{\includegraphics[width=0.09\textwidth]{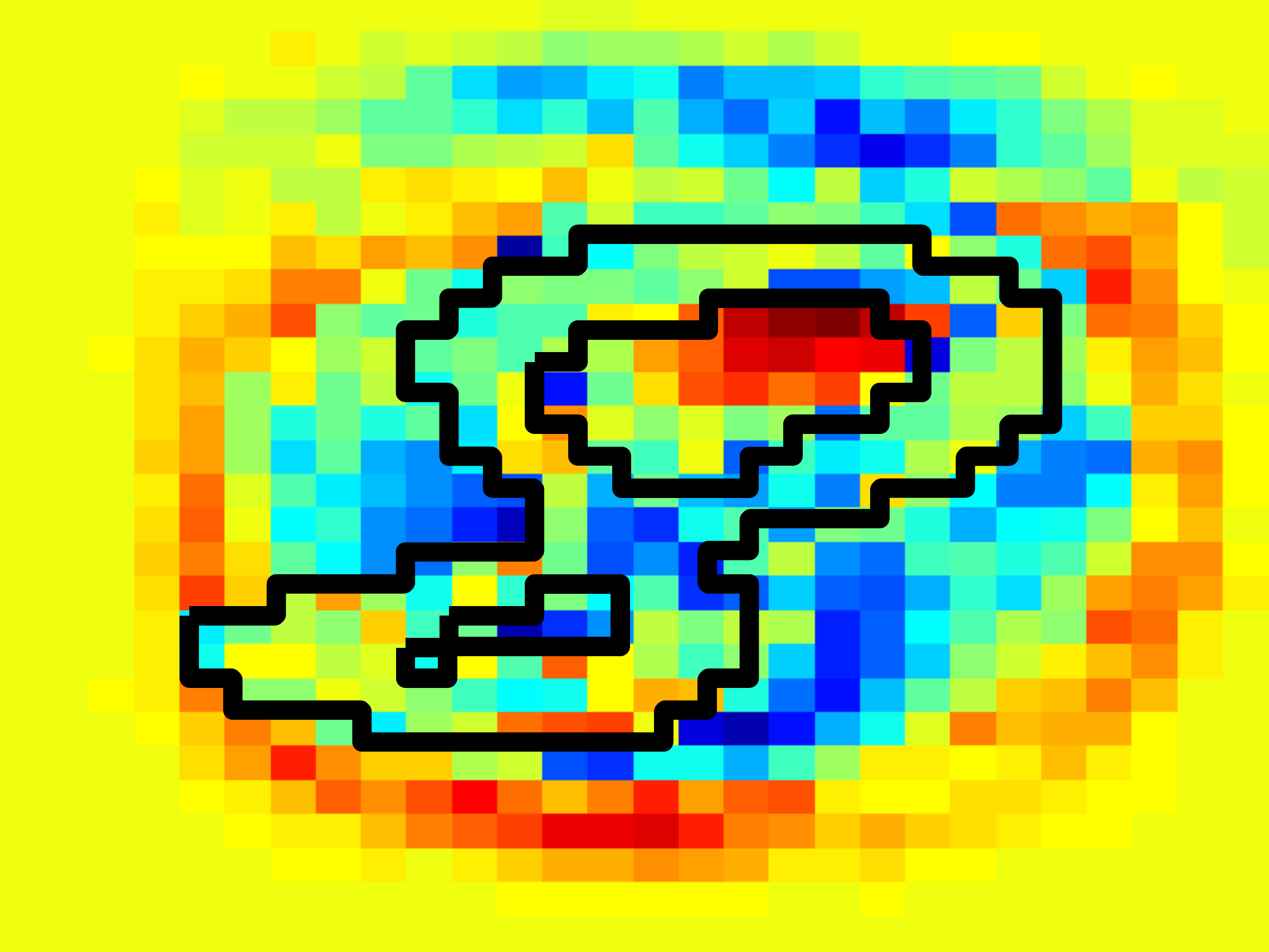}}&
		\subcaptionbox*{}{\includegraphics[width=0.09\textwidth]{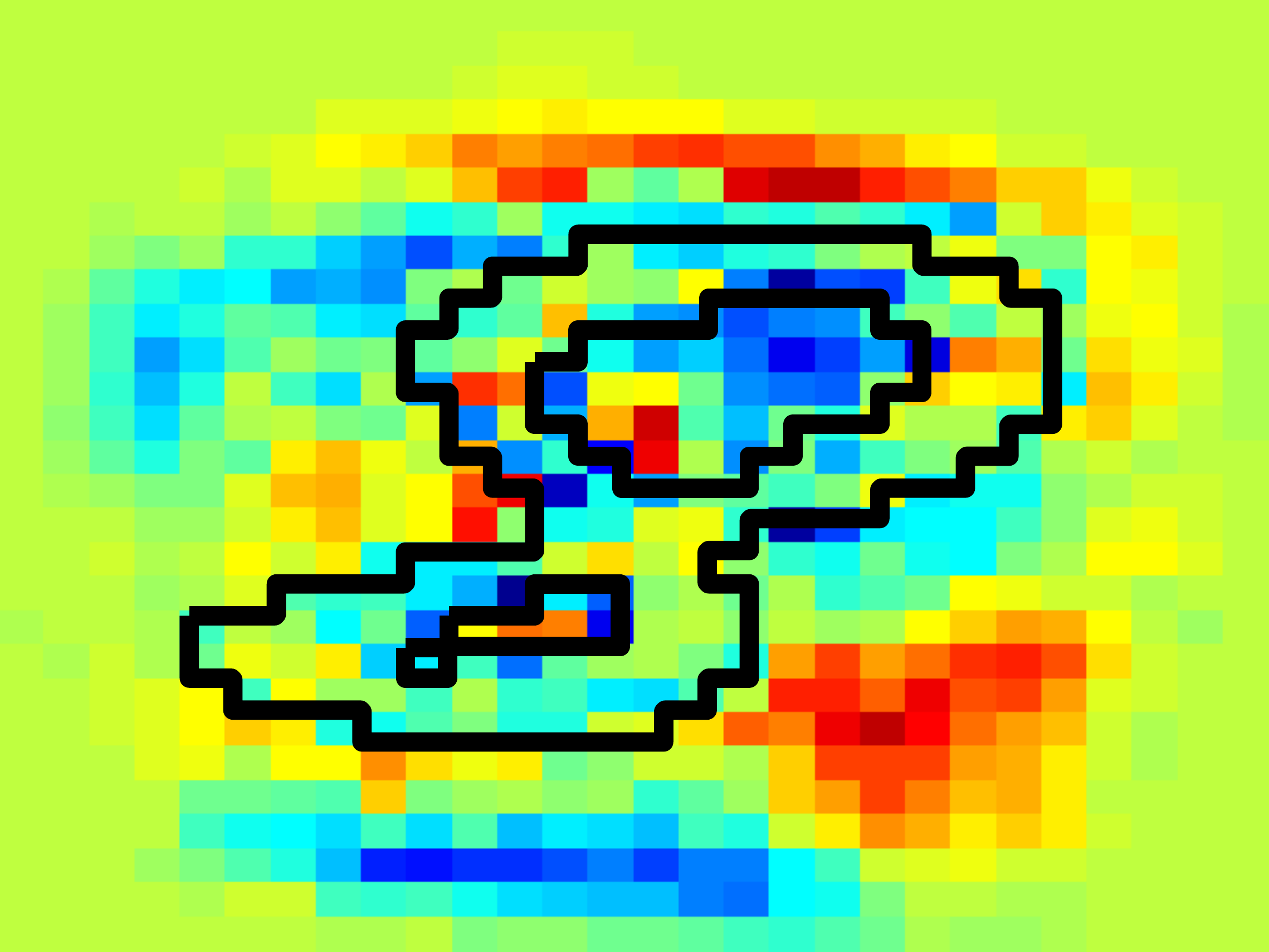}}&
		\subcaptionbox*{}{\includegraphics[width=0.09\textwidth]{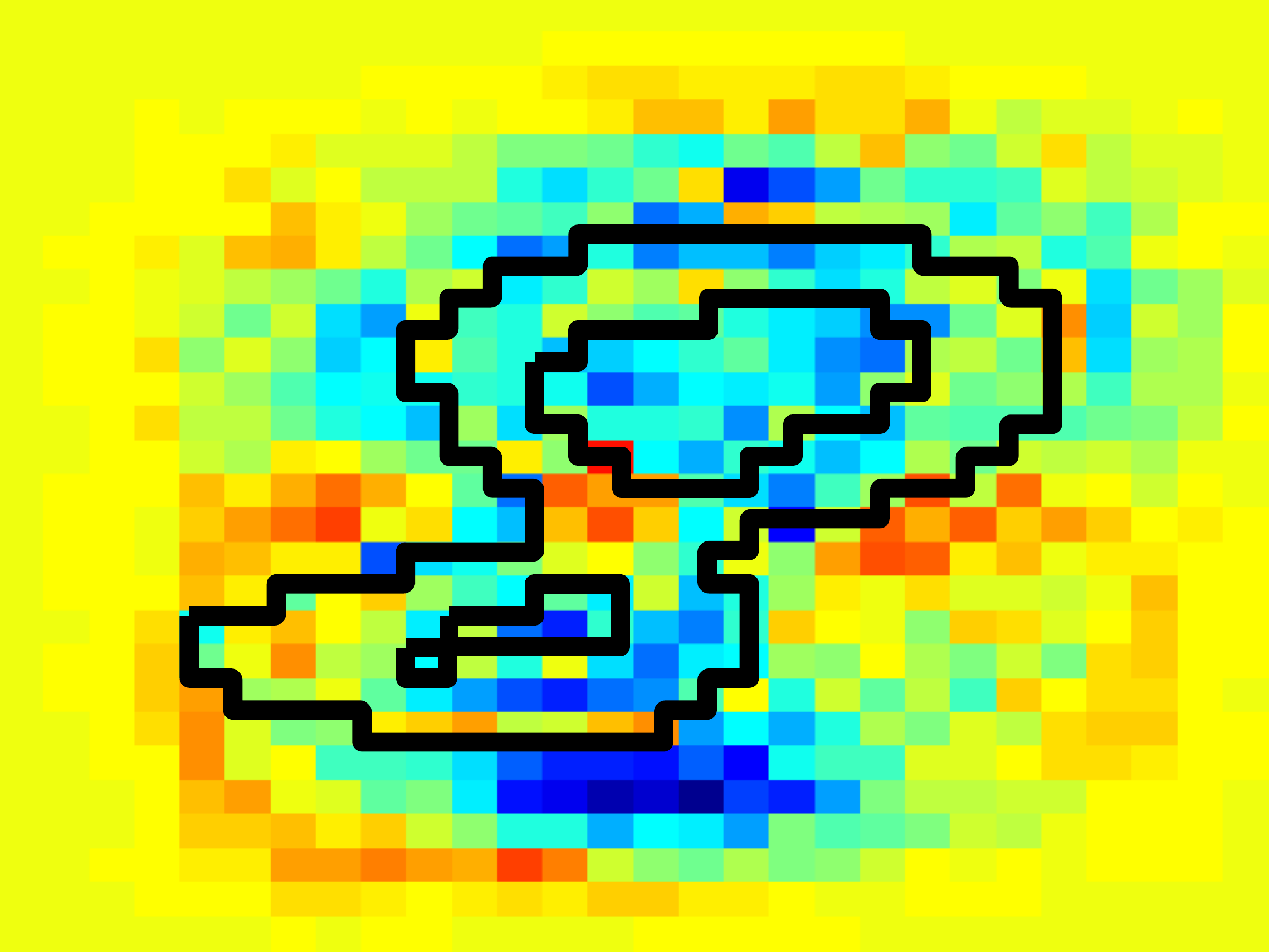}}&
		\subcaptionbox*{}{\includegraphics[width=0.09\textwidth]{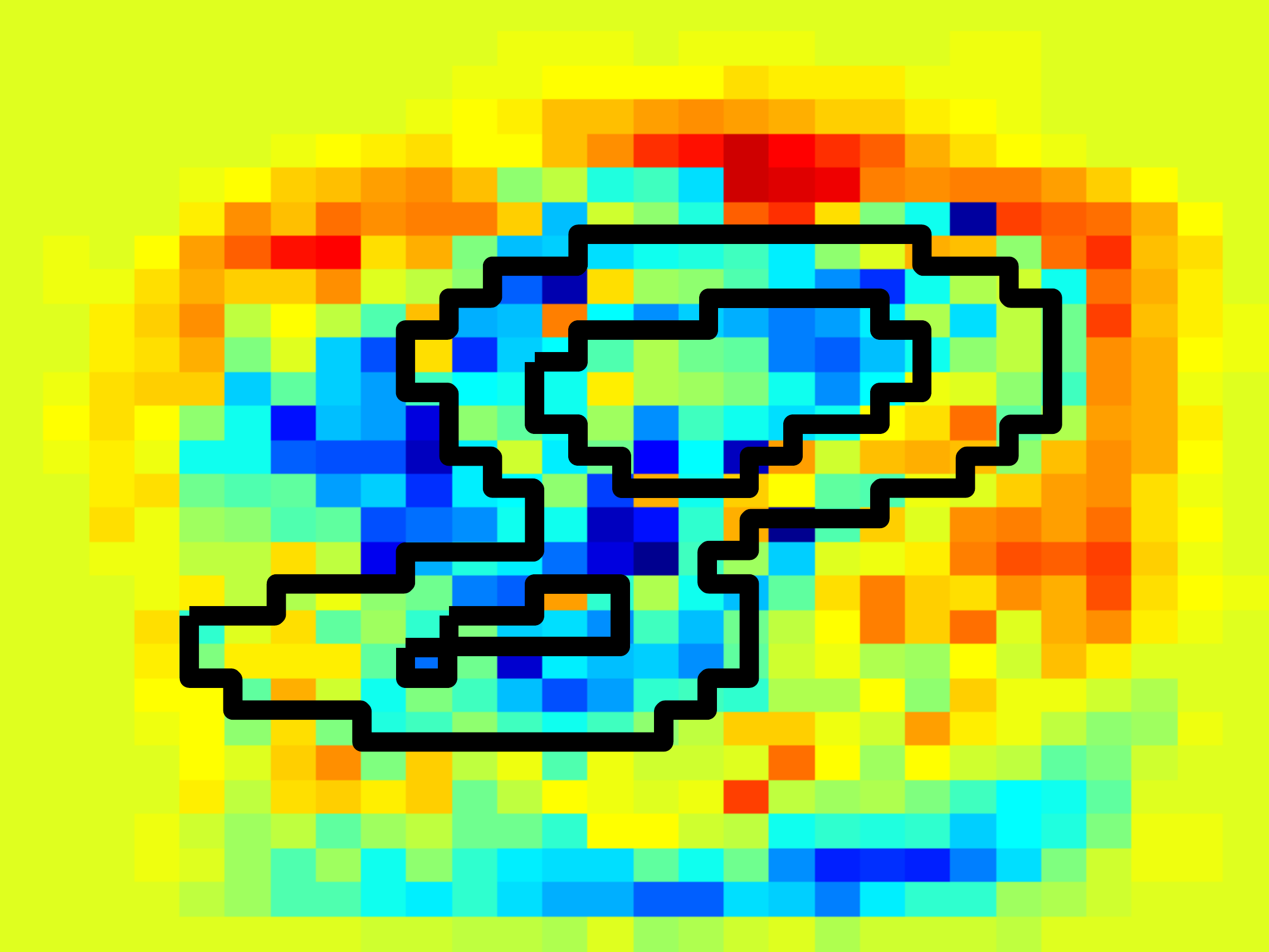}}
		\\[-1em]
		&0.85&0.70&1.08&0.79&0.89&0.87&0.87&0.52&1.19&0.94\\[0.5em]\hline
		
		\subcaptionbox*{}{\includegraphics[width=0.09\textwidth]{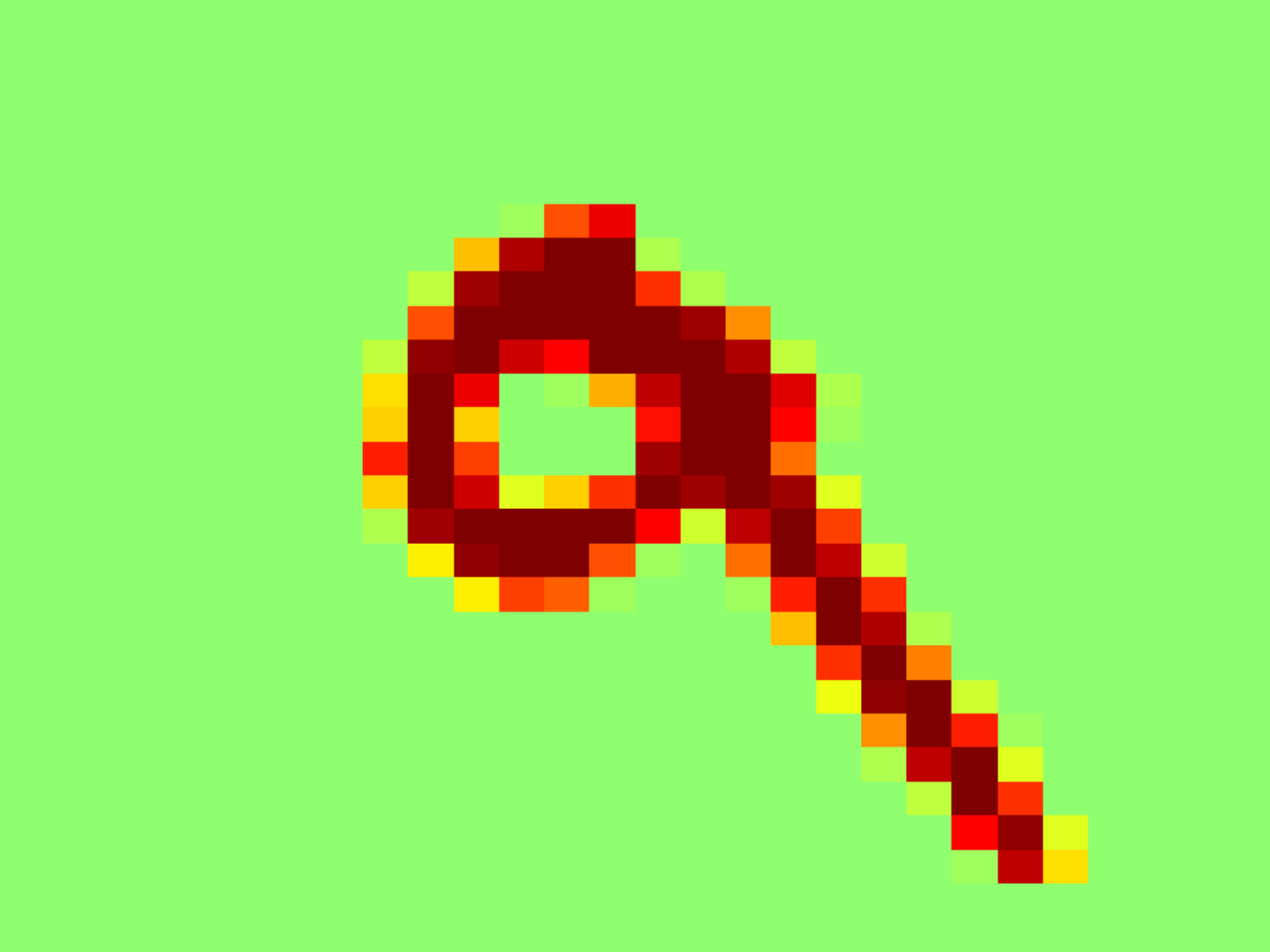}}&
		\subcaptionbox*{}{\includegraphics[width=0.09\textwidth]{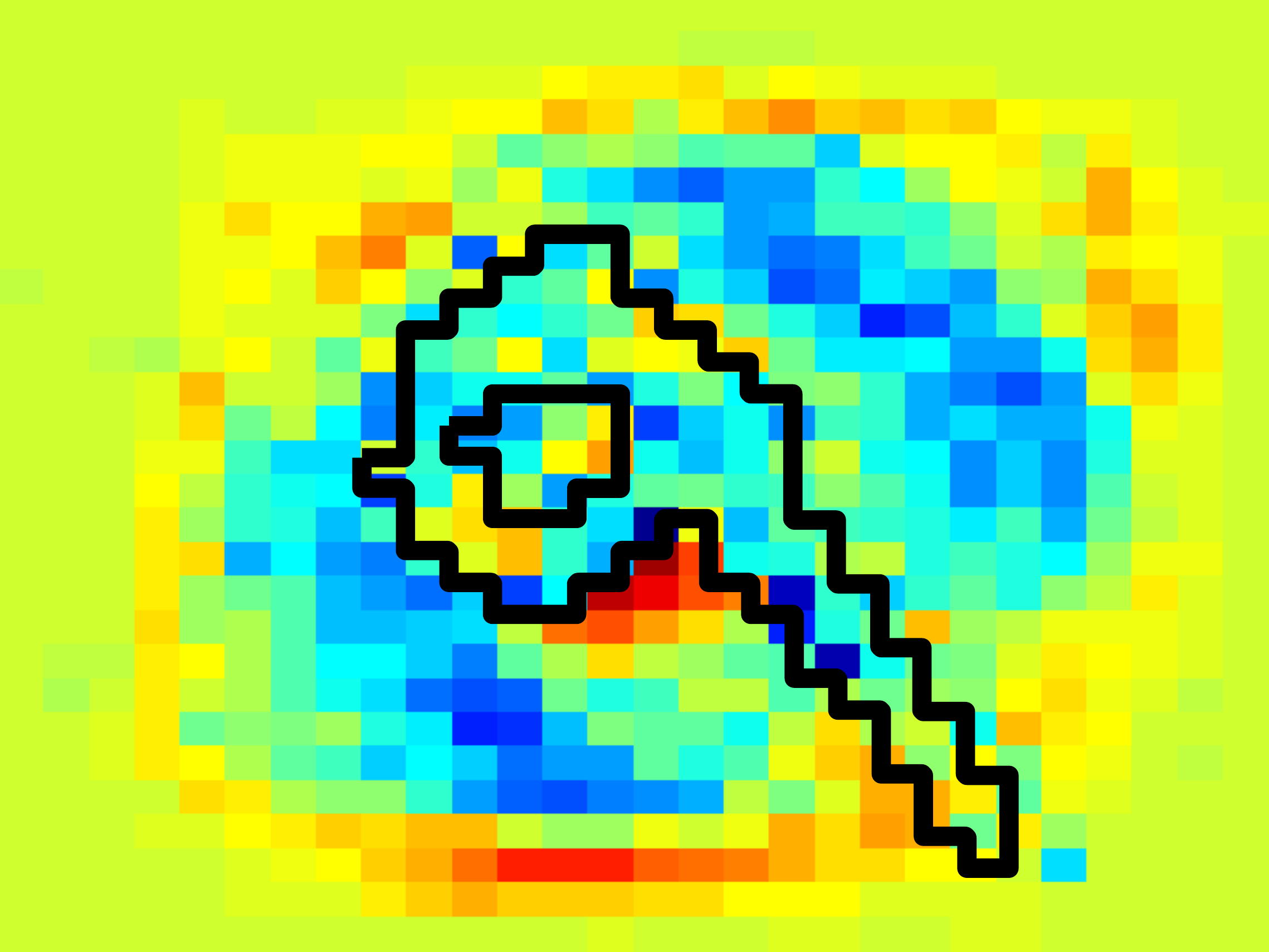}}&
		\subcaptionbox*{}{\includegraphics[width=0.09\textwidth]{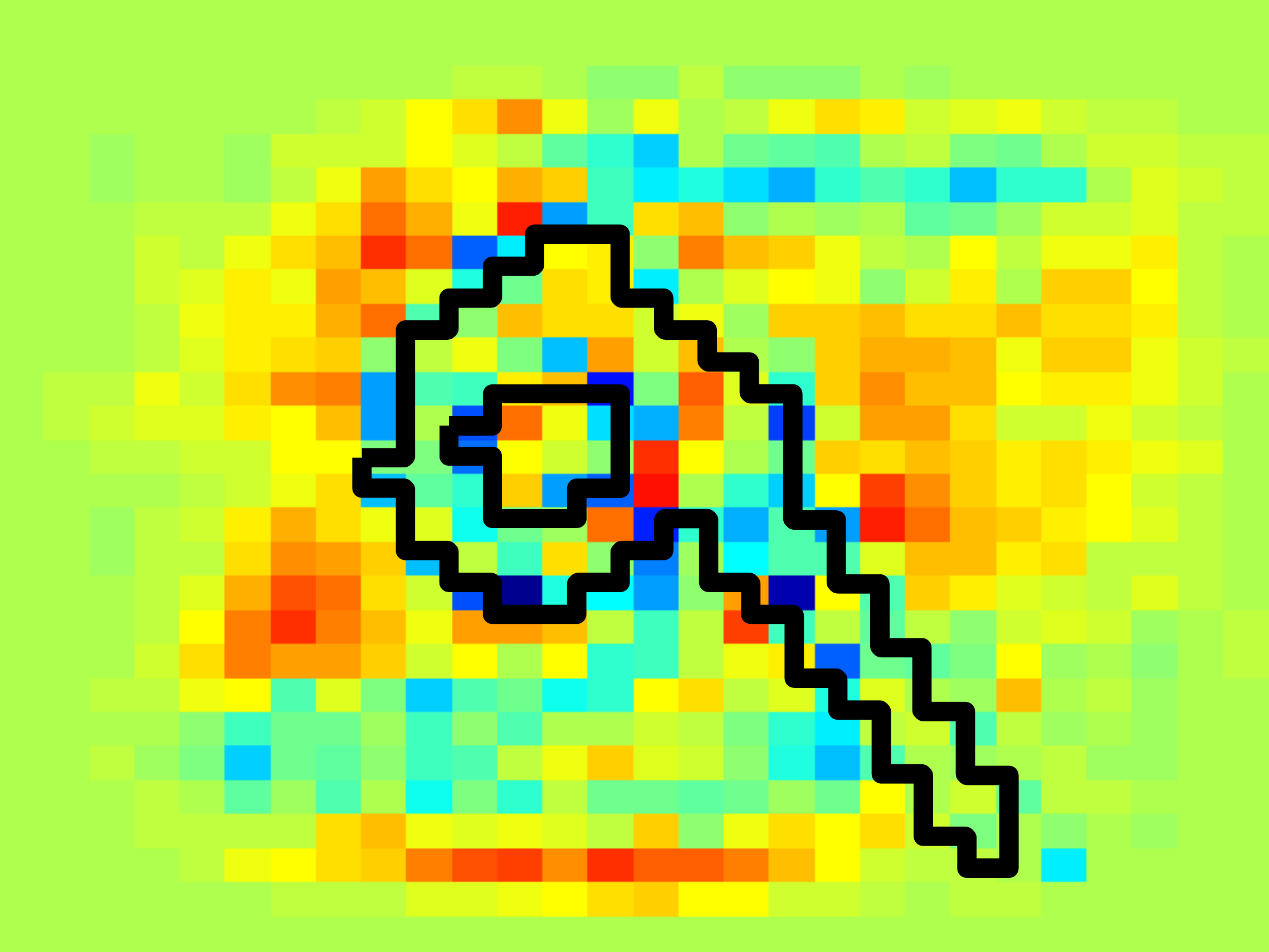}}&
		\subcaptionbox*{}{\includegraphics[width=0.09\textwidth]{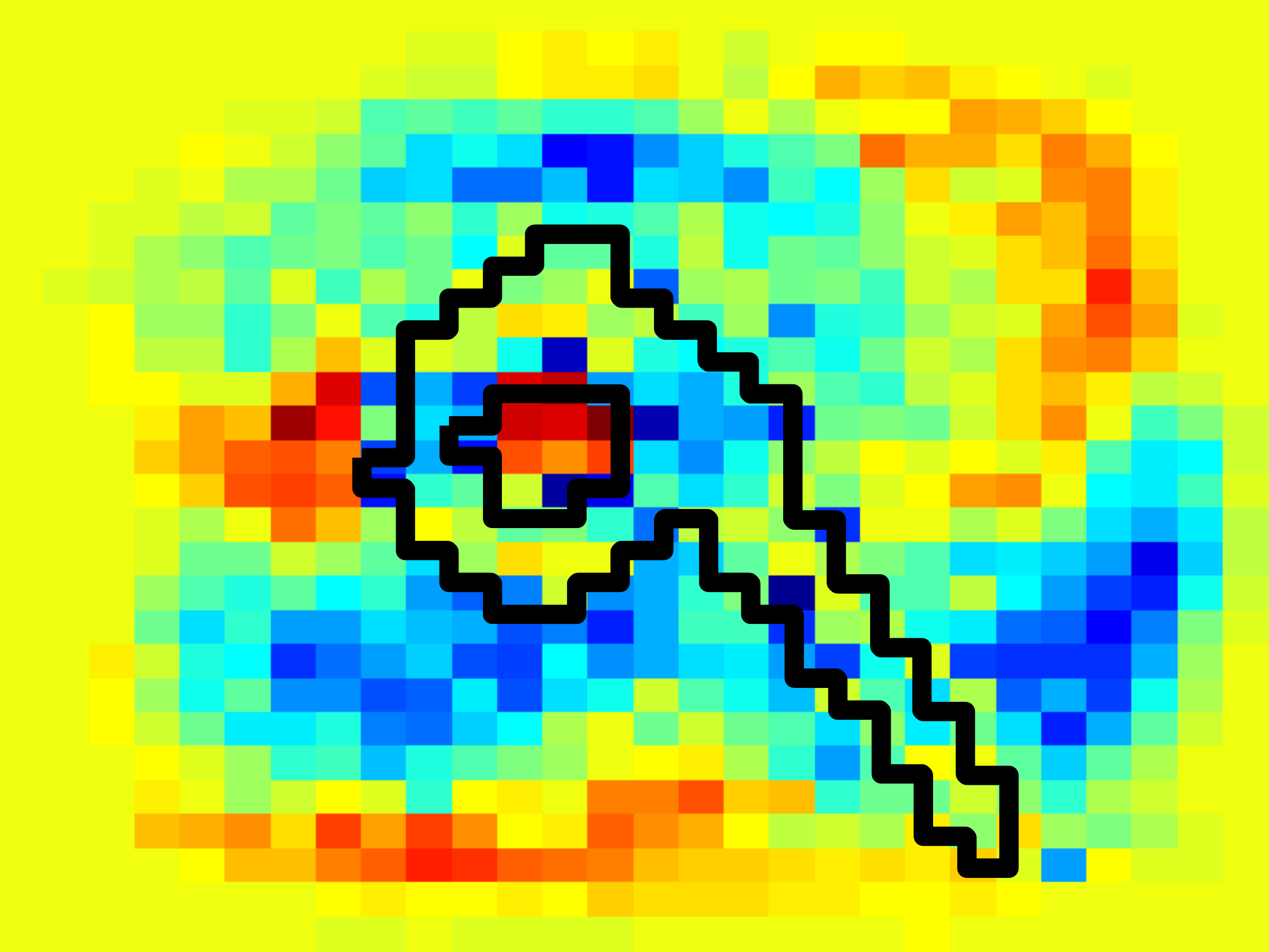}}&
		\subcaptionbox*{}{\includegraphics[width=0.09\textwidth]{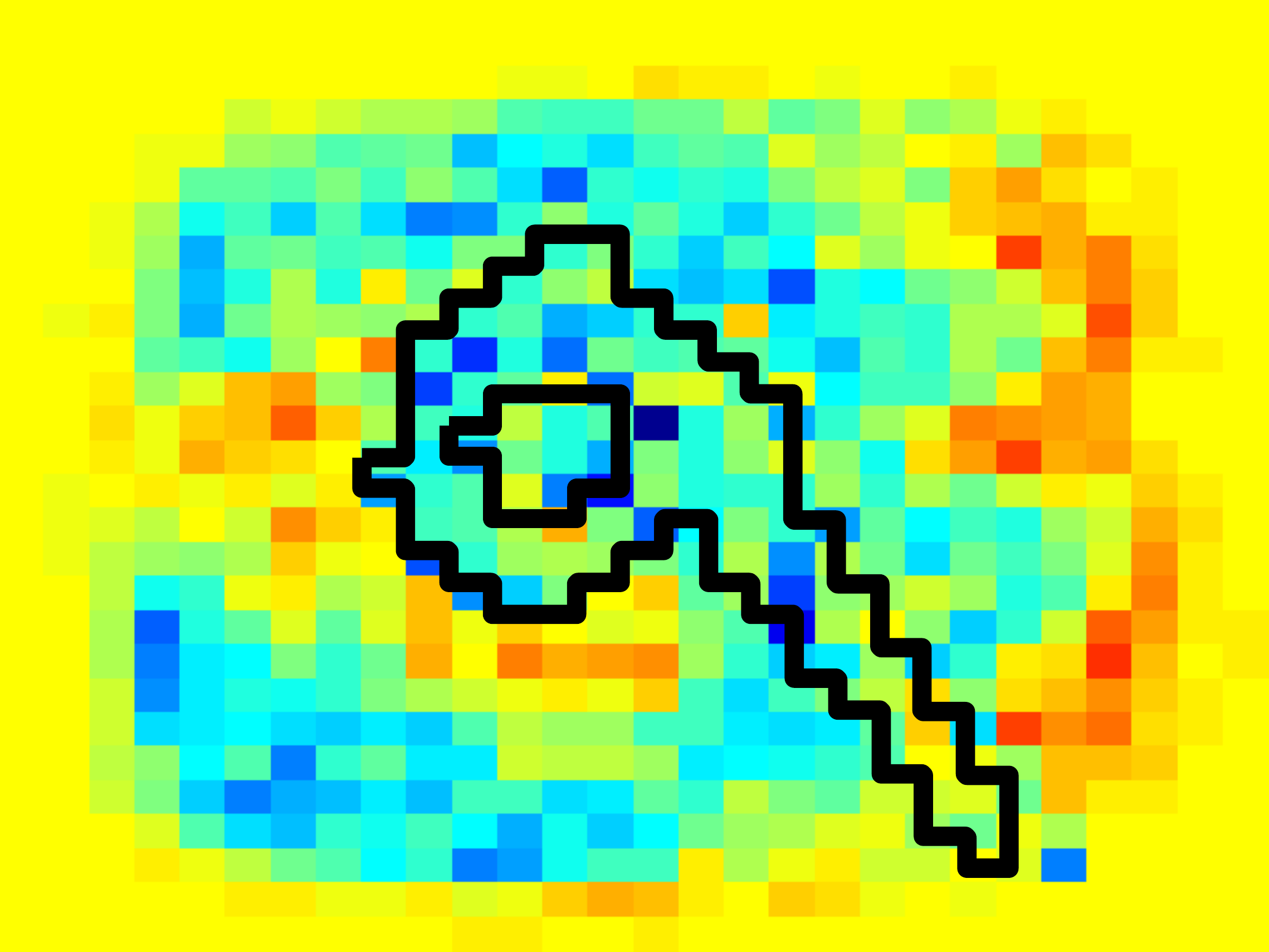}}&
		\subcaptionbox*{}{\includegraphics[width=0.09\textwidth]{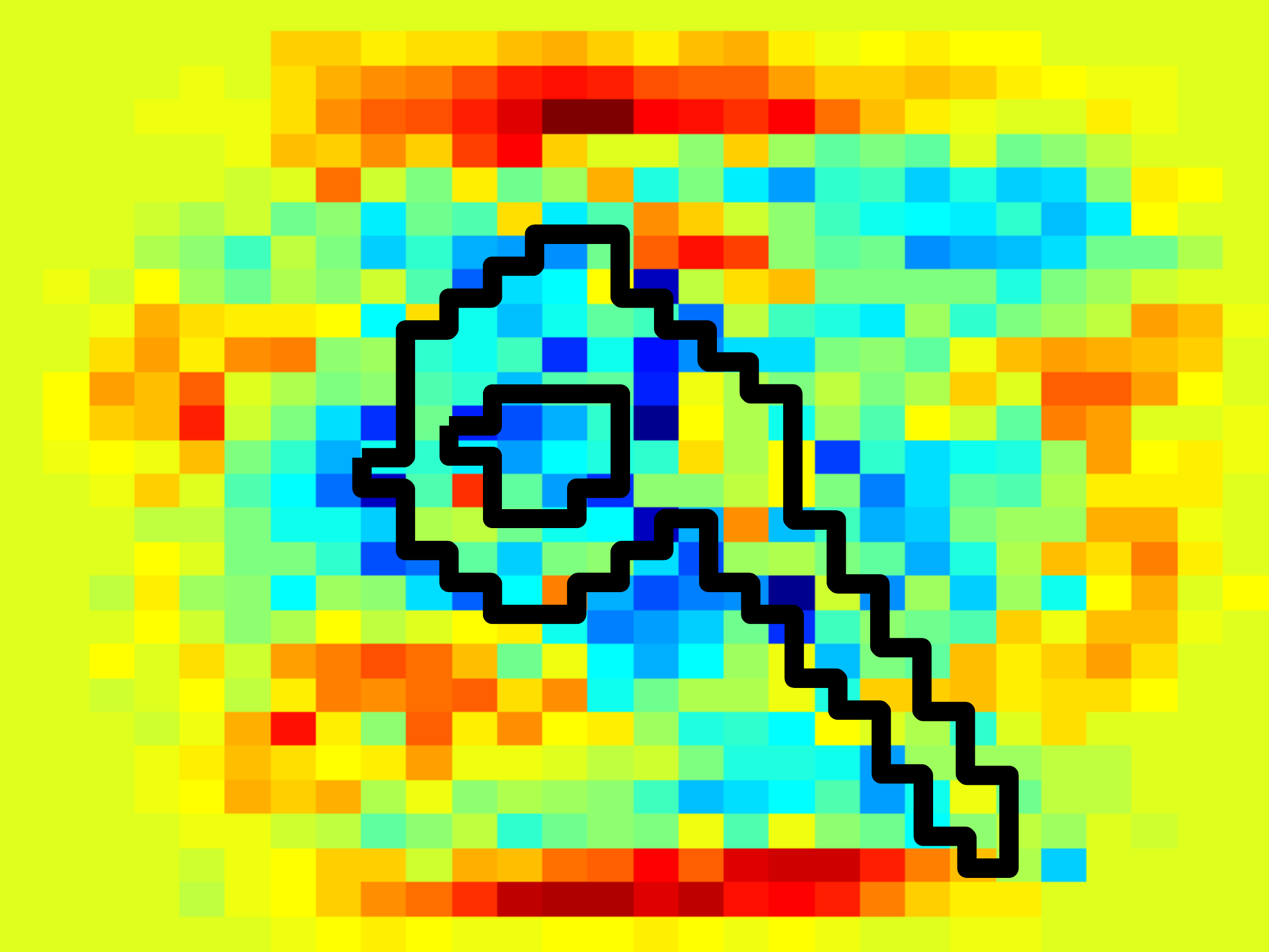}}&
		\subcaptionbox*{}{\includegraphics[width=0.09\textwidth]{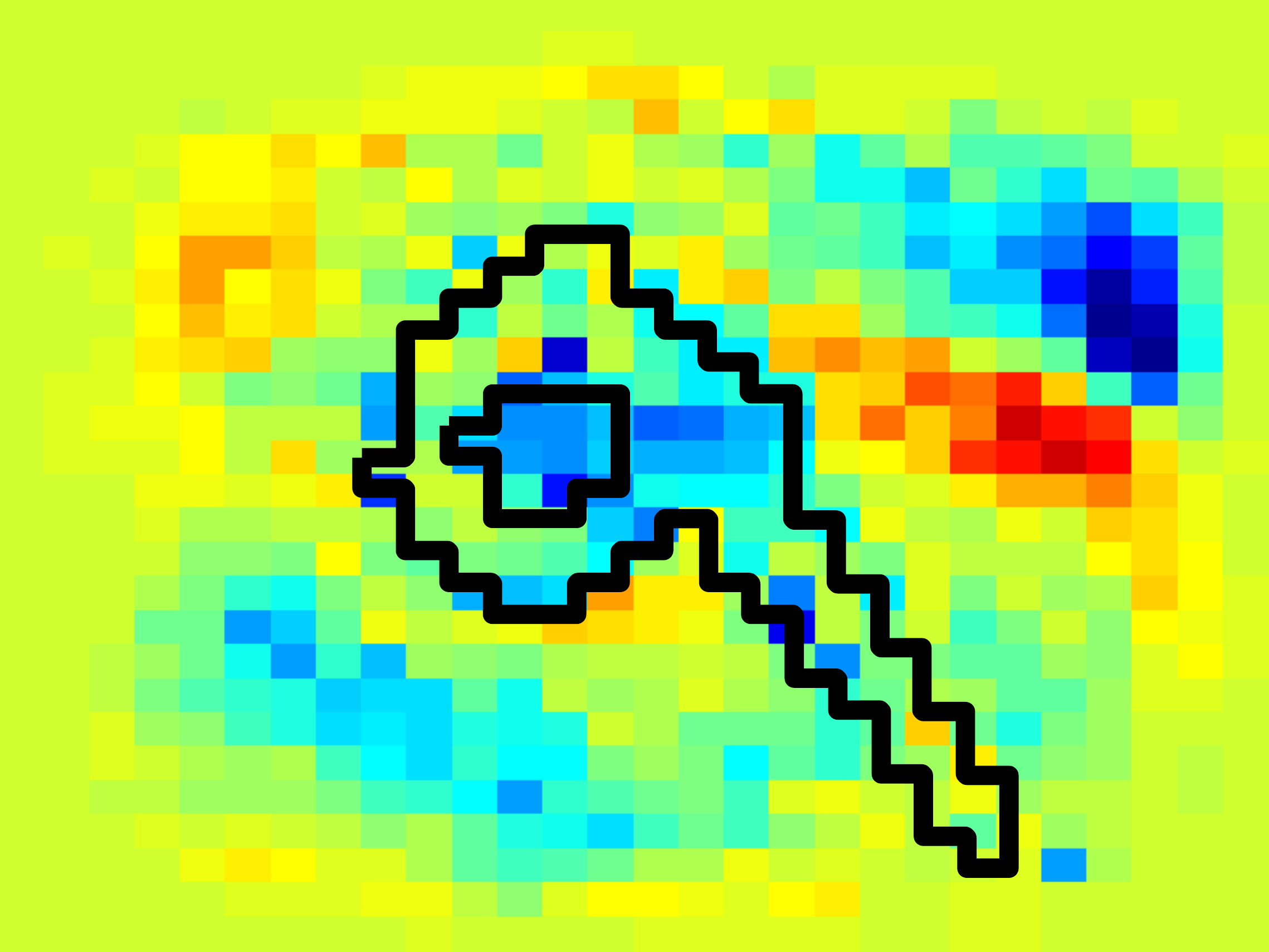}}&
		\subcaptionbox*{}{\includegraphics[width=0.09\textwidth]{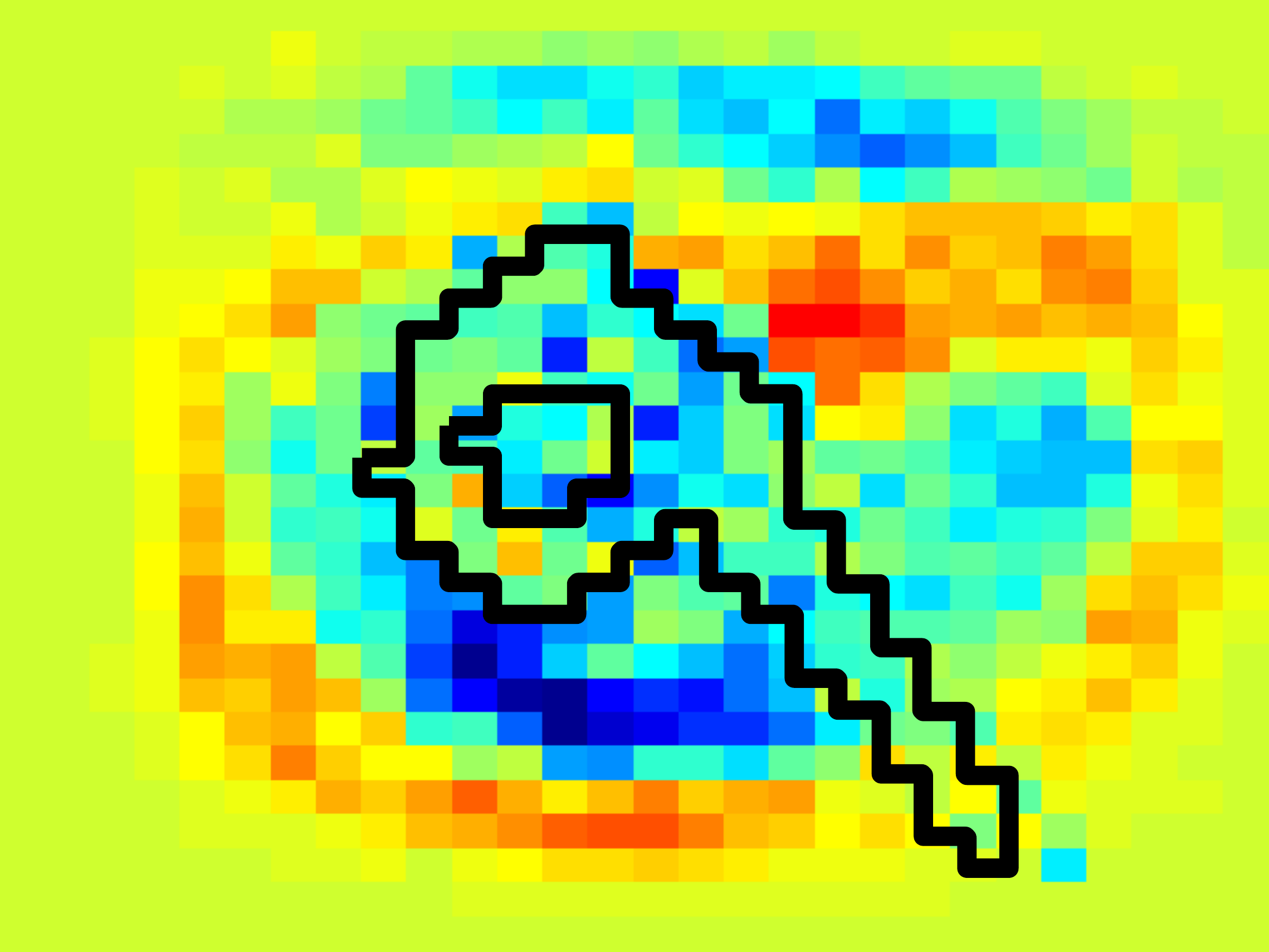}}&
		\subcaptionbox*{}{\includegraphics[width=0.09\textwidth]{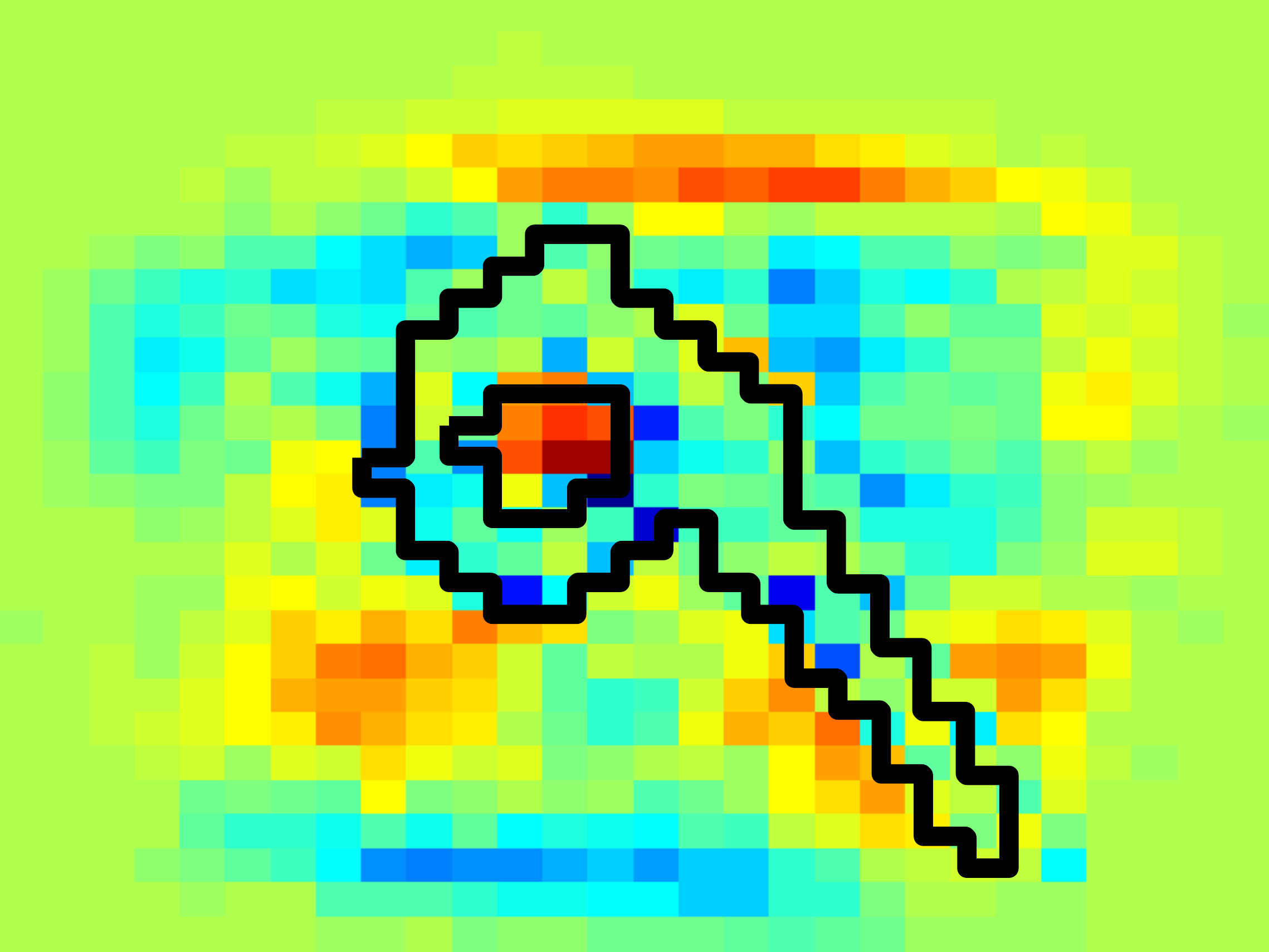}}&
		\subcaptionbox*{}{\includegraphics[width=0.09\textwidth]{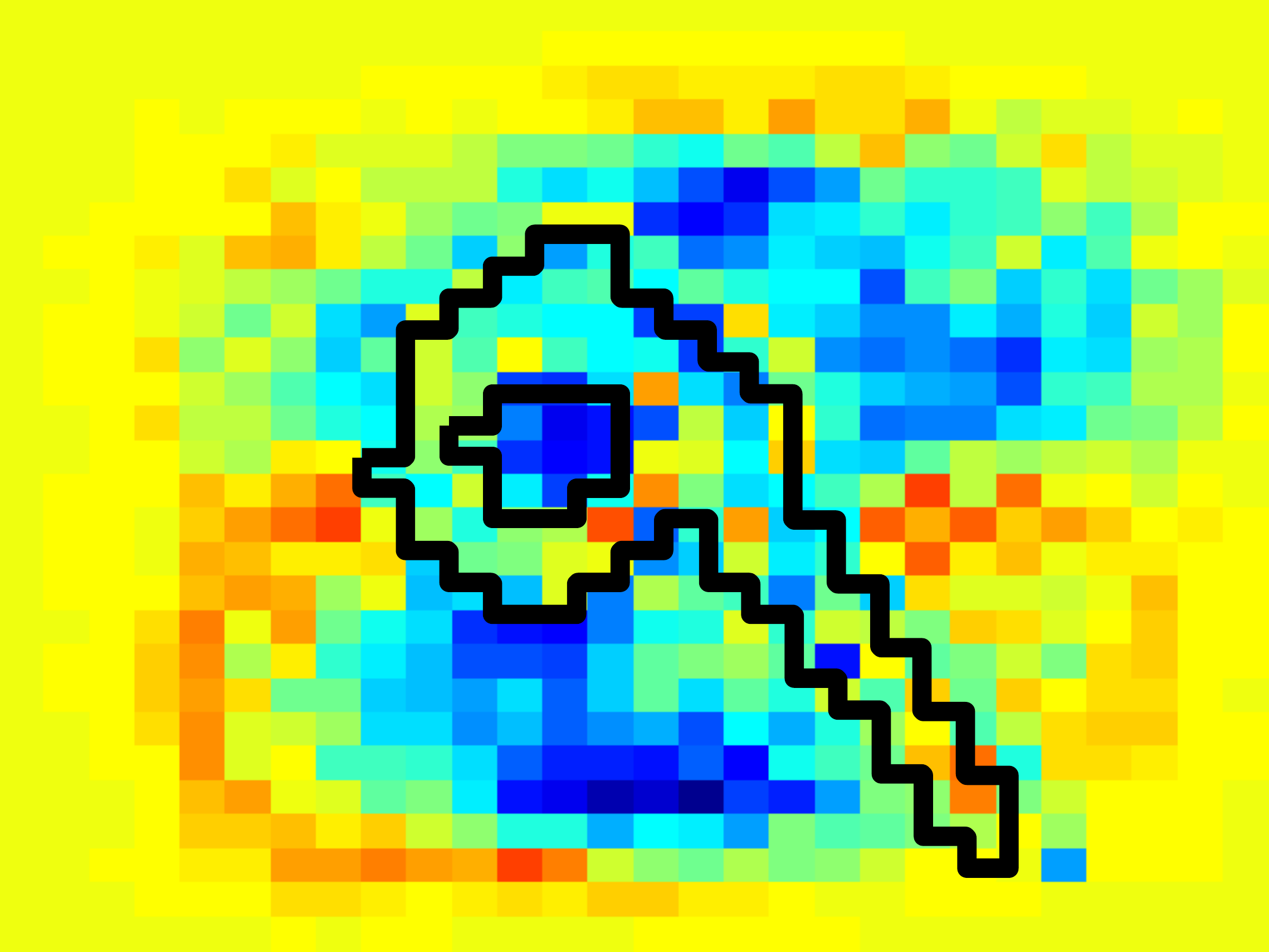}}&
		\subcaptionbox*{}{\includegraphics[width=0.09\textwidth]{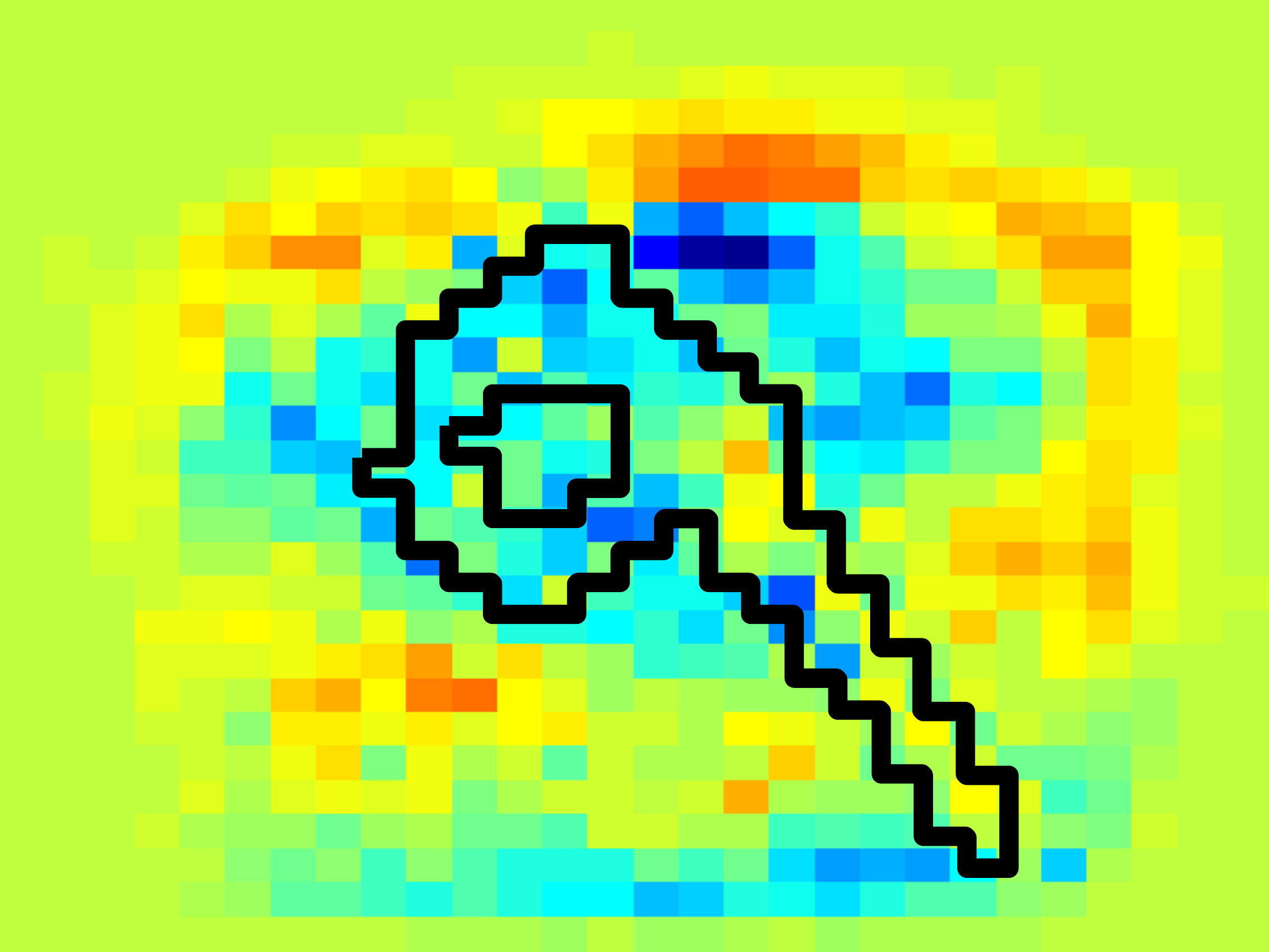}}
		\\[-1em]
		&0.47&0.89&0.65&0.84&0.95&0.80&0.72&0.64&0.66&1.02\\[0.5em]\hline
\end{tabular}
}

\caption{Correct classifications and its corresponding feature strength heatmap and aggregated feature strength values for each output class}
\label{fig:FSF_allClass}
\end{figure*}

Fig~\ref{fig:SEF_to_FSF} shows the synaptic efficacy functions and the extracted feature strength function of one input feature for one of the MNIST class (digit 7). Here $\hat{t}_o$ is chosen as $1.66ms$. It may also be noted that the synaptic efficacy functions beyond $\hat{t}_o$ (right side to $\hat{t}_o$) are not used to extract the FSF, as the $H\big(\hat{t}_o-s_i^r\big)$ acts as a low-pass filter in the equation~\ref{equation:FSF}. From Fig~\ref{fig:FSF}, the influence of that feature on the classification can be described. The feature strength is positive for those feature values within the range of $[0.4,0.6]$ or $[0.75,0.88]$, thereby increasing the likelihood of predicting the class label as class 7 (digit 7). However, the collective feature strength value from all the features are used to predict the class label as in equation~\ref{equation:classification_FSF}.

Figure~\ref{fig:FSF_allClass} shows the examples of correct classification for all the input digits ($0~to~9$). The first column shows the image of the input sample used and the subsequent columns show the feature strength heatmaps for the given input image and the aggregated feature strength values for each output class. In each row, the output classes corresponding to the correct classes produce the highest aggregated feature strength values for the inputs. This implies a correct classification. Values of heatmap are in the ranges of $[-0.05,0.05]$, where `blue', `green' and `red' colours represent $-0.05$, $0$ and $0.05$ respectively. The hue between `green' and `red' in the heatmap corresponds to positive feature strength values and hue between `blue' and `green' corresponds to negative feature strength values.

The reasoning for these predictions can be explained by the feature strength values. It can be seen clearly in the second row (the row corresponding to input digit `$1$') in figure~\ref{fig:FSF_allClass}, that each output neuron has a region with negative feature strengths (blue regions) that looks similar to the output class digit. That region acts as the template to match the input samples. Whenever an input image aligns well with the template, the aggregated feature strength increases as some parts of the template become positive (boosting). On the other hand, the aggregated feature strength decreases (weakening) if there is a mismatch. This boosting and weakening mechanism ensures that for a given output class, the correct class samples get higher aggregated feature strengths to make a correct prediction. The same interpretation can be transformed into the time domain where MC-SEFRON classifier actually makes the prediction.

\section{Conclusions}
\label{section:conclusion}

In this paper, a novel knowledge encoding method to extract knowledge from a trained Multi-Class SEFRON classifier and its interpretation have been presented. The knowledge encoding method ensures the consistency between the classification in the time domain and feature domain. First, the earlier developed binary-class SEFRON classifier is extended to handle multi-class classification problems. In the MC-SEFRON classifier, input data is encoded into spike patterns using the population encoding scheme. Binary-class SEFRON's learning rule (modified STDP rule) is used to train the MC-SEFRON classifier. Weights in an MC-SEFRON classifier are time-varying functions. The weighted postsynaptic potentials in the time domain are transformed into the feature domain as functions of features using the knowledge encoding method. Those transformed functions in the feature domain are referred to as Feature Strength Functions (FSF). A set of FSF for each class represents the knowledge extracted from MC-SEFRON classifier for the corresponding class.  FSFs enable one to easily interpret the prediction of the classifier. The correctness of the FSF (extracted knowledge) is measured by the classification accuracy when used directly. Aggregated values of feature strengths that are sampled from FSFs for a given input is used for classification. Performance of MC-SEFRON and FSFs have been validated using ten UCI machine learning datasets and the MNIST dataset. MC-SEFRON classifier is trained on the given dataset and the knowledge is extracted to interpret the predictions of the MC-SEFRON classifier. The classification accuracy obtained using the FSFs indicates that the loss of performance using the extracted knowledge is minimal. Hence, one can conclude that the logical explanation provided in the feature domain for the predictions of MC-SEFRON classifier is reliable. 

\section*{Acknowledgement}
This work was supported by the Science and Engineering Research Council of A*STAR (Agency for Science, Technology and Research), Singapore.

\bibliographystyle{model5-names}\biboptions{authoryear}
\bibliography{Sefron_multiclass}

\end{document}